%% file: main.tex
\documentclass{article}
\makeatletter\def\input@path{{./}}\makeatother
\usepackage[accepted]{icml2025}   %
\makeatletter\renewcommand{\ICML@appearing}{Preprint.}\makeatother
\makeatletter\renewcommand{\printAffiliationsAndNotice}[1]{\stepcounter{@affiliationcounter}%
{\let\thefootnote\relax\footnotetext{\raggedright\hspace*{-\footnotesep}\ifdefined\isaccepted #1\fi%
\forloop{@affilnum}{1}{\value{@affilnum} < \value{@affiliationcounter}}{\textsuperscript{\arabic{@affilnum}}%
\ifcsname @affilname\the@affilnum\endcsname\csname @affilname\the@affilnum\endcsname\else
{\bf AUTHORERR: Missing \textbackslash{}icmlaffiliation.}\fi}.
\ifdefined\paperemails Email: \paperemails.\fi
\\\Notice@String}}}\makeatother
\usepackage[utf8]{inputenc}
\usepackage[T1]{fontenc}
\usepackage{amsmath,amssymb,amsthm}
\usepackage{booktabs}
\usepackage{array}
\usepackage{longtable}
\usepackage{placeins}
\usepackage{graphicx}
\usepackage{xcolor}
\usepackage[expansion=false]{microtype}
\usepackage{hyperref}
\definecolor{mydarkblue}{rgb}{0,0.08,0.45}
\hypersetup{colorlinks=true, linkcolor=mydarkblue, citecolor=mydarkblue, filecolor=mydarkblue, urlcolor=mydarkblue,
  pdfborder=0 0 0, pdftitle={Infinite Worlds from One Example: What Local Generators Must Share}, pdfauthor={Amin Najafgholizadeh and Mohsen Afsharchi}}
\graphicspath{{./}{figs/}}
\newtheorem{theorem}{Theorem}
\newtheorem{proposition}{Proposition}
\newtheorem{corollary}{Corollary}
\newtheorem{definition}{Definition}
\newcommand{\Z}{\mathbb{Z}}

\icmltitlerunning{Infinite worlds from one example}

\begin{document}
\input{tables/macros.tex}
\input{tables/macros_matched.tex}
\input{tables/macros_baselines.tex}
\input{tables/macros_cert.tex}
\input{tables/macros_v12.tex}

\input{tables/macros_pending.tex}
\twocolumn[
\icmltitle{Infinite Worlds from One Example: What Local Generators Must Share}
\begin{icmlauthorlist}
\icmlauthor{Amin Najafgholizadeh}{znu}
\icmlauthor{Mohsen Afsharchi}{znu}
\end{icmlauthorlist}
\icmlaffiliation{znu}{Department of Computer Engineering, University of Zanjan}   %
\gdef\paperemails{\mbox{amin.ngz@znu.ac.ir}; \mbox{afsharchi@znu.ac.ir}}   %
\icmlkeywords{procedural content generation, wave function collapse, neural cellular automata, local algorithms, factors of i.i.d.}
\vskip 0.3in
]
\printAffiliationsAndNotice{}

\begin{abstract}
Seeded noise lets games generate any region of an unbounded world on demand. We ask what such a generator needs when
every small window must come from one example, the constraint behind Wave Function Collapse. The difficulty is that
local constraints can require agreement over arbitrary distances: in a brick wall, each row keeps its place between the mortar lines. We study local generators, each cell a fixed function of the randomness near
it. When the constraints carry a quantity that varies across the example, no generator that applies one rule,
ignoring position, to independent randomness at every cell can be both exactly valid and faithful to the example's statistics,
however far each cell reads. At radius $r$, staying faithful forces forbidden windows at a density of at least order
$1/r$, and a construction attains this order. To vary the quantity between exactly valid worlds, a generator needs
randomness shared along its direction. We give a finitely checkable condition under which row types
drawn from the seed and the position modulo a fixed period suffice for exact validity without learning. A search over
the example's patches certifies these quantities and names the input each needs. Across \nPublic\ public examples of
Wave Function Collapse and \fxNExamples\ from other implementations, it certifies only examples where a learned
generator on independent randomness stays far from valid after repair. Given that input, a neural cellular
automaton and a diffusion model become valid on \certInputMethodValid\ and \ddThreeValid\ of the
\certInputCount\ certified examples that receive one, and on none from independent randomness. Valid means at most one forbidden window in a thousand after repair. With octaves of seeded noise elsewhere, the automaton is valid on
\nMethodAll\ of the \nPublic\ public examples and \fxMethodValid\ of the \fxNExamples\ from other implementations, and
repair keeps random access.
\end{abstract}

\input{sec_intro}
\input{sec_setting}
\input{sec_theory}
\input{sec_cert}
\input{sec_generators}
\input{sec_evidence}     %

\input{sec_results}      %
\input{sec_diffusion}    %
\input{sec_related}
\input{sec_limits}
\input{sec_conclusion}

\bibliographystyle{icml2025}
\bibliography{refs}

\appendix
\onecolumn
\input{app_roadmap}
\FloatBarrier
\input{app_setting}
\FloatBarrier
\input{app_proofs}
\FloatBarrier
\input{app_converse}
\FloatBarrier
\input{app_procedure}
\FloatBarrier
\input{app_method}
\FloatBarrier
\input{app_setup}
\FloatBarrier

\input{app_tables}
\FloatBarrier
\input{app_evidence}
\FloatBarrier

\input{app_mincoinrun}
\FloatBarrier
\input{app_smile}
\FloatBarrier
\input{app_diffusion}
\FloatBarrier

\input{app_results}
\FloatBarrier
\input{app_record}
\FloatBarrier
\input{supp_tables}
\FloatBarrier
\end{document}

%% file: tables/macros.tex
\newcommand{\mainR}{32}
\newcommand{\nSamples}{52}
\newcommand{\nPone}{7}
\newcommand{\nPoneOf}{52}
\newcommand{\nPtwo}{50}
\newcommand{\nPtwoOf}{52}
\newcommand{\medVsixteenRawRound}{3\%}
\newcommand{\medVsixteenRaw}{3.1\%}
\newcommand{\medVsixteenRep}{2.8\%}
\newcommand{\nVsixteenLeFive}{35}
\newcommand{\nVsixteenLeOne}{8}
\newcommand{\nRthirtytwoCollapsed}{0}
\newcommand{\nRthirtytwoLegs}{0}
\newcommand{\nRthirtytwoBetter}{0}
\newcommand{\medTrainMin}{56}
\newcommand{\medWallBig}{30}
\newcommand{\ctrlColthreeVsixteen}{0.9\%}
\newcommand{\ctrlColthreeVsixtyfour}{0.7\%}
\newcommand{\ctrlColthreeRatio}{0.80}
\newcommand{\ctrlStripesVsixteen}{6.4\%}
\newcommand{\ctrlStripesVsixtyfour}{4.7\%}
\newcommand{\ctrlStripesRatio}{0.74}
\newcommand{\ctrlChessVsixteen}{5.9\%}
\newcommand{\ctrlChessVsixtyfour}{3.6\%}
\newcommand{\ctrlChessRatio}{0.61}
\newcommand{\ctrlProperthreeVsixteen}{0.5\%}
\newcommand{\ctrlProperthreeVsixtyfour}{0.4\%}
\newcommand{\ctrlProperthreeRatio}{0.81}
\newcommand{\bestSample}{Font}
\newcommand{\bestVsixteen}{0.1\%}
\newcommand{\worstSample}{RedMaze}
\newcommand{\worstVsixteen}{12.3\%}
\newcommand{\nCoinEditOf}{56}
\newcommand{\nRules}{56}
\newcommand{\nSweepSeeds}{1}
\newcommand{\medVsixtyfourRaw}{0.0362}
\newcommand{\medVsixtyfourRep}{0.0276}
\newcommand{\nSeams}{56}
\newcommand{\nCoinEdit}{56}
\newcommand{\nMonotone}{19}
\newcommand{\nCollapsed}{0}
\newcommand{\nMone}{6}
\newcommand{\nMoneOf}{12}
\newcommand{\medMended}{$1.2\cdot10^{-3}$}
\newcommand{\medChanged}{4.1\%}
\newcommand{\mendedAngular}{0.15\%}
\newcommand{\mendedCave}{0.00\%}
\newcommand{\mendedChess}{6.09\%}
\newcommand{\mendedCity}{2.68\%}
\newcommand{\mendedColthree}{0.76\%}
\newcommand{\mendedColfour}{40.47\%}
\newcommand{\mendedKnot}{0.09\%}
\newcommand{\mendedMaze}{0.00\%}
\newcommand{\mendedPaths}{0.02\%}
\newcommand{\mendedQud}{1.97\%}
\newcommand{\mendedRooms}{0.03\%}
\newcommand{\mendedSand}{0.00\%}
\newcommand{\mendedSewers}{1.04\%}
\newcommand{\mendedSkewtwo}{2.57\%}
\newcommand{\mendedStripes}{6.20\%}
\newcommand{\mendedWall}{8.83\%}
\newcommand{\nMfour}{16}
\newcommand{\nMfourOf}{16}
\newcommand{\variantFinalSixteen}{0.504}
\newcommand{\variantPatternSixteen}{2.55}
\newcommand{\nMendAll}{52}
\newcommand{\nMendAllGood}{16}
\newcommand{\nMendAllPartial}{8}
\newcommand{\nMendAllBad}{28}
\newcommand{\nCollapsedFirst}{0}
\newcommand{\nRulesFirst}{56}
\newcommand{\nRthirtytwoCollapsedFirst}{9}
\newcommand{\nRthirtytwoLegsFirst}{52}
\newcommand{\nRthirtytwoBetterFirst}{8}
\newcommand{\medFirstSixteen}{5.9\%}
\newcommand{\nPoneFirst}{2}
\newcommand{\nPoneFirstOf}{52}
\newcommand{\medTrainMinFirst}{19}
\newcommand{\medWallBigFirst}{13}
\newcommand{\nPtwoFirst}{49}
\newcommand{\nPtwoFirstOf}{52}
\newcommand{\nMonotoneFirst}{11}
\newcommand{\ratioFinalFirst}{0.52}
\newcommand{\spearmanFinalFirst}{0.94}
\newcommand{\nSpearmanFinalFirst}{52}
\newcommand{\slopeCave}{-1.35}
\newcommand{\slopeSixtyfourCave}{-0.59}
\newcommand{\slopeChess}{-0.91}
\newcommand{\slopeSixtyfourChess}{-0.90}
\newcommand{\slopeColthree}{-0.89}
\newcommand{\slopeSixtyfourColthree}{-0.76}
\newcommand{\slopeFont}{-0.78}
\newcommand{\slopeSixtyfourFont}{-0.29}
\newcommand{\slopeHogs}{-1.07}
\newcommand{\slopeSixtyfourHogs}{-0.60}
\newcommand{\slopePaths}{-0.96}
\newcommand{\slopeSixtyfourPaths}{-0.54}
\newcommand{\slopeStripes}{-0.94}
\newcommand{\slopeSixtyfourStripes}{-0.88}
\newcommand{\slopeTown}{-0.54}
\newcommand{\slopeSixtyfourTown}{-0.17}
\newcommand{\slopeWall}{-0.71}
\newcommand{\slopeSixtyfourWall}{-0.45}
\newcommand{\nMendEight}{21}
\newcommand{\nMendEightRows}{52}
\newcommand{\nMendFourOnly}{16}
\newcommand{\nMendEightOnly}{5}
\newcommand{\nMendPartialEight}{10}
\newcommand{\nMendWallEight}{21}
\newcommand{\mendFourShare}{a third}
\newcommand{\nConstSat}{44}
\newcommand{\nConstSatOf}{52}
\newcommand{\nConstSatWall}{15}
\newcommand{\nNonConst}{8}
\newcommand{\nNonConstWall}{6}
\newcommand{\nonConstNames}{Maze, Mazelike, Nested, RedMaze, ScaledMaze, SimpleMaze, SmileCity and Wall}

%% file: tables/macros_matched.tex
\newcommand{\nPtwoM}{32}
\newcommand{\nPtwoMOf}{37}
\newcommand{\nPtwoT}{32}
\newcommand{\nRatioAboveTwo}{3}
\newcommand{\medRuleKL}{0.048}
\newcommand{\medWfcKL}{0.121}
\newcommand{\medRatioKL}{0.39}
\newcommand{\medIndepKL}{1.37}
\newcommand{\nPtwoMSmall}{42}
\newcommand{\nPtwoMSmallOf}{51}
\newcommand{\nPtwoTSmall}{49}
\newcommand{\nRatioAboveTwoSmall}{2}
\newcommand{\medRuleKLSmall}{0.079}
\newcommand{\medWfcKLSmall}{0.210}
\newcommand{\medRatioKLSmall}{0.49}
\newcommand{\medIndepKLSmall}{3.56}
\newcommand{\nPtwoMMid}{43}
\newcommand{\nPtwoMMidOf}{50}
\newcommand{\medRatioKLMid}{0.40}
\newcommand{\nPtwoMExt}{43}
\newcommand{\nPtwoMExtOf}{49}
\newcommand{\nBtSolved}{10}
\newcommand{\nBtOf}{12}
\newcommand{\nBtSolvedAll}{12}
\newcommand{\nBtOfAll}{15}

%% file: tables/macros_baselines.tex
\newcommand{\nBone}{11 of 12}
\newcommand{\nBoneSixtyfour}{10 of 12}
\newcommand{\nBtwo}{8 of 12}
\newcommand{\nBtwoSixtyfour}{8 of 12}
\newcommand{\nBthree}{3 of 12}
\newcommand{\nBthreeSixtyfour}{3 of 12}
\newcommand{\nBfour}{11 of 12}
\newcommand{\medRatioMHraw}{3.9}
\newcommand{\medRatioMHrawThirtytwo}{5.2}
\newcommand{\medRatioFonerep}{5.7}
\newcommand{\medRatioFonerepThirtytwo}{7.6}
\newcommand{\medRatioMHmended}{6.8}
\newcommand{\nFalsifier}{0 of 12}
\newcommand{\floorRatioLo}{four}
\newcommand{\floorRatioHi}{eight}

%% file: tables/macros_cert.tex
\newcommand{\certRadius}{172}
\newcommand{\certExamples}{Cave and Maze}
\newcommand{\certSeeds}{3}
\newcommand{\certRegion}{the full canvas}
\newcommand{\certRegionWidth}{1024}
\newcommand{\certPeriodThree}{112}
\newcommand{\certPvalThreeCanvas}{$2^{-8}$}
\newcommand{\certPvalThree}{$2^{-24}\approx 6\cdot10^{-8}$}
\newcommand{\certPeriodTwo}{340}
\newcommand{\certPvalTwo}{$2^{-9}\approx 0.002$}
\newcommand{\certSandPval}{$2^{-8}$}
\newcommand{\certWindows}{3\,133\,452}
\newcommand{\certBound}{9.6\cdot10^{-7}}
\newcommand{\certPeriod}{36}
\newcommand{\certPeriodOne}{21}

%% file: tables/macros_v12.tex
\newcommand{\certResistantCount}{13}%
\newcommand{\certResistantOf}{21}%
\newcommand{\certResistantWords}{thirteen}%
\newcommand{\certOthersCount}{0}%
\newcommand{\resistMazeNames}{RedMaze, ScaledMaze and SimpleMaze}%
\newcommand{\certOthersOf}{31}%
\newcommand{\certMendedOf}{21}%
\newcommand{\certPartialOf}{10}%
\newcommand{\certMendedCertified}{0}%
\newcommand{\certCtwoCount}{4}%
\newcommand{\certCtwoOf}{6}%
\newcommand{\certCthreePoints}{57}%
\newcommand{\certUncertifiedResistant}{8}%
\newcommand{\certUncertifiedResistantWords}{eight}%
\newcommand{\certUncertifiedNames}{Cat, ColoredCity, Lake, Nested, ScaledMaze, Sewers, Skew1 and Skew2}%
\newcommand{\certOnewayCount}{1}%
\newcommand{\certOnewayWords}{one}%
\newcommand{\certSideTestCap}{1.5}%
\newcommand{\certSideTestPMin}{7}%
\newcommand{\certSideTestPMax}{11}%
\newcommand{\townP}{6}%
\newcommand{\townValues}{2}%
\newcommand{\townMassBig}{0.67}%
\newcommand{\townMassSmall}{0.33}%
\newcommand{\townKappaF}{0.22}%
\newcommand{\townKappaFFrac}{2/9}%
\newcommand{\townKappaZero}{0.146}%
\newcommand{\townKappaFive}{0.116}%
\newcommand{\townFloor}{$1.3\cdot10^{-4}$}%
\newcommand{\townRowPeriod}{3}%
\newcommand{\constSatCount}{44}%
\newcommand{\constSatOf}{52}%
\newcommand{\certConstCount}{9}%
\newcommand{\certConstWords}{nine}%
\newcommand{\certNonConstCount}{4}%
\newcommand{\certInvariantCount}{8}%
\newcommand{\certPhaseCount}{4}%
\newcommand{\certFisherP}{$3\cdot10^{-5}$}%
\newcommand{\certFisherPAll}{$8\cdot10^{-6}$}%
\newcommand{\certResistantRegCount}{12}%
\newcommand{\certTargetForcedWalls}{12}%
\newcommand{\certTargetForcedSix}{6}%
\newcommand{\certAllCount}{14}%
\newcommand{\certPthreeCount}{12}%
\newcommand{\certTargetForcedCount}{15}%
\newcommand{\certTargetForcedSlackCount}{15}%
\newcommand{\certTargetForcedMinSlack}{0.047}%
\newcommand{\certTargetForcedMinSlackExample}{Skyline2}%
\newcommand{\certBricksMajor}{0.875}%
\newcommand{\certFullSimpleMaze}{0.75}%
\newcommand{\certSpecSimpleMaze}{0.25}%
\newcommand{\certFullRedMaze}{0.75}%
\newcommand{\certSpecRedMaze}{0.25}%
\newcommand{\certFullFabric}{0.60}%
\newcommand{\certSpecFabric}{0.20}%
\newcommand{\certFloorMin}{$4.5\cdot10^{-4}$}%
\newcommand{\certFloorMax}{$1.1\cdot10^{-2}$}%
\newcommand{\certFloorMinExample}{3Bricks}%
\newcommand{\certFloorMaxExample}{RedMaze}%
\newcommand{\certDeltaAboveCount}{5}%
\newcommand{\certDeltaAboveWords}{five}%
\newcommand{\certDeltaAboveNames}{Flowers, MoreFlowers, Platformer, Skyline and Skyline2}%
\newcommand{\certDeltaAboveOf}{10}%
\newcommand{\certBigPNames}{3Bricks and SmileCity}%
\newcommand{\townDeltaLo}{0.18}%
\newcommand{\townDeltaHi}{0.25}%
\newcommand{\lpEtaZero}{0.01}%
\newcommand{\lpKappaEtaMin}{0.030}%
\newcommand{\lpKappaEtaMinExample}{Skyline2}%
\newcommand{\lpKappaEtaPositive}{14}%
\newcommand{\lpSkylineTwoKappaFive}{0.047}%
\newcommand{\lpSkylineTwoIneq}{0.015}%
\newcommand{\lpEtaRule}{0.001}%
\newcommand{\lpRuleDeltaStillCount}{3}%
\newcommand{\lpBigPFreeNames}{none}%
\newcommand{\lpBigPBoundNames}{3Bricks}%
\newcommand{\lpBigPBoundKappa}{0.051}%
\newcommand{\lpBigPBoundKappaF}{0.125}%
\newcommand{\lpRuleDeltaStillNames}{3Bricks, Platformer and Skyline2}%
\newcommand{\certFloorRepairMin}{$3.4\cdot10^{-4}$}%
\newcommand{\certFloorRepairMax}{$8.2\cdot10^{-3}$}%
\newcommand{\certRepairOverFloorMin}{9}%
\newcommand{\certRepairOverFloorMax}{181}%
\newcommand{\certRepairOverFloorMinExample}{SimpleWall}%
\newcommand{\certRepairOverFloorMaxExample}{3Bricks}%
\newcommand{\certRepairRateMin}{3\%}%
\newcommand{\certRepairRateMax}{11\%}%
\newcommand{\certResidualOverFloorMin}{51}%
\newcommand{\certResidualOverFloorMax}{556}%
\newcommand{\colfourWindows}{966}%
\newcommand{\colfourProperWindows}{9\,612}%
\newcommand{\colfourShare}{10\%}%
\newcommand{\certFloorBelowCount}{5}%
\newcommand{\certFloorBelowWords}{five}%
\newcommand{\certFloorBelowNames}{3Bricks, Platformer, Skyline2, SmileCity and Town}%
\newcommand{\satSampledDefects}{150}%
\newcommand{\satCoinSeed}{4242}%
\newcommand{\satCanvas}{256}%
\newcommand{\satFailedBlocks}{700}%
\newcommand{\satFailedBlocksSolved}{0}%
\newcommand{\satWallFailedBlocks}{509}%
\newcommand{\satWallFailedBlocksSolved}{0}%
\newcommand{\satControlValidSolved}{160}%
\newcommand{\satControlValidOf}{160}%
\newcommand{\satControlPhaseUnsat}{100}%
\newcommand{\satControlPhaseOf}{100}%
\newcommand{\satGlucoseAgree}{75}%
\newcommand{\satGlucoseOf}{75}%
\newcommand{\satRuleMinusRandomLo}{23}%
\newcommand{\satRuleMinusRandomHi}{34}%
\newcommand{\satRuleMinusRandomCount}{4}%
\newcommand{\satRuleMinusRandomOf}{5}%
\newcommand{\satAngularDiff}{0}%
\newcommand{\satRuleUnsatLo}{77\%}%
\newcommand{\satRuleUnsatHi}{96\%}%
\newcommand{\satRandomUnsatLo}{43\%}%
\newcommand{\satRandomUnsatHi}{69\%}%
\newcommand{\satProfileWithin}{37}%
\newcommand{\satProfileOf}{46}%
\newcommand{\satProfileLo}{-16}%
\newcommand{\satProfileHi}{17}%
\newcommand{\satWallsPooledSat}{1.6\%}%
\newcommand{\satWallsNeverSat}{5}%
\newcommand{\satOriginalInstances}{13\,716}%
\newcommand{\satOriginalCave}{100\%}%
\newcommand{\satOriginalMaze}{81\%}%
\newcommand{\satOriginalWallUnsat}{100\%}%
\newcommand{\satOriginalLeftSat}{2.6\%}%
\newcommand{\satOriginalLeftN}{117}%
\newcommand{\satOriginalUnresolved}{0\%}%
\newcommand{\satPfourCount}{3}%
\newcommand{\satLeftCity}{0.0\%}%
\newcommand{\satLeftSkewtwo}{4.7\%}%
\newcommand{\satLeftSewers}{2.5\%}%
\newcommand{\witCoverageWall}{98\%}%
\newcommand{\witLinesWall}{14}%
\newcommand{\witCoverageSimpleWall}{100\%}%
\newcommand{\witLinesSimpleWall}{13}%
\newcommand{\witCoverageFlowers}{90\%}%
\newcommand{\witLinesFlowers}{8}%
\newcommand{\witCoverageSkyline}{65\%}%
\newcommand{\witLinesSkyline}{3}%
\newcommand{\witCoverageHeld}{3}%
\newcommand{\witCoverageOf}{4}%
\newcommand{\witSoundSat}{490}%
\newcommand{\witSoundWitnessed}{0}%
\newcommand{\witTotal}{3\,145}%
\newcommand{\witTargetedSolved}{0}%
\newcommand{\witBtwoCount}{0}%
\newcommand{\witControlSolved}{0}%
\newcommand{\witControlsCoverage}{0}%
\newcommand{\witSecondaryFound}{9}%
\newcommand{\witSecondaryOf}{3\,235}%
\newcommand{\witSkylineRhoFour}{48\%}%
\newcommand{\witSkylineRhoSixteen}{82\%}%
\newcommand{\wxRhoLo}{4}%
\newcommand{\wxRhoHi}{16}%
\newcommand{\wxRhoMid}{8}%
\newcommand{\wxRowInvCountWords}{eight}%
\newcommand{\wxShareRowInvMin}{64\%}%
\newcommand{\wxShareRowInvMinName}{Skyline}%
\newcommand{\wxShareRowInvMinNear}{65\%}%
\newcommand{\wxShareRowInvMax}{100\%}%
\newcommand{\wxShareRowInvMaxName}{SimpleWall}%
\newcommand{\wxOtherShareLo}{68\%}%
\newcommand{\wxOtherShareHi}{99\%}%
\newcommand{\wxSolvRowInv}{107}%
\newcommand{\wxSolvAll}{362}%
\newcommand{\wxUncertCountWords}{eight}%
\newcommand{\wxUncertPfour}{0.4\%}%
\newcommand{\wxUncertPfourMax}{2\%}%
\newcommand{\wxUncertPfourMaxName}{Cat}%
\newcommand{\wxPhaseLevelLo}{26\%}%
\newcommand{\wxPhaseLevelHi}{38\%}%
\newcommand{\wxSmileLevelMissWords}{one}%
\newcommand{\wxTownOrder}{60\%}%
\newcommand{\wxNegRegions}{6\,240}%
\newcommand{\wxPosEach}{150}%
\newcommand{\wxTrivPhaseLo}{97\%}%
\newcommand{\wxShareSixteenMin}{76\%}%
\newcommand{\wxShareSixteenMinName}{MoreFlowers}%
\newcommand{\shareRowCutLo}{4.6}%
\newcommand{\shareRowCutHi}{98}%
\newcommand{\shareRowKLLo}{2.4}%
\newcommand{\shareRowKLHi}{3.2}%
\newcommand{\shareRowCutLoR}{5}%
\newcommand{\shareRowCutHiR}{100}%
\newcommand{\shareRowKLLoR}{2}%
\newcommand{\shareRowKLHiR}{3}%
\newcommand{\shareResCtrlSimpleWall}{0.046}%
\newcommand{\shareResRowSimpleWall}{0.0005}%
\newcommand{\shareKLCtrlSimpleWall}{0.016}%
\newcommand{\shareKLRowSimpleWall}{0.053}%
\newcommand{\shareResCtrlSkyline}{0.066}%
\newcommand{\shareResRowSkyline}{0.005}%
\newcommand{\shareKLCtrlSkyline}{0.236}%
\newcommand{\shareKLRowSkyline}{0.576}%
\newcommand{\shareResCtrlWall}{0.094}%
\newcommand{\shareResRowWall}{0.021}%
\newcommand{\shareKLCtrlWall}{0.048}%
\newcommand{\shareKLRowWall}{0.138}%
\newcommand{\shareEoneCount}{6}%
\newcommand{\shareEtwoCount}{4}%
\newcommand{\shareEOf}{7}%
\newcommand{\sharePhaseEditCells}{0}%
\newcommand{\shareCaveKLCtrl}{0.026}%
\newcommand{\shareCaveKLRow}{0.028}%
\newcommand{\complDoneCount}{1}%
\newcommand{\complDtwoCount}{1}%
\newcommand{\complDthreeCount}{0}%
\newcommand{\pilotKLb}{0.016}%
\newcommand{\pilotKLstar}{0.021}%
\newcommand{\pilotRowKLStart}{0.053}%
\newcommand{\pilotRowKLBest}{0.036}%
\newcommand{\pilotRowKLStartRound}{0.05}%
\newcommand{\pilotRowKLBestRound}{0.04}%
\newcommand{\pilotOrdKLRound}{0.02}%
\newcommand{\pilotRowResLo}{$3.6\cdot10^{-5}$}%
\newcommand{\pilotRowResHi}{$5.1\cdot10^{-4}$}%
\newcommand{\pilotRowResRatioLo}{89}%
\newcommand{\pilotRowResRatioHi}{1265}%
\newcommand{\pilotSwitchRowLo}{0.02}%
\newcommand{\pilotSwitchRowHi}{0.09}%
\newcommand{\pilotSwitchOrdLo}{3.1}%
\newcommand{\pilotSwitchOrdHi}{3.2}%
\newcommand{\pilotRedrawRowLo}{0.12}%
\newcommand{\pilotRedrawRowHi}{0.13}%
\newcommand{\pilotRedrawOrd}{0.40}%
\newcommand{\pilotRedrawRatio}{0.32}%
\newcommand{\pilotColRes}{0.038}%
\newcommand{\pilotColKL}{0.063}%
\newcommand{\pilotColSwitch}{2.5}%
\newcommand{\pilotColRedraw}{0.31}%
\newcommand{\pilotColOverRow}{83}%
\newcommand{\pilotOrdRes}{0.046}%
\newcommand{\pilotTilingLo}{0.61}%
\newcommand{\pilotTilingHi}{0.65}%
\newcommand{\replayNormG}{0.04}%
\newcommand{\replayNormB}{0.08}%
\newcommand{\replayRawG}{0.61}%
\newcommand{\replayIid}{0.57}%
\newcommand{\replayRepetitionG}{0}%
\newcommand{\replayRepetitionShift}{1}%
\newcommand{\replayReuseCanvases}{1}%
\newcommand{\replayReuseCanvasesOf}{168}%
\newcommand{\replayReusePositions}{2}%
\newcommand{\replayReuseExample}{Village}%
\newcommand{\replayZeroCount}{16}%
\newcommand{\replayAlignedG}{0.36}%
\newcommand{\replayAlignedB}{0.41}%
\newcommand{\replayAlignedRatio}{0.87}%
\newcommand{\replayLocalCount}{21}%
\newcommand{\replayLocalOf}{21}%
\newcommand{\replayLocalMin}{99.9\%}%
\newcommand{\replayVfourCount}{14}%
\newcommand{\replayVfourTV}{13}%
\newcommand{\replayVfourAligned}{6}%
\newcommand{\replayVfourOrig}{21}%
\newcommand{\replayPopulation}{21}%
\newcommand{\replayCostRatio}{0.08}%
\newcommand{\replayCostRatioInv}{13}%
\newcommand{\replayRigidCount}{5}%
\newcommand{\replayRigidNames}{Chess, Fabric, RedMaze, SimpleMaze and Stripes}%
\newcommand{\replayTVLo}{0.06}%
\newcommand{\replayTVHi}{0.40}%
\newcommand{\replayWallReplayBLo}{0.65}%
\newcommand{\replayWallReplayBHi}{0.84}%
\newcommand{\replayWallReplayGLo}{0.28}%
\newcommand{\replayWallReplayGHi}{0.61}%
\newcommand{\replayWallAlignedBLo}{0.16}%
\newcommand{\replayWallAlignedBHi}{0.38}%
\newcommand{\replayWallAlignedGLo}{0.38}%
\newcommand{\replayWallAlignedGHi}{0.70}%
\newcommand{\replayWallDefectLo}{4\%}%
\newcommand{\replayWallDefectHi}{9\%}%
\newcommand{\raAlignedA}{0.40}%
\newcommand{\raAlignedG}{0.36}%
\newcommand{\raNormA}{0.10}%
\newcommand{\raNormG}{0.04}%
\newcommand{\raBetterCount}{17}%
\newcommand{\raBetterReplay}{17}%
\newcommand{\raBetterAligned}{8}%
\newcommand{\raOf}{21}%
\newcommand{\raTVG}{0.20}%
\newcommand{\raTVA}{0.21}%
\newcommand{\raTVBetterG}{13}%
\newcommand{\raCostRatio}{12}%
\newcommand{\raCostRatioLo}{6}%
\newcommand{\raCostRatioHi}{56}%
\newcommand{\raWallNormALo}{0.43}%
\newcommand{\raWallNormAHi}{0.67}%
\newcommand{\raWallNormGLo}{0.02}%
\newcommand{\raWallNormGHi}{0.09}%
\newcommand{\floorCheckCovered}{32}%
\newcommand{\floorCheckViolations}{0}%
\newcommand{\floorCheckMinSlack}{2.60}%
\newcommand{\floorRowMinSlack}{0.77}%
\newcommand{\floorRowNegative}{0}%
\newcommand{\floorPhaseNegative}{8}%
\newcommand{\floorRadiusRaw}{32}%
\newcommand{\floorRadiusRepair}{44}%
\newcommand{\certClassResidMin}{0.024}%
\newcommand{\mapRuleSeed}{4242}%
\newcommand{\mendOneKRoundChess}{6\%}%
\newcommand{\mendOneKRoundStripes}{6\%}%
\newcommand{\mendOneKRoundWall}{9\%}%
\newcommand{\mendOneKRoundQud}{2\%}%
\newcommand{\mendOneKRoundColthree}{0.8\%}%
\newcommand{\mendOneKRoundCity}{3\%}%
\newcommand{\mendSeamExamples}{16}%
\newcommand{\mendSeamExamplesOf}{16}%
\newcommand{\mendReplicateRules}{3}%
\newcommand{\pilotRowResRatioLoR}{89}%
\newcommand{\pilotRowResRatioHiR}{1\,300}%
\newcommand{\floorCheckMinSlackR}{2.6}%
\newcommand{\witSkylineRhoFourAll}{49\%}%
\newcommand{\witSkylineRhoSixteenAll}{85\%}%
\newcommand{\witFlowersRhoFourAll}{79\%}%
\newcommand{\wideMendNew}{5}%
\newcommand{\wideMendNewWords}{five}%
\newcommand{\wideMendNames}{SimpleKnot, Lines, City, Water and TrickKnot}%
\newcommand{\nMendSixteen}{26}%
\newcommand{\nPartialSixteen}{5}%
\newcommand{\widePartialNames}{Qud, Mazelike, Link, Mountains and Link2}%
\newcommand{\wideRatio}{0.36}%
\newcommand{\wideCertRatio}{1.0}%
\newcommand{\wideNormLo}{0.02}%
\newcommand{\wideNormHi}{0.07}%
\newcommand{\wideChangedLo}{14\%}%
\newcommand{\wideChangedHi}{43\%}%
\newcommand{\wideRowSkyline}{0.0039}%
\newcommand{\wideRowWall}{0.021}%
\newcommand{\wideCertMendedWords}{none}%
\newcommand{\wideUncertMendedWords}{none}%
\newcommand{\robustSeeds}{eight}%
\newcommand{\wideRowSimpleMax}{$5.1\cdot10^{-4}$}%
\newcommand{\hybValidCanvases}{156}%
\newcommand{\hybValidOf}{156}%
\newcommand{\hybNrepH}{0.10}%
\newcommand{\hybNrepA}{0.09}%
\newcommand{\hybNrepG}{0.04}%
\newcommand{\hybMidpoint}{0.07}%
\newcommand{\hybAlignedH}{0.400}%
\newcommand{\hybAlignedA}{0.405}%
\newcommand{\hybCertNrepH}{0.62}%
\newcommand{\hybCertNrepA}{0.63}%
\newcommand{\hybCertBar}{0.47}%
\newcommand{\hybCostOverA}{1.2}%
\newcommand{\hybCostOverG}{16}%
\newcommand{\segRatioLo}{3.4}%
\newcommand{\segRatioHi}{282}%
\newcommand{\segRatioFailing}{3Bricks and Town}%
\newcommand{\segSlopeLo}{$-$1.36}%
\newcommand{\segSlopeHi}{$-$1.20}%
\newcommand{\segSlopeEdgeLo}{$-$1.01}%
\newcommand{\segSlopeEdgeHi}{$-$0.94}%
\newcommand{\segRateMed}{0.031}%
\newcommand{\segRuleRepairMed}{0.069}%
\newcommand{\segBelowRuleCount}{13}%
\newcommand{\segRepSixtyFour}{0.66}%
\newcommand{\iidConnFlatSet}{10}%
\newcommand{\msSimpleWallResidual}{$5.5\cdot10^{-4}$}%
\newcommand{\rowSimpleWallResidual}{$4.7\cdot10^{-4}$}%
\newcommand{\flatSimpleWallResidual}{0.046}%
\newcommand{\noTermSimpleWallResidual}{$4.5\cdot10^{-4}$}%
\newcommand{\msWallResidual}{0.026}%
\newcommand{\rowWallResidual}{0.021}%
\newcommand{\flatWallResidual}{0.094}%
\newcommand{\noTermWallResidual}{0.012}%
\newcommand{\msSkylineResidual}{0.011}%
\newcommand{\rowSkylineResidual}{0.0053}%
\newcommand{\rowSkylineResidualR}{0.005}%
\newcommand{\flatSkylineResidual}{0.066}%
\newcommand{\noTermSkylineResidual}{0.016}%
\newcommand{\strongTermWallResidual}{0.10}%
\newcommand{\msWallsFound}{2}%
\newcommand{\msWallsFoundOf}{3}%
\newcommand{\msWallsFoundWords}{two}%
\newcommand{\msSimpleWallAligned}{0.30}%
\newcommand{\flatSimpleWallAligned}{0.37}%
\newcommand{\msWallAligned}{0.60}%
\newcommand{\flatWallAligned}{0.69}%
\newcommand{\msSimpleWallNrep}{0.15}%
\newcommand{\flatSimpleWallNrep}{0.02}%
\newcommand{\msVariationKeptCount}{3}%
\newcommand{\msVariationKeptOf}{5}%
\newcommand{\msWallKLRatioLo}{3}%
\newcommand{\msWallKLRatioHi}{6}%
\newcommand{\msSkylineKLRatio}{1.5}%
\newcommand{\msCaveKL}{0.014}%
\newcommand{\flatCaveKL}{0.026}%
\newcommand{\msRoomsKL}{0.087}%
\newcommand{\flatRoomsKL}{0.19}%
\newcommand{\msCaveCorrLength}{5.7}%
\newcommand{\flatCaveCorrLength}{6.3}%
\newcommand{\msRoomsCorrBeyond}{128}%
\newcommand{\flatRoomsCorrLength}{9.7}%
\newcommand{\msRoomsGiant}{0.95}%
\newcommand{\flatRoomsGiant}{0.20}%
\newcommand{\msPilotUncertValid}{1}%
\newcommand{\msPilotUncertValidNames}{Sewers}%
\newcommand{\msPilotUncertCutMedian}{2.7}%
\newcommand{\msPilotUncertCutLo}{1.7}%
\newcommand{\msPilotUncertCutHi}{107}%
\newcommand{\msPilotUncertKLRatio}{1.5}%
\newcommand{\noTermPilotUncertValid}{2}%
\newcommand{\noTermPilotUncertValidNames}{Sewers and Skew2}%
\newcommand{\noTermPilotUncertCutMedian}{4.6}%
\newcommand{\noTermPilotUncertKLRatio}{1.7}%
\newcommand{\strongTermPilotUncertValid}{1}%
\newcommand{\strongTermPilotUncertValidNames}{ColoredCity}%
\newcommand{\msPilotUncertOf}{8}%
\newcommand{\msTermEffectCount}{2}%
\newcommand{\msTermEffectOf}{5}%
\newcommand{\msPilotExamples}{13}%
\newcommand{\msPilotExamplesWords}{thirteen}%
\newcommand{\nMendM}{27}%
\newcommand{\nMendMWords}{twenty-seven}%
\newcommand{\nMendMOf}{52}%
\newcommand{\msNewMendedCount}{7}%
\newcommand{\msNewMendedWords}{seven}%
\newcommand{\msNewMended}{City, Flowers, Lines, Mazelike, MoreFlowers, Platformer and Sewers}%
\newcommand{\msNewFromPartial}{City, Lines and Mazelike}%
\newcommand{\msNewFromPartialWords}{three}%
\newcommand{\msNewFromResistant}{Flowers, MoreFlowers, Platformer and Sewers}%
\newcommand{\msNewFromResistantWords}{four}%
\newcommand{\msNewCertified}{Flowers, MoreFlowers and Platformer}%
\newcommand{\msNewCertifiedCount}{3}%
\newcommand{\msNewCertifiedWords}{three}%
\newcommand{\msNewUncertified}{Sewers}%
\newcommand{\msUncertStillWords}{seven}%
\newcommand{\msKept}{20}%
\newcommand{\msKeptOf}{21}%
\newcommand{\msLost}{Cats}%
\newcommand{\msLostResidual}{0.004}%
\newcommand{\msCutMedian}{4.4}%
\newcommand{\msCutMedianWords}{four}%
\newcommand{\msCutOf}{21}%
\newcommand{\msCutCertifiedMedian}{19}%
\newcommand{\msCutUncertifiedMedian}{2.7}%
\newcommand{\msCutStillMedian}{3.1}%
\newcommand{\msKLMendedMedian}{0.11}%
\newcommand{\flatKLMendedMedian}{0.16}%
\newcommand{\msNrepOnFlatSetMedian}{0.07}%
\newcommand{\flatNrepOnFlatSetMedian}{0.04}%
\newcommand{\msKLBetterCount}{14}%
\newcommand{\msKLNewMedian}{0.14}%
\newcommand{\flatKLNewMedian}{0.05}%
\newcommand{\msAlignedOnFlatSetMedian}{0.36}%
\newcommand{\flatAlignedOnFlatSetMedian}{0.36}%
\newcommand{\msLocalEditMedian}{0.59\%}%
\newcommand{\flatLocalEditMedian}{1.5\%}%
\newcommand{\msLocalEditRatio}{2.1}%
\newcommand{\msCoinEditCellsMedian}{16}%
\newcommand{\flatCoinEditCellsMedian}{161}%
\newcommand{\msSeamsExact}{52}%
\newcommand{\msLinesGiant}{0.30}%
\newcommand{\flatLinesGiant}{0.80}%
\newcommand{\noTermMendCount}{28}%
\newcommand{\noTermAdds}{Qud, SimpleWall and Skew2}%
\newcommand{\noTermLacks}{Flowers and Platformer}%
\newcommand{\noTermKept}{20}%
\newcommand{\noTermCutMedian}{5.1}%
\newcommand{\noTermKLMendedMedian}{0.13}%
\newcommand{\noTermNrepOnFlatSetMedian}{0.07}%
\newcommand{\coinPairs}{24}%
\newcommand{\coinPerPair}{2}%
\newcommand{\coinChannels}{48}%
\newcommand{\coinPerPairWords}{two}%
\newcommand{\coinScaleMax}{4}%
\newcommand{\coinPeriod}{16}%
\newcommand{\coinScaleList}{0, 2, 3, 4}%
\newcommand{\coinBlockSides}{1, 4, 8, 16}%
\newcommand{\coinSquareSides}{4, 8 and 16}%
\newcommand{\coinFinePairs}{6}%
\newcommand{\coinCoarsePairs}{18}%
\newcommand{\coinLinePairs}{9}%
\newcommand{\coinSquarePairs}{3}%
\newcommand{\coinFlat}{8}%
\newcommand{\coinFlatWords}{eight}%
\newcommand{\varFrac}{0.1}%
\newcommand{\varFracWords}{a tenth}%
\newcommand{\varFracStrong}{0.3}%
\newcommand{\varFracStrongWords}{three tenths}%
\newcommand{\msBatch}{8}%
\newcommand{\msBatchBricks}{2}%
\newcommand{\bricksPatterns}{423}%
\newcommand{\msTrainMin}{98}%
\newcommand{\msLR}{$2\cdot10^{-3}$}%
\newcommand{\msLRFloor}{5\%}%
\newcommand{\msSteps}{16\,000}%
\newcommand{\msWfcCount}{33}%
\newcommand{\msWfcOf}{50}%
\newcommand{\wfcCountFlat}{43}%
\newcommand{\msWfcCountSmall}{27}%
\newcommand{\msWfcCountSmallOf}{51}%
\newcommand{\msWfcCountExt}{37}%
\newcommand{\msWfcCountExtOf}{49}%
\newcommand{\msRawKLRatioSmall}{1.6}%
\newcommand{\msRawKLRatioMid}{1.4}%
\newcommand{\msRawKLRatioExt}{1.3}%
\newcommand{\msConnFlatSet}{16}%
\newcommand{\flatConnFlatSet}{15}%
\newcommand{\msConnOwn}{17}%
\newcommand{\iidConnOwn}{10}%
\newcommand{\msConnFlatSetOf}{19}%
\newcommand{\msConnOwnOf}{22}%
\newcommand{\msCorrLagM}{32}%
\newcommand{\msCorrLagFlatLo}{9}%
\newcommand{\msCorrLagFlatHi}{10}%
\newcommand{\msTimeRatio}{0.99}%
\newcommand{\msTimeCaveM}{568}%
\newcommand{\msTimeCaveFlat}{603}%
\newcommand{\msTimeThreads}{4}%
\newcommand{\msVPopulation}{27}%
\newcommand{\msVRigid}{0}%
\newcommand{\msNrepMedian}{0.07}%
\newcommand{\msRepetitionMedian}{0}%
\newcommand{\shiftRepetitionMedianM}{1}%
\newcommand{\msAlignedMedian}{0.36}%
\newcommand{\blockAlignedMedianM}{0.39}%
\newcommand{\msAlignedRatio}{0.92}%
\newcommand{\msLocalFineMedian}{0.75\%}%
\newcommand{\msLocalAllMedian}{1.1\%}%
\newcommand{\flatLocalMedianV}{1.4\%}%
\newcommand{\msLocalAllValidCount}{27}%
\newcommand{\msLocalAllValidMin}{92\%}%
\newcommand{\replayCostRatioInvM}{16}%
\newcommand{\msTVMedian}{0.19}%
\newcommand{\msRobustAllEight}{26}%
\newcommand{\msRobustBelowSeven}{none}%
\newcommand{\msRobustCertifiedAbove}{4}%
\newcommand{\nMendMSixteen}{32}%
\newcommand{\nMendMSixteenWords}{thirty-two}%
\newcommand{\msSixteenNew}{Qud, SimpleKnot, SimpleWall, TrickKnot and Water}%
\newcommand{\msSixteenNewCount}{5}%
\newcommand{\msSixteenLost}{none}%
\newcommand{\nMendMTwelve}{31}%
\newcommand{\msWideRatio}{0.49}%
\newcommand{\msWideWallLo}{0.013}%
\newcommand{\msWideWallHi}{0.059}%
\newcommand{\msCutInvariantMedian}{53}%
\newcommand{\msCutPhaseLo}{1.0}%
\newcommand{\msCutPhaseHi}{3.1}%
\newcommand{\msPhaseNames}{Fabric, RedMaze, SimpleMaze and SmileCity}%
\newcommand{\msWallsFoundOfWords}{three}%
\newcommand{\msRobustNotAll}{Flowers}%
\newcommand{\msTVBelowBlockCount}{16}%
\newcommand{\latticeMedianM}{0.11}%
\newcommand{\latticeMedianFlat}{0.029}%
\newcommand{\latticeRatio}{4.2}%
\newcommand{\latticeWithinTwice}{3}%
\newcommand{\latticeOf}{27}%
\newcommand{\latticeCity}{0.72}%
\newcommand{\latticeNullM}{0.043}%
\newcommand{\latticeNullFlat}{0.022}%
\newcommand{\latticeWithinOwnNull}{9}%
\newcommand{\bsOneAligned}{0.31}%
\newcommand{\bsOneNrep}{0.37}%
\newcommand{\bsTwoAligned}{0.37}%
\newcommand{\bsTwoNrep}{0.11}%
\newcommand{\bsOneRuleMoreVaried}{21}%
\newcommand{\bsTwoRuleLowerNrep}{19}%
\newcommand{\bsTwoRuleLowerNrepUncert}{16}%
\newcommand{\bsTwoUncertOf}{24}%
\newcommand{\bsTwoSolverMoreVaried}{17}%
\newcommand{\bsTwoSolverMoreVariedUncert}{17}%
\newcommand{\bsTwoRepMax}{0.0014}%
\newcommand{\bsTwoRepMaxName}{Village}%
\newcommand{\bsOneAlignedRatio}{1.18}%
\newcommand{\bsOneRuleLowerNrep}{26}%
\newcommand{\bsTwoAlignedRatio}{0.98}%
\newcommand{\seamExamples}{30}%
\newcommand{\seamCanvases}{90}%
\newcommand{\seamDiagCount}{20}%
\newcommand{\nCCEightM}{27}%
\newcommand{\nCCEightFlat}{21}%
\newcommand{\nCCSixteenM}{35}%
\newcommand{\nCCSixteenFlat}{27}%
\newcommand{\nCCM}{40}%
\newcommand{\nCCFlat}{28}%
\newcommand{\ccLeadEight}{6}%
\newcommand{\ccLeadEightWords}{six}%
\newcommand{\ccLeadSixteen}{8}%
\newcommand{\ccLeadSixteenWords}{eight}%
\newcommand{\ccLeadTwentyFour}{12}%
\newcommand{\ccLeadTwentyFourWords}{twelve}%
\newcommand{\iidCCOf}{25}%
\newcommand{\iidCCValid}{0}%
\newcommand{\iidCCResidualMin}{0.64}%
\newcommand{\iidCCResidualMax}{1.0}%
\newcommand{\iidAllOf}{52}%
\newcommand{\iidAllValid}{0}%
\newcommand{\iidAllResidualMin}{0.38}%
\newcommand{\ccNewM}{13}%
\newcommand{\ccNewMWords}{thirteen}%
\newcommand{\ccNewMNames}{3Bricks, Cat, Cats, ColoredCity, Link, Nested, Qud, SimpleKnot, SimpleWall, Skew2, Town, TrickKnot and Water}%
\newcommand{\ccNewFlat}{7}%
\newcommand{\ccNewFlatWords}{seven}%
\newcommand{\ccNewFlatNames}{City, Lines, Link, Qud, SimpleKnot, TrickKnot and Water}%
\newcommand{\ccLostM}{none}%
\newcommand{\ccRobustM}{36}%
\newcommand{\ccRobustFlat}{28}%
\newcommand{\ccRobustLost}{3Bricks, Cat, Cats and Town}%
\newcommand{\ccCertValidM}{6}%
\newcommand{\ccCertValidMWords}{six}%
\newcommand{\ccCertValidMNames}{3Bricks, Flowers, MoreFlowers, Platformer, SimpleWall and Town}%
\newcommand{\ccCertValidMRobust}{4}%
\newcommand{\ccCertValidMRobustWords}{four}%
\newcommand{\ccCertValidFlat}{0}%
\newcommand{\ccChangedEight}{0.071}%
\newcommand{\ccChangedOld}{0.12}%
\newcommand{\ccChangedNew}{0.31}%
\newcommand{\ccChangedFlat}{0.23}%
\newcommand{\ccChangedMax}{0.52}%
\newcommand{\ccNrepMedian}{0.065}%
\newcommand{\ccRepetitionMedian}{0}%
\newcommand{\ccAlignedMedian}{0.39}%
\newcommand{\ccKLMedian}{0.13}%
\newcommand{\ccNrepFlatMedian}{0.046}%
\newcommand{\ccAlignedFlatMedian}{0.37}%
\newcommand{\ccKLFlatMedian}{0.12}%
\newcommand{\ccKLCommonM}{0.10}%
\newcommand{\ccKLCommonFlat}{0.12}%
\newcommand{\ccNrepCommonM}{0.061}%
\newcommand{\ccNrepCommonFlat}{0.046}%
\newcommand{\ccCommonOf}{28}%
\newcommand{\ccLocalFineMedian}{3.3\%}%
\newcommand{\ccLocalAllFlatMedian}{6.4\%}%
\newcommand{\ccLocalValidMin}{99.6\%}%
\newcommand{\ccConnOwn}{22}%
\newcommand{\ccConnOwnOf}{33}%
\newcommand{\seamCCCanvases}{120}%
\newcommand{\seamCCAgreeSixtyFour}{10}%
\newcommand{\seamCCRunSixtyFour}{120}%
\newcommand{\seamCCAgreeNinetySix}{67}%
\newcommand{\seamCCRunNinetySix}{120}%
\newcommand{\seamCCAgreeOneTwentyEight}{81}%
\newcommand{\seamCCRunOneTwentyEight}{120}%
\newcommand{\seamCCAgreeOneSixty}{99}%
\newcommand{\seamCCRunOneSixty}{39}%
\newcommand{\seamCCAgreeOneNinetyTwo}{117}%
\newcommand{\seamCCRunOneNinetyTwo}{39}%
\newcommand{\seamCCAgreeTwoTwentyFour}{119}%
\newcommand{\seamCCRunTwoTwentyFour}{39}%
\newcommand{\seamCCDisagreeNames}{Cat, Link and Nested}%
\newcommand{\seamCCWorstExample}{Cat}%
\newcommand{\seamCCWorstCells}{442}%
\newcommand{\seamCCMarginMost}{192}%
\newcommand{\mendBoundEight}{248}%
\newcommand{\mendBoundCC}{760}%
\newcommand{\mendStepEight}{31}%
\newcommand{\mendStepCC}{95}%
\newcommand{\certRadiusCC}{804}%
\newcommand{\smoothEightM}{24}%
\newcommand{\smoothSixteenM}{34}%
\newcommand{\smoothCCM}{34}%
\newcommand{\latticeBlockStageTwo}{0.10}%
\newcommand{\latticeSmoothStageTwo}{0.060}%
\newcommand{\trainSeedCounts}{27, 28 and 28}%
\newcommand{\ccWallResidualLo}{0.0034}%
\newcommand{\ccWallResidualHi}{0.055}%
\newcommand{\ccSkylineResidual}{0.0020}%
\newcommand{\ccExactZeroM}{27}%
\newcommand{\ccExactZeroFlat}{23}%
\newcommand{\ccUncertValidMWords}{five}%
\newcommand{\ccUncertValidMNames}{Cat, ColoredCity, Nested, Sewers and Skew2}%
\newcommand{\ccUncertValidFlat}{0}%
\newcommand{\trainSeedCountsNoTerm}{28, 28 and 24}%
\newcommand{\smoothPilotLatticeM}{0.11}%
\newcommand{\smoothPilotLatticeBlock}{0.46}%
\newcommand{\smoothWallEight}{0.0030}%
\newcommand{\smoothWallSixteen}{0.0022}%
\newcommand{\seamAgreeMarginEight}{64}%
\newcommand{\chunkAreaReachCC}{48}%
\newcommand{\chunkAreaMarginCC}{6.3}%
\newcommand{\chunkAreaMarginEight}{2.3}%
\newcommand{\ccChangedTwentyFourAll}{0.20}%
\newcommand{\ccChangedSixteenAll}{0.14}%
\newcommand{\tileCCValid}{23}%
\newcommand{\tileCCCertValid}{none}%
\newcommand{\tileRepetitionMedian}{0.46}%
\newcommand{\tileNrepMedian}{0.098}%
\newcommand{\tileChangedMedian}{0.081}%
\newcommand{\tileNotRuleNames}{Skew1}%
\newcommand{\ruleNotTileCount}{18}%
\newcommand{\onePassM}{32}%
\newcommand{\onePassFlat}{25}%
\newcommand{\onePassLead}{7}%
\newcommand{\onePassLeadWords}{seven}%
\newcommand{\onePassLostM}{3Bricks, Cat, Cats, ColoredCity, Nested, Skew2, Town and Village}%
\newcommand{\onePassLostMCount}{8}%
\newcommand{\onePassLostMCountWords}{eight}%
\newcommand{\onePassLostFlat}{Link, Qud and TrickKnot}%
\newcommand{\onePassLostFlatCount}{3}%
\newcommand{\onePassLostFlatCountWords}{three}%
\newcommand{\onePassGained}{none}%
\newcommand{\secondPassShareM}{13\%}%
\newcommand{\radCompOf}{28}%
\newcommand{\radCompMedianM}{3}%
\newcommand{\radCompLowerQuartileM}{2}%
\newcommand{\radCompUpperQuartileM}{5}%
\newcommand{\radCompMeanM}{4.3}%
\newcommand{\radCompCompletedM}{3243}%
\newcommand{\radCompFailedM}{394}%
\newcommand{\radCompMedianFlat}{3}%
\newcommand{\radCompLowerQuartileFlat}{2}%
\newcommand{\radCompUpperQuartileFlat}{6}%
\newcommand{\radCompMeanFlat}{4.8}%
\newcommand{\radCompCompletedFlat}{3615}%
\newcommand{\radCompFailedFlat}{434}%
\newcommand{\radCompSmallerOn}{8}%
\newcommand{\radCompEqualOn}{15}%
\newcommand{\radCompLargerOn}{5}%
\newcommand{\periodShareSolverTwo}{0.62}%
\newcommand{\periodShareSolverOne}{0.72}%
\newcommand{\periodShareRuleEight}{0.65}%
\newcommand{\periodShareRuleTwentyFour}{0.66}%
\newcommand{\periodShareReplay}{1.00}%
\newcommand{\periodChanceSolver}{0.54}%
\newcommand{\periodChanceRule}{0.57}%
\newcommand{\periodExcessSolverTwo}{0.023}%
\newcommand{\periodExcessSolverOne}{0.14}%
\newcommand{\periodExcessRule}{0.036}%
\newcommand{\periodExcessRuleTwentyFour}{0.033}%
\newcommand{\periodSolverAboveRule}{6}%
\newcommand{\periodSolverAboveRuleOf}{27}%
\newcommand{\periodSolverAboveRuleBar}{18}%
\newcommand{\periodSolverAboveRuleEightSeeds}{7}%
\newcommand{\periodExcessSolverAboveRule}{14}%
\newcommand{\mhCCValid}{20}%
\newcommand{\mhCCValidWords}{twenty}%
\newcommand{\mhCCOnlyNames}{Link2, Mountains and Skew1}%
\newcommand{\mhCCOnlyCount}{3}%
\newcommand{\mhCCOnlyCountWords}{three}%
\newcommand{\mhCCChangedMedian}{0.044}%
\newcommand{\ruleNotMhCount}{23}%
\newcommand{\ruleNotMhCountWords}{twenty-three}%
\newcommand{\mhCCNotRuleCount}{3}%
\newcommand{\mhCCCertValid}{none}%
\newcommand{\uncertInvalidMNames}{Lake, Link2, Mountains, ScaledMaze and Skew1}%
\newcommand{\uncertInvalidMCount}{5}%
\newcommand{\uncertInvalidMCountWords}{five}%
\newcommand{\uncertOpenNames}{Lake and ScaledMaze}%
\newcommand{\mendCPUEight}{9.5}%
\newcommand{\mendCPUTwentyFour}{28}%
\newcommand{\mendCPUTwentyFourRaw}{28}%
\newcommand{\mendCPURatio}{3.1}%
\newcommand{\mendCPURatioRaw}{3.1}%
\newcommand{\genCPUMedian}{13}%
\newcommand{\ruleCostRatioTwentyFour}{1.8}%
\newcommand{\mendShareTwentyFour}{69\%}%
\newcommand{\replayCostRatioInvMTwentyFour}{10}%
\newcommand{\replayCostRatioInvMTwentyFourRaw}{10}%
\newcommand{\raCostRatioTwentyFour}{9}%
\newcommand{\raCostRatioTwentyFourLo}{1.6}%
\newcommand{\raCostRatioTwentyFourHi}{51}%
\newcommand{\raCostRatioRuleTwentyFour}{9}%
\newcommand{\raCostRatioRuleEight}{13}%
\newcommand{\certPatchMax}{12}%
\newcommand{\certCap}{200\,000}%
\newcommand{\methodNCourseWords}{eight}%
\newcommand{\methodNPhaseWords}{four}%
\newcommand{\methodNOct}{40}%
\newcommand{\methodCourseNames}{3Bricks, Flowers, MoreFlowers, Platformer, SimpleWall, Skyline, Skyline2 and Wall}%
\newcommand{\methodPhaseNames}{Fabric, RedMaze, SimpleMaze and SmileCity}%
\newcommand{\bricksCertP}{4}%
\newcommand{\bricksTypesWords}{two}%
\newcommand{\rowInvCountWords}{eight}%
\newcommand{\rowInvCount}{8}%
\newcommand{\courseBandMult}{16}%
\newcommand{\courseRunMaxPairs}{30}%
\newcommand{\courseRunMaxRows}{31}%
\newcommand{\courseHashBits}{62}%
\newcommand{\courseSolverRunSkyline}{4}%
\newcommand{\courseSolverRunSkylineTwo}{3}%
\newcommand{\courseSolverRunPlatformer}{4}%
\newcommand{\courseAutomatonRun}{2}%
\newcommand{\courseWarmFrac}{0.3}%
\newcommand{\courseWarmFracWords}{30 percent}%
\newcommand{\courseWarmUntil}{4\,800}%
\newcommand{\phaseDFabric}{5}%
\newcommand{\phaseDRedMaze}{4}%
\newcommand{\phaseDSimpleMaze}{4}%
\newcommand{\phaseDSmileCity}{2}%
\newcommand{\ruleWidth}{64}%
\newcommand{\trainRoundsLo}{8}%
\newcommand{\trainRoundsHi}{32}%
\newcommand{\repairRounds}{4}%
\newcommand{\repairRoundsWords}{four}%
\newcommand{\mendRadiusMain}{24}%
\newcommand{\mendBlockMain}{48}%
\newcommand{\mendPassesWords}{two}%
\newcommand{\mendClassesWords}{four}%
\newcommand{\wallRowsFollow}{100\%}%
\newcommand{\courseSampleBands}{200}%
\newcommand{\courseSampleSeedsWords}{five}%
\newcommand{\platCoursesEachWords}{sixteen}%
\newcommand{\platCourseShort}{32}%
\newcommand{\platCourseLong}{33}%
\newcommand{\platBand}{1{,}040}%
\newcommand{\simpleWallRunMax}{11}%
\newcommand{\simpleWallExampleRuns}{2 and 3}%
\newcommand{\thrWallMisplaced}{half}%
\newcommand{\wallMisplacedPooled}{0.62}%
\newcommand{\wallMisplacedSeeds}{0.72, 0.10 and 0.54}%
\newcommand{\witRuleWall}{18\%}%
\newcommand{\witRuleSkyline}{100\%}%
\newcommand{\wallSecondReach}{9}%
\newcommand{\thrOnePassM}{34}%
\newcommand{\thrOnePassFlat}{27}%
\newcommand{\roundsList}{4, 8, 16 and 32}%
\newcommand{\roundsFewest}{4}%
\newcommand{\roundsCounts}{20, 32, 39 and 40}%
\newcommand{\roundsFourCount}{20}%
\newcommand{\roundsFourBar}{30}%
\newcommand{\mhSweeps}{64}%
\newcommand{\thrMetropolis}{28}%
\newcommand{\thrMetropolisFalsifier}{36}%
\newcommand{\seamMarginBase}{128}%
\newcommand{\seamMarginNext}{192}%
\newcommand{\seamMarginsWide}{256 and 320}%
\newcommand{\seamDiffered}{39}%
\newcommand{\seamDifferedAgreeNext}{38}%
\newcommand{\seamAgreedBase}{81}%
\newcommand{\seamAllMargin}{256}%
\newcommand{\thrSeamWide}{320}%
\newcommand{\thrGridPhase}{3}%
\newcommand{\glPhaseValid}{2}%
\newcommand{\glPhaseValidNames}{RedMaze and SimpleMaze}%
\newcommand{\glSkylineSpan}{0.0021 to 0.0037}%
\newcommand{\glFabricRate}{0.19}%
\newcommand{\glSmileRate}{0.018}%
\newcommand{\glChangedCave}{0.034}%
\newcommand{\glChangedRooms}{0.14}%
\newcommand{\glChangedCity}{0.43}%
\newcommand{\glChangedWall}{0.21}%
\newcommand{\glChangedSkylineTwo}{0.26}%
\newcommand{\glRawWall}{0.095}%
\newcommand{\glRawSkylineTwo}{0.027}%
\newcommand{\glFixedNames}{Cave, City, Fabric, RedMaze, SimpleMaze, SmileCity and Wall}%
\newcommand{\glFixedCountWords}{seven}%
\newcommand{\glTriedCountWords}{ten}%
\newcommand{\thrTenfold}{tenfold}%
\newcommand{\thrWallValidTwo}{two}%
\newcommand{\csWallRawSpan}{0.18 to 0.20}%
\newcommand{\csWallSpan}{0.17 to 0.20}%
\newcommand{\csWallChanged}{0.0019}%
\newcommand{\csSkylineSpan}{0.0067 to 0.011}%
\newcommand{\csSkylineTwoSpan}{0.0008 to 0.0029}%
\newcommand{\csWallRowMatch}{0\%}%
\newcommand{\csWellFormedRate}{0.0052}%
\newcommand{\thrRowsFollow}{95\%}%
\newcommand{\cwWallRawSpan}{0.0079 to 0.0096}%
\newcommand{\cwWallChanged}{0.045}%
\newcommand{\cwSkylineSeeds}{0, 0.0017 and 0.0021}%
\newcommand{\cwSkylineTwoSeeds}{0, 0.0025 and 0.0014}%
\newcommand{\cwWallRawSpanOne}{0.0095 to 0.013}%
\newcommand{\cwWallChangedOne}{0.056}%
\newcommand{\cwSkylineSeedsOne}{0, 0.0013 and 0.0013}%
\newcommand{\cwSkylineTwoSeedsOne}{0.0006, 0.0042 and 0.0022}%
\newcommand{\cwSkyRowsMin}{99.8\%}%
\newcommand{\mpSmileSpan}{0.0071 to 0.010}%
\newcommand{\mpSmileSpanOne}{0.022 to 0.042}%
\newcommand{\smileWrongWidthLo}{0.99}%
\newcommand{\smileWrongWidthHi}{1.00}%
\newcommand{\openLongLo}{94\%}%
\newcommand{\openLongHi}{99\%}%
\newcommand{\smileCertPmax}{10}%
\newcommand{\smileCertPbeyond}{12}%
\newcommand{\thrSmilePeriod}{8}%
\newcommand{\thrSmilePatch}{9}%
\newcommand{\octCertValidWords}{six}%
\newcommand{\octCertValid}{6}%
\newcommand{\octCertValidNames}{3Bricks, Flowers, MoreFlowers, Platformer, SimpleWall and Town}%
\newcommand{\certCountStrict}{11}%
\newcommand{\certStrictNames}{Fabric, Flowers, MoreFlowers, Platformer, RedMaze, SimpleMaze, SimpleWall, Skyline, Skyline2, Town and Wall}%
\newcommand{\certOwnNames}{3Bricks, Fabric, Flowers, MoreFlowers, Platformer, RedMaze, SimpleMaze, SimpleWall, Skyline, Skyline2, SmileCity, Town and Wall}%
\newcommand{\certPendingWords}{none}%
\newcommand{\certOctPendingNames}{none}%
\newcommand{\thrFresh}{2}%
\newcommand{\rigidFromP}{4}%
\newcommand{\rigidReach}{12}%
\newcommand{\mazeWindows}{16}%
\newcommand{\mazeBranching}{2}%
\newcommand{\fabricPatchesLo}{19}%
\newcommand{\fabricPatchesHi}{769}%
\newcommand{\thrSeedsWithin}{3}%
\newcommand{\thrSeedsUncert}{two}%
\newcommand{\mendRadiusLo}{8}%
\newcommand{\mendRadiusMid}{16}%
\newcommand{\descStrictLo}{2}%
\newcommand{\descStrictMid}{2}%
\newcommand{\rawValidFlat}{0}%
\newcommand{\repairValidFlat}{1}%
\newcommand{\rawValidOct}{1}%
\newcommand{\repairValidOct}{2}%
\newcommand{\rawValidMethod}{2}%
\newcommand{\repairValidMethod}{3}%
\newcommand{\thrSeamMethod}{256}%
\newcommand{\fOneWall}{0.75 to 0.76}%
\newcommand{\fOneSkyline}{0.95}%
\newcommand{\fOneSkylineTwo}{0.99}%
\newcommand{\replayRawSpan}{0.027 to 0.030}%
\newcommand{\replayChangedSpan}{0.14 to 0.18}%
\newcommand{\thrCopierRep}{0.5}%
\newcommand{\copierNames}{Wall, SimpleWall, Platformer, Skyline2, Skyline, Flowers and MoreFlowers}%
\newcommand{\copierCountWords}{seven}%
\newcommand{\copierCount}{7}%
\newcommand{\copierNoneNamesFinal}{3Bricks}%
\newcommand{\copierRawZeroCount}{7}%
\newcommand{\copierRowCopyMin}{1.00}%
\newcommand{\copierRepLongLo}{0.41}%
\newcommand{\copierRepLongHi}{0.70}%
\newcommand{\copierRepWallMax}{0.009}%
\newcommand{\copierRepWallBelow}{one in a hundred}%
\newcommand{\copierCoverageHMin}{1.00}%
\newcommand{\copierTVWallLo}{0.036}%
\newcommand{\copierTVWallHi}{0.044}%
\newcommand{\copierTVSimpleWallHi}{0.011}%
\newcommand{\copierRepLo}{0.64}%
\newcommand{\copierRepHi}{0.70}%
\newcommand{\copierChangedWallLo}{0\%}%
\newcommand{\copierChangedWallHi}{0\%}%
\newcommand{\ruleChangedWallLo}{3\%}%
\newcommand{\ruleChangedWallHi}{6\%}%
\newcommand{\ruleTVWallLo}{0.35}%
\newcommand{\ruleTVWallHi}{0.38}%
\newcommand{\ruleRepHWallLo}{0.011}%
\newcommand{\ruleRepHWallHi}{0.021}%
\newcommand{\ruleRowCopyWall}{0.22}%
\newcommand{\octRowCopyLo}{0.043}%
\newcommand{\octRowCopyHi}{0.30}%
\newcommand{\octRowCopyRawLo}{0.008}%
\newcommand{\octRowCopyRawHi}{0.31}%
\newcommand{\jointOffsetsWallLo}{3}%
\newcommand{\jointOffsetsWallHi}{4}%
\newcommand{\jointOffsetsWallAllowed}{4}%
\newcommand{\bricksRunCopy}{7}%
\newcommand{\bricksRunNewLo}{8}%
\newcommand{\bricksRunNewHi}{31}%
\newcommand{\phaseLookupNames}{Fabric, RedMaze and SimpleMaze}%
\newcommand{\phaseLookupCountWords}{three}%
\newcommand{\mhTypedSweeps}{256}%
\newcommand{\mhTypedValidNames}{Wall, SimpleWall, Flowers and MoreFlowers}%
\newcommand{\mhTypedValidCountWords}{four}%
\newcommand{\mhTypedFailMin}{0.067}%
\newcommand{\mhTypedTVWall}{0.50}%
\newcommand{\mhTypedReach}{4\,608}%
\newcommand{\mhLongSweeps}{1\,024}%
\newcommand{\mhLongFactorWords}{sixteen}%
\newcommand{\mhLongOfWords}{nineteen}%
\newcommand{\mhLongValidCount}{0}%
\newcommand{\mhLongOctValidCount}{17}%
\newcommand{\mhUncertTotal}{20}%
\newcommand{\crValidWords}{seven}%
\newcommand{\crRedrawnLo}{16\%}%
\newcommand{\crRedrawnHi}{37\%}%
\newcommand{\crRepMax}{$2\cdot10^{-5}$}%
\newcommand{\crMultiRowWords}{none}%
\newcommand{\crTvBelowRuleCountWords}{six}%
\newcommand{\crTvAboveRuleNames}{Platformer}%
\newcommand{\crTvWallLo}{0.13}%
\newcommand{\crTvWallHi}{0.14}%
\newcommand{\crTvBlockDiffMax}{0.02}%
\newcommand{\crTvBlockExamplesWords}{six}%
\newcommand{\crTvBlockWall}{0.12}%
\newcommand{\crAlignedWall}{0.48}%
\newcommand{\crNrepWall}{0.09}%
\newcommand{\crNrepBelowCopierWords}{all seven}%
\newcommand{\crCpuLo}{46}%
\newcommand{\crCpuHi}{457}%
\newcommand{\ruleCourseValidWords}{seven}%
\newcommand{\ruleCourseInvalidNames}{3Bricks}%
\newcommand{\ruleBelowMHCountWords}{all eight}%
\newcommand{\ruleRepOtherHi}{$2\cdot10^{-5}$}%
\newcommand{\ruleRepHTallHi}{0}%
\newcommand{\ruleRepHTallWords}{six}%
\newcommand{\fabricPhaseMass}{0.60}%
\newcommand{\fabricOctK}{6}%
\newcommand{\fabricOctDenRaw}{194}%
\newcommand{\fabricOctFloorRaw}{$3.0\cdot10^{-3}$}%
\newcommand{\fabricOctFloorRepair}{$1.6\cdot10^{-3}$}%
\newcommand{\fabricOctMendPeriod}{96}%
\newcommand{\fabricOctFloorMended}{$1.5\cdot10^{-4}$}%
\newcommand{\fabricOctFloorRawWorst}{$1.8\cdot10^{-3}$}%
\newcommand{\nPublic}{52}%
\newcommand{\nMethodTwentyFour}{45}%
\newcommand{\nCertMethodWords}{eleven}%
\newcommand{\nOctUncertValid}{34}%
\newcommand{\nUncertOf}{39}%
\newcommand{\wallRawMethodWords}{one in a hundred}%
\newcommand{\wallChangedMethod}{4\%}%
\newcommand{\rewrittenMedianMethod}{18\%}%
\newcommand{\rewrittenMedianMethodFrac}{0.18}%
\newcommand{\wallChangedMethodFrac}{0.045}%
\newcommand{\rawValidMethodAll}{4}%
\newcommand{\rawValidMethodAllWords}{four}%
\newcommand{\rawValidMethodAllNames}{City, Fabric, RedMaze and SimpleMaze}%
\newcommand{\repairValidMethodAll}{6}%
\newcommand{\repairValidMethodAllWords}{six}%
\newcommand{\repairValidMethodAllNames}{City, Fabric, Font, Platformer, RedMaze and SimpleMaze}%
\newcommand{\certMethodNames}{Fabric, Flowers, MoreFlowers, Platformer, RedMaze, SimpleMaze, SimpleWall, Skyline, Skyline2, Town and Wall}%
\newcommand{\nMethodHarvested}{52}%
\newcommand{\nMethodRobust}{41}%
\newcommand{\methodValidNotRobustNames}{Cat, Cats, Skyline2 and Town}%
\newcommand{\nMethodPendingWords}{none}%
\newcommand{\methodPendingNames}{none}%
\newcommand{\methodExactZero}{34}%
\newcommand{\nRigidWords}{two}%
\newcommand{\rigidNames}{RedMaze and SimpleMaze}%
\newcommand{\roundsValidFour}{20}%
\newcommand{\roundsValidEight}{32}%
\newcommand{\roundsValidSixteen}{39}%
\newcommand{\roundsValidThirtyTwo}{40}%
\newcommand{\roundsRawFour}{0.092}%
\newcommand{\roundsRawThirtyTwo}{0.020}%
\newcommand{\seamAllCanvases}{120}%
\newcommand{\seamOneDiffersMargin}{224}%
\newcommand{\seamAllCanvasWide}{768}%
\newcommand{\seamAllChunkWide}{64}%
\newcommand{\seamAllGap}{96}%
\newcommand{\seamGapFiveTwelve}{32}%
\newcommand{\seamGapTwoTwentyFour}{16}%
\newcommand{\seamChunkFiveTwelve}{64}%
\newcommand{\seamChunkTwoTwentyFour}{32}%
\newcommand{\chunkAreaMarginAll}{9}%
\newcommand{\raOverRuleChunkAtMargin}{1.0}%
\newcommand{\raOverRuleChunkAtMarginLo}{0.2}%
\newcommand{\raOverRuleChunkAtMarginHi}{4.9}%
\newcommand{\samplerFloorResLo}{0.75}%
\newcommand{\samplerFloorResHi}{0.99}%
\newcommand{\smileWidthMax}{30}%
\newcommand{\smileHeightMax}{24}%
\newcommand{\smilePeriodsNoMouth}{3, 5, 6, 8, 9, \ldots}%
\newcommand{\smilePeriodsMouth}{8, 13, 16, 18, 19, 21, 22, \ldots}%
\newcommand{\smileShiftNarrow}{five}%
\newcommand{\uncertOpenSeedOneNames}{ScaledMaze}%
\newcommand{\trainSeedOneCount}{38}%
\newcommand{\trainSeedOneCountWords}{thirty-eight}%
\newcommand{\trainSeedTwoCount}{38}%
\newcommand{\trainSeedTwoCountWords}{thirty-eight}%
\newcommand{\trainSeedAllCount}{36}%
\newcommand{\trainSeedAnyCount}{41}%
\newcommand{\trainSeedCertAllNames}{Flowers, Platformer, SimpleWall and Town}%
\newcommand{\trainSeedCertAllWords}{four}%
\newcommand{\trainSeedCertTwoNames}{3Bricks and MoreFlowers}%
\newcommand{\trainSeedCertTwoWords}{two}%
\newcommand{\trainSeedUncertInvalidWords}{three}%
\newcommand{\seamMethodOf}{6}%
\newcommand{\seamMethodOfWords}{six}%
\newcommand{\seamMethodAgree}{6}%
\newcommand{\seamMethodMargin}{256}%
\newcommand{\seamMethodDiffering}{0}%
\newcommand{\seamMethodWallChangedLo}{102}%
\newcommand{\seamMethodWallChangedHi}{220}%
\newcommand{\courseOtherOfWords}{seven}%
\newcommand{\courseOtherValidWords}{six}%
\newcommand{\courseOtherValidNames}{Flowers, MoreFlowers, Platformer, SimpleWall, Skyline and Skyline2}%
\newcommand{\courseOtherInvalidNames}{3Bricks}%
\newcommand{\mazeMethodValidWords}{both}%
\newcommand{\courseOtherAlignedLo}{0.22}%
\newcommand{\courseOtherAlignedHi}{0.48}%
\newcommand{\courseOtherNrepLo}{0.11}%
\newcommand{\courseOtherNrepHi}{0.37}%
\newcommand{\courseOtherNrepHiName}{Platformer}%
\newcommand{\certMethodInvalidNames}{3Bricks and SmileCity}%
\newcommand{\certMethodInvalidWords}{two}%
\newcommand{\threeBricksWorst}{0.0011}%
\newcommand{\threeBricksOverWords}{one}%
\newcommand{\smileMethodResLo}{0.0071}%
\newcommand{\smileMethodResHi}{0.010}%
\newcommand{\ofExamplesWords}{twelve}%
\newcommand{\ofAboveCountWords}{eleven}%
\newcommand{\ofBelowNames}{3Bricks}%
\newcommand{\ofSpreadCountWords}{six}%
\newcommand{\ofSpreadNames}{Fabric, Flowers, RedMaze, SimpleMaze, SimpleWall and Wall}%
\newcommand{\ofConcMoreNames}{MoreFlowers and Skyline}%
\newcommand{\ofExampleBelowNames}{3Bricks, Platformer, Skyline2 and SmileCity}%
\newcommand{\ofConcLessNames}{RedMaze}%
\newcommand{\ofRatioMedianRaw}{24}%
\newcommand{\ofRatioMedianRepair}{29}%
\newcommand{\initScale}{0.5}%
\newcommand{\reinjectScale}{0.1}%
\newcommand{\trainCanvas}{48}%
\newcommand{\gradClip}{1}%
\newcommand{\tempStart}{1}%
\newcommand{\tempEnd}{0.2}%
\newcommand{\tempFallFracWords}{70 percent}%
\newcommand{\lamF}{1}%
\newcommand{\lamH}{0.1}%
\newcommand{\lamV}{4}%
\newcommand{\lamM}{2}%
\newcommand{\hardWarmUntil}{4\,800}%
\newcommand{\hardWarmFracWords}{30 percent}%
\newcommand{\thrWindowsMain}{64}%
\newcommand{\klFloor}{$10^{-12}$}%
\newcommand{\editCanvas}{192}%
\newcommand{\mendedFloorDen}{1660}%
\newcommand{\mendedFloorMax}{$4.6\cdot10^{-4}$}%
\newcommand{\mendedFloorMaxName}{RedMaze}%
\newcommand{\octPendingValidWords}{none}%
\newcommand{\octPendingValidNames}{none}%
\newcommand{\nCertHarvestedWords}{thirteen}%
\newcommand{\certDeltaAboveOfWords}{ten}%
\newcommand{\chunkAreaSmallAll}{81}%
\newcommand{\runCapRows}{31}%
\newcommand{\wallMethodAligned}{0.41}%
\newcommand{\wallMethodNrep}{0.09}%
\newcommand{\wallRABSAligned}{0.23}%
\newcommand{\wallRABSNrep}{0.67}%
\newcommand{\wallMethodRepWords}{none}%
\newcommand{\csWallRowMatchWords}{none}%
\newcommand{\wallRowsFollowWords}{every}%
\newcommand{\seamRetrainsExamples}{3Bricks, Flowers, MoreFlowers, Platformer, RedMaze, SimpleMaze, SimpleWall, Skyline and Skyline2}%
\newcommand{\seamRetrainsOf}{27}%
\newcommand{\seamRetrainsAgree}{27}%
\newcommand{\seamRetrainsDiffering}{0}%
\newcommand{\seamMethodAllKernelsWords}{eleven}%
\newcommand{\seamMethodAllOf}{33}%
\newcommand{\seamMethodAllAgree}{33}%
\newcommand{\seamMethodAllVerdict}{met}%
\newcommand{\seamMethodAllChangedHiPct}{44\%}%
\newcommand{\seamMethodAllChangedHiName}{Skyline}%
\newcommand{\seamMethodAllMendedWords}{eight}%
\newcommand{\heldoutWallTValid}{yes}%
\newcommand{\heldoutWallTValidWords}{valid}%
\newcommand{\heldoutWallTRaw}{0.023}%
\newcommand{\heldoutWallTChanged}{0.15}%
\newcommand{\heldoutWallTFollow}{100\%}%
\newcommand{\heldoutSimpleWallTValid}{yes}%
\newcommand{\heldoutSimpleWallTValidWords}{valid}%
\newcommand{\heldoutSimpleWallTRaw}{0.018}%
\newcommand{\heldoutSimpleWallTChanged}{0.079}%
\newcommand{\heldoutSimpleWallTFollow}{100\%}%
\newcommand{\heldoutWallHTwoValid}{yes}%
\newcommand{\heldoutWallHTwoValidWords}{valid}%
\newcommand{\heldoutWallHTwoRaw}{0}%
\newcommand{\heldoutWallHTwoChanged}{0}%
\newcommand{\heldoutWallHTwoFollow}{100\%}%
\newcommand{\heldoutTransposedValidWords}{both}%
\newcommand{\heldoutValidCountWords}{three}%
\newcommand{\smileMPEightValidWords}{not valid}%
\newcommand{\smileMPEightResLo}{0.14}%
\newcommand{\smileMPEightResHi}{0.16}%
\newcommand{\baSkylinesValidWords}{both}%
\newcommand{\baSkylinesFollow}{100\%}%
\newcommand{\baRetrainValidWords}{all four}%
\newcommand{\baRetrainFollow}{100\%}%
\newcommand{\baSkylinesSeedOneInvalidNames}{Skyline2}%
\newcommand{\baSkylinesSeedOneWorst}{0.0029}%
\newcommand{\mazeTiledWords}{both}%
\newcommand{\smileMPEightSeedOneResLo}{0.22}%
\newcommand{\smileMPEightSeedOneResHi}{0.23}%
\newcommand{\smileMPEightDistinctRaw}{8}%
\newcommand{\flatSamplerValidNames}{Skyline and Wall}%
\newcommand{\flatSamplerValidCountWords}{two}%
\newcommand{\phaseOneCoordValidNames}{Fabric}%
\newcommand{\phaseOneCoordValidCountWords}{one}%
\newcommand{\wallNoColorTermValidWords}{valid}%
\newcommand{\wallNoColorTermRaw}{0.024}%
\newcommand{\wallNoColorTermChanged}{0.15}%
\newcommand{\wallNoColorTermFollow}{0.0\%}%
\newcommand{\skylineNoLowerLimitValidWords}{valid}%
\newcommand{\skylineLowerLimitRows}{25}%
\newcommand{\skylineNoLowerLimitShortShare}{0.65}%
\newcommand{\methodSeedsCertOfWords}{twelve}%
\newcommand{\methodSeedZeroCertCount}{11}%
\newcommand{\methodSeedZeroCertCountWords}{eleven}%
\newcommand{\methodSeedOneCertCount}{10}%
\newcommand{\methodSeedOneCertCountWords}{ten}%
\newcommand{\methodSeedTwoCertCount}{11}%
\newcommand{\methodSeedTwoCertCountWords}{eleven}%
\newcommand{\methodAllSeedsCertCount}{10}%
\newcommand{\methodAllSeedsCertCountWords}{ten}%
\newcommand{\methodAllSeedsCertNames}{Fabric, Flowers, MoreFlowers, Platformer, RedMaze, SimpleMaze, SimpleWall, Skyline, Town and Wall}%
\newcommand{\wallKappaFFrac}{2/3}%
\newcommand{\wallTightRatio}{12}%
\newcommand{\wallTightRmin}{5}%
\newcommand{\wallZeroOfWords}{eleven}%
\newcommand{\wallTVCanvasesWords}{six}%
\newcommand{\wallTVElevenLo}{0.35}%
\newcommand{\wallTVElevenHi}{0.39}%
\newcommand{\wallTVElevenHiNear}{0.38}%
\newcommand{\wallTVElevenWords}{eleven}%
\newcommand{\wallFloorWindowsMin}{11}%
\newcommand{\simpleWallOctTVHi}{0.18}%
\newcommand{\simpleWallKappaF}{0.53}%
\newcommand{\pbMethodNames}{Fabric, RedMaze, SimpleMaze, SimpleWall and Wall}%
\newcommand{\pbFloorWindowsMin}{11}%
\newcommand{\pbOctNames}{SimpleWall}%
\newcommand{\freshFlatCount}{28}%
\newcommand{\freshMethodKept}{44}%
\newcommand{\freshMethodLostNames}{Cat}%
\newcommand{\certCountDecided}{3}%
\newcommand{\certStrictNamesDecided}{Fabric, Town and Wall}%
\newcommand{\certOwnNamesDecided}{Fabric, SmileCity, Town and Wall}%
\newcommand{\certOctPendingNamesDecided}{3Bricks, Flowers, MoreFlowers, Platformer and SimpleWall}%
\newcommand{\nCertMethodAllWords}{eleven}%
\newcommand{\nCertMethodAll}{11}%
\newcommand{\nMethodAll}{45}%
\newcommand{\certGainedOverOctWords}{six}%
\newcommand{\certGainedOverOctNames}{Fabric, RedMaze, SimpleMaze, Skyline, Skyline2 and Wall}%
\newcommand{\certLostToOctNames}{3Bricks}%
\newcommand{\methodRobustCertCountWords}{nine}%
\newcommand{\methodNotRobustCertNames}{Skyline2 and Town}%
\newcommand{\certInputCount}{12}%
\newcommand{\certInputCountWords}{twelve}%
\newcommand{\certInputMethodValid}{10}%
\newcommand{\certInputMethodValidWords}{ten}%
\newcommand{\certInputOctValid}{5}%
\newcommand{\certInputOctValidWords}{five}%
\newcommand{\inputAloneValidWords}{ten}%
\newcommand{\inputAloneValid}{10}%
\newcommand{\inputAloneInvalidWords}{two}%
\newcommand{\shapedNames}{3Bricks, Fabric, Skyline, Skyline2 and Wall}%
\newcommand{\shapedCountWords}{five}%
\newcommand{\notShapedCountWords}{eight}%
\newcommand{\notShapedValidCountWords}{seven}%
\newcommand{\notShapedInvalidNames}{SmileCity}%
\newcommand{\notShapedOctValidCountWords}{five}%
\newcommand{\notShapedOctValidNames}{Flowers, MoreFlowers, Platformer, SimpleWall and Town}%
\newcommand{\notShapedGainedNames}{RedMaze and SimpleMaze}%
\newcommand{\townThmDelta}{0.33}%
\newcommand{\townThmKappa}{0.023}%
\newcommand{\thmCanvasesWords}{eight}%
\newcommand{\thmLPOtherNames}{SimpleWall, Platformer, RedMaze, SimpleMaze and Fabric}%
\newcommand{\thmLPOtherCountWords}{five}%
\newcommand{\thmSpreadNames}{Flowers, MoreFlowers and Skyline}%
\newcommand{\thmSpreadWindowsMinWords}{four}%
\newcommand{\thmSeparatedCountWords}{nine}%
\newcommand{\thmAnyDepthCountWords}{six}%
\newcommand{\thmSpreadCountWords}{three}%
\newcommand{\thmWallWindows}{26}%
\newcommand{\thmVfrNames}{Wall, SimpleWall, RedMaze, SimpleMaze and Fabric}%
\newcommand{\thmVfrWindowsMin}{11}%
\newcommand{\thmOneCanvasCountWords}{three}%
\newcommand{\thmRawRepWords}{none}%
\newcommand{\thmRawMultiRowWords}{none}%
\newcommand{\clCleanMin}{5\%}%
\newcommand{\clCleanMinName}{3Bricks}%
\newcommand{\clCleanSkylineLo}{21\%}%
\newcommand{\clCleanSkylineHi}{39\%}%
\newcommand{\clCleanRowsMin}{4}%
\newcommand{\clCleanPatchesTotal}{1\,495\,989}%
\newcommand{\clCleanRowsTotal}{4\,178}%
\newcommand{\wallExBrickLength}{8}%
\newcommand{\wallExJointPct}{11}%
\newcommand{\wallMethodBrickMedianLo}{14}%
\newcommand{\wallMethodBrickMedianHi}{16}%
\newcommand{\wallMethodBrickPNinetyLo}{41}%
\newcommand{\wallMethodBrickPNinetyHi}{47}%
\newcommand{\wallMethodJointPct}{4}%
\newcommand{\wallMethodRawJointPct}{5}%
\newcommand{\wallMendJointLossPctLo}{18}%
\newcommand{\wallMendJointLossPctHi}{21}%
\newcommand{\wallMethodJointBoundPct}{70}%
\newcommand{\wallMethodJointWindowPctLo}{14}%
\newcommand{\wallMethodJointWindowPctHi}{16}%
\newcommand{\wallExJointWindowPct}{41}%
\newcommand{\wallBlockBrickMedian}{8.00}%
\newcommand{\wallBlockJointPct}{9}%
\newcommand{\wallMHBrickMedianLo}{26}%
\newcommand{\wallMHBrickMedianHi}{33}%
\newcommand{\wallMHJointPct}{2}%
\newcommand{\wallMethodCourseRowsWords}{200}%
\newcommand{\wallMethodRepHHi}{0.012}%
\newcommand{\wallBlockRepHLo}{0.034}%
\newcommand{\wallBlockRepHHi}{0.045}%
\newcommand{\wallCopierRepHLo}{0.87}%
\newcommand{\ccFlatResistValidWords}{none}%
\newcommand{\ccFlatPartialValidWords}{seven}%
\newcommand{\ccFlatPartialOfWords}{ten}%
\newcommand{\ccFlatPartialNotNames}{Link2, Mazelike and Mountains}%
\newcommand{\ccFlatResistClosest}{0.004}%
\newcommand{\ccFlatResistClosestName}{ScaledMaze}%
\newcommand{\crBrickPNinetyRange}{18--19}%
\newcommand{\wallMethodBrickMedianRange}{14--16}%
\newcommand{\wallMethodBrickPNinetyRange}{41--47}%
\newcommand{\crBrickMedianRange}{8}%
\newcommand{\refLayoutExamplesWords}{seven}%
\newcommand{\refLayoutCanvases}{107}%
\newcommand{\wallLayoutCanvasesWords}{six}%
\newcommand{\wallLayoutAllCanvasesWords}{eleven}%
\newcommand{\wallLayoutFixedRows}{41}%
\newcommand{\wallLayoutRowsPerStrip}{96}%
\newcommand{\wfcTVWallLo}{0.26}%
\newcommand{\wfcTVWallHi}{0.26}%
\newcommand{\raTVWallLo}{0.11}%
\newcommand{\raTVWallHi}{0.13}%
\newcommand{\blockTVWallLo}{0.13}%
\newcommand{\blockTVWallHi}{0.13}%
\newcommand{\wallMethodTypeMarginal}{0.014}%
\newcommand{\blockChunkSeconds}{293}%
\newcommand{\raChunkSeconds}{245}%
\newcommand{\wallMethodChunkSeconds}{5}%
\newcommand{\wallMethodMendSeconds}{1}%
\newcommand{\thrUniformCopierRep}{0.3}%
\newcommand{\thrJointOffsets}{two}%
\newcommand{\thrJointRep}{0.05}%
\newcommand{\thrCourseCoverage}{0.9}%
\newcommand{\thrRuleRepH}{0.2}%
\newcommand{\thrTypedValid}{four}%
\newcommand{\thrLongValid}{8}%
\newcommand{\thrLongFalsifier}{14}%
\newcommand{\thrMethodSeedsWithin}{2}%
\newcommand{\thrMethodSeedsFalsifier}{three}%
\newcommand{\thrOwnFloorCount}{10}%
\newcommand{\thrPastFalsifier}{two}%
\newcommand{\thrVZeroRep}{0.05}%
\newcommand{\thrVZeroReplay}{0.2}%
\newcommand{\thrVZeroCanvases}{two of three}%
\newcommand{\thrHeldMazePeriod}{3}%
\newcommand{\thrHeldMazePatch}{3 or 4}%
\newcommand{\thrDoorMazePatch}{3}%
\newcommand{\thrHeldMazeAligned}{0.1}%
\newcommand{\heldMazePassageWords}{two}%
\newcommand{\doorMazeOpeningWords}{one}%
\newcommand{\uniformCopierRepLo}{0.41}%
\newcommand{\uniformCopierRepHi}{0.58}%
\newcommand{\jointFewCanvasesWords}{one}%
\newcommand{\jointFewExampleNames}{SimpleWall}%
\newcommand{\jointFewKindsWords}{two}%
\newcommand{\courseCoverageMin}{1.00}%
\newcommand{\wallCopierRepHHi}{0.90}%
\newcommand{\mhTypedSixtyFourCountWords}{three}%
\newcommand{\mhTypedSixtyFourMoreFlowersWords}{one}%
\newcommand{\mhTypedSixtyFourFirstReadWords}{two}%
\newcommand{\mhRuleOnlyNames}{Platformer, Skyline2 and Skyline}%
\newcommand{\mhOctOnlyNames}{Platformer and 3Bricks}%
\newcommand{\mhLongValidWords}{none}%
\newcommand{\methodSeedTwoOthersValid}{10}%
\newcommand{\methodSeedTwoOthersWords}{eleven}%
\newcommand{\methodSeedTwoOthersValidWords}{ten}%
\newcommand{\methodSeedTwoWallWords}{valid}%
\newcommand{\methodSeedOneLostNames}{Skyline2}%
\newcommand{\methodSeedsTripWords}{none}%
\newcommand{\methodSeedsInvalidAllNames}{3Bricks}%
\newcommand{\ofEpsDenMin}{2.6}%
\newcommand{\ofTieSeedsWords}{one}%
\newcommand{\ofTieCellsWords}{one}%
\newcommand{\ofTieCellsRepair}{10}%
\newcommand{\ofConcLessRule}{0.2496}%
\newcommand{\ofConcLessExample}{0.25}%
\newcommand{\pbSimpleWallTV}{0.21}%
\newcommand{\pbMazeTV}{0.004}%
\newcommand{\pbMazeKappaF}{0.75}%
\newcommand{\pbFivePastWords}{none}%
\newcommand{\pbFiveClosestName}{MoreFlowers}%
\newcommand{\pbFiveClosestTV}{0.18}%
\newcommand{\pbFiveClosestKappa}{0.17}%
\newcommand{\pbWallSeedOneTVLo}{0.36}%
\newcommand{\pbWallSeedOneTVHi}{0.39}%
\newcommand{\pbWallSeedTwoWords}{past it as well}%
\newcommand{\pbWallSeedsVerdict}{met}%
\newcommand{\pbWallSeedTwoTVLo}{0.23}%
\newcommand{\pbWallSeedTwoTVHi}{0.25}%
\newcommand{\pbFreshNames}{Fabric, RedMaze, SimpleMaze, SimpleWall and Wall}%
\newcommand{\pbFlatPastNames}{none}%
\newcommand{\freshMethodVerdict}{met}%
\newcommand{\heldMazeReachLo}{3}%
\newcommand{\heldMazeReach}{6}%
\newcommand{\doorMazeReach}{7}%
\newcommand{\doorMazeKappa}{2/3}%
\newcommand{\doorMazeD}{3}%
\newcommand{\doorMazeOctFloor}{0.0044}%
\newcommand{\thrVZeroRows}{0.5}%
\newcommand{\vZeroWallValidWords}{valid on all eight coin seeds}%
\newcommand{\vZeroWallMendedHi}{0}%
\newcommand{\vZeroWallTVLo}{0.26}%
\newcommand{\vZeroWallTVHi}{0.26}%
\newcommand{\vZeroWallAligned}{0.45}%
\newcommand{\vZeroOneVerdict}{met}%
\newcommand{\vZeroWallRepHi}{0}%
\newcommand{\vZeroWallRepWords}{none}%
\newcommand{\vZeroWallNrep}{0.09}%
\newcommand{\vZeroTwoFalsifier}{not tripped}%
\newcommand{\vZeroTwoVerdict}{met}%
\newcommand{\vZeroWallRowCopyHi}{0.22}%
\newcommand{\vZeroWallRepHHi}{0.029}%
\newcommand{\vZeroRowsVerdict}{does not copy rows}%
\newcommand{\vZeroSkylineValidWords}{valid on all eight coin seeds}%
\newcommand{\vZeroSkylineMendedHi}{0}%
\newcommand{\vZeroSkylineRepHi}{0}%
\newcommand{\vZeroSkylineNrep}{0.12}%
\newcommand{\vZeroSkylineAligned}{0.51}%
\newcommand{\vZeroCmpWallNoTVLo}{0.26}%
\newcommand{\vZeroCmpWallNoTVHi}{0.26}%
\newcommand{\vZeroCmpWallNoChangedLo}{0.12}%
\newcommand{\vZeroCmpWallNoChangedHi}{0.16}%
\newcommand{\vZeroCmpWallNoNrep}{0.09}%
\newcommand{\vZeroCmpWallWithTVLo}{0.35}%
\newcommand{\vZeroCmpWallWithTVHi}{0.38}%
\newcommand{\vZeroCmpWallWithChangedLo}{0.03}%
\newcommand{\vZeroCmpWallWithChangedHi}{0.06}%
\newcommand{\vZeroCmpWallWithNrep}{0.09}%
\newcommand{\vZeroCmpSkylineNoTVLo}{0.25}%
\newcommand{\vZeroCmpSkylineNoTVHi}{0.26}%
\newcommand{\vZeroCmpSkylineNoChangedLo}{0.56}%
\newcommand{\vZeroCmpSkylineNoChangedHi}{0.59}%
\newcommand{\vZeroCmpSkylineNoNrep}{0.12}%
\newcommand{\vZeroCmpSkylineWithTVLo}{0.32}%
\newcommand{\vZeroCmpSkylineWithTVHi}{0.35}%
\newcommand{\vZeroCmpSkylineWithChangedLo}{0.26}%
\newcommand{\vZeroCmpSkylineWithChangedHi}{0.44}%
\newcommand{\vZeroCmpSkylineWithNrep}{0.12}%
\newcommand{\vZeroCmpWallNoChangedMin}{0.12}%
\newcommand{\vZeroCmpWallWithChangedMax}{0.06}%
\newcommand{\ruleChangedSkylinePctLo}{26\%}%
\newcommand{\ruleChangedSkylinePctHi}{44\%}%
\newcommand{\heldMazeOctValidWords}{valid on all eight coin seeds}%
\newcommand{\heldMazeOctMendedHi}{$3.9\cdot10^{-4}$}%
\newcommand{\heldMazeOctRawLo}{0.0091}%
\newcommand{\heldMazeOctRawHi}{0.012}%
\newcommand{\heldMazeOctRep}{0}%
\newcommand{\hmTwoVerdict}{met}%
\newcommand{\heldMazeAligned}{0.40}%
\newcommand{\hmThreeVerdict}{met}%
\newcommand{\doorMazeMethodValidWords}{valid on all eight coin seeds}%
\newcommand{\doorMazeMethodMendedHi}{0}%
\newcommand{\doorMazeMethodRawHi}{0}%
\newcommand{\doorMazeMethodRep}{0}%
\newcommand{\doorMazeMethodNrep}{0.78}%
\newcommand{\hdTwoVerdict}{met}%
\newcommand{\doorMazeAligned}{0.105}%
\newcommand{\doorMazeAlignedR}{0.10}%
\newcommand{\hdThreeVerdict}{met}%
\newcommand{\doorMazeOctValidWords}{not valid}%
\newcommand{\doorMazeOctMendedHi}{0.059}%
\newcommand{\doorMazeOctRawLo}{0.049}%
\newcommand{\doorMazeOctRawHi}{0.061}%
\newcommand{\doorMazeOctAboveFloorWords}{three}%
\newcommand{\doorMazeOctAligned}{0.35}%
\newcommand{\heldoutAllCountWords}{five}%
\newcommand{\heldoutAllValidCountWords}{five}%
\newcommand{\heldoutAllValidCount}{5}%
\newcommand{\heldOctWallTValidWords}{not valid}%
\newcommand{\heldOctSimpleWallTValidWords}{valid on all eight coin seeds}%
\newcommand{\heldOctWallHTwoValidWords}{valid on all eight coin seeds}%
\newcommand{\heldOctWallTMendedLo}{$5.4\cdot10^{-3}$}%
\newcommand{\heldOctWallHTwoHeightsPct}{42}%
\newcommand{\heldMethodWallHTwoHeightsWords}{every}%
\newcommand{\heldoutInputNeededCountWords}{two}%
\newcommand{\blockRefBudgetWords}{twelve}%
\newcommand{\earlyConfigRounds}{16}%
\newcommand{\earlyConfigUpdates}{8\,000}%
\newcommand{\dirExtraChannelsWords}{eight}%
\newcommand{\statWeightUpdates}{4\,000}%
\newcommand{\statWeightRowList}{3, 10 or 30}%
\newcommand{\statWeightOrdList}{1, 3 or 10}%
\newcommand{\statWeightLR}{$2\cdot10^{-4}$}%
\newcommand{\mendGroupWide}{24}%
\newcommand{\mendGroupNarrow}{8}%
\newcommand{\tcTransportedWords}{twelve}%
\newcommand{\tcCertifiedWords}{thirteen}%
\newcommand{\tcUncertOf}{39}%
\newcommand{\tcRadiusMin}{8}%
\newcommand{\tcLadderMax}{128}%
\newcommand{\tcRadiusMax}{512}%
\newcommand{\tcTorusSmall}{256}%
\newcommand{\tcTorusMid}{512}%
\newcommand{\tcTorusLarge}{2048}%
\newcommand{\tcCanvases}{468}%
\newcommand{\tcCanvasesAmend}{87}%
\newcommand{\tcFourBar}{ten}%
\newcommand{\tcOneVerdict}{met}%
\newcommand{\tcTwoVerdict}{met}%
\newcommand{\tcThreeVerdict}{met}%
\newcommand{\tcFourVerdict}{met}%
\newcommand{\tcFiveVerdict}{met}%
\newcommand{\tcShareDevMax}{4\%}%
\newcommand{\tcFloorPairs}{62}%
\newcommand{\tcFloorRatioAllLo}{7.2}%
\newcommand{\tcFloorRatioAllHi}{133}%
\newcommand{\tcFloorRatioLo}{8.6}%
\newcommand{\tcFloorRatioHi}{120}%
\newcommand{\tcFloorRatioLoName}{SimpleWall}%
\newcommand{\tcFloorRatioHiName}{3Bricks}%
\newcommand{\tcWallFloorRatio}{9.5}%
\newcommand{\tcWallFloorRatioR}{512}%
\newcommand{\tcWallEpsLarge}{0.0062}%
\newcommand{\tcWallFloorLarge}{0.00065}%
\newcommand{\tcWallFloorRatioMin}{8.4}%
\newcommand{\tcWallFloorRatioMax}{11}%
\newcommand{\tcKappaRatioLo}{8.7}%
\newcommand{\tcKappaRatioHi}{173}%
\newcommand{\tcKappaRatioLoName}{SimpleMaze}%
\newcommand{\tcKappaRatioHiName}{3Bricks}%
\newcommand{\tcKappaRatioHiAbout}{170}%
\newcommand{\tcKappaSpreadMax}{1.2}%
\newcommand{\tcKappaSpreadMaxName}{SimpleWall}%
\newcommand{\tcWallKappaRatio}{9.6}%
\newcommand{\tcWallKappaFloorLarge}{0.00065}%
\newcommand{\tcWallKappaRatioMin}{9.5}%
\newcommand{\tcWallKappaRatioMax}{11}%
\newcommand{\tcEpsOverQMedian}{0.70}%
\newcommand{\tcEpsOverQLo}{0.50}%
\newcommand{\tcEpsOverQHi}{0.88}%
\newcommand{\tcREpsMedian}{2.4}%
\newcommand{\tcREpsLo}{1.7}%
\newcommand{\tcREpsHi}{3.4}%
\newcommand{\tcRuleRadius}{32}%
\newcommand{\tcRuleOverCopyMedian}{1.1}%
\newcommand{\tcRuleOverCopyLo}{0.56}%
\newcommand{\tcRuleOverCopyHi}{1.4}%
\newcommand{\tcRuleOverCopyWithinTwo}{12}%
\newcommand{\tcRuleOverCopyWithinTwoWords}{twelve}%
\newcommand{\tcRuleOverCopyWithinTwoOf}{all twelve}%
\newcommand{\tcCertValidAnyWords}{none}%
\newcommand{\tcTheoryValidAnyWords}{none}%
\newcommand{\tcThreeValidNames}{none}%
\newcommand{\tcUncertValidSixtyFour}{19}%
\newcommand{\tcUncertValidSixtyFourWords}{nineteen}%
\newcommand{\tcPublicOf}{52}%
\newcommand{\tcAutoUncertValidThirtyTwo}{28}%
\newcommand{\tcAutoAllValidThirtyTwo}{28}%
\newcommand{\tcUncertValidThirtyTwo}{16}%
\newcommand{\tcAllValidThirtyTwo}{16}%
\newcommand{\tcBothValidThirtyTwo}{16}%
\newcommand{\tcCopyOnlyThirtyTwo}{0}%
\newcommand{\tcAutoOnlyThirtyTwo}{12}%
\newcommand{\tcTilesValid}{23}%
\newcommand{\tcBothValidSixtyFour}{19}%
\newcommand{\tcCopyOnlySixtyFour}{0}%
\newcommand{\tcTilesOnlySixtyFour}{4}%
\newcommand{\tcWallRMendSixtyFour}{2.6}%
\newcommand{\tcWallRMendTwoFiftySix}{3.0}%
\newcommand{\tcWallMendedTwoFiftySix}{0.012}%
\newcommand{\tcWallEpsTwoFiftySix}{0.013}%
\newcommand{\tcWallRMendRatioTwoFiftySix}{1.2}%
\newcommand{\tcWallRMendFiveTwelve}{2.7}%
\newcommand{\tcWallMendedFiveTwelve}{0.0054}%
\newcommand{\tcWallEpsFiveTwelve}{0.0062}%
\newcommand{\tcWallRMendRatioFiveTwelve}{1.1}%
\newcommand{\tcContrastRadius}{64}%
\newcommand{\tcControlsWords}{ten}%
\newcommand{\tcCopyOnlyThirtyTwoWords}{none}%
\newcommand{\tcWallFloorBelowThrR}{332}%
\newcommand{\tcWallKappaRatioAbout}{ten}%
\newcommand{\tcRemCertMax}{44\%}%
\newcommand{\tcRemCertMaxName}{SimpleWall}%
\newcommand{\tcRemUncertMin}{51\%}%
\newcommand{\tcRemUncertMinName}{ColoredCity}%
\newcommand{\tcRemCertMedian}{28\%}%
\newcommand{\tcRemUncertMedian}{98\%}%
\newcommand{\tcRemResistMin}{51\%}%
\newcommand{\tcResistInvalidSixtyFourWords}{eight}%
\newcommand{\tcRemCertMaxThirtyTwo}{29\%}%
\newcommand{\tcRemCertMaxThirtyTwoName}{Flowers}%
\newcommand{\tcRemUncertMinThirtyTwo}{15\%}%
\newcommand{\tcRemUncertMinThirtyTwoName}{Cat}%
\newcommand{\tcControlsValidOneTwentyEightWords}{six}%
\newcommand{\tcResistValidOneTwentyEightNames}{Skew2}%
\newcommand{\tcResistValidSixtyFourWords}{none}%
\newcommand{\tcWallREpsSixtyFour}{3.1}%
\newcommand{\tcWallREpsFiveTwelve}{3.2}%
\newcommand{\tcOneBand}{15\%}%
\newcommand{\tcFiveFactorWords}{two}%
\newcommand{\tcDeltaOverQMax}{0.87}%
\newcommand{\tcRadiusLargeMin}{256}%
\newcommand{\tcWallPassRadius}{2\,700}%
\newcommand{\ddUpdates}{20\,000}%
\newcommand{\ddLevels}{50}%
\newcommand{\ddRadius}{300}%
\newcommand{\ddRadiusPerLevel}{6}%
\newcommand{\ddTrainings}{78}%
\newcommand{\ddModelsA}{50}%
\newcommand{\ddModelsB}{28}%
\newcommand{\ddCertWords}{thirteen}%
\newcommand{\ddTransWords}{twelve}%
\newcommand{\ddCtrlWords}{ten}%
\newcommand{\ddPilotBarWords}{three}%
\newcommand{\ddChannels}{128}%
\newcommand{\ddBlocksWords}{five}%
\newcommand{\ddTrainLevels}{1\,000}%
\newcommand{\ddCosineOffset}{0.008}%
\newcommand{\ddBetaMax}{0.999}%
\newcommand{\ddCrop}{64}%
\newcommand{\ddBatch}{32}%
\newcommand{\ddLR}{$2\cdot10^{-4}$}%
\newcommand{\ddEMA}{0.999}%
\newcommand{\ddDropShare}{10\%}%
\newcommand{\ddGuidance}{2}%
\newcommand{\ddPilotUpdatesHigh}{60\,000}%
\newcommand{\ddPilotLevelsLow}{20}%
\newcommand{\ddParamsM}{1.25 to 1.27}%
\newcommand{\ddOneVerdict}{met}%
\newcommand{\ddTwoVerdict}{not met}%
\newcommand{\ddThreeVerdict}{not met}%
\newcommand{\ddFourVerdict}{met}%
\newcommand{\ddFiveVerdict}{met}%
\newcommand{\ddSixVerdict}{met}%
\newcommand{\ddFloorVerdict}{met}%
\newcommand{\ddSeamVerdict}{met}%
\newcommand{\ddOneOf}{26}%
\newcommand{\ddOneValid}{0}%
\newcommand{\ddOneValidNames}{none}%
\newcommand{\ddTwoBar}{four}%
\newcommand{\ddTwoOf}{6}%
\newcommand{\ddTwoOfWords}{six}%
\newcommand{\ddTwoNames}{Cave, Maze, Knot, Rooms, Office and City}%
\newcommand{\ddTwoValid}{2}%
\newcommand{\ddTwoValidWords}{two}%
\newcommand{\ddTwoValidNames}{Cave and Knot}%
\newcommand{\ddTwoInvalidWords}{four}%
\newcommand{\ddThreeBar}{eight}%
\newcommand{\ddThreeOf}{12}%
\newcommand{\ddRepWindow}{16}%
\newcommand{\ddRepBar}{0.05}%
\newcommand{\ddEscapeBar}{0.5}%
\newcommand{\ddThreeValid}{8}%
\newcommand{\ddThreeValidWords}{eight}%
\newcommand{\ddThreeCounted}{5}%
\newcommand{\ddThreeCountedWords}{five}%
\newcommand{\ddThreeEscapes}{3}%
\newcommand{\ddThreeEscapeNames}{Fabric, RedMaze and SimpleMaze}%
\newcommand{\ddThreeRepAbove}{0}%
\newcommand{\ddThreeRepAboveNames}{none}%
\newcommand{\ddFourFactorWords}{ten}%
\newcommand{\ddFourBar}{eight}%
\newcommand{\ddFourTenfold}{12}%
\newcommand{\ddFourTenfoldWords}{twelve}%
\newcommand{\ddFourNotNames}{none}%
\newcommand{\ddFiveOf}{8}%
\newcommand{\ddFiveAsPredicted}{8}%
\newcommand{\ddFiveAgainstNames}{none}%
\newcommand{\ddSixBarWords}{half}%
\newcommand{\ddSixLevelsHigh}{100}%
\newcommand{\ddSixLevelsLow}{20}%
\newcommand{\ddSixWall}{0.90}%
\newcommand{\ddSixSimpleWall}{0.90}%
\newcommand{\ddSixFabric}{0.89}%
\newcommand{\ddSixCave}{0.79}%
\newcommand{\ddSweepLevels}{10, 20, 50 and 100}%
\newcommand{\ddSweepSize}{2\,048}%
\newcommand{\ddFloorOf}{24}%
\newcommand{\ddFloorRegRadius}{120}%
\newcommand{\ddFloorHeld}{24}%
\newcommand{\ddFloorMinRatio}{120}%
\newcommand{\ddFloorMinRatioName}{SimpleWall at seed 0}%
\newcommand{\ddSeamChunk}{64}%
\newcommand{\ddSeamOf}{4\,096}%
\newcommand{\ddSeamBarWords}{one cell in ten thousand}%
\newcommand{\ddSeamWholeOf}{65\,536}%
\newcommand{\ddSeamMax}{0}%
\newcommand{\ddSeamHalfMax}{0}%
\newcommand{\ddSeamDiffSingleWords}{none}%
\newcommand{\ddSeamDiffHalfWords}{none}%
\newcommand{\ddHalfSingleShareLo}{0.021\%}%
\newcommand{\ddHalfSingleShareHi}{0.22\%}%
\newcommand{\ddCtrlInvalidOf}{8}%
\newcommand{\ddCtrlInvalidNames}{Sewers, Skew2, Lake and Nested}%
\newcommand{\ddCtrlInvalidValid}{0}%
\newcommand{\ddCtrlInvalidValidWords}{none}%
\newcommand{\ddRawOverAutoMedianA}{2.2}%
\newcommand{\ddRawOverAutoMinA}{1.2}%
\newcommand{\ddRawOverAutoMaxA}{22}%
\newcommand{\ddChangedMedianA}{7.3\%}%
\newcommand{\ddChangedAutoMedianA}{13\%}%
\newcommand{\ddRawOverAutoMedianB}{0.51}%
\newcommand{\ddRawOverAutoMinB}{0}%
\newcommand{\ddRawOverAutoMaxB}{1.8}%
\newcommand{\ddChangedMedianB}{9.3\%}%
\newcommand{\ddChangedAutoMedianB}{10\%}%
\newcommand{\ddModelsReadOf}{78 of 78}%
\newcommand{\ddCopyRadius}{128}%
\newcommand{\ddCopyBound}{0.031}%
\newcommand{\ddRawOverCopyMedianA}{7.5}%
\newcommand{\ddRawOverCopyMedianB}{1.3}%
\newcommand{\ddDdimOfA}{50}%
\newcommand{\ddDdimValidA}{4}%
\newcommand{\ddDdimOfB}{28}%
\newcommand{\ddDdimValidB}{18}%
\newcommand{\ddGuidOfB}{28}%
\newcommand{\ddGuidValidB}{19}%
\newcommand{\ddPrimValidA}{4}%
\newcommand{\ddPrimValidB}{20}%
\newcommand{\ddAlignedMedianB}{0.34}%
\newcommand{\ddNrepMedianB}{0.12}%
\newcommand{\ddAlignedMethodMedianB}{0.36}%
\newcommand{\ddNrepMethodMedianB}{0.14}%
\newcommand{\ddWallRawA}{0.22}%
\newcommand{\ddWallMendedA}{0.22}%
\newcommand{\ddWallRawOverAutoA}{2.0}%
\newcommand{\ddWallChangedA}{0.14\%}%
\newcommand{\ddWallOwnFloorA}{0.0012}%
\newcommand{\ddWallRawOverOwnFloorA}{175}%
\newcommand{\ddWallRawOverFloorA}{198}%
\newcommand{\ddWallRawOverCopyA}{8.4}%
\newcommand{\ddWallRawB}{0.012}%
\newcommand{\ddWallMendedB}{0}%
\newcommand{\ddWallRawOverAutoB}{0.11}%
\newcommand{\ddWallChangedB}{8.4\%}%
\newcommand{\ddWallOwnFloorB}{0.0011}%
\newcommand{\ddWallRawOverOwnFloorB}{11}%
\newcommand{\ddWallRawOverFloorB}{11}%
\newcommand{\ddWallRawOverCopyB}{0.45}%
\newcommand{\ddWallFloor}{0.0011}%
\newcommand{\ddWallAutoRaw}{0.11}%
\newcommand{\ddWallAutoChanged}{2\%}%
\newcommand{\ddWallMethodRaw}{0.0087}%
\newcommand{\ddWallMethodChanged}{4.5\%}%
\newcommand{\ddWallRawOverMethodB}{1.3}%
\newcommand{\ddCaveRawOverRule}{19}%
\newcommand{\ddCaveChanged}{15\%}%
\newcommand{\ddCaveChangedRule}{0.45\%}%
\newcommand{\ddCertRawOverAutoMinA}{1.4}%
\newcommand{\ddCertRawOverAutoMaxA}{3.1}%
\newcommand{\ddCtrlRawOverAutoMedianA}{7.4}%
\newcommand{\ddRemovedCertMedianA}{3.2\%}%
\newcommand{\ddRemovedCtrlMedianA}{84\%}%
\newcommand{\ddSixMin}{0.89}%
\newcommand{\ddFourZeroWords}{nine}%
\newcommand{\ddMethodValidTransWords}{ten}%
\newcommand{\ddThreeCountedNames}{Flowers, MoreFlowers, Platformer, SimpleWall and Wall}%
\newcommand{\ddThreeInvalidMethodValidNames}{Skyline and Skyline2}%
\newcommand{\ddThreeInvalidChangedMin}{38\%}%
\newcommand{\ddTransNonPhaseWords}{nine}%
\newcommand{\ddMethodValidNonPhaseWords}{seven}%
\newcommand{\ddSkyRawOverMethodLo}{4.7}%
\newcommand{\ddSkyRawOverMethodHi}{5.7}%
\newcommand{\ddSkyRawOverMethodAboutWords}{five}%
\newcommand{\ddCountedRepMax}{0.0059}%
\newcommand{\ddCountedRepMaxWords}{fewer than one in a hundred}%
\newcommand{\ddThreeEscapeWords}{three}%
\newcommand{\ddRowTestRows}{17\,409}%
\newcommand{\ddRowTestCopiesWords}{none}%
\newcommand{\ddRowTestCanvasesWords}{16}%
\newcommand{\ddRowTestWallCourseLo}{3\%}%
\newcommand{\ddRowTestWallCourseHi}{4\%}%
\newcommand{\recMilli}{$10^{-3}$}%
\newcommand{\recCenti}{$10^{-2}$}%
\newcommand{\recHalf}{half}%
\newcommand{\recTwice}{twice}%
\newcommand{\recThird}{a third}%
\newcommand{\recTenth}{a tenth}%
\newcommand{\recFifth}{a fifth}%
\newcommand{\softShareBar}{60\%}%
\newcommand{\softSpeedBar}{100}%
\newcommand{\softSpeedRead}{1000}%
\newcommand{\curveRounds}{64}%
\newcommand{\curveOldBar}{0.65}%
\newcommand{\curveRankBar}{0.8}%
\newcommand{\radLadderTop}{64}%
\newcommand{\radBandLo}{-1.3}%
\newcommand{\radBandHi}{-0.7}%
\newcommand{\radMendBar}{-2}%
\newcommand{\radColBar}{-1}%
\newcommand{\radSteep}{-1.5}%
\newcommand{\threeDBar}{10\%}%
\newcommand{\threeDBarTwo}{30\%}%
\newcommand{\threeDRoundsLo}{8}%
\newcommand{\threeDRoundsHi}{16}%
\newcommand{\statsSmallBar}{38}%
\newcommand{\statsSmallTolBar}{44}%
\newcommand{\statsMidBar}{three quarters}%
\newcommand{\statsBigBar}{26}%
\newcommand{\statsBigTolBar}{40}%
\newcommand{\statsSd}{two}%
\newcommand{\statsBtBar}{8}%
\newcommand{\certCtwoBar}{3}%
\newcommand{\certCthreeBar}{30}%
\newcommand{\certCfourBar}{17}%
\newcommand{\certCfiveBar}{four of five}%
\newcommand{\certTargetBar}{10}%
\newcommand{\certTargetOf}{12}%
\newcommand{\mendDivBar}{10\%}%
\newcommand{\mendCellsBar}{20\%}%
\newcommand{\mendNineBar}{9}%
\newcommand{\mendRadiusBar}{2}%
\newcommand{\mendDefectsBar}{60\%}%
\newcommand{\mendBestWords}{four}%
\newcommand{\mendBlocksBar}{20\%}%
\newcommand{\wideNewBar}{4}%
\newcommand{\widePartialOf}{10}%
\newcommand{\wideUncertBar}{3}%
\newcommand{\satCaveBar}{95\%}%
\newcommand{\satWallBar}{30\%}%
\newcommand{\satUndecidedBar}{5\%}%
\newcommand{\satDiffBar}{10}%
\newcommand{\satDiffCount}{3 of 5}%
\newcommand{\satFallBar}{30}%
\newcommand{\satFlatBar}{10}%
\newcommand{\witBar}{75\%}%
\newcommand{\witCountBar}{3 of 4}%
\newcommand{\witFreeBar}{30}%
\newcommand{\wxPatchSmall}{3}%
\newcommand{\wxPatchBig}{4}%
\newcommand{\wxUncertBar}{5\%}%
\newcommand{\complBar}{3}%
\newcommand{\complOf}{4}%
\newcommand{\complDthreeBar}{20}%
\newcommand{\shareBar}{5}%
\newcommand{\shareKLBar}{1.5}%
\newcommand{\crFiveBar}{five}%
\newcommand{\msWallsBar}{2}%
\newcommand{\msVarBar}{4}%
\newcommand{\msAlignedSlack}{0.05}%
\newcommand{\msNrepSlack}{0.1}%
\newcommand{\msKLBar}{1.5}%
\newcommand{\msCorrBar}{1.5}%
\newcommand{\msUncertBar}{3}%
\newcommand{\msCountBar}{24}%
\newcommand{\msKeepBar}{20}%
\newcommand{\msCutBar}{twofold}%
\newcommand{\msKLMendBar}{1.25}%
\newcommand{\msNrepBar}{0.05}%
\newcommand{\msSixteenBar}{30}%
\newcommand{\msSeedsBar}{seven}%
\newcommand{\tsCountBar}{24}%
\newcommand{\tsKeepBar}{22}%
\newcommand{\tsDiffBar}{3}%
\newcommand{\bsBar}{0.8}%
\newcommand{\latticeBar}{20}%
\newcommand{\smLessBar}{one}%
\newcommand{\smTwoBar}{two}%
\newcommand{\ccSmoothLowerBar}{9}%
\newcommand{\ccSmoothOf}{13}%
\newcommand{\ccValidEightBar}{1}%
\newcommand{\ccValidSixteenBar}{3}%
\newcommand{\ccSixteenBar}{35}%
\newcommand{\ccTwentyFourBar}{38}%
\newcommand{\ccFlatBar}{34}%
\newcommand{\mvTwofold}{twofold}%
\newcommand{\mvOneBar}{1}%
\newcommand{\mvWallBar}{two of three}%
\newcommand{\mvRuleBar}{42}%
\newcommand{\mvFlatBar}{30}%
\newcommand{\reRuleBar}{35}%
\newcommand{\reFlatBar}{27}%
\newcommand{\reNrepBar}{0.10}%
\newcommand{\reAlignedBar}{0.30}%
\newcommand{\tileValidBar}{30}%
\newcommand{\tileCertBar}{3}%
\newcommand{\hwBar}{9}%
\newcommand{\hwOf}{12}%
\newcommand{\hwTenfold}{tenfold}%
\newcommand{\replayBar}{0.5}%
\newcommand{\connBar}{0.8}%
\newcommand{\seamCCMarginBar}{192}%
\newcommand{\seamCCShareBar}{90\%}%
\newcommand{\mendTestFailed}{38\%}%
\newcommand{\fxThreeBar}{21}%
\newcommand{\softCollapsed}{30}%
\newcommand{\softLegs}{50}%
\newcommand{\softRepairEight}{8\%}%
\newcommand{\softRepairSixteen}{6\%}%
\newcommand{\softSpeedMedian}{440}%
\newcommand{\softSpeedSlowNames}{Chess and Water}%
\newcommand{\varRandomMono}{3}%
\newcommand{\varRandomOf}{4}%
\newcommand{\varFinalMono}{none}%
\newcommand{\threeDMaze}{69\%}%
\newcommand{\threeDPipes}{5\%}%
\newcommand{\threeDMazeTwo}{49\%}%
\newcommand{\threeDChessLo}{34\%}%
\newcommand{\threeDChessHi}{26\%}%
\newcommand{\mendEarlyCount}{5}%
\newcommand{\mendEarlyChanged}{8.6\%}%
\newcommand{\seamCCWorstMargin}{224}%
\newcommand{\glueHoldsCount}{7}%
\newcommand{\glueHoldsWords}{seven}%
\newcommand{\glueFailNames}{3Bricks}%
\newcommand{\glueShiftZeroCount}{6}%
\newcommand{\glueShiftZeroWords}{six}%
\newcommand{\glueAllShiftsNames}{Wall}%
\newcommand{\glueConfigsMax}{120}%
\newcommand{\glueConfigsTotalMax}{17{,}425}%
\newcommand{\glueSubsetMax}{992}%
\newcommand{\glueWallFreeTransitions}{21}%
\newcommand{\glueWallFreeNonExample}{0}%
\newcommand{\glueWallAllTransitions}{81}%
\newcommand{\glueWallAllNonExample}{18}%
\newcommand{\glueBricksFreeRunWords}{seven}%
\newcommand{\glueBricksRunLimitMax}{31}%
\newcommand{\glueBricksBandWalksWithRealizationWords}{one}%
\newcommand{\glOneNames}{WallT, SimpleWallT and WallH2}%
\newcommand{\glOneCountWords}{three}%
\newcommand{\glOneCanvasesWords}{nine}%
\newcommand{\lrWorlds}{32}%
\newcommand{\lrCanvas}{128}%
\newcommand{\lrCanvasSmallCountWords}{five}%
\newcommand{\lrCanvasSmall}{96}%
\newcommand{\lrDistMaxSmall}{48}%
\newcommand{\lrDistMidSmall}{12}%
\newcommand{\lrDistMax}{64}%
\newcommand{\lrDistMid}{16}%
\newcommand{\lrShuffles}{20}%
\newcommand{\lrCertOrder}{11}%
\newcommand{\lrCertOrderWords}{eleven}%
\newcommand{\lrCertOf}{12}%
\newcommand{\lrCertOfWords}{twelve}%
\newcommand{\lrCertBar}{10}%
\newcommand{\lrCertMissName}{Skyline2}%
\newcommand{\lrSkylineTwoRatio}{0.45}%
\newcommand{\lrUncertOrder}{0}%
\newcommand{\lrUncertOrderWords}{none}%
\newcommand{\lrUncertBar}{5}%
\newcommand{\lrNonResOrder}{1}%
\newcommand{\lrNonResOrderWords}{one}%
\newcommand{\lrNonResOf}{31}%
\newcommand{\lrNonResBar}{6}%
\newcommand{\lrNonResOrderNames}{Maze}%
\newcommand{\lrNonResOrderDir}{antidiagonal}%
\newcommand{\lrNonResNotMeasuredNames}{Cats}%
\newcommand{\lrTownDirs}{rows}%
\newcommand{\lrUncertFallFloorWords}{seven}%
\newcommand{\lrUncertFallMinName}{Cat}%
\newcommand{\lrUncertVanishNames}{Nested and Skew2}%
\newcommand{\lrNonResKeepMoreNames}{Mountains}%
\newcommand{\lrCertKeepHalfWords}{eleven}%
\newcommand{\glTwoControlNames}{SimpleMaze, SimpleWall, RedMaze and Fabric}%
\newcommand{\glTwoCarriedNames}{Cat and ColoredCity}%
\newcommand{\glTwoNoneNames}{Lake, Skew1 and Skew2}%
\newcommand{\glTwoUndecidedNames}{Nested, ScaledMaze and Sewers}%
\newcommand{\glTwoAgree}{3}%
\newcommand{\glTwoAgreeWords}{three}%
\newcommand{\glTwoAgreeBar}{six}%
\newcommand{\glTwoSlopeLo}{$-$1.3}%
\newcommand{\glTwoSlopeHi}{$-$0.7}%
\newcommand{\glTwoSlopeNone}{$-$1.6}%
\newcommand{\glTwoSpearman}{0.67}%
\newcommand{\glTwoBendFrom}{32}%
\newcommand{\glTwoBendTo}{64}%
\newcommand{\glTwoCatSlope}{$-$1.27}%
\newcommand{\glTwoColoredCitySlope}{$-$1.18}%
\newcommand{\mwRadiusA}{48}%
\newcommand{\mwRadiusB}{96}%
\newcommand{\mwBlockA}{96}%
\newcommand{\mwBlockB}{192}%
\newcommand{\mwBoundA}{1528}%
\newcommand{\mwBoundB}{3064}%
\newcommand{\mwSeedsWords}{eight}%
\newcommand{\mwExamplesWords}{seven}%
\newcommand{\mwValidANames}{Skew1 and ScaledMaze}%
\newcommand{\mwValidAWords}{two}%
\newcommand{\mwCountA}{47}%
\newcommand{\mwInvalidAWords}{five}%
\newcommand{\mwInvalidANames}{3Bricks, Lake, Link2, Mountains and SmileCity}%
\newcommand{\mwValidBNewNames}{Mountains and SmileCity}%
\newcommand{\mwValidBWords}{four}%
\newcommand{\mwInvalidBNames}{Lake and Link2}%
\newcommand{\mwKilledNames}{3Bricks}%
\newcommand{\mwBricksOverA}{six}%
\newcommand{\mwBricksOverR}{five}%
\newcommand{\mwSmileRatioA}{0.32}%
\newcommand{\mwFarRatioMedian}{1.02}%
\newcommand{\mwCellsBLo}{31\%}%
\newcommand{\mwCellsBHi}{50\%}%
\newcommand{\mwTvRatioBNew}{2.1}%
\newcommand{\mwSkewTvA}{0.30}%
\newcommand{\mwSkewTvR}{0.21}%
\newcommand{\mwSmileCellsB}{50\%}%
\newcommand{\mwSmileTvB}{0.33}%
\newcommand{\mwSmileTvR}{0.16}%
\newcommand{\mwFourN}{three}%
\newcommand{\mwTwoBar}{0.7}%
\newcommand{\mwFiveBar}{0.1}%
\newcommand{\mwFiveMax}{0.08}%
\newcommand{\fxNExamples}{27}%
\newcommand{\fxNExamplesWords}{twenty-seven}%
\newcommand{\fxNRepos}{8}%
\newcommand{\fxNReposWords}{eight}%
\newcommand{\fxNCertified}{5}%
\newcommand{\fxNCertifiedWords}{five}%
\newcommand{\fxCertifiedNames}{GbDiamond, GbMountain, GbLattice, GbTriangles and GbZigzag}%
\newcommand{\fxNInput}{4}%
\newcommand{\fxNInputWords}{four}%
\newcommand{\fxInputNames}{GbMountain, GbLattice, GbTriangles and GbZigzag}%
\newcommand{\fxRefusedNames}{GbDiamond}%
\newcommand{\fxNUncert}{22}%
\newcommand{\fxNUncertWords}{twenty-two}%
\newcommand{\fxNResistant}{9}%
\newcommand{\fxNResistantWords}{nine}%
\newcommand{\fxNCertResistant}{5}%
\newcommand{\fxNCertResistantWords}{five}%
\newcommand{\fxNCertNotResistant}{0}%
\newcommand{\fxNCertNotResistantWords}{none}%
\newcommand{\fxNMendable}{13}%
\newcommand{\fxNMendableWords}{thirteen}%
\newcommand{\fxNPartial}{5}%
\newcommand{\fxNPartialWords}{five}%
\newcommand{\fxNNotResistant}{18}%
\newcommand{\fxNNotResistantWords}{eighteen}%
\newcommand{\fxMethodValid}{21}%
\newcommand{\fxMethodValidWords}{twenty-one}%
\newcommand{\fxMethodRobust}{20}%
\newcommand{\fxMethodRobustWords}{twenty}%
\newcommand{\fxMethodInvalidNames}{GreedyA, GreedyB, GreedyC, GbDiamond, KrCave and DbPathway}%
\newcommand{\fxMethodValidInput}{4}%
\newcommand{\fxMethodValidInputWords}{four}%
\newcommand{\fxOctValidInput}{1}%
\newcommand{\fxOctValidInputWords}{one}%
\newcommand{\fxMethodInputPhrase}{all four}%
\newcommand{\fxInputDecidesNames}{GbLattice, GbTriangles and GbZigzag}%
\newcommand{\fxTiledNames}{GbLattice, GbTriangles and GbZigzag}%
\newcommand{\fxNTiledWords}{three}%
\newcommand{\fxInputOtherNames}{GbMountain}%
\newcommand{\fxOctFragileNames}{GbMountain}%
\newcommand{\fxSeedsValidInput}{8 of 8}%
\newcommand{\fxSeedsPhrase}{and so are its rules trained from training seeds 1 and 2}%
\newcommand{\fxOctValidAll}{18}%
\newcommand{\fxOctValidAllWords}{eighteen}%
\newcommand{\fxCtrlValidAll}{17}%
\newcommand{\fxCtrlValidAllWords}{seventeen}%
\newcommand{\fxOctValidUncert}{17}%
\newcommand{\fxOctValidUncertWords}{seventeen}%
\newcommand{\fxTileValidUncert}{16}%
\newcommand{\fxTileValidUncertWords}{sixteen}%
\newcommand{\fxIidValidUncert}{0}%
\newcommand{\fxIidValidUncertWords}{none}%
\newcommand{\fxOctOverTilePhrase}{by a single example}%
\newcommand{\fxLostToCtrlNames}{GreedyB and DbPathway}%
\newcommand{\fxTcValidCert}{0}%
\newcommand{\fxTcValidCertWords}{none}%
\newcommand{\fxCheckMinRatioWords}{six}%
\newcommand{\fxCheckMinRatio}{6.2}%
\newcommand{\fxVerdictOne}{HELD}%
\newcommand{\fxVerdictTwo}{HELD}%
\newcommand{\fxVerdictThree}{HELD}%
\newcommand{\fxVerdictFour}{HELD}%
\newcommand{\fxVerdictFive}{HELD}%
\newcommand{\fxVerdictSix}{HELD}%
\newcommand{\fxVerdictCheck}{HELD}%
\newcommand{\sxCanvasesWords}{twelve}%
\newcommand{\sxCanvasWide}{2048}%
\newcommand{\sxChunkWide}{64}%
\newcommand{\sxEdgeGap}{992}%
\newcommand{\sxMarginList}{256, 384, 512 and 760}%
\newcommand{\sxOctaveNames}{Cat, Link and Nested}%
\newcommand{\sxDiffering}{0}%

%% file: sec_intro.tex
\section{Introduction}
\label{sec:intro}

Games generate unbounded worlds from seeded noise. A terrain height is a hash of its coordinates and a seed, so any
region can be computed on its own and in any order, and machines that share the seed make the same world. When every
small window of a map must come from a designer's example, worlds are made by search instead. Wave Function
Collapse \citep{gumin2016wfc} fills a canvas in a data-dependent order and restarts when the constraints run into a contradiction. Model synthesis
\citep{merrell2009model} modifies a valid model one overlapping block at a time, so that a contradiction costs one
block, and block solvers built on it \citep{boris2021infinite,kleineberg2023infinite} give random access by re-solving
blocks over a fixed periodic background, which they keep where a block cannot be solved. Where validity fixes a row's
place, as in a brick wall, their worlds keep the background's rows. Learned generators with random access, such as InfiniteDiffusion for terrain
\citep{goslin2026infinite}, carry no hard constraint.

\begin{figure*}[t]\centering\includegraphics[width=\textwidth]{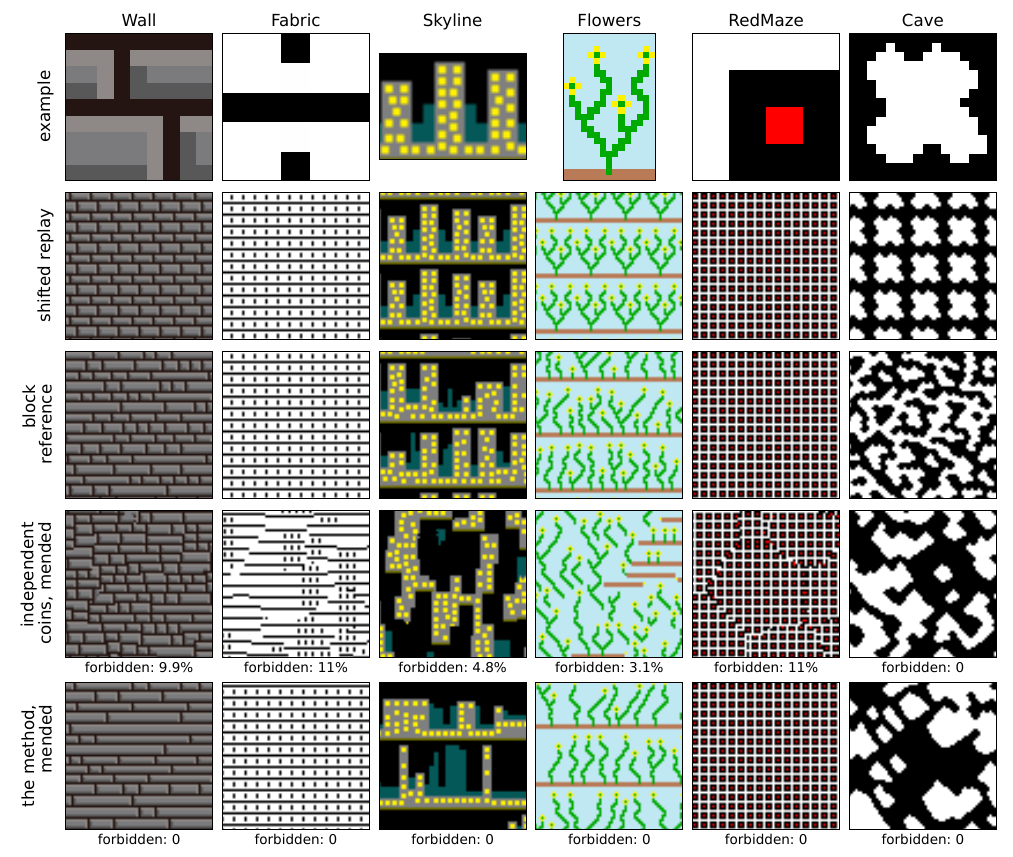}
\caption{Worlds from one example. Each column shows a public example and the central $64\times64$ cells of a
$256\times256$ canvas from shifted replay (the example tiled from a random offset), the block reference (a block solver
on blocks placed at random), the rule on independent coins and the method (the learned rule given the input of
Section~\ref{sec:cert}), both mended, with the share of forbidden windows on the whole canvas. On the first five
examples validity carries a quantity over any distance, an invariant along rows or a phase. Independent coins leave
forbidden windows on all five; given the quantity, the method leaves none, and on Fabric and RedMaze its world is the
tiled example. Cave carries no such quantity, and both rules
are valid on it.}
\label{fig:gallery}\end{figure*}

We study random access through locality: each cell is a fixed function of the randomness near it, which may include
coins shared along whole rows or on blocks that meet its neighborhood (Section~\ref{sec:ra}). A cellular automaton run
for a fixed number of rounds is local in this sense, and so is a diffusion sampler whose convolutional denoiser draws
fresh noise at every step. A generator may also read its position modulo a fixed period; one that reads its position
with no period, or randomness at unboundedly many scales, can also give random access but lies outside this model. We
ask what a local generator must be given, whatever its architecture, to make worlds whose every
window comes from one example (Figure~\ref{fig:gallery}).

\begin{figure*}[t]\centering\includegraphics[width=\textwidth]{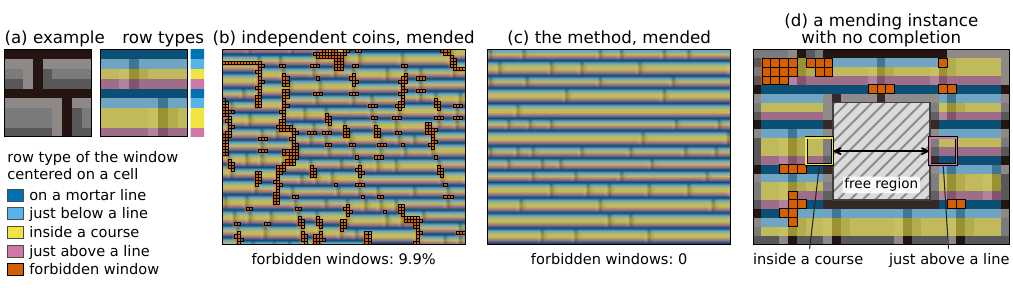}
\caption{The row type on Wall, a row's place relative to the horizontal mortar lines, which validity keeps constant
along every row. (a) The example and the row type of each of its rows. (b) The rule on independent coins after repair
and mending, each cell tinted by the row type of the window centered on it: the type changes along rows, and the
forbidden windows sit where it changes. (c) The method, given row types drawn from the seed, at the same coin seed:
every row keeps its type, and no window is forbidden. (d) A mending instance with no completion
(Appendix~\ref{app:witness}): of the cells freed around a forbidden window (hatched), two frozen patches on one row
carry row types that validity cannot join, a \emph{witness} (Appendix~\ref{app:pwitness}).}
\label{fig:quantity}\end{figure*}

A brick wall shows what independent coins, which give each cell randomness of its own, cannot do
(Figure~\ref{fig:quantity}). No window of the example lets a mortar row step up or end, so a row's place among the
mortar lines is a quantity that validity carries along the whole row. A local generator on independent coins decides it
at distant cells of a row independently, and wherever they disagree a forbidden window lies between them.
\citet{bergmann2017psgan} observed the same of textures: a generator on independent noise with a limited receptive
field cannot align distant regions. For any
quantity that validity carries along an axis and the statistics make vary, a generator of radius $r$ on independent
coins leaves forbidden windows at a density of order at least $1/r$ unless its statistics stray from the example's. At
any depth, that is, however far each cell reads, no such generator is both free of forbidden windows and faithful to the
statistics (Section~\ref{sec:theory}). This holds even
where one repeated color satisfies every constraint, so that the constraints alone are solvable in constant time
\citep{brandt2017lcl}. The order cannot be improved: the min-coin copy, which copies the example over regions drawn from
the coins, stays within a constant factor of the bound, and a rule learned on independent coins leaves forbidden windows
at about the copy's rate (Figure~\ref{fig:mincoin}).

The example alone shows whether the theorems apply. A search over the overlap graphs of the example's locally valid
patches finds quantities that validity carries, and we call an example with one \emph{certified}
(Section~\ref{sec:cert}). We call a generator valid on an example when, after mending, a local re-solve around each
forbidden window, at most one window in a thousand stays forbidden on each canvas. On \certResistantOf\ public
examples, which we call resistant, a rule learned on independent coins stays far from valid. The search certifies
\certResistantCount\ of them and none of the others. Where mending fails on a certified example, at the radii we
examined, the re-solve mostly holds a witness, two frozen patches whose values of the quantity no valid filling can join
(Figure~\ref{fig:quantity}(d)), and the min-coin copy, mended the same way, is valid on no certified example at radii
up to \tcLadderMax\ (Section~\ref{sec:resist}).

A shared input escapes the bound, and random access allows one. An invariant along rows is supplied as row types drawn
band by band from the seed with the example's own transitions, and a phase as the coordinates modulo its period, both at
bounded work per cell (Section~\ref{sec:cert}). Coins shared on blocks of bounded length help only as a larger radius
would, unless the generator ties the quantity to their grid (Proposition~\ref{prop:blocks}). The position alone also
escapes the bound but gives one world, the example tiled from absolute coordinates. A generator free of forbidden windows, at any depth, can vary the quantity between worlds only through randomness shared
along whole lines (Theorem~\ref{thm:shared}). Where the
example's rows glue, a condition decided by a finite computation on the example, that is also enough: a generator that
reads only its band's coin and its position leaves no forbidden window for any value of the coins
(Theorem~\ref{thm:glue}).

The rows glue on \glueHoldsWords\ of the \rowInvCountWords\ examples with a row invariant, all but 3Bricks. These
are the examples where the copier, which fills the sampled rows with stretches of the example's rows, exists and leaves
no forbidden window (Section~\ref{sec:cert}). With the copier, or a lookup of the example at
the coordinates modulo the period, the input carries validity by itself, without learning or mending, on
\inputAloneValidWords\ of the \certInputCountWords\ certified examples that receive one.

A learned local generator can use the input as well. A small neural cellular automaton
\citep{mordvintsev2020growing}, learned from the single example and given this input, is valid after mending on the
same \certInputMethodValidWords\ certified examples, all but \certMethodInvalidNames. On the octaves of seeded noise
alone, which share coins along whole rows as well as on blocks, it is valid on \certInputOctValidWords\ of the
\certInputCountWords, and on independent coins on none. On the \methodRobustCertCountWords\ certified examples where it
is valid on all \thmCanvasesWords\ canvases, no canvas keeps a forbidden window, while no generator on independent
coins with the same statistics could be free of them, at any depth on most of these examples and at the radius of the
whole pipeline on the rest (Section~\ref{sec:valid}). A convolutional diffusion model given the input is valid on
\ddThreeValidWords\ of the \certInputCountWords, \ddThreeCountedWords\ of them without copying the example
(Section~\ref{sec:diffusion}).

Outside the corpus on which we built the procedure, the automaton is valid on all \heldoutAllValidCountWords\ examples
we held out, \heldoutInputNeededCountWords\ of them only with the input. On \fxNExamples\ examples taken from other
implementations of Wave Function Collapse and related generators, the search again certifies only examples that resist
independent coins, and the automaton, given the input where a certificate exists and the octaves elsewhere, is valid on
\fxMethodValid\ of them, against \fxOctValidAll\ on the octaves alone (Section~\ref{sec:heldout}).

Where a copier exists, learning is not what makes the worlds valid or new. The automaton composes its rows instead of
copying them, and on Wall its canvases lie where Theorem~\ref{thm:nogo} excludes the law of every generator on
independent coins (Figure~\ref{fig:bound}), but a block solver that re-solves the copier's canvases does the same
without learning (Section~\ref{sec:learning}). Learning adds something of its own where nothing is certified, as the
prior from which a mender starts. From a rule learned on the octaves, one mender makes \nOctUncertValid\ of the
\tcUncertOf\ uncertified examples valid, more than from any prior without learning that we gave it, though a block
solver started from the tiled example needs no prior and is valid there by construction.

With the octaves where nothing is certified and the input elsewhere, the automaton, which we call the method, is valid
on \nMethodAll\ of the \nPublic\ public examples. Most of that validity comes with the mender, which rewrites a median
of \rewrittenMedianMethod\ of the cells where the method ends valid. Regions mended separately agree once each carries
a margin as wide as a mending change can travel, so the method keeps random access (Section~\ref{sec:access}).

Our contributions are the lower bound, its counterpart at any depth and a construction that attains its order
(Section~\ref{sec:theory}); a necessary condition on what a local generator free of forbidden windows must share to vary the quantity between
worlds, and a sufficient one decided on the example; the certificate and the input built from it, which keeps random access
(Section~\ref{sec:cert}); and a measured account of what a learned rule adds to that input (Section~\ref{sec:results}).
Appendix~\ref{app:record} gives, for most experiments, the outcome we expected before the run beside the outcome
observed.

%% file: sec_setting.tex
\section{Setting}
\label{sec:ra}

A world is a coloring $x:\Z^2\to[C]$ in $C$ colors, and one example decides which $N\times N$ windows it may contain and
how often. The coloring $x$ is \emph{$K$-valid} if every window lies in the set $K\subseteq[C]^{N\times N}$
of allowed windows, and a window outside $K$ is \emph{forbidden}. As in Wave Function Collapse's overlapping model
\citep{gumin2016wfc}, $K$ holds the windows of the example, read with wrap-around and without rotations or reflections,
and their frequencies $f$ are its \emph{window statistics}, so the plane tiled with the example is $K$-valid with
statistics exactly $f$. A coloring is measured by its density of forbidden windows and the total variation of its
window frequencies from $f$, forbidden windows pooled as one outcome. The examples are the \nSamples\ public
bitmap examples of the WaveFunctionCollapse repository, none chosen for this study, read at $N=3$. A
generator is \emph{valid} on an example when, after mending, at most one window in a thousand is forbidden on each
canvas of the protocol (Section~\ref{sec:setup}), a weaker demand than $K$-validity, which the theorems concern.

\emph{Coins} are uniform numbers hashed from the seed and the coordinates, shared over blocks as in seeded noise. A
\emph{scale pair} $a=(i,j)$, with $i,j\in\{0,1,2,\dots\}\cup\{\infty\}$, puts the cell $v=(x,y)$ in the block of
$v^{a}=(\lfloor x/2^{i}\rfloor,\lfloor y/2^{j}\rfloor)$, of $2^{i}\times2^{j}$ cells fixed in absolute coordinates
(rows, columns or the plane when a scale is $\infty$); for a finite set $\mathcal A$ of pairs, each block $w$ of each
$a\in\mathcal A$ carries independent coins $\omega_{a,w}$, and cell $v$ reads
$\omega(v)=(\omega_{a,v^{a}})_{a\in\mathcal A}$.

\begin{definition}[random-access generator]\label{def:ra}
A \emph{random-access generator of radius $r$ on the scale pairs $\mathcal A$} is a map from the coins to a coloring $X$
of the form $X_v=g\big((\omega(v+u))_{\|u\|_\infty\le r}\big)$ for one fixed function $g$. \emph{In the general form},
$g$ also reads $v\bmod p$, the position of $v$ modulo a fixed period $p$ in each coordinate, and $\omega(v)$ also lists
the coins of the block that holds $v$ in each of finitely many further lattices of blocks fixed in absolute coordinates,
every block with coins of its own.
\end{definition}

\begin{definition}[generator on independent coins]\label{def:indep}
When $\mathcal A=\{(0,0)\}$, every cell has coins of its own, the vectors $(\omega_v)_{v\in\Z^2}$ are i.i.d., and
$X_v=g\big((\omega_{v+u})_{\|u\|_\infty\le r}\big)$. Such a generator commutes with every translation, and $X_v$ is a
fixed function of the coins within $\ell_\infty$ distance $r$ of $v$. In the language of ergodic theory it is a block
factor of an i.i.d.\ process, that is, a finitary factor whose coding radius is bounded
\citep{holroyd2017finitary,grebik2023grids,berlow2025separating}.
\end{definition}

A random-access generator is \emph{local}: any region follows from the coins within distance $r$ of it at a fixed cost
per cell, so regions computed apart agree cell for cell, and editing a region's coins changes the output only within
that distance. The inputs of Section~\ref{sec:cert}, row types drawn band by band from the seed and the coordinates
modulo a period, fit the general form, and so does
the mended output, with a \emph{reach}, beyond which no coin changes a cell's color, in place of the radius
(Appendix~\ref{app:setting}).

%% file: sec_theory.tex
\section{Limits of local generators}
\label{sec:theory}

When validity carries a quantity along an axis and the example's statistics make it vary, no local generator on
independent coins is both free of forbidden windows and faithful to the statistics, at any depth; at radius $r$ a
faithful one's density of forbidden windows falls as $1/r$ at best, an order that is attained. A generator free of forbidden
windows that escapes these bounds carries the quantity as a function of its position and of randomness shared along the axis.

\paragraph{Quantities carried by validity.}
A $P\times P$ patch, $P\ge N$, is \emph{locally valid} if all its $N\times N$ windows lie in $K$, and $V_P$ is the set
of such patches. For an axis direction $e$, the \emph{overlap graph} $G_{P,e}$ on $V_P$ has an edge from $a$ to $b$ when
$b$ can follow $a$ one step along $e$, agreeing with it on their overlap. A \emph{transported quantity} is a map
$\varphi$ on $V_P$ with a bijection $\sigma$ of its values such that $\varphi(b)=\sigma(\varphi(a))$ on every edge. For
an \emph{invariant}, $\varphi$ is the weakly connected component and $\sigma$ the identity, so it cannot change along a
line parallel to $e$ without a forbidden window; along rows it is the \emph{row type} (Figure~\ref{fig:quantity}). For a
\emph{phase}, $\sigma$ adds one modulo $d$, where $d\ge2$ divides the signed length of every closed walk in a weak
component; in a checkerboard it is the column's parity. The \emph{full labeling} gives each locally valid patch its weak
component $c$ and its level modulo $d_c$, the greatest common divisor of these lengths in $c$
(Appendix~\ref{app:labeling}). A \emph{one-way quantity}, the strongly connected component of $G_{P,e}$, can move along
an edge to a component it reaches but never return.

\begin{theorem}[a floor from a transported quantity]\label{thm:floor}
Let $g$ be a generator on independent coins of radius $r$ (Definition~\ref{def:indep}) with output $X$, and let $\varphi$ be a
transported quantity along $e$ at patch size $P$. Let $p_s$ be the probability that the $P\times P$ patch of $X$ at a
cell is locally valid and has $\varphi=s$. Then the density $\varepsilon$ of forbidden windows of $X$ satisfies
\[
\varepsilon\;\ge\;\frac{1-\max_s p_s}{(2r+P+1)(P-N+1)^2}.
\]
\end{theorem}

The theorem binds every such generator, whatever its architecture or training, among them the learned rule of
Section~\ref{sec:method} on each cell's own coins alone, the \emph{independent-coin rule}. Patches $2r+P$ apart along
$e$ read disjoint coins, so their values of $\varphi$ agree after that many steps of $\sigma$ with probability at most
$\max_s p_s$, as validity between them would require (Appendix~\ref{app:pfloor}).

\paragraph{Defects against fidelity.}
The bound vanishes only if $\varphi$ is concentrated on one value, which the example's statistics forbid. When $P=N$, a
generator within total variation $\delta$ of $f$ has $p_s\le f(\varphi=s)+\delta$, and Theorem~\ref{thm:floor} becomes
an inequality between validity, fidelity and radius:
\begin{equation}\label{eq:vfr}
\varepsilon\,(2r+N+1)\;+\;\delta\;\ge\;\kappa_f,\qquad \kappa_f=1-\max_s f(\varphi=s).
\end{equation}
For any $P$, a linear program gives $\kappa(\delta)$, one minus the largest share of one value of $\varphi$ under a
stationary distribution on $V_P$ whose $N\times N$ marginal lies within $\delta$ of $f$, with $\kappa(0)=\kappa_f$ and
$\kappa(\delta)\ge\kappa_f-\delta$ when $P=N$ (Appendix~\ref{app:lp}).

\paragraph{Rare changes.}
A quantity that the example lets change, but rarely, still sets a floor at finite radius. Let $J\subset K$ be a set of
windows with total frequency $\theta=f(J)$ in the example, and let $\varphi$ be a transported quantity along $e$ at
$P=N$ with $K\setminus J$ in place of $K$, so that along a line it departs from $\sigma$ only at a window of $J$. On a
wall, a window that let a mortar row step up would be such a change.

\begin{corollary}[rare changes]\label{cor:slip}
Let $g$ be a generator on independent coins of radius $r$ whose output has a density $\varepsilon$ of forbidden windows
and window statistics within total variation $\delta$ of $f$. With
$\kappa_J=1-\max_s f(\{w\in K\setminus J:\varphi(w)=s\})$,
\[
(\varepsilon+\delta+\theta)\,(2r+N+1)\;\ge\;\kappa_J .
\]
\end{corollary}

The proof runs Theorem~\ref{thm:floor} on $K\setminus J$ in place of $K$, with the windows of $J$ counted as forbidden
(Appendix~\ref{app:pslip}); when $J$ is empty, the same proof gives inequality~\eqref{eq:vfr}. A generator with
$\varepsilon=\delta=0$ needs $2r+N+1\ge\kappa_J/\theta$, so a quantity that may change at a share $\theta$ of windows
requires coordination over distances of at least order $1/\theta$.

\paragraph{Unbounded radius.}
A \emph{factor of i.i.d.\ coins} is $X=F(\omega)$ for a measurable map $F$ that commutes with every translation of
$\Z^2$, such as a generator on independent coins or a rule run for any number of rounds.

\begin{theorem}[no generator driven by independent coins]\label{thm:nogo}
Let $X$ be a factor of i.i.d.\ coins that is $K$-valid almost surely, and let $\varphi$ be a transported or one-way
quantity along an axis direction $e$, under the full labeling of Appendix~\ref{app:proofs} or, for a one-way quantity,
under the labeling by strongly connected components. Then almost surely
$\varphi$ takes one and the same value at every cell, and that value lies in a component without a phase. Hence if
$\kappa(\delta)>0$, the window statistics of $X$ lie farther than $\delta$ from $f$; for $P=N$ their total variation
from $f$ is at least $\kappa_f$.
\end{theorem}

Its proof uses ergodicity in place of a radius, which makes the value that a line free of forbidden windows carries
almost surely one constant (Appendix~\ref{app:proofs}).

\begin{corollary}[convolutional diffusion samplers]\label{cor:diffusion}
Let $X$ be the output of a diffusion sampler run on the whole plane without padding, whose denoiser is convolutional
with a finite receptive field at each step and whose noise is drawn independently at every cell and step, however many
steps it takes. If $X$ is $K$-valid almost surely and $\kappa(\delta)>0$ for a transported quantity of the example, the
window statistics of $X$ lie farther than $\delta$ from $f$; for $P=N$ their total variation from $f$ is at least
$\kappa_f$.
\end{corollary}

Run without padding, the sampler commutes with every translation, so its output is a factor of i.i.d.\ coins and
Theorem~\ref{thm:nogo} applies; zero padding would let it read its position \citep{xu2021positional}.

\paragraph{Local algorithms.}
A $K$-valid coloring is a locally checkable labeling of the grid \citep{brandt2017lcl}. Some such labelings admit
generators on independent coins whose defects fall exponentially in the radius, while on agreement constraints such as
checkerboards none does better than order $1/r$ (Proposition~\ref{prop:classes} in Appendix~\ref{app:proofs}, after
\citet{brandt2017lcl} and \citet{grebik2023grids}). Both theorems rest on classical arguments for factors of i.i.d.\
processes and add a trade between defects, fidelity and radius that binds even constraints a constant coloring
satisfies.

\paragraph{Coins shared on blocks.}
The octaves of seeded noise share coins over blocks, which both theorems exclude.

\begin{proposition}[coins shared on blocks]\label{prop:blocks}
Let $X$ commute with the periods of a lattice of blocks fixed in absolute coordinates, and let $X_v$ be a function of
the independent coins within distance $r$ of $v$ and of coins attached to the blocks that meet that box. (a) If every
block has length at most $\ell$ along $e$, Theorem~\ref{thm:floor} holds with $2r+\ell$ in place of $2r$, where
$\varepsilon$ is the density of forbidden windows and $p_s$ is taken at the cell where it is largest. (b) With $K$ and
$f$ read from the example with wrap-around, copying the tiled example on every block of an $\ell\times\ell$ lattice,
under a uniform translation drawn from the block's coin, reads one coin per cell and gives $\varepsilon\le2(N-1)/\ell$,
with window statistics within $2(N-1)/\ell$ of $f$. (c) If the lattice has period $T$ along $e$ and $X_v$ may also read coins
shared along whole lines and over the plane, a floor of the same order holds on the components of the full labeling
whose period $d_c$ does not divide $T$, with a multiple of $T$ in place of $\ell$ (stated in full in Appendix~\ref{app:pblocks}).
\end{proposition}

A coin shared over a length $\ell$ along $e$ thus adds about $\ell/2$ to the radius while the quantity stays spread at
every cell, and one shared along a whole line escapes (a). Part (c) binds a rule that fixes the quantity by the position
that blocks fixed in absolute coordinates reveal.

\paragraph{The floor's order is attained.}
The \emph{min-coin copy} copies the example as (b) does, over regions drawn from independent coins: every cell $v$ draws
a priority, uniform on $[0,1]$, and a translation $t_v$ of the tiled example $Y$, uniform over one period, and copies
the translate drawn by the cell $m_v$ of lowest priority within distance $r$ of $v$: $X_v=Y(v-m_v+t_{m_v})$. Forbidden
windows occur only where the regions of cells that share $m_v$, each a copy of the tiled example, meet.

\begin{figure}[t]\centering\includegraphics[width=\linewidth]{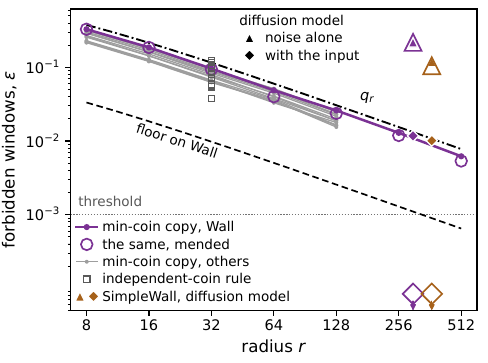}
\caption{The min-coin copy's share $\varepsilon$ of forbidden windows before mending against its radius $r$: gray lines for
the other certified examples with a transported quantity, the purple line for Wall, open circles for Wall after
mending. Every line stays below the bound $q_r$ of Proposition~\ref{prop:mincoin} (dash-dotted) and falls as $1/r$, on
Wall at about \tcWallKappaRatioAbout\ times the floor that Theorem~\ref{thm:floor} sets for a generator with the
example's statistics (dashed); the dotted line is the threshold of validity. Squares mark the independent-coin rule
before repair and mending, at its own radius $r=\tcRuleRadius$. Triangles mark the diffusion model of
Section~\ref{sec:diffusion} from noise alone and diamonds the same model given the certificate's input, at its radius
$r=\ddRadius$, filled before mending and open after (purple for Wall, brown for SimpleWall; an arrow means that mending left no forbidden window). Of the model's markers only the triangles come
from a generator on independent coins, which the floor binds; the diamonds read an input shared along whole rows.}
\label{fig:mincoin}\end{figure}
\begin{proposition}[the floor's order is attained]\label{prop:mincoin}
Let $N\ge2$, and read $K$ and $f$ from the example with wrap-around, as in Section~\ref{sec:ra}. For every
radius $r\ge N$, the min-coin copy is a generator on independent coins of radius $r$, and its density of forbidden
windows $\varepsilon$ and the total variation $\delta$ of its window statistics from $f$ satisfy
\[
\begin{aligned}
&\varepsilon\;\le\;q_r,\qquad \delta\;\le\;q_r,\qquad\text{where}\\
&q_r\;=\;1-\Bigl(\frac{2r+2-N}{2r+N}\Bigr)^{2}\;<\;\frac{4(N-1)}{2r+N}.
\end{aligned}
\]
\end{proposition}

Here $q_r$ is the probability that the cells of a window do not all copy from one cell; when they do, the window is one
of the tiling at a uniform translation (Appendix~\ref{app:mincoin}). Inequality~\eqref{eq:vfr} sets the copy, whose
statistics lie within $q_r$ of $f$, a floor of $(\kappa_f-q_r)/(2r+N+1)$, and once $4(N-1)<\kappa_f(2r+N)$ the copy
stays within a multiple of that floor that falls toward $4(N-1)/\kappa_f$, \wallTightRatio\ on Wall. The order $1/r$
of Theorem~\ref{thm:floor} therefore cannot be improved, only its constant. In Figure~\ref{fig:mincoin} the copy sits
at up to about \tcKappaRatioHiAbout\ times the floor on the certified examples with a transported quantity
(Appendix~\ref{app:mincoinrun}), and the independent-coin rule lies within a factor of two of the copy at its own
radius on \tcRuleOverCopyWithinTwoOf\ of them. The diffusion model from noise lies far above the copy at its own radius,
so its radius alone does not explain its rate (Appendix~\ref{app:diffusion}).

\paragraph{What a generator must share.}
For a generator in the general form of Definition~\ref{def:ra} and an axis direction $e$, the coins \emph{shared along
$e$}, $\mathcal G_e$, are those of its blocks that are unbounded along $e$: whole lines parallel to $e$, groups of such
lines, and the plane. Let $\Lambda$ be the lattice of translations in multiples of the period $p$ that map every lattice
of blocks to itself, and $T$ its period along $e$.

\begin{theorem}[what a generator free of forbidden windows must share]\label{thm:shared}
Let $X$ be a random-access generator in the general form, or any measurable function of its coins and of the position
modulo $p$ that commutes with the translations in $\Lambda$, and let $\varphi$ be a transported or one-way quantity
along an axis direction $e$. (a) If $X$ is $K$-valid almost surely, then at every cell the value of $\varphi$ is almost
surely a function of $\mathcal G_e$ and of the cell, and along $e$ this function repeats with period $T$. (b) If
$\varphi$ is transported and $X$ has radius $r$, the floor of Theorem~\ref{thm:floor} holds given $\mathcal G_e$, with
the largest length along $e$ of the other blocks it reads added to $2r$ (stated in full in Appendix~\ref{app:cvfloor}).
\end{theorem}

Part (a) is Theorem~\ref{thm:nogo} run given $\mathcal G_e$: translation by $Te$ fixes those coins and moves the others,
which stay independent, and no assumption on the statistics enters. On finite grids the same dichotomy separates
shared from private randomness \citep{balliu2025shared}. Part (b) is Theorem~\ref{thm:floor} run given
$\mathcal G_e$ and contains it and Proposition~\ref{prop:blocks}(a), so a shared input lowers the floor only through
what it reveals about the quantity (Appendix~\ref{app:converse}).

Without coins shared along $e$, a generator free of forbidden windows thus lays the quantity out the same
way in almost every world, even if it reads its position. Since the layout repeats with period $T$, a phase whose period
does not divide $T$ cannot be carried without defects at any depth (Corollary~\ref{cor:rigid} in
Appendix~\ref{app:converse}). Functions of the coordinates, such as the example tiled from absolute coordinates, satisfy
(a) but repeat one world. For the layout of the quantity to differ between worlds, a generator free of forbidden windows therefore needs an
input shared along the quantity's direction that, with the position, fixes the quantity at every cell: for an invariant, values shared along whole lines; for a phase, a period
$T$ that is a multiple of its own and a shared coin that shifts it. Section~\ref{sec:cert} draws an allowed sequence of
row types band by band from the seed, which a rule on coins shared along whole rows would have to do itself
(Appendix~\ref{app:rowseq}).

The same transport limits editing a valid world. If two $K$-valid colorings agree outside a finite region, a
transported quantity takes the same value on both at every cell, since along each line parallel to $e$ it is carried in
from outside. A local edit therefore keeps every row's type, and changing a row's type means editing every patch along
that row, in both directions without end. This is why block solvers over a fixed background keep its rows.

%% file: sec_cert.tex
\section{Finding the quantity and supplying it}
\label{sec:cert}

The example alone shows what a local generator must be given: a search of its locally valid patches finds the quantity
that validity carries, before anything is trained, and its kind names an input that supplies it and keeps random
access.

\paragraph{The search.}
For each axis direction $e$, and for the two diagonals, the search builds $G_{P,e}$ for $P$ from $N$ upward until the number of locally valid
patches passes \certCap\ (Appendix~\ref{app:search}). It reports an invariant when two weak components hold patches of
the example, a phase when a weak component $c$ that holds some of them has $d_c\ge2$, and a one-way quantity when two
strongly connected components hold them. A quantity found this way, together with $\kappa(0)>0$, is a
\emph{certificate}: it proves a floor for every generator on independent coins whose statistics lie within $\delta$ of
the example's whenever $\kappa(\delta)>0$ (Section~\ref{sec:theory}), and an example with one is \emph{certified}. The
search is one-sided: an uncertified example may have a floor that a larger patch or another kind of quantity would
reveal, and a certified one a second quantity, as SmileCity does (Appendix~\ref{app:smile}).

\paragraph{The input.}
The kind of the example's first certificate, under a fixed tie-break (Appendix~\ref{app:procedure}), decides what the rule reads besides the octaves:
\begin{itemize}
\item an invariant: the type of each line along its axis, from the row-type sampler;
\item a phase of period $d$: the phase channels of period $d$;
\item a one-way quantity or an invariant along a diagonal, which have no input of their own, or no certificate at all:
the octaves alone.
\end{itemize}

\paragraph{The row-type sampler.}
Along rows, where every certified invariant of the corpus runs, the type of row $y$ is the certificate's label of the
strip of $P$ rows from $y-\lfloor P/2\rfloor$. The sampler's
states are the pairs of consecutive types and its steps the triples that the example shows. It cuts the rows into bands
of $L$ rows starting at the multiples of $L$, fixes the state at the first row of each band to an \emph{anchor}, and
draws the other states from the example's transition frequencies, conditioned on returning to the anchor at the band's
end, so the anchor falls on the same rows in every world, a constraint that an offset hashed from the seed would lift
(Section~\ref{sec:limits}). The draw reads only
the seed and the band's index, so the state of any row costs work bounded by $L$, and the rule receives it as one-hot
channels (Appendix~\ref{app:procedure}).

\paragraph{When the row types are enough.}
On most examples with a row invariant, the row types and the position are enough to vary the quantity
between worlds with no forbidden window. Call a sequence
of $L+P-1$ rows of the example, each at some horizontal shift, a \emph{realization} of the types drawn for a band
if all its windows lie in $K$ and its strips of $P$ rows give the band's rows those types (Appendix~\ref{app:cvglue}).
The example's rows \emph{glue} if there is a strip $Z$ of $P-1$ such rows for which every type sequence that the
sampler can draw for a band has a realization that begins and ends with $Z$.

\begin{theorem}[when the row types are enough]\label{thm:glue}
Let the example carry an invariant along rows, with the row-type sampler drawing bands of $L\ge P-1$ rows and $P\le5$,
as on every example here, and let its rows glue with the strip $Z$. Let a generator fill the rows of each band with the
first $L$ rows of a realization of the band's types that begins and ends with $Z$, chosen by a fixed rule from the coin
from which the sampler draws the band's types. It is a random-access generator in the general form that reads only that
coin and its position modulo a period, and its output is $K$-valid for every value of the coins.
\end{theorem}

Two consecutive bands both carry $Z$ on the rows where they meet, so every window of the output lies within one
realization (Appendix~\ref{app:converse}). The copier, which also fills each band with stretches of the example's rows,
chains its offsets toward the next band's in place of a fixed strip, so it is valid by construction, not by the
theorem (Section~\ref{sec:valid}). Gluing is decided by a finite computation on the example (Appendix~\ref{app:cvdecide}). The rows glue on every example with a row invariant but 3Bricks, and these are the
examples where the copier exists; on them the computation came after the copier's results and explains them. On the \glOneCountWords\ held-out walls it came first, and there too the rows glue
and the copier leaves no forbidden window.

\paragraph{The phase channels.}
For a phase of period $d$ the rule receives one-hot channels of $x\bmod d$ and of $y\bmod d$ in absolute coordinates.
As functions of the coordinates, they keep random access and exact seams.

\paragraph{The octaves.}
Besides its own coins, every cell reads coins of every scale pair $(i,j)$ with $i,j\in\{\coinScaleList,\infty\}$ other
than $(0,0)$. Their blocks run from squares of side \coinSquareSides\ to whole rows, columns and the plane, the finite ones nested
in one lattice of period $T=\coinPeriod$, and where nothing is certified they give coarse structure, as octave noise
does for terrain.

The failures of \shapedCountWords\ certified examples, \shapedNames, shaped parts of this procedure
(Appendix~\ref{app:procedure}); the other certified examples test it within the corpus, the held-out ones as it stood
when they ran, and those from other implementations as it stands (Section~\ref{sec:heldout}).

%% file: sec_generators.tex
\section{Generators and protocol}
\label{sec:generators}
\label{sec:method}
\label{sec:setup}

The tiled example is valid and has the example's window statistics exactly, so validity and fidelity alone cannot show
that a generator does anything. We compare the learned generators with generators that show what copying and mending
achieve without learning (Table~\ref{tab:generators}), and measure copying and variation by normalized agreement with
the tiled example, repetition of its patches and aligned variation between seeds, except where every valid world is a
translate of the tiling (Appendix~\ref{app:measures}).

\begin{table}[t]\centering\scriptsize\setlength{\tabcolsep}{4pt}\def\rrcell{\raggedright\arraybackslash\hspace{0pt}}%
\caption{The generators without learning that the text names (Appendix~\ref{app:references} defines every
generator). All but the block reference and Wave Function Collapse are random access, after mending too by the margin
of Section~\ref{sec:access}. Shifted replay, the block reference, the random-access block solver and Wave Function
Collapse skip the mender.}
\label{tab:generators}
\begin{tabular}{@{}>{\rrcell}p{0.25\linewidth}>{\rrcell}p{0.71\linewidth}@{}}
\toprule
generator & role\\
\midrule
the min-coin copy & on each cell's own coins, the floor's order (Proposition~\ref{prop:mincoin})\\
shifted replay & the tiled example at one offset drawn from the seed: valid and faithful, one world up to
translation\\
the i.i.d.\ draw & each cell drawn from its own coins: copies nothing; what mending makes from noise\\
block-shifted tiles & one offset per block of a fixed lattice: copies with defects on block boundaries\\
the copier & the row types filled with the example's rows, one coin per stretch: the input alone\\
the Metropolis filler & written by hand, on the row types and hashed coins: composition on the input\\
the block reference & shifted replay re-solved on blocks placed at random, after \citet{boris2021infinite} and
\citet{kleineberg2023infinite}: no random access\\
the random-access block solver & the same on a lattice fixed in absolute coordinates: random access, valid by
construction\\
the re-solved copier & the copier's canvas re-solved on that lattice: composition on the input, valid by
construction\\
Wave Function Collapse & global search over the whole canvas: the reference for window statistics\\
\bottomrule
\end{tabular}
\end{table}

\begin{figure*}[t]\centering\includegraphics[width=\textwidth]{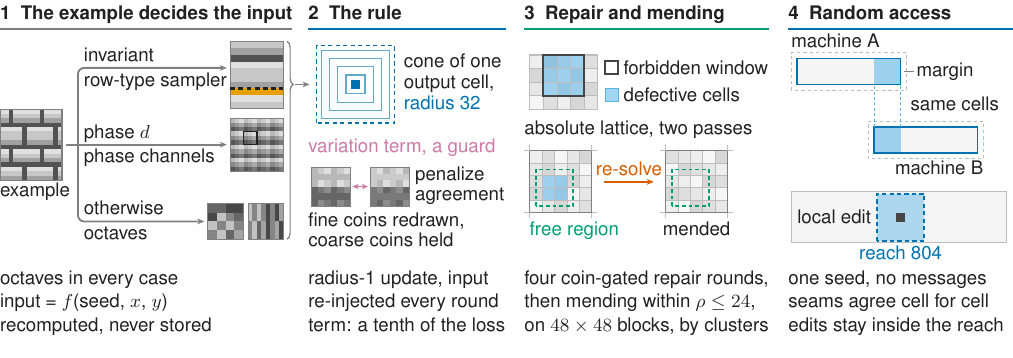}
\caption{The method. (1) The certificate decides the input: row types drawn band by band for an invariant along rows
(dashed: a band boundary; orange: the anchor), the coordinates modulo $d$ for a phase of period $d$, and in every case
the octaves. (2) The rule, a radius-one update run for \mainR\ rounds and trained with a variation term against copying
the example on coarse blocks. (3) Repair recolors defective cells, those covered by a forbidden window, and mending
re-solves the free region around each remaining defect, block by block on a lattice fixed in absolute coordinates.
(4) Regions mended separately agree once each carries a margin that covers how far a mending change can travel, and a
local edit, which overrides the coins read in a region, changes no cell farther from that region than the pipeline's
reach, the distance shown.}
\label{fig:schematic}\end{figure*}

\paragraph{The learned automaton.}
The method's rule is a neural cellular automaton \citep{mordvintsev2020growing} learned from the single example
(Figure~\ref{fig:schematic}): a radius-one update run without padding for \mainR\ rounds, so a local generator of
radius \mainR\ on the input of Section~\ref{sec:cert}. Training asks every round for allowed windows with the example's
frequencies, and a variation term penalizes agreement between two runs that differ only in their fine coins, against the
shortcut of Proposition~\ref{prop:blocks}(b); rules trained without it do not copy either
(Appendix~\ref{app:sharing}). Then \repairRoundsWords\ repair rounds recolor cells under forbidden windows with coins
hashed from the seed and the position (Appendix~\ref{app:method}). The \emph{octave rule} is the same construction on
the octaves alone; the independent-coin rule, which reads no shared coins and so cannot take that shortcut, was trained
before we added the variation term and has none.

\paragraph{The diffusion model.}
A convolutional diffusion model is trained on each example by Gaussian diffusion on the one-hot colors
\citep{ho2020ddpm}, with the standard loss and no term for forbidden windows. It samples with one step of its unpadded
denoiser at each of \ddLevels\ noise levels, under noise hashed from the seed, step and coordinates, so a cell depends
on independent noise within \ddRadius\ cells. It is trained from two training seeds, from noise alone on the certified
examples and on controls, and with the input of Section~\ref{sec:cert}, without the octaves, wherever a certificate
names one (Appendix~\ref{app:diffusion}).

\paragraph{Mending.}
Mending re-solves what repair leaves in two passes over a lattice of $\mendBlockMain\times\mendBlockMain$ blocks fixed
in absolute coordinates. It frees the cells around each defect, out to the \emph{mending radius} \mendRadiusMain, and
re-solves them with Wave Function Collapse's propagation, cluster by cluster where a block's defects cannot be completed
together (Appendix~\ref{app:mender}). It thus rewrites part of the output of its \emph{prior}, the generator it starts
from (Section~\ref{sec:learning}). No coin farther than \certRadiusCC\ cells from a cell can change its mended color,
the \emph{reach} of the whole pipeline: the rule's radius with its repair rounds, plus the \mendBoundCC\ cells that a
mending change can travel.

\paragraph{Protocol.}
Validity is judged on three $256\times256$ canvases per example, drawn on coin seeds that every generator shares, the
\emph{development seeds}. A robustness test adds five canvases on the \emph{robustness seeds}, and a last test three on
\emph{fresh seeds} that no decision used; the rules counted are trained from training seed 0
(Table~\ref{tab:protocols}). The examples fall into three classes by what mending within radius \mendRadiusLo, without cluster completion, makes of
the independent-coin rule on one canvas: it leaves at most one window in a thousand forbidden on the \emph{mendable}
examples, at most one in a hundred on the \emph{partial} ones, and more on the \emph{resistant} ones, which
\emph{resist independent coins}. Every certified example lies far from these boundaries, with at least \certClassResidMin\ of its windows
forbidden.

%% file: sec_evidence.tex
\section{Results}
\label{sec:results}

\subsection{Independent coins fail where the certificate finds a quantity}
\label{sec:resist}

The certificate picks out the examples on which independent coins fail. It certifies \certResistantCount\ of the
\certResistantOf\ examples that resist independent coins (Section~\ref{sec:setup}, Table~\ref{tab:cert}) and none of the
\certOthersOf\ others (Figure~\ref{fig:corpus}). Against the \certMendedOf\ examples that independent coins mend, leaving
out the partial ones, a one-sided Fisher exact test gives $p=$~\certFisherPAll\ if the examples are counted as
independent, which variants of one picture are not (Appendix~\ref{app:evidence}). On
examples from other implementations, which played no part in designing the search, it again certifies only examples
that resist (Section~\ref{sec:heldout}).

Once mended, the rule on independent coins is valid on most uncertified examples, and on no certified one at any
mending radius we tried (Appendix~\ref{app:mendall}). The theorems explain this failure without forcing it at our threshold:
mending gives a generator on independent coins the pipeline's reach as its radius (Proposition~\ref{prop:blocks}(a)),
and at that reach the floor for a generator with the example's statistics lies below the threshold on every certified
example (Appendix~\ref{app:scopefloors}).

The failure is not peculiar to learning. The min-coin copy, which attains the floor's order, is valid after mending on none of the certified examples at radii up to \tcLadderMax, while at radius \tcContrastRadius\ it is already valid on
\tcUncertValidSixtyFour\ examples that do not resist (Appendix~\ref{app:mincoinrun}). On Wall it is still invalid at radius \tcRadiusMax, and if its rate kept falling as $1/r$ it would reach the threshold
only near radius \tcWallPassRadius.

Sampled from noise alone, the diffusion model fails on every certified example but also on most of the uncertified
controls that the independent-coin rule makes valid, so at its training budget it cannot test the certificate's
selectivity (Table~\ref{tab:diffusion}).

Mending stalls on the certified examples for the reason the theorems give. At the radii we examined, most instances it
cannot complete hold a witness, two frozen patches on one line whose values of the quantity no valid filling can join
(Figure~\ref{fig:quantity}(d), Table~\ref{tab:witness}).

The \certUncertifiedResistantWords\ resistant examples without a certificate show no quantity carried along a line. They stay
uncertified when the search widens to larger patches, knight-move directions and tall strips (Appendix~\ref{app:cert}).
In the worlds of Wave Function Collapse, a global solver, the dependence between $3\times3$ windows persists to
\lrDistMax\ cells along the quantity's direction on every certified example but \lrCertMissName, and along no direction
on the \certUncertifiedResistantWords\ without one (Appendix~\ref{app:longrange}). Whatever constrains them is not a quantity carried along a line, at least
not one visible at the patch sizes and distances we measured.

\subsection{Local generators given the certificate's input}
\label{sec:valid}
\label{sec:alone}
\label{sec:mending}

\begin{figure*}[t]\centering\includegraphics[width=\textwidth]{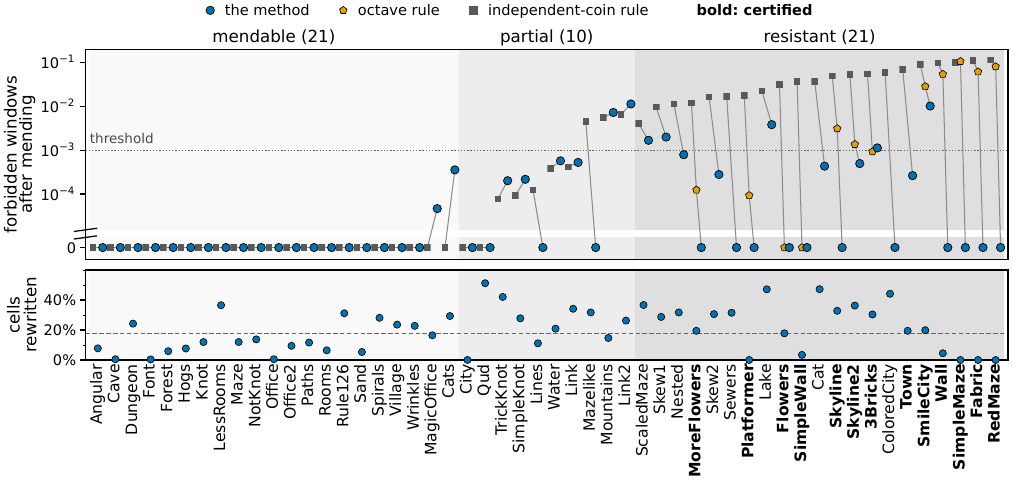}
\caption{Every public example after mending: the share of forbidden windows on the worst of three canvases for the
independent-coin rule (square), the octave rule (pentagon) and the method (circle); on Town and the uncertified examples
the method is the octave rule, and only its circle is drawn. The dotted line marks validity, one forbidden window in a
thousand; exact zeros sit below the break in the axis. The lower row gives the share of the method's cells that mending
rewrites, and the dashed line its median over the examples the method makes valid. The examples are grouped by class
(Section~\ref{sec:setup}) and ordered within each by the independent-coin rule; names in bold carry a certificate.}
\label{fig:corpus}\end{figure*}

\begin{figure}[t]\centering\includegraphics[width=\linewidth]{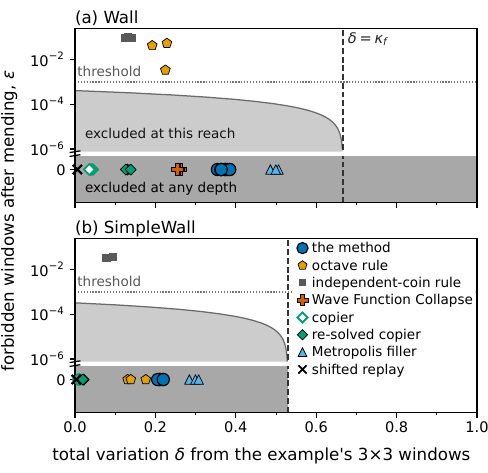}
\caption{Wall and SimpleWall against the bounds. Each point is one canvas mended within radius \mendRadiusMain, at the
total variation $\delta$ of its $3\times3$ window statistics from the example's and its share $\varepsilon$ of forbidden
windows; exact zeros sit below the break in the axis, and the dotted line is the threshold of validity.
Inequality~\eqref{eq:vfr} excludes the light region for every generator on independent coins at the pipeline's reach,
$r=\certRadiusCC$, and Theorem~\ref{thm:nogo} the dark band at any depth: the zero level left of $\delta=\kappa_f$ on
SimpleWall, and all of it on Wall, where the program of Section~\ref{sec:theory} stays positive at every $\delta$. On
Wall the method's canvases sit in the dark band, as do those of both copiers, the Metropolis filler and shifted replay,
which read inputs shared along whole rows or over the plane, and of Wave Function Collapse, a global solver; the
independent-coin and octave rules lie above the threshold. On SimpleWall the octave rule, whose octaves include coins
shared along whole rows, has no forbidden window either.}
\label{fig:bound}\end{figure}

The input often carries validity by itself, on \inputAloneValidWords\ of the \certInputCountWords\ certified examples
that receive one. The copier, which exists where the rows glue (Section~\ref{sec:cert}), leaves no forbidden window on
any of its \copierCountWords\ examples even before mending (Table~\ref{tab:copier}). Where the example is one period
of its phase, as on \phaseLookupNames, reading it at the coordinates the phase channels give yields the tiled example;
on SmileCity, whose example spans several periods, it does not (Appendix~\ref{app:smile}).

Given the input, the method is valid on \nCertMethodAll\ of the \certResistantCount\ certified examples, all of
which defeat the rule on independent coins (Figure~\ref{fig:corpus}). The octaves, which include coins shared along whole rows, already give \octCertValidWords\ of them; the input adds \certGainedOverOctWords, Wall among them, and loses
\certLostToOctNames. On Fabric, by Proposition~\ref{prop:blocks}(c), every rule on the octaves that matches the
example's statistics has a floor above the threshold before mending, though not after (Appendix~\ref{app:scopefloors}),
while given the phase channels the rule leaves nothing to mend. With the octaves where nothing is certified, the method
is valid on \nMethodAll\ of the \nPublic\ public examples, against \nCCFlat\ for the rule on independent coins.

On most of the certified examples the theorems set the method's worlds beyond the reach of independent coins. On \methodRobustCertCountWords\ certified
examples no canvas of the method keeps a forbidden window (Table~\ref{tab:thm}, Figure~\ref{fig:bound}). On Wall and
\thmLPOtherCountWords\ others the program of Section~\ref{sec:theory} stays positive at the canvases' largest distance from the
example's statistics, so no generator on independent coins free of forbidden windows comes as close at any depth
(Theorem~\ref{thm:nogo}); on \phaseLookupNames\ the method's world is the tiled example, which no such generator reaches
in any case. On the other \thmSpreadCountWords\ the statistics lie too far for Theorem~\ref{thm:nogo}. There Theorem~\ref{thm:floor}
still requires a generator on independent coins whose most common row type is no more common than in the method's
worlds to leave at least \thmSpreadWindowsMinWords\ forbidden windows per canvas in expectation, but at the pipeline's
reach it forces each one only somewhere along a stretch longer than a canvas is wide, so a clean canvas is weak
evidence there.

\phantomsection\label{sec:heldout}The method holds up on examples that shaped nothing in the procedure. It is valid on every
certified example whose failures played no part in designing it (Section~\ref{sec:cert}) but \notShapedInvalidNames,
and on all
\heldoutAllValidCountWords\ examples we drew or transposed for the test: three walls, two of them transposed so that their invariant runs along columns, and two mazes, one certified. On transposed Wall and on the certified maze the octaves
alone fail (Appendix~\ref{app:heldout}). It leaves \certMethodInvalidNames\ invalid: on SmileCity the rule mixes two
lattices of faces along the rows, which no finite patch certifies (Appendix~\ref{app:smile}), and on 3Bricks \threeBricksOverWords\ of its three canvases stays just above the threshold (Appendix~\ref{sec:open}).

Examples that we neither drew nor tuned against test the procedure from further outside. We ran it unchanged on \fxNExamples\ example images from
\fxNReposWords\ other open-source implementations of Wave Function Collapse and related generators
(Appendix~\ref{app:freshex}), and the method is valid on \fxMethodValid\ of them, more than the octave rule or
independent coins (Table~\ref{tab:fresh}). The certified examples carry the difference: the method is valid on all
\fxNInputWords\ that receive an input and the octave rule on \fxOctValidInputWords, though on \fxNTiledWords\ of them,
each one period of a pattern, the method's world is the tiled example. Where nothing is certified the method is the
octave rule, which does about as well there as tiles of the example shifted block by block under the same mender.

%% file: sec_results.tex
\subsection{What a learned rule adds}
\label{sec:learning}
\label{sec:replay}

What does a learned rule add to an input that already carries validity? Where a copier exists, the input alone makes
the worlds valid, and a block solver without learning, run on the copier's canvases, makes them new as well. The copier
is valid on all \copierCountWords\ examples where it exists (Section~\ref{sec:valid}) but copies: every row of its
canvases is a row of the example up to a horizontal shift (Table~\ref{tab:copier}). The rule composes its rows
instead: on Wall none of the rows inside its courses is a row of the example (Table~\ref{tab:wallbricks}), and on
every example with a row invariant none of its $16\times16$ patches that hold no forbidden window before mending occurs
in the tiled example (Table~\ref{tab:clean}). The composition is therefore the rule's own, and on Flowers,
MoreFlowers and Skyline, where mending touches almost every patch, the mender's as well.

A block solver without learning composes as well once it starts from the copier's canvases. The re-solved copier
re-solves every block of such a canvas in two passes on the random-access block solver's lattice. It starts from a
canvas without forbidden windows, and each re-solve keeps every window it touches allowed and always has a completion,
the block's current content, so it leaves no forbidden window on any seed. As in the rule's worlds,
\crMultiRowWords\ of its rows with more than one color is a row of the example, and its statistics lie closer to the
example's than the rule's on all but \crTvAboveRuleNames\ (Table~\ref{tab:copier}).

The layout of rows is the sampler's, which draws a new sequence of row types from each seed. Every filler that
reads it follows it, while the block solvers run from the tiled example keep the tiling's sequence on every seed, since a block
re-solved under frozen surroundings cannot change a row's type (Appendix~\ref{app:replay}). On Wall none of the
method's canvases has the tiling's sequence and no two agree up to a shift; only the rows around the band anchors
share their type on every canvas (Figure~\ref{fig:layout}). The method's normalized
agreement with the tiling is \wallMethodNrep, where an i.i.d.\ draw scores zero and the tiling one, against
\wallRABSNrep\ for the random-access block solver (Table~\ref{tab:wallvar}).

On Wall the rule's statistics lie farther from the example's than either copier's, mostly because its bricks run
longer (Table~\ref{tab:wallbricks}). A sampler written by hand composes less well: the Metropolis filler, which
composes on the row types and reads far beyond the rule's radius, is valid on \mhTypedValidCountWords\ of the examples with a row invariant, and the rule on
\ruleCourseValidWords\ (Appendix~\ref{app:copier}).

Learning adds something of its own where nothing is certified, as the prior from which one mender starts. From the octave rule the mender makes \nOctUncertValid\ of the \nUncertOf\ uncertified examples valid, and from any
prior without learning that we gave it no more than \tileCCValid, the count of block-shifted tiles
(Appendices~\ref{app:floors} and~\ref{app:mincoinrun}). On the examples from other implementations the octave rule
leads block-shifted tiles where nothing is certified only \fxOctOverTilePhrase\ (Section~\ref{sec:heldout}). The block
solvers need no prior and are valid on the uncertified examples by construction, and where the octave rule is valid at
mending radius \mendRadiusLo, mostly on uncertified examples, the random-access block solver varies about as much as
the rule and keeps a little more of the tiling's layout (Appendix~\ref{app:replay}).

Most of the rule's validity comes with the mender. Before mending the method is valid on \rawValidMethodAll\ of the
\nPublic\ examples, \rawValidMethodAllNames. On the examples it makes valid, mending rewrites a median of
\rewrittenMedianMethod\ of the cells (Figure~\ref{fig:corpus}), and almost every rewritten cell would make a window
forbidden if its old color returned (Appendix~\ref{app:rulemap}).

\subsection{Random access after mending}
\label{sec:access}

Mending keeps random access once each chunk carries a margin as wide as a mending change can travel, at most
\mendBoundCC\ cells. Every input channel is a function of the seed and the coordinates, so chunks agree before mending,
and the mender seeds every solve by the seed, its block's absolute coordinates and the cluster's index
(Appendix~\ref{app:mender}). A narrower margin sufficed on every canvas we tested: at \seamAllMargin\ cells each agrees cell for cell with its
separately mended chunks (Appendix~\ref{app:seam}). On \sxCanvasesWords\ canvases \sxCanvasWide\ cells wide, whose
edge lies beyond a mending change's reach of the chunk, so that the chunk sees the values of the unbounded world, the
agreement holds at each margin we tried, from \seamAllMargin\ to \mendBoundCC\ cells. At \seamAllMargin\ cells a chunk of $256^2$ cells generates and mends \chunkAreaMarginAll\ times its own area, and \chunkAreaReachCC\
times at the margin the construction guarantees; the ratio falls as chunks grow.

A mender of longer reach makes more examples valid, at a wider margin. Mending within radius \mwRadiusA\ instead of \mendRadiusMain\ widens the guaranteed margin from \mendBoundCC\ to
\mwBoundA\ cells and adds \mwValidANames, valid on all \mwSeedsWords\ canvases and without copying, for \mwCountA\ of
the \nPublic\ public examples. A mender close to a global solver also makes \mwValidBNewNames\ valid, with window
statistics about \mwTvRatioBNew\ times as far from the example's, and still leaves \mwInvalidBNames\ invalid
(Appendix~\ref{app:mwide}).

%% file: sec_diffusion.tex
\subsection{A diffusion model given the input}
\label{sec:diffusion}

Does the input help a learned generator of another kind? It does for the convolutional diffusion model of
Section~\ref{sec:generators}. From noise alone, where Theorem~\ref{thm:floor} binds it at its radius and
Corollary~\ref{cor:diffusion} at any number of steps, the model is valid on none of the \ddTransWords\ certified
examples that receive an input (Section~\ref{sec:resist}). Given the input, it is valid after mending on
\ddThreeValidWords\ of them at both training seeds. It follows the input: on all \ddFourTenfoldWords\ the
quantity switches value along its direction at least \ddFourFactorWords\ times less often than from noise alone, and on
Wall its share of forbidden windows before mending falls from \ddWallRawA\ to \ddWallRawB, near the method's.

Where the model is valid and its worlds can differ from the tiling, it does not copy. On \ddThreeEscapeNames, whose
valid worlds are essentially translates of the tiling (Appendix~\ref{app:rigid}), its world is the tiled example, as the method's is. On the other \ddThreeCountedWords\ it
makes valid, few of its $16\times16$ patches occur in the tiled example and none of its rows with more than one color
is a row of the example, while every row of the copier's canvases is (Table~\ref{tab:diffusion}). It falls short of
the method only on \ddThreeInvalidMethodValidNames, where it leaves about \ddSkyRawOverMethodAboutWords\ times as many
forbidden windows before mending as the method and stays invalid after mending (Appendix~\ref{app:diffusion}).

%% file: sec_related.tex
\section{Related work}
\label{sec:related}

\paragraph{Constrained content.}
Wave Function Collapse \citep{gumin2016wfc} and model synthesis \citep{merrell2009model,merrell2011model} are read as
constraint solving \citep{karth2017wfc} and extended with a hierarchy \citep{nie2023nested} or with learned
constraints or orders of collapse \citep{bhaumik2026learning,yiu2025markovian}, and neural cellular automata enforce
the constraints with a mixed-integer program \citep{zhang2023scalable}, all by global search; our generators are local.

\paragraph{Random access.}
Non-parametric texture synthesis \citep{efros1999texture} is sequential. Wang tiles
\citep{cohen2003wang,wei2004tile,lagae2006corner}, by-example noise \citep{heitz2018noise} and parallel synthesis
\citep{lefebvre2005parallel} compute any part of an unbounded texture on its own, and InfinityGAN
\citep{lin2022infinitygan} and InfiniteDiffusion \citep{goslin2026infinite} learn random access without hard
constraints. Constrained worlds are grown in the order of exploration \citep{kleineberg2019city} or re-solved
block by block \citep{boris2021infinite,kleineberg2023infinite}, after model synthesis \citep{merrell2009model}, over a
known-good background that \citet{boris2021infinite} makes optional at the cost of its fallback for failed blocks. Here
an input named by the example lets local generators meet the constraint.

\paragraph{Local computation.}
The $K$-valid colorings form a subshift of finite type \citep{lind1995symbolic}. Classifications of locally
checkable labelings on grids \citep{brandt2017lcl,grebik2023grids,berlow2025separating} and local mending
\citep{balliu2022mending} describe what rules of bounded radius with coins can satisfy
\citep[after][]{luby1986,jeavons2016feedback}. \citet{balliu2025shared} separate shared from private randomness with
grids on which every row must output one bit: with private coins and too few rounds to see a whole row, its nodes
can agree only by fixing the bit by their vertical position, while with shared randomness each node reads the shared bit
at its row. Theorem~\ref{thm:shared} finds this dichotomy for every quantity that validity carries along an axis, at
any depth, with the example's statistics in place of their adversarial input. Finitary codings
\citep{holroyd2017finitary,spinka2020finitary} let the
radius be random but almost surely finite, and finitely dependent colorings \citep{holroyd2016finitely} ask only for
independence at a distance. Local computation
algorithms \citep{rubinfeld2011lca}, local access to huge random objects \citep{goldreich2010huge,biswas2020local} and
perfect sampling of infinite Gibbs measures \citep{anand2022perfect} answer local queries about one consistent sample.
Height functions and tiling groups \citep{conway1990tiling,thurston1990conway} carry quantities like ours along paths and
give criteria for whether a region can be tiled; we find ours by searching the example. These concern satisfiability or a given measure, not
an example's statistics.

\paragraph{Local samplers.}
Short-range denoisers can miss long-range structure \citep[Section~6]{montanari2023sampling}: a local denoiser succeeds
once its radius exceeds the data's Markov length by a logarithmic factor, and fails where that length diverges
\citep{hu2025local}. Trained
denoisers are local in practice \citep{niedoba2025mechanistic,lukoianov2025locality}, and by locality and equivariance
convolutional diffusion models sample locally consistent mosaics of training patches \citep{kamb2024creativity}. By Corollary~\ref{cor:diffusion}, such a sampler, run on the whole plane from independent
noise, cannot be both free of forbidden windows and faithful where validity carries a quantity that the statistics make
vary. Sampling can need far more rounds than constructing: a uniform independent set needs rounds of the order of the
diameter on some graphs, though the empty set is one \citep{feng2017sampled}. Gibbs measures have related limits
\citep{vandenberg1999finitary}, and for diffusion models,
generating scale by scale eases the sampling of long-range structure \citep{guth2022wavelet}.

\paragraph{Shared inputs.}
Seeded noise sums octaves of interpolated gradients hashed on integer lattices
\citep{perlin1985image,perlin2002improving,lagae2010survey}, and our octaves are such noise without interpolation.
Learned generators of unbounded images draw multiscale randomness too \citep{skorokhodov2021aligning,henzler2020neural}.
\citet{bergmann2017psgan} observed that a texture generator on independent noise with a limited receptive field
\citep{jetchev2016texture} is a mixing process that cannot align distant regions, and gave theirs a global channel and periodic channels with a random
phase, which for a phase is the input Theorem~\ref{thm:shared} asks for. Under a hard constraint we bound the defects
that independent noise costs, find in the example which shared input a generator needs, and show that a generator
free of forbidden windows can vary the quantity between worlds only through randomness shared along the axis. The
row-type sampler combines the recursive method of random generation
\citep{nijenhuis1978combinatorial,flajolet1994calculus} with counter-based random numbers \citep{salmon2011parallel},
and the certificate sets the period of our phase channels and the sampler's automaton.

\paragraph{Learning from one image.}
Neural cellular automata learn update rules from one target
\citep{mordvintsev2020growing,niklasson2021textures} or generate levels under quality-diversity search
\citep{earle2022nca}, SinGAN \citep{shaham2019singan} and TOAD-GAN \citep{awiszus2020toadgan} learn from one image
or level, and a diffusion sampler with a closed-form patch denoiser needs no training \citep{qiu2026training}. Run for a fixed number of steps on coins indexed by position, a texture automaton is a local generator with
random access, as ours is, but none of these works holds its output to a hard constraint.

%% file: sec_limits.tex
\section{Limitations}
\label{sec:limits}

We built the certificate search, the inputs and the training recipe on the public examples, and widened the search
after its first count (Section~\ref{sec:resist}), so the counts on them show agreement on the corpus the procedure was
built on rather than a test on new examples. Even the certified examples whose failures shaped none of it test it only within
that corpus, and on most of them the octave rule alone is valid as well. The tests from outside it are the
\heldoutAllCountWords\ examples we drew or transposed, where the input decides the outcome on
\heldoutInputNeededCountWords, and the \fxNExamples\ examples from other implementations, where it decides
the outcome only on \fxTiledNames, each one period of a pattern and all from one repository (Section~\ref{sec:heldout}).

The certificate is one-sided: an uncertified example may carry a quantity that the search does not cover, and a
certified example may carry a second quantity that no input of the procedure fixes, as SmileCity does
(Section~\ref{sec:cert}). The worlds of a global solver show no persistent dependence on the resistant examples without
a certificate, but that test can miss a quantity, as it misses the certified Skyline2's (Appendix~\ref{app:longrange}). Theorem~\ref{thm:glue} needs the example's rows to glue, which
they do not on 3Bricks; whether other rows realize its row types there, and which condition is necessary as well as
sufficient, is open (Appendix~\ref{app:converse}). The procedure has no input for a one-way quantity or for an invariant along a diagonal, so on Town and on
\fxRefusedNames\ the method is the octave rule, and every certified invariant in the corpus runs along rows, so only the transposed walls
test the procedure along columns (Section~\ref{sec:heldout}).

The phase channels fix the layout to the coordinates, so on Fabric and the mazes every coin seed we drew gives the same canvas,
though worlds with their statistics leave a seed little to choose beyond the translate (Appendix~\ref{app:rigid}). An
offset hashed from the seed, a coin shared over the plane and so allowed by Theorem~\ref{thm:shared}, would let the seed
place the phase or the anchor rows of the row-type sampler and keep random access; we did not train with one. Mending rewrites cells with Wave Function Collapse's
completions, and we have not divided the credit for validity between the rule and the mender on each example
(Section~\ref{sec:learning}). A mender of longer reach adds valid examples only at a wider margin, and on
\mwInvalidBNames\ none that we ran gives valid worlds (Section~\ref{sec:access}). Validity allows up to one forbidden window in a thousand on each canvas, a threshold we
chose; Figure~\ref{fig:corpus} shows the examples on which no canvas keeps one.

Each count rests on one training run per example and on three canvases. Trained from two more seeds, the method stays
valid at both on \methodAllSeedsCertCountWords\ of the \methodSeedZeroCertCountWords\ certified examples it makes valid
at the first, and at each new seed the octave rule is valid on slightly fewer examples than at the first
(Table~\ref{tab:seeds}). Judged on \thmCanvasesWords\ canvases instead of three, the method is no longer valid on
\methodNotRobustCertNames\ (Appendix~\ref{app:rulemap}). At the reach of the whole pipeline, clean canvases set the method apart from
generators on independent coins only weakly (Section~\ref{sec:valid}). The diffusion model ran at one
training budget (Appendix~\ref{app:diffusion}). We ran one window size, $N=3$, without symmetry variants, and only in
two dimensions but for one maze of voxels (Appendix~\ref{app:further}).

%% file: sec_conclusion.tex
\section{Conclusion}
\label{sec:conclusion}

A local generator needs more than independent coins to make the worlds that some examples allow. When validity carries
a quantity along an axis and the example's statistics make it vary, no generator on independent coins is both free of
forbidden windows and faithful to the statistics, however far each cell reads, and at a finite radius its defects fall
only as the inverse of the radius, an order that is attained. A certificate computed from the example finds such a
quantity on most of the public examples that resist independent coins and on none of the
others, and on examples from other implementations it again certifies only examples that resist. An input that distant cells share escapes the bound without giving up random access, and at any depth a generator free of forbidden windows needs one to vary the quantity between worlds; where the example's rows glue, as on every example with a
row invariant but 3Bricks, the row types drawn from the seed and the position are enough. Copying the example along this
input already gives valid worlds on all but \inputAloneInvalidWords\ of the certified examples that receive it, and a
learned cellular automaton or a convolutional diffusion model given it becomes valid on most of them, where on
independent coins it is valid on none. Learning adds something of its own where nothing is certified, as the prior from
which a mender starts, though a block solver started from the tiled example is valid there by construction. The learned
rule needs a mender that rewrites part of each world, and regions mended separately are guaranteed to agree once each carries a margin as
wide as a mending change can travel. What stays open is structure that the certificate does not name, such as the lattice of
SmileCity's faces, an input for a quantity that can change only one way or that runs along a diagonal, and a condition for worlds free of forbidden
windows where the rows do not glue, as on 3Bricks.

%% file: app_roadmap.tex
\paragraph{The appendices.} Appendices~\ref{app:setting} to~\ref{app:converse} hold random access in detail and the
proofs. Appendices~\ref{app:procedure} to~\ref{app:setup} describe the certificate search, the method and the
protocols. Appendices~\ref{app:cert} to~\ref{app:results} give the detailed results: the certificate on every example,
more on the evidence of Section~\ref{sec:results}, the failure of independent coins, SmileCity, the diffusion model and
further results. Appendix~\ref{app:record} sets each expectation we fixed before a run against what happened, and
Appendix~\ref{supp:tables} holds the tables that only the appendices cite.

%% file: app_setting.tex
\section{Random access in detail}
\label{app:setting}

\paragraph{One setting for every example.}
The repository sets the window size and the symmetry variants per example. We read every example periodically, at
window size $N=3$ and with no symmetry variants, one setting for all, so that no example gets a setting chosen for it.
The images are the repository's samples; its README credits the people and games they come from.

\paragraph{Periods.}
Let $T$ be the longest finite side of the blocks of the pairs in $\mathcal A$. A shift by a vector of $T\Z^2$ maps every
block of every pair onto a block of the same pair, so it permutes the coins without changing their law, and the output
commutes with it. Since block boundaries are fixed in absolute coordinates, a generator on the scale pairs alone is
translation-equivariant under these shifts and in general under no other; one in the general form commutes with the
shifts by common multiples of $T$ and $p$ that map every further lattice onto itself. In practice the coins are hashes
of the seed, the coarsened cell and a channel index, computed where they are needed. A rule that applies a radius-one
update for $R$ rounds to coins alone is a random-access generator of radius $R$; ours reads \coinPairs\ scale pairs
besides the cell's own coins, with period $T=\coinPeriod$ (Section~\ref{sec:cert}). The row-type sampler draws the row
types of a band of $L$ rows from that band's coins at a cost proportional to $L$, and the phase channels are functions
of the coordinates, so both fit the general form of Definition~\ref{def:ra}.

\paragraph{The three properties.}
Any region $S$ is computed from the coins read on $S$ and on a margin of width $r$, with a fixed number of hashes per
cell wherever $S$ lies, a coin of whole rows being one hash however long the rows. Independent computations of
overlapping regions hash the same coins at the same cells, so they agree cell for cell: seams are exact, and many
machines generate one world with nothing shared but the seed. An edit that \emph{overrides the coins read in $S$},
giving the cells of $S$ new values of every coin they read while every other cell keeps its coins, changes the output
only within distance $r$ of $S$, so edits are contained; redrawing a coin that a whole block shares can instead change
the output anywhere within distance $r$ of the block.

\paragraph{The mended output.}
A random-access generator must satisfy $K$ from what it sees within distance $r$ of each cell, and the last step of our
pipeline reaches farther. The mender (Section~\ref{sec:method}) re-solves the defects that the rule leaves, block
by block on a lattice fixed in absolute coordinates and with randomness hashed from the seed and the block coordinates,
so its output depends on a cell's position modulo twice the block size as well as on the coins, and a change it makes
can travel beyond the rule's radius. The mended output has a \emph{reach}, a distance beyond which no coin can change a
cell's mended color, and we call such a construction \emph{random access in the extended sense}: it is a generator in
the general form of Definition~\ref{def:ra} with its reach in place of the radius $r$, so the three properties hold
with the reach as the margin, and Proposition~\ref{prop:blocks} covers it. Appendix~\ref{app:mender} bounds the reach
of our pipeline, and Section~\ref{sec:access} measures the smaller margin at which every canvas we tested agrees.

%% file: app_proofs.tex
\section{Proofs}
\label{app:proofs}

Throughout, $K\subseteq[C]^{N\times N}$ is the set of allowed windows, a coloring is $x:\Z^2\to[C]$, $Q^P_v(x)$ is the
$P\times P$ patch of $x$ whose corner is $v$, and $W_v(x)=Q^N_v(x)$ is the window at $v$. A generator on independent coins
of radius $r$ is a translation-equivariant map from i.i.d.\ coins $(\omega_v)$ to $X$ with $X_v$ a fixed function of the
coins within $\ell_\infty$ distance $r$ of $v$ (Definition~\ref{def:indep}). The law of $X$ is then translation-invariant.

\subsection{The full labeling}
\label{app:labeling}

For a weakly connected component $c$ of $G_{P,e}$, give each edge $(a,b)$ the weight $+1$ and let $d_c$ be the greatest
common divisor of the signed lengths of the closed walks of $c$ that may traverse edges in either direction ($d_c=0$
when every such walk has signed length zero, so that $c$ has no directed cycle). Fixing a root, every patch of $c$ gets a level modulo $d_c$, and $\varphi(a)=(c,\text{level})$
with $\sigma(c,i)=(c,i+1\bmod d_c)$ is a transported quantity on all of $V_P$, with no exception for edges that enter
or leave a strongly connected component. Invariants are the case of several components; phases the case $d_c\ge2$.
Table~\ref{tab:cert} reports constants for this labeling.

\subsection{\texorpdfstring{Theorem~\ref{thm:floor}}{Theorem 1}}
\label{app:pfloor}

Let $k=2r+P$ and $u_j=je$ for $j=0,\dots,k$. Write $A$ for the event that the patches $Q^P_{u_j}(X)$ are locally valid
for every $j$. On $A$, each pair of consecutive patches agrees on its overlap, so it is an edge of $G_{P,e}$ and
$\varphi(Q^P_{u_{j+1}})=\sigma(\varphi(Q^P_{u_j}))$; hence $\varphi(Q^P_{u_k})=\sigma^k(\varphi(Q^P_{u_0}))$. The patch
at $u_0$ is a function of the coins in the box $[-r,P-1+r]$ along $e$, and the patch at $u_k$ of the coins in
$[k-r,k+P-1+r]$; since $k\ge 2r+P$ the two boxes are disjoint, and the two patches are independent. Writing $p^{(v)}$
for the law of $\varphi$ at $v$ on the event that the patch is locally valid, translation invariance gives
$p^{(u_0)}=p^{(u_k)}=p$, so
\[
\Pr[A]\;\le\;\sum_s p_s\,p_{\sigma^k(s)}\;\le\;\max_t p_t\sum_s p_s\;\le\;\max_t p_t .
\]
If $A$ fails, some patch on the segment contains a forbidden window. A patch contains $(P-N+1)^2$ windows, so
$\Pr[Q^P_v\notin V_P]\le(P-N+1)^2\varepsilon$, and a union bound over the $k+1$ patches gives
$1-\max_t p_t\le\Pr[A^c]\le(k+1)(P-N+1)^2\varepsilon$, which is the claim. \qed

For a one-way quantity, write $s\leadsto t$ when component $t$ is reachable from component $s$ in $G_{P,e}$, with
$s\leadsto s$. The same steps give, on $A$, that $\varphi(Q^P_{u_k})$ is reachable from $\varphi(Q^P_{u_0})$, so
$\Pr[A]\le\sum_{s\leadsto t}p_s\,p_t$ by independence, and the union bound gives
\[
\varepsilon\;\ge\;\frac{1-\sum_{s\leadsto t}p_s\,p_t}{(2r+P+1)(P-N+1)^2}.
\]
For an invariant or a phase the only pairs allowed are $t=\sigma^k(s)$, the sum is at most $\max_s p_s$, and this is the
bound of Theorem~\ref{thm:floor}. Appendix~\ref{app:lp} bounds the numerator for a one-way quantity from below by the
constant $\kappa(\delta)$.

\subsection{\texorpdfstring{Inequality~\eqref{eq:vfr}}{Inequality (1)}}

For $P=N$ the patch is the window, so $p_s=h(\{w\in K:\varphi(w)=s\})$, where $h$ is the law of $W_0(X)$ on all of
$[C]^{N\times N}$. Since $f$ puts no mass outside $K$, $h(B)\le f(B)+\delta$ for every set $B$ of windows, with
$\delta$ the total variation between $h$ and $f$; pooling the forbidden windows into one outcome, as
Section~\ref{sec:ra} does, leaves this total variation unchanged, for the same reason. Hence $\max_s p_s\le\max_s f(\varphi=s)+\delta$, and
Theorem~\ref{thm:floor} with $P=N$ gives $\varepsilon(2r+N+1)\ge\kappa_f-\delta$. \qed

\subsection{\texorpdfstring{Corollary~\ref{cor:slip}}{The corollary on rare changes}}
\label{app:pslip}

The proof of Theorem~\ref{thm:floor} uses only the overlap graph of the locally valid patches, so it holds with
$K_0=K\setminus J$ in place of $K$, a window being forbidden for $K_0$ when it lies outside $K_0$. Let $h$ be the law of
the window of $X$ at a cell, and write $D=2r+N+1$ and $B_s=\{w\in K_0:\varphi(w)=s\}$. The windows outside $K_0$ have
density $\varepsilon+h(J)$, and at $P=N$ the theorem gives $(\varepsilon+h(J))D\ge1-h(B_s)$ for the value $s$ with the
largest $h(B_s)$. As in the proof of inequality~\eqref{eq:vfr}, $h(B)\le f(B)+\delta$ for every set $B$ of windows in
$K$. Applied to $J\cup B_s$, this gives $h(B_s)\le\theta+(1-\kappa_J)+\delta-h(J)$, so
$\varepsilon D\ge\kappa_J-\theta-\delta-(D-1)h(J)$; applied to $J$, it gives $h(J)\le\theta+\delta$. Hence
$\varepsilon D\ge\kappa_J-(\theta+\delta)D$, which is the claim. When $J$ is empty, $h(J)=0$ and the first step with
$h(B_s)\le1-\kappa_f+\delta$ gives inequality~\eqref{eq:vfr}. \qed

\subsection{The linear program}
\label{app:lp}

Let $P\ge N$. A distribution $q$ on $V_P$ is \emph{stationary} if every $(P-1)\times P$ pattern has the same mass as the
first $P-1$ rows of a patch and as its last $P-1$ rows, and every $P\times(P-1)$ pattern the same mass as the first
$P-1$ columns and as the last $P-1$ columns. The $N\times N$ window then has the same law at every position in the
patch, and we call it the $N\times N$ marginal of $q$. For a translation-invariant random coloring $X$ without defects,
the law of $Q^P_0(X)$ is a stationary distribution on $V_P$, since the patches at $0$ and at a unit vector have the same
law and share all but one row or column, and its $N\times N$ marginal is the window law of $X$. The constant
$\kappa(\delta)$ of Section~\ref{sec:theory} is one minus the largest occupancy $q(\varphi=s)$ over the values $s$ and the
stationary $q$ whose $N\times N$ marginal lies within $\delta$ of $f$ in total variation, a linear program for each
value. Such an $X$ whose window statistics lie within $\delta$ of $f$ therefore has $\max_sp_s\le1-\kappa(\delta)$.

\emph{A one-way quantity.} When $\varphi$ is the strongly connected component, let $f_s$ be the example's share of
patches in component $s$. The example's constant is $\kappa_f=1-\sum_{s\leadsto t}f_sf_t$, and for such a quantity
$\kappa(\delta)$ names a lower bound on $1-\sum_{s\leadsto t}p_sp_t$ that the program gives: the sum, over ordered
pairs of components visited by the example with $t$ not reachable from $s$, of the products of the smallest occupancies
the program allows to $s$ and to $t$. It is a lower bound because $\sum_sp_s\le1$ gives
$1-\sum_{s\leadsto t}p_sp_t\ge\sum_{s\not\leadsto t}p_sp_t$, with equality when no patch has a defect, and because for a
coloring without defects whose window statistics lie within $\delta$ of $f$ each $p_s$ is at least the smallest
occupancy the program allows to $s$.

\emph{With defects.} For $P\ge N$, let $q$ be the law of $Q^P_0(X)$ on all $P\times P$ colorings and $q_V$ its restriction
to $V_P$, of mass $1-\beta$ with $\beta\le(P-N+1)^2\varepsilon$. Translation invariance makes $q$ exactly stationary.
Dropping the patches outside $V_P$, of total mass $\beta$, lowers each side of these equalities by $\beta$ in total, so
summed over the $(P-1)\times P$ sub-patterns the masses that $q_V$ gives them from the two sides differ by at most
$2\beta$, the same holds over the $P\times(P-1)$ sub-patterns, and summed over both kinds they differ by at most
$4\beta$; and the $N\times N$ marginal of $q_V$ is within $\delta+\beta$ of $f$ in total variation. Let
$\kappa(\delta,\eta)$ be one minus the largest occupancy of any single value of $\varphi$ over nonnegative $q$ on $V_P$
with total mass in $[1-\eta,1]$, stationarity violated by at most $2\eta$ summed over both kinds of sub-pattern, and
$N\times N$ marginal within $\delta+\eta$ of $f$. It is a linear program, decreasing in $\eta$, and
$\kappa(\delta,0)=\kappa(\delta)$ is the constant above. The law $q_V$ is feasible at $\eta=2\beta$, so
$\max_s p_s\le1-\kappa(\delta,2\beta)$, and Theorem~\ref{thm:floor} gives, for any $\eta_0\ge0$,
\[
\varepsilon\ge\frac{1}{(P-N+1)^2}\min\Big\{\frac{\eta_0}{2},\frac{\kappa(\delta,\eta_0)}{2r+P+1}\Big\}.
\]
The program over exactly stationary measures, $\eta=0$, is what Table~\ref{tab:cert} reports. The table has
\certAllCount\ rows with an invariant or a phase: the \certInputCount\ certified examples with one, and Chess and
Stripes, two controls whose only $K$-valid colorings are the two phases of a period-two pattern
(Appendix~\ref{app:further}). With defects allowed at $\eta_0=\lpEtaZero$ and $\delta=0.05$, the program stays positive
on \lpKappaEtaPositive\ of these rows, with a smallest value of \lpKappaEtaMin\ on \lpKappaEtaMinExample, so a generator
whose statistics lie within that slack has a defect density of order $1/r$ with that constant. For $P=N$ the program is
stricter than inequality~\eqref{eq:vfr}, which gives $\kappa_f-\delta$: on Skyline2, $\kappa(0.05)$ is
\lpSkylineTwoKappaFive\ against $\kappa_f-0.05=\lpSkylineTwoIneq$. \qed

\subsection{\texorpdfstring{Theorem~\ref{thm:nogo}}{Theorem 2}}

Let $\tau$ be the translation by $e$ and $\tau'$ the unit translation orthogonal to it. A factor of i.i.d.\ coins is
mixing under the action of $\Z^2$, since the coins are, and mixing passes to factors. As the $\Z^2$ action is mixing,
every sub-action generated by a single nonzero translation is mixing as well, and hence ergodic; in particular
$\tau^D$ is ergodic for every $D\ge1$. On the event, of probability one, that $X$ is $K$-valid, every patch is locally valid and consecutive patches
along $e$ form edges of $G_{P,e}$.

\emph{Invariants and phases.} Under the full labeling $\varphi=(c,z)$ with $\sigma(c,z)=(c,z+1\bmod d_c)$.
Let $\psi(v)=\sigma^{-j}(\varphi(Q^P_v(X)))$ for $v=je+v'$ with $v'$ on the line through the origin orthogonal to $e$;
along every line parallel to $e$, $\psi$ is constant. Let $\lambda$ be its value on the line through the origin. A
component with $d_c=0$ has no directed cycle and so carries no bi-infinite walk. The patches along a line parallel to
$e$ in a $K$-valid coloring form such a walk, so they lie in one component, with $d_c\ge1$, whose levels modulo $d_c$
are finitely many, and $\lambda$ takes finitely many values; let
$D$ be a common multiple of the $d_c\ge1$, so that $\sigma^D$ is the identity. Then $\lambda(\tau^DX)=\lambda(X)$, so
each event $\{\lambda=s\}$ is $\tau^D$-invariant and has probability zero or one: $\lambda$ is almost surely a
constant $s_0$. Since $\tau X$ has the law of $X$ and $\lambda(\tau X)=\sigma(\lambda(X))$, also $\sigma(s_0)=s_0$,
which forces $d_{c_0}=1$ for the component $c_0$ of $s_0$. The line through $\tau'^{\,m}$ of the origin has the law of
the line through the origin, so its value is $s_0$ almost surely too, for every $m$, and $\varphi=s_0$ at every cell
almost surely.

\emph{One-way quantities.} Let $\varphi$ be the strongly connected component, and $m$ the number of components. Along
a line it never returns to a component it has left, so it changes at most $m-1$ times on the whole line. If $q$ is the
probability that it changes between $v$ and $v+e$, the expected number of changes on a segment of $n$ steps is $nq$ by
stationarity, and it is at most $m-1$ for every $n$; so $q=0$, and $\varphi$ is constant along lines almost surely. It
is then an invariant, and the argument above makes it a single value at every cell.

\emph{Statistics.} With $\varphi$ concentrated on one value, $\max_sp_s=1$, while every stationary distribution
without defects whose window marginal is within $\delta$ of $f$ has $\max_sp_s\le1-\kappa(\delta)$ (for a one-way
quantity, $1-\sum_{s\leadsto t}p_sp_t\ge\kappa(\delta)$, and concentration makes the left side zero). The law of $X$
is such a distribution when its window statistics are within $\delta$ of $f$, so $\kappa(\delta)>0$ rules this out.
For $P=N$, $p_s\le f(\varphi=s)+\delta$ as in the proof of inequality~\eqref{eq:vfr}, so $1\le1-\kappa_f+\delta$, that
is, $\delta\ge\kappa_f$. \qed

\subsection{\texorpdfstring{Proposition~\ref{prop:blocks}}{Proposition 1}}
\label{app:pblocks}

In full, part (c) reads as follows. \emph{Suppose every block has length at most $T$ along $e$ and the lattice has
period $T$ along $e$, where a coin shared along a whole line orthogonal to $e$, or along a group of consecutive such
lines, counts as a block, and let $X_v$ also read coins shared along whole lines parallel to $e$, or along groups of
consecutive such lines, and over the whole plane. Take $\varphi$ under the full labeling, let $D\ge1$ be a common
multiple of the positive values of $d_c$, let $\mu$ be the probability, averaged over the cells of one period of the
lattice, that a cell's patch is locally valid and lies in a component $c$ whose $d_c$ does not divide $T$, $d_c=0$
included. Then}
\[
\varepsilon\;\ge\;\frac{\mu}{2\,\bigl(2r+(D+1)T+P\bigr)\,(P-N+1)^2}.
\]

Part (a) gives no floor when the quantity is fixed by position, and blocks fixed in absolute coordinates tell a rule
its position, since it sees the coins change at a block boundary. A rule that leaves no forbidden window must fix the
quantity by position. Two cells far apart along $e$ read disjoint sets of coins, apart from coins shared along whole
lines or over the plane, so given those shared coins their values of the quantity are independent; validity along the
line between them makes each value determine the other; and two independent values that determine each other are
constant. Each value is therefore fixed by the cell's position within the lattice's period and by the shared coins, so
the law of the layout repeats with period $T$ along $e$, which a phase of period $d$ allows without defects only when
$d$ divides $T$. When it does not, (c) gives a floor of the same order as (a), in proportion to the mass $\mu$ of the
components whose period does not divide $T$; when $P=N$ and the window statistics lie within $\delta$ of the example's,
$\mu$ is at least the example's mass less $\delta$. A coin shared along a whole line, $\ell=\infty$, escapes (a): two
cells of the line read the same coin however far apart they are, so they can agree on an invariant along that line with
no defect between them, while different lines carry different values. The construction in (b) copies the example on
every block, so it says nothing about variation.

(a) Follow the proof of Theorem~\ref{thm:floor} with $k=\ell+2r+P$. The patch at $u_0$ reads the independent coins in the
box $[-r,P-1+r]$ along $e$ and the coins of the blocks that meet it; the patch at $u_k$ those of $[k-r,k+P-1+r]$. A
block meeting both boxes would cover the $k-2r-P+2=\ell+2$ cells from $P-1+r$ to $k-r$, which no block of length at most
$\ell$ does, so the two patches are again independent. The law of $X$ is now invariant only under the lattice's periods, so
the laws $p^{(u_0)}$ and $p^{(u_k)}$ may differ; with $p^*$ the largest $p^{(v)}_s$ over cells and values,
$\Pr[A]\le\sum_sp^{(u_0)}_s\,p^{(u_k)}_{\sigma^k(s)}\le p^*$. Average the union bound over the starting cell $u$ in
one period of the lattice: the average of $\Pr[Q^P_v\notin V_P]$ over a period is at most $(P-N+1)^2\varepsilon$, with
$\varepsilon$ the density of forbidden windows, so $1-p^*\le(k+1)(P-N+1)^2\varepsilon$. The one-way version and
inequality~\eqref{eq:vfr} follow as before, with $\delta$ the largest distance from $f$ of the window statistics at a
single cell.

(b) Since $K$ and $f$ are read from the example with wrap-around, tiling the plane with the example leaves no forbidden
window and gives the example's window frequencies. A window inside one block is a window of the tiling at a uniformly
random translation, whose law is $f$, and it is allowed. A window meets two blocks only if its corner lies within $N-1$ cells of a block's boundary along one of the
two axes, which holds for a share at most $2(N-1)/\ell$ of cells. Hence $\varepsilon\le2(N-1)/\ell$, and the window law
averaged over cells is within $2(N-1)/\ell$ of $f$ in total variation.

(c) Let $\mathcal G$ be the coins shared along whole lines parallel to $e$ or along groups of consecutive such lines, and
over the plane.
All coins are independent, so given $\mathcal G$ the other coins are still independent with their own laws. The
translation by $Te$ fixes every coin of $\mathcal G$ and permutes the other coins, cell coins to cell coins and the coin of
every block, a group of lines orthogonal to $e$ included, to the coin of its translate, a block of the same kind, since
the lattice has period $T$ along $e$; it therefore preserves their joint
law, given $\mathcal G$ as well. As $X$ commutes with that translation, the conditional probabilities
$q_s(v)=\Pr[Q^P_v\in V_P,\ \varphi(Q^P_v)=s\mid\mathcal G]$ satisfy $q_s(v+Te)=q_s(v)$. Let $k$ be the smallest
integer with $k\equiv1\pmod D$ and $kT\ge2r+T+P$; then $kT+1\le2r+(D+1)T+P$, since $k$ lies within $D-1$ of
$\lceil(2r+T+P)/T\rceil$. Let $u_k=u_0+kTe$. As in (a), the patches at $u_0$ and $u_k$ read disjoint sets of coins
outside $\mathcal G$, so they are independent given $\mathcal G$. On the event $A$ that every patch on the segment
between them is locally valid, $\varphi(Q^P_{u_k})=\sigma^{kT}(\varphi(Q^P_{u_0}))$. Hence, with $q=q(u_0)$,
\[
\Pr[A\mid\mathcal G]\;\le\;\sum_s q_s\,q_{\sigma^{kT}(s)} .
\]
On a component with $d_c\ge1$, $\sigma^{kT}=\sigma^T$, since $kT\equiv T$ modulo $d_c$. If $d_c$ divides $T$, $\sigma^{kT}$ fixes every value, and the terms add to at most the component's mass
$Q_c=\sum_{s\in c}q_s$. If it does not, $\sigma^{kT}=\sigma^T$ shifts the level by $T\bmod d_c\neq0$; if
$d_c=0$, the level is an integer and $\sigma^{kT}$ shifts it by $kT\ge1$. In both cases $\sigma^{kT}$
maps the component's values one to one and moves every value, and for any such map $\pi$,
$\bigl(\sum_s q_s\bigr)^2\ge\sum_s q_s^2+\sum_s q_s\,q_{\pi(s)}\ge2\sum_s q_s\,q_{\pi(s)}$: the pairs
$(s,\pi(s))$ are distinct and off the diagonal, and $q_s\,q_{\pi(s)}\le(q_s^2+q_{\pi(s)}^2)/2$. Those terms add to at most
$Q_c^2/2\le Q_c/2$. So $\Pr[A\mid\mathcal G]\le1-\mu_{\mathcal G}/2$, with $\mu_{\mathcal G}$ the conditional mass at $u_0$
of the components whose $d_c$ does not divide $T$, and taking expectations, $\Pr[A^c]$ is at least half the probability that the patch at $u_0$ is
locally valid with $\varphi$ in such a component. Averaging the union bound over $u_0$ in one period of the lattice,
as in (a), gives $\mu/2\le(kT+1)(P-N+1)^2\varepsilon$, a floor slightly sharper than the statement's, which
Appendix~\ref{app:scopefloors} evaluates. \qed

\subsection{\texorpdfstring{Proposition~\ref{prop:mincoin}}{Proposition 2}}
\label{app:mincoin}

Let $\omega_v=(U_v,t_v)$, with $U_v$ uniform on $[0,1]$ and $t_v$ uniform over one period of the tiling $Y$, all
independent. Let $B_r(v)=\{v+u:\|u\|_\infty\le r\}$, let $m_v$ be the cell of smallest priority in $B_r(v)$ (ties have
probability zero), and let $X_v=Y(v-m_v+t_{m_v})$. Then $X_v$ is a fixed function of the coins in $B_r(v)$, and shifting
all coins by a vector shifts $X$ by the same vector, so the min-coin copy is a generator on independent coins of radius
$r$.

Fix a window, the $N\times N$ square $W$ of cells with corner $w$. The boxes $B_r(v)$, $v\in W$, have as their union a
square $\mathcal U$ of side $2r+N$ and as their intersection a square $\mathcal I$ of side $2r+2-N$, which is not empty
since $r\ge N$. Let $z$ be the cell of smallest priority in $\mathcal U$. If $z\in\mathcal I$, then $z$ lies in every
box and has the smallest priority in each, so $m_v=z$ for every $v\in W$. Conversely, if $m_v=z'$ for every $v\in W$,
then $z'\in\mathcal I$ and $z'$ has the smallest priority in every box, hence in $\mathcal U$, so $z'=z$. The priorities
are independent and continuous, so $z$ is uniform on $\mathcal U$, and the cells of $W$ copy from one cell with
probability $|\mathcal I|/|\mathcal U|=1-q_r$. On that event the window of $X$ at $w$ is the window of $Y$ at
$w-z+t_z$. Given the priorities, which fix $z$, the translation $t_z$ is uniform over the period and independent of
them, so the window is a window of the tiling at a uniform translation: it is allowed and its law is $f$. Hence
$\varepsilon\le q_r$, and the law $h$ of the window is $(1-q_r)f+q_rg$ for some law $g$, so
$\|h-f\|_{\mathrm{TV}}=q_r\|g-f\|_{\mathrm{TV}}\le q_r$. With $a=(2r+2-N)/(2r+N)$, which lies in $(0,1)$ since $2\le N\le r$,
$q_r=(1-a)(1+a)<2(1-a)=4(N-1)/(2r+N)$.

For the comparison in Section~\ref{sec:theory}, let $P=N$. Every generator on independent coins of radius $r$ whose
window statistics lie within $q_r$ of $f$ has $\varepsilon\ge(\kappa_f-q_r)/(2r+N+1)$ by inequality~\eqref{eq:vfr}. The
map $q\mapsto q(2r+N+1)/(\kappa_f-q)$ increases on $[0,\kappa_f)$, so whenever $\kappa_f(2r+N)>4(N-1)$ the ratio of the
min-coin copy's $\varepsilon$ to that floor is less than
\[
\frac{4(N-1)\,(2r+N+1)}{\kappa_f\,(2r+N)-4(N-1)},
\]
which tends to $4(N-1)/\kappa_f$ as $r$ grows. \qed

\subsection{What the bounds bind in our pipeline}
\label{app:scopefloors}

Theorems~\ref{thm:floor} and~\ref{thm:nogo} bind generators on independent coins, among them the independent-coin rule
of Section~\ref{sec:theory}, with or without its repair rounds. They do not bind the method. Its row types and the
octaves' coins of whole rows are shared along whole lines, which escape both theorems and
Proposition~\ref{prop:blocks}(a), and its phase channels, the coordinates modulo the certified period $d$, are an input of period $d$, which the
certified phase's period divides, so Proposition~\ref{prop:blocks}(c) sets no floor on that phase. The octaves, the coins on the scale pairs that our rule reads
(Section~\ref{sec:cert}), fall under Proposition~\ref{prop:blocks}(c) as a whole: their coins of finite scale lie on
blocks nested in one lattice of period $T=\coinPeriod$, and their other coins are shared along whole rows, whole columns
or the plane. Among the certified examples the resulting floor binds only Fabric, whose phase has period \phaseDFabric,
which does not divide $T$. There it lies above the threshold of validity before mending, and below it after. Once
mended, no output has a floor from Section~\ref{sec:theory} above that threshold on any certified example, so no result
there requires the mended independent-coin rule to be invalid; where we searched for witnesses, they show why it is
(Section~\ref{sec:resist}).

These conclusions rest on the following values. The rule runs \mainR\ rounds of a radius-one
update, so its radius is \floorRadiusRaw, and each of its \repairRoundsWords\ repair rounds adds $N$ cells, for a radius
of \floorRadiusRepair. Mending at the main radius works on blocks of side $\ell=\mendBlockMain$, and its eight class
steps carry a change at most \mendBoundCC\ cells (Appendix~\ref{app:mender}), so the reach of the whole pipeline is
$r=\certRadiusCC$ cells. The threshold of validity is one forbidden window in a thousand (Section~\ref{sec:ra}).

\emph{Fabric under the octaves.} Fabric's phase has period \phaseDFabric\ and is read off single windows, so $P=N$. The
octaves fall under Proposition~\ref{prop:blocks}(c) with $T=\coinPeriod$ and $D=\phaseDFabric$, and the components
whose period does not divide $T$ hold \fabricPhaseMass\ of the example's windows, which is $\mu$ for a rule whose window
statistics match the example's. At radius \floorRadiusRaw, $k=\fabricOctK$ and $2(kT+1)(P-N+1)^2=\fabricOctDenRaw$, so
such a rule leaves at least \fabricOctFloorRaw\ of its windows forbidden, and with its repair rounds, at radius
\floorRadiusRepair, at least \fabricOctFloorRepair; both lie above the threshold. The mended output commutes only with
shifts by common multiples of $T$ and of twice the mender's block, so with that period, \fabricOctMendPeriod\ cells, in
place of $T$ and with the pipeline's reach as the radius the floor falls to \fabricOctFloorMended, below the threshold.

\emph{The mended independent-coin rule.} The mender's randomness is hashed from the seed and the coordinates of its
blocks, of side $\ell$ in both passes, so the mended output of the independent-coin rule falls under
Proposition~\ref{prop:blocks}(a) with the pipeline's reach as the radius. Its denominator is
$(2r+\ell+P+1)(P-N+1)^2$, which is \mendedFloorDen\ when $P=N$, and its numerator is at most one, so at this reach the
bound of (a) cannot reach the threshold whatever the output's statistics. For an output that spreads the quantity at
every cell as the example does and has its statistics, the numerator is the example's own spread, which on every row of
Table~\ref{tab:cert} with an invariant or a phase is the column $\kappa(0)$, equal to $\kappa_f$ when $P=N$. The floor
computed from that column is not itself a column of the table, whose floor at radius \floorRadiusRaw\ is that of
Theorem~\ref{thm:floor} for the raw rule. Its largest value over those rows, Chess and Stripes included, is
\mendedFloorMax, on \mendedFloorMaxName.

\subsection{Two classes of constraints}

Section~\ref{sec:theory} cites the following two bounds. Let $\varepsilon(r)$ be the smallest density of forbidden
windows over generators on independent coins of radius $r$.

\begin{proposition}[two classes]\label{prop:classes}
(a) If the locally checkable labeling of $K$ is solvable in $O(\log^* n)$ rounds on $n\times n$ toroidal grids, then
$\varepsilon(r)\le Ce^{-cr}$ for constants $C,c>0$ that depend on $K$. (b) If $K$ is an agreement constraint, that is,
for some $P\ge N$ every locally valid $P\times P$ patch is the restriction of one of $m\ge2$ translates of a periodic
pattern and distinct translates differ on every block of $P\times(P-1)$ or $(P-1)\times P$ cells, then
$\varepsilon(r)\ge1/(2(2r+P+1)(P-N+1)^2)$.
\end{proposition}

\emph{Proof of (a).} \citet{brandt2017lcl} show that such a $K$ admits an algorithm in the normal form $A'\circ S_k$: a
maximal independent set (MIS) of the $k$-th power of the grid, followed by a finite function of the MIS within a
constant radius $\rho_0$. Run the MIS with fresh priorities in every round: in each round every undecided vertex draws a
uniform priority from its coins; a vertex whose priority is below those of all its undecided neighbors (in the power
graph, of degree $\Delta=(2k+1)^2-1$) joins the set, and its neighbors are removed. Whatever the history, the vertex's
own priority is one of at most $\Delta+1$ fresh independent draws in its undecided closed neighborhood, so it is such a
local minimum with probability at least $1/(\Delta+1)$; hence it is still undecided after $t$ rounds with probability
at most $(1-1/(\Delta+1))^t$. After $t$ rounds the decided vertices form an independent set that is maximal except in
the neighborhoods of the vertices still undecided; decide those arbitrarily and apply $A'$. A cell is defective only if
some vertex within distance $\rho_0+k$ of it is undecided, since $A'$ reads the set within radius $\rho_0$ and the set
can fail to be maximal only within distance $k$ of an undecided vertex, so the defect density is at most
$C(1-1/(\Delta+1))^t$ with $C=(2\rho_0+2k+1)^2$, and the construction reads coins within radius $O(kt)=r$. \citet{brandt2017lcl} prove the normal
form correct on all sufficiently large tori; since $A'$ is a finite-radius function of an MIS of the power grid, and any
finite MIS configuration of the plane extends to an MIS of a large torus (copy it with a margin of width $k$ onto a
torus large enough that the copy does not wrap onto itself, then add vertices outside the region greedily until the set
is maximal), the rule is correct on the plane as well. The
known rate is faster: a finitary-factor solution whose coding radius exceeds $r$ with probability at most
$1/\mathrm{tower}(cr)$ exists \citep[Cor.~3.3]{grebik2023grids}, and cutting it off at radius $r$ gives
$\varepsilon(r)\le C/\mathrm{tower}(cr)$.

\emph{Proof of (b).} The phase of a locally valid patch, the index of the translate it comes from, is a transported
quantity along every lattice direction, with $\sigma$ the permutation of translates induced by the unit shift. Since
$m\ge2$, some unit vector $e$ is not a period of the pattern, so $\sigma$ along $e$ has no fixed point. Translation
invariance gives $\Pr[\text{phase}(v)=i]=\Pr[\text{phase}(v+e)=i]$, and the two events exclude each other, because on
both of them the phase at $v+e$ would be $\sigma(i)\ne i$. Each therefore has probability at most $\tfrac12$, and
$\max_i p_i\le\tfrac12$. Theorem~\ref{thm:floor} along $e$ gives the claim. \qed

The same argument shows that a forced edit on an agreement constraint can need an unbounded repair: pin a cell to a
phase its neighborhood does not have, and any valid completion within radius $\rho$ would have to change phase across
the boundary of the ball, which the transport rules out. This is the witness of Appendix~\ref{app:pwitness} in its
simplest form.

\subsection{Witnesses}
\label{app:pwitness}

A transported quantity can also prove that a particular mending instance has no completion. A \emph{mending instance}
frees the cells of a region $F$ and fixes all others, and a completion is valid if every window that touches $F$ is
allowed. Let $a$ and $b$ be locally valid patches at $u$ and $u+ke$, entirely outside $F$, such that every window of
the patches $Q^P_{u+je}$, $0\le j\le k$, either touches $F$ or lies in fixed cells and is allowed. In any valid
completion every one of these patches is then locally valid, so by the argument of Theorem~\ref{thm:floor}
$\varphi(b)=\sigma^k(\varphi(a))$. If $\varphi(b)\ne\sigma^k(\varphi(a))$, no valid completion exists, and we call the
pair a \emph{witness}: it proves without any search that the instance has no completion (Figure~\ref{fig:quantity}(d);
Appendix~\ref{app:witness} counts the witnesses on the failed instances). \qed

\subsection{A sequence of row types at finite radius}
\label{app:rowseq}

Coins shared along whole rows can carry an invariant along rows at finite radius, once the rule turns them into an
allowed sequence of row types, one in which every two consecutive types can be consecutive in a coloring free of
forbidden windows. Suppose some type $\tau$ can follow itself, as the sky above a skyline can. Fix finitely many
\emph{stretches}, sequences of types $x_1,\dots,x_m$ such that $\tau,x_1,\dots,x_m,\tau$ is allowed, such as a course of
bricks, and let $M$ be the length of the longest. Give each row $y$ a uniform coin $U_y$ shared along the row, and call
$y$ a \emph{start} if $U_y<U_{y'}$ for every other row $y'$ within distance $M$; ties have probability zero. The rows
$y+1$ to $y+m$ after each start $y$ take a stretch chosen by the coin of row $y$, and every other row takes $\tau$. Of
two rows within distance $M$ of each other at most one is a start, so two starts lie more than $M$ rows apart, the
stretches never overlap, and every stretch is preceded and followed by a row of type $\tau$; the sequence is therefore
allowed. Whether a row is a start depends on the coins of the rows within distance $M$ of it, and the type of row $z$
depends on which of the rows $z-M$ to $z-1$ are starts and on their coins, so it is a function of the coins of the rows
within distance $2M$ of $z$. The construction gives the sequence no particular law, and the method does not use it.

\subsection{The certificate search}
\label{app:search}

On every public example where the search of Section~\ref{sec:cert} reports a phase, and on Chess and Stripes, $d_c$
equals the period of the strongly connected component that holds the example's patches in $c$, the greatest common
divisor of its cycle lengths, which $d_c$ in general only divides, since every directed cycle is a closed walk. We also
searched the two diagonals. Two patches one step apart along a diagonal leave two corner cells of their union outside
both, so an edge there asks the whole box of side $P+1$ to be locally valid. On the public examples the diagonals certify
none that the axis directions leave uncertified; among the examples from other implementations they certify
\fxRefusedNames\ (Appendix~\ref{app:freshex}). Appendix~\ref{app:procedure} gives the order in which the search tries patch sizes
and directions.

%% file: app_converse.tex
\ifdefined\lemma\else\newtheorem{lemma}{Lemma}\fi
\section{Proofs of the characterization}
\label{app:converse}

Theorems~\ref{thm:floor} and~\ref{thm:nogo} bound generators whose distant cells share no randomness. Run given the
coins that a generator shares along the quantity's direction, their two arguments prove Theorem~\ref{thm:shared}, and
where the example's rows glue, a generator that reads the row types band by band, and its position, needs nothing more
(Theorem~\ref{thm:glue}).

\subsection{The setting}
\label{app:cvsetting}

We read Definition~\ref{def:ra} as follows. The input of a generator in the general form consists of finitely many
\emph{families} of coins. A family is a lattice of blocks fixed in absolute coordinates, every block a rectangle,
possibly unbounded in one direction or in both; each block carries a vector of coins, and these vectors are i.i.d.\
within a family and independent across families. The cells' own coins form one family, whose blocks are single cells,
and so does every scale pair of Section~\ref{sec:ra} and every further lattice of Definition~\ref{def:ra}. Cell $v$
reads $v\bmod p$ and the coins of the blocks that meet the box $v+[-r,r]^2$.

Fix an axis direction $e$. The \emph{extent} of a family along $e$ is the length of its blocks along $e$: one for the
cells, $2^i$ along rows and $2^j$ along columns for the scale pair $(i,j)$, one or $w$ for whole lines orthogonal to $e$ or groups of $w$
consecutive such lines, and infinite for whole lines parallel to $e$, groups of such lines and the plane. The coins
\emph{shared along $e$}, $\mathcal G_e$, are the coins of the families of infinite extent along $e$, and we use the
same symbol for the $\sigma$-field they generate. All other coins belong to families of finite extent, and $\ell$ in
Theorem~\ref{thm:shared}(b) is the largest of these extents. Throughout, a function of some coins means a measurable
function of them, and equalities between random variables hold almost surely.

For a translation $t$ of $\Z^2$ write $\tau_t$ for the translation of colorings, $(\tau_tx)_v=x_{v+t}$, and $\theta_t$
for the translation of coins, which gives every block $B$ the coins of $B+t$. As in Section~\ref{sec:theory}, $\Lambda$
is the lattice of translations $t\in p\Z^2$ that map the blocks of every family to blocks of the same family; it has
full rank, and $T$ is the least positive integer with $Te\in\Lambda$. For $t\in\Lambda$, cell $v$ of $X(\theta_t\omega)$
reads the coins that cell $v+t$ of $X(\omega)$ reads, and $v\bmod p=(v+t)\bmod p$, so
\begin{equation}\label{eq:cvequi}
X(\theta_t\omega)\;=\;\tau_t\,X(\omega)\qquad(t\in\Lambda).
\end{equation}
This is what it means for $X$ to commute with the translations in $\Lambda$. Besides the radius and the blocks read,
which the counting bounds use, it is all that the proofs below use of $X$, so the arguments from ergodicity hold for
every measurable function of the coins and of the position modulo $p$ with this property. Since $\theta_t$ permutes the blocks of each family, it preserves the law of the
coins, so the law of $X$ is invariant under $\tau_t$ for $t\in\Lambda$, and in general under no other translation.
A function $a$ on the cells that is periodic under $\Lambda$, such as the probability that the window at $v$ is
forbidden, has an average $\operatorname{avg}_va(v)$ over the cells of one period of $\Lambda$ that does not depend on
the period chosen, and $\operatorname{avg}_va(v+u)=\operatorname{avg}_va(v)$ for every $u\in\Z^2$. The density of
forbidden windows is $\varepsilon=\operatorname{avg}_v\Pr[W_v(X)\notin K]$. For a random quantity $a(v)$ attached to
each cell, such as $1-\max_sq_s(v)$ in Theorem~\ref{thm:shared}(b), we write
$\overline{a}=\operatorname{avg}_v\mathbb E[a(v)]$; its expectation is periodic under $\Lambda$ as well, since
$\theta_t$ maps the coins of $\mathcal G_e$ to coins of $\mathcal G_e$.

\subsection{\texorpdfstring{Theorem~\ref{thm:shared}(b): the floor given the shared coins}{Theorem 3(b)}}
\label{app:cvfloor}

In full, part (b) reads as follows. \emph{If $\varphi$ is transported, $X$ has radius $r$ and every block it reads
whose coins are not in $\mathcal G_e$ has length at most $\ell$ along $e$, a cell's own coins counting as a block of
length one, then}
\[
\varepsilon\;\ge\;\frac{\overline{1-\max_s q_s}}{(2r+P+\ell)(P-N+1)^2},
\]
\emph{where $q_s(v)$ is the probability given $\mathcal G_e$ that the $P\times P$ patch of $X$ at $v$ is locally valid
and has $\varphi=s$, and the bar takes the expectation, averaged over the cells of one period of $\Lambda$.}

The counting argument of Theorem~\ref{thm:floor} needs two patches far apart along $e$ to be independent. Given
$\mathcal G_e$ they are, once their distance exceeds the extent of every other block.

\begin{lemma}\label{lem:condind}
Let $X$ have radius $r$, let every family outside $\mathcal G_e$ have extent at most $\ell$ along $e$, and let
$k\ge2r+P+\ell-1$. For every cell $v$, the patches $Q^P_v(X)$ and $Q^P_{v+ke}(X)$ are independent given $\mathcal G_e$.
\end{lemma}

\emph{Proof.} The patch at $v$ is a function of $v\bmod p$, of the coins of $\mathcal G_e$ and of the other coins of the
blocks that meet the box $v+[-r,P-1+r]^2$, and the patch at $v+ke$ is a function of the same kind for the box shifted
by $ke$. Counted from $v$ along $e$, the two boxes cover the positions $-r$ to $P-1+r$ and $k-r$ to $k+P-1+r$. A block
that meets both is a rectangle with a cell in each, so along $e$ it covers every position from $P-1+r$ to $k-r$, which
are $k-2r-P+2\ge\ell+1$ positions, more than any block outside $\mathcal G_e$ covers. The two patches therefore read disjoint
sets of coins outside $\mathcal G_e$, and these coins are independent of one another and of $\mathcal G_e$. \qed

As in Theorem~\ref{thm:shared}, $q_s(v)$ is the probability given $\mathcal G_e$ that $Q^P_v(X)$ is locally valid and
has $\varphi=s$.

\begin{proposition}[the floor given the shared coins, sharp form]\label{prop:sharp}
Under the assumptions of Lemma~\ref{lem:condind}, for a transported quantity $\varphi$ along $e$ with bijection
$\sigma$,
\begin{equation}\label{eq:cvsharp}
\begin{aligned}
&\varepsilon\,(k+1)(P-N+1)^2\\
&\qquad\ge\;\operatorname{avg}_v\,\mathbb E\Bigl[1-\sum_s q_s(v)\,q_{\sigma^k(s)}(v+ke)\Bigr],
\end{aligned}
\end{equation}
and for a one-way quantity the same holds with $\sum_{s\leadsto t}q_s(v)\,q_t(v+ke)$ in place of the sum.
\end{proposition}

\emph{Proof.} Let $A$ be the event that the $k+1$ patches $Q^P_{v+je}(X)$, $0\le j\le k$, are all locally valid. On
$A$ consecutive patches agree on their overlap, so they are edges of $G_{P,e}$, and
$\varphi(Q^P_{v+ke}(X))=\sigma^k(\varphi(Q^P_v(X)))$. By Lemma~\ref{lem:condind},
\[
\begin{aligned}
\Pr[A\mid\mathcal G_e]&\le\sum_s\Pr\bigl[\varphi(Q^P_v)=s,\\
&\qquad\quad\varphi(Q^P_{v+ke})=\sigma^k(s)\bigm|\mathcal G_e\bigr]\\
&=\sum_s q_s(v)\,q_{\sigma^k(s)}(v+ke),
\end{aligned}
\]
where each event includes that both patches are locally valid. If $A$ fails, one of the $k+1$ patches holds a forbidden window, and
a patch holds $(P-N+1)^2$ windows, so $\Pr[A^c]\le\sum_{j=0}^{k}\sum_u\Pr[W_{v+je+u}(X)\notin K]$, with $u$ running
over the corners of the windows of a patch. Averaged over $v$ in one period of $\Lambda$, each of the
$(k+1)(P-N+1)^2$ terms becomes $\varepsilon$. Hence
$\operatorname{avg}_v\mathbb E\bigl[1-\sum_sq_s(v)\,q_{\sigma^k(s)}(v+ke)\bigr]\le\operatorname{avg}_v\Pr[A^c]
\le(k+1)(P-N+1)^2\varepsilon$. For a one-way quantity, on $A$ the component at $v+ke$ is reachable from the one at $v$,
which gives the second form in the same way. \qed

\emph{Proof of Theorem~\ref{thm:shared}(b).} Since $\sigma^k$ is a bijection,
$\sum_sq_s(v)\,q_{\sigma^k(s)}(v+ke)\le\max_sq_s(v)\sum_tq_t(v+ke)\le\max_sq_s(v)$. With $k=2r+P+\ell-1$,
inequality~\eqref{eq:cvsharp} becomes the bound of Theorem~\ref{thm:shared}(b). \qed

For a one-way quantity the sum over the pairs $s\leadsto t$ can exceed $\max_sq_s(v)$, so its floor is the second form
of Proposition~\ref{prop:sharp}, as in Appendix~\ref{app:pfloor}.

\paragraph{Defects, fidelity and what is shared.}
Let $p_s(v)=\mathbb E[q_s(v)]$ be the probability that the patch at $v$ is locally valid with $\varphi=s$. The
\emph{gain} at $v$,
\[
I(v)\;=\;\mathbb E\bigl[\max_sq_s(v)\bigr]-\max_sp_s(v)\;\ge\;0,
\]
is how much knowing $\mathcal G_e$ raises the chance of guessing the value of $\varphi$ at $v$. Since
$\mathbb E[1-\max_sq_s(v)]=1-\max_sp_s(v)-I(v)$, Theorem~\ref{thm:shared}(b) reads
$\varepsilon(2r+P+\ell)(P-N+1)^2+\operatorname{avg}_vI(v)\ge\operatorname{avg}_v\bigl(1-\max_sp_s(v)\bigr)$. When $P=N$,
let $\delta_v$ be the total variation between the law of the window $W_v(X)$ and $f$, with all forbidden windows
pooled into one outcome. As in the proof of inequality~\eqref{eq:vfr}, $p_s(v)\le f(\varphi=s)+\delta_v$, so
$1-\max_sp_s(v)\ge\kappa_f-\delta_v$, and with $\bar\delta=\operatorname{avg}_v\delta_v$ and
$\bar I=\operatorname{avg}_vI(v)$,
\begin{equation}\label{eq:cvthree}
\varepsilon\,(2r+N+\ell)\;+\;\bar\delta\;+\;\bar I\;\ge\;\kappa_f .
\end{equation}
A generator can thus meet the example's spread $\kappa_f$ with forbidden windows, with window laws that differ from $f$
at each position, or with coins shared along $e$ that reveal the quantity. For a generator that commutes with every
translation and shares nothing, $\bar I=0$, $\ell=1$ and $\bar\delta$ is the distance $\delta$ of its statistics from
$f$, so~\eqref{eq:cvthree} is inequality~\eqref{eq:vfr}. In general $\bar\delta$ averages the distance at each position
and can exceed the distance of the averaged statistics: the example tiled from absolute coordinates has the example's
statistics, yet at each cell a window fixed in advance, and it meets~\eqref{eq:cvthree} through $\bar\delta$ alone.

\paragraph{The earlier floors as cases.}
When nothing is shared, $\mathcal G_e$ is trivial and $\ell=1$, and a generator on independent coins has $q_s(v)=p_s$
at every cell, so Theorem~\ref{thm:shared}(b) is Theorem~\ref{thm:floor}. When no block is unbounded along $e$ and
blocks have extent at most $\ell$, as in Proposition~\ref{prop:blocks}(a), $\mathcal G_e$ is again trivial and
$q_s(v)=p_s(v)\le p^*$, the largest of these probabilities over cells and values, so part (b) gives the floor of that
proposition with $2r+P+\ell$ in place of $2r+\ell+P+1$: its proof (Appendix~\ref{app:pblocks}) separates the two patches
by one step more than they need.

Proposition~\ref{prop:blocks}(c) works by a different mechanism. There a rule may fix the quantity by its position given
the shared coins, so that $\max_sq_s(v)=1$ wherever the patch is locally valid and the numerator of part (b) vanishes.
Defects are forced instead because the layout repeats with period $T$ while $\sigma^T$ moves every value of a phase
whose period does not divide $T$. The sharp form~\eqref{eq:cvsharp} sees this. Translation by $Te$ fixes every coin of
$\mathcal G_e$ and permutes the others without changing their law, so by~\eqref{eq:cvequi} $q_s(v+Te)=q_s(v)$. Take
the separation $kT$ of Appendix~\ref{app:pblocks}, with $k\equiv1$ modulo a common multiple $D$ of the positive $d_c$ and
$kT\ge2r+T+P$; this exceeds $2r+P+\ell-1$, since there every block outside $\mathcal G_e$ has length at most $T$ along
$e$. Then~\eqref{eq:cvsharp} reads
$\varepsilon(kT+1)(P-N+1)^2\ge\operatorname{avg}_v\mathbb E\bigl[1-\sum_sq_s(v)\,q_{\sigma^{kT}(s)}(v)\bigr]$, and the
argument of Appendix~\ref{app:pblocks} bounds the sum by $1-\mu_{\mathcal G}/2$, with $\mu_{\mathcal G}$ the
conditional mass defined there, which gives Proposition~\ref{prop:blocks}(c) in the sharper form proved there, with
$T$ the period of $\Lambda$ along $e$. The max form alone gives nothing in this case.

\subsection{\texorpdfstring{Theorem~\ref{thm:shared}(a): the quantity is a function of the shared coins}{Theorem 3(a)}}
\label{app:cvergodic}

Theorem~\ref{thm:nogo} uses ergodicity in place of a radius, and given $\mathcal G_e$ the remaining coins are still
ergodic under translation along $e$.

\emph{The coins given $\mathcal G_e$.} Write the coins as $\omega=(\gamma,\omega')$, with $\gamma$ the coins of
$\mathcal G_e$ and $\omega'$ the others; the two are independent. The coins of $\omega'$ are indexed by the cells, the
bounded blocks, and the lines orthogonal to $e$ and groups of them. The translation $\theta_{Te}$ maps each of these
indices to another of the same family and fixes none, while it maps every block of infinite extent along $e$ to itself
and so fixes every coin of $\gamma$. Choose one index in every orbit of $\theta_{Te}$ and let $\xi_n$ collect the coins
of the $n$-th translates of the chosen indices. The $\xi_n$, $n\in\Z$, are i.i.d., and $\theta_{Te}$ shifts the
sequence $(\xi_n)$ by one place, so under $\theta_{Te}$ the coins $\omega'$ form a Bernoulli shift. A Bernoulli shift
is mixing, so $\theta_{DTe}$ is ergodic on $\omega'$ for every $D\ge1$. By~\eqref{eq:cvequi},
$X(\gamma,\theta_{Te}\omega')=\tau_{Te}X(\gamma,\omega')$ for every $\gamma$, and the law of $\omega'$ does not depend
on $\gamma$. Let $E$ be the event that $X$ is $K$-valid. It has probability one, so for almost every $\gamma$ it has
probability one in $\omega'$, and it is invariant under $\theta_{Te}$, since a translate of a $K$-valid coloring is
$K$-valid.

\emph{Proof of part (a) for a transported quantity.} Fix a cell $v'$ on the line through the origin orthogonal to $e$,
and let $\lambda(X)=\varphi(Q^P_{v'}(X))$. On $E$ the patches along the line through $v'$ parallel to $e$ are locally
valid and consecutive ones are edges of $G_{P,e}$, so
\begin{equation}\label{eq:cvline}
\varphi\bigl(Q^P_{v'+je}(X)\bigr)\;=\;\sigma^j\bigl(\lambda(X)\bigr)\qquad(j\in\Z).
\end{equation}
These patches form a bi-infinite walk in the finite graph $G_{P,e}$, so some patch $a$ occurs twice, $m\ge1$ steps
apart, and $\varphi(a)=\sigma^m(\varphi(a))$: the values along the line lie in a finite orbit of $\sigma$. Since
$\varphi$ takes finitely many values on $V_P$, finitely many finite orbits meet them; let $D\ge1$ be a common multiple
of their lengths, so that $\sigma^D$ fixes every value that $\lambda$ takes on $E$. Under the full labeling these orbits
are the levels of the components with $d_c\ge1$, as in the proof of Theorem~\ref{thm:nogo}, and if $\sigma$ has finite
order, $D$ can be that order. By~\eqref{eq:cvline} with $j=T$, $\lambda(\tau_{Te}X)=\sigma^T(\lambda(X))$ on $E$, so
$\lambda(\tau_{DTe}X)=\lambda(X)$ on $E$. Fix $\gamma$ outside a null set. The function
$\omega'\mapsto\lambda(X(\gamma,\omega'))$ is invariant under $\theta_{DTe}$ on the invariant event $E$ of probability
one, so by ergodicity it is almost surely a constant $h(v',\gamma)$, a measurable function of $\gamma$, since it is the
value $s$ with $\Pr[\lambda(X)=s\mid\mathcal G_e]=1$. As $\omega'$ and $\theta_{Te}\omega'$ have the same law,
$\lambda(X(\gamma,\theta_{Te}\omega'))=\sigma^T(\lambda(X(\gamma,\omega')))$ has the law of $\lambda(X(\gamma,\omega'))$;
both are constants, so $\sigma^T(h(v',\gamma))=h(v',\gamma)$. Put $h_v=\sigma^j\circ h(v',\cdot)$ for $v=v'+je$.
By~\eqref{eq:cvline}, $\varphi(Q^P_v(X))=h_v(\mathcal G_e)$ almost surely, at all cells at once since they are
countably many, with $h_{v+e}=\sigma\circ h_v$ and $h_{v+Te}=\sigma^T\circ h_v=h_v$. \qed

\emph{Proof of part (a) for a one-way quantity.} Let $\varphi$ be the strongly connected component and $m$ the number
of components. Along a line free of forbidden windows $\varphi$ never returns to a component it has left, so it changes
at most $m-1$ times on the whole line. Let $c_j=1$ if $X\in E$ and the components at $v'+je$ and $v'+(j+1)e$ differ,
and $c_j=0$ otherwise. Given $\gamma$ the law of $X$ is invariant under $\tau_{Te}$, so
$\mathbb E[c_{j+T}\mid\gamma]=\mathbb E[c_j\mid\gamma]$, and
$n\sum_{j=0}^{T-1}\mathbb E[c_j\mid\gamma]=\mathbb E\bigl[\sum_{j=0}^{nT-1}c_j\bigm|\gamma\bigr]\le m-1$ for every $n$.
Hence every $c_j$ vanishes almost surely: $\varphi$ is constant along each line parallel to $e$, and
$\lambda(\tau_{Te}X)=\lambda(X)$ almost surely. Ergodicity of $\theta_{Te}$ given $\gamma$, applied to the events
$\{\lambda(X)=s\}$, which are invariant up to null sets, makes $\lambda(X)$ a function $h(v',\mathcal G_e)$, and
$h_v=h(v',\cdot)$ at every cell $v$ of the line. \qed

\paragraph{Order parameters.}
The same argument needs no patch. Let $\Omega'$ be a measurable set of $K$-valid colorings with $\tau_e\Omega'=\Omega'$,
$V$ a finite set and $\sigma$ a bijection of $V$. A measurable map $\Phi:\Omega'\to V$ is an \emph{order parameter
along $e$} if $\Phi(\tau_ex)=\sigma(\Phi(x))$ for every $x\in\Omega'$. The value $\lambda$ above is one, with $\Omega'$
the $K$-valid colorings and $V$ the union of the finite orbits of $\sigma$ that meet the values of $\varphi$, but an
order parameter need not be a function of any finite patch.

\begin{proposition}[order parameters]\label{prop:order}
Let $X$ be as in Theorem~\ref{thm:shared} with $X\in\Omega'$ almost surely, and let $\Phi$ be an order parameter along
$e$ on $\Omega'$. Then $\Phi(X)=h(\mathcal G_e)$ almost surely for a measurable $h$ with $\sigma^T\circ h=h$.
\end{proposition}

\emph{Proof.} Let $D$ be the order of $\sigma$. The event $\{X\in\Omega'\}$ is invariant under $\theta_{Te}$, and on it
$\Phi(\tau_{DTe}X)=\sigma^{DT}(\Phi(X))=\Phi(X)$. As in the proof of part (a), ergodicity of $\theta_{DTe}$ given
$\gamma$ makes $\Phi(X)$ almost surely a constant $h(\gamma)$, and the invariance of the law of $\omega'$ under
$\theta_{Te}$ gives $\sigma^T(h(\gamma))=h(\gamma)$. \qed

\paragraph{SmileCity's lattice of faces.}
SmileCity's example has eight rows and eight columns. Its first and last rows and columns are black, and the six by six
cells inside them hold a face in which no cell is black. In a coloring, call a \emph{face run} a run of non-black cells
in a row with a black cell at each end; in the tiled example every face run is six cells wide and starts at a column of
one residue class modulo eight. Let $\Omega_1$ be the set of $K$-valid colorings that have a face run and in which all
face runs, in all rows, start at columns of one residue class modulo eight, and let $\Phi(x)\in\Z/8$ be that class, the
\emph{offset} of the faces of $x$ along rows. Write $e_1$ and $e_2$ for the unit steps along rows and along columns.
Translating a coloring along the rows moves its face runs and translating it along the columns does not, so $\Omega_1$
is invariant under both translations and
\[
\Phi(\tau_{e_1}x)=\Phi(x)-1,\qquad \Phi(\tau_{e_2}x)=\Phi(x)\qquad(x\in\Omega_1).
\]
So $\Phi$ is an order parameter along rows, with $\sigma$ the step $s\mapsto s-1$ of order eight, and along columns,
with $\sigma$ the identity. Over all $K$-valid colorings no finite patch fixes this lattice
(Appendix~\ref{app:smile}), which is why the certificate search cannot find it.

What both the coins shared along rows and those shared along columns determine is determined by the coins they have
in common.

\begin{lemma}\label{lem:meet}
Let $U$, $U'$ and $\Pi$ be independent random elements of standard Borel spaces, and $Z$ a random variable with
finitely many values. If $Z=a(U,\Pi)$ and $Z=b(U',\Pi)$ almost surely for measurable $a$ and $b$, then $Z=c(\Pi)$ almost
surely for a measurable $c$.
\end{lemma}

\emph{Proof.} The law of $(U,U',\Pi)$ is the product of the three laws, so by Fubini's theorem, for almost every value
$\pi$ of $\Pi$, $a(U,\pi)=b(U',\pi)$ almost surely. Fix such a $\pi$ and a value $s$. The events $\{a(U,\pi)=s\}$ and
$\{b(U',\pi)=s\}$ are independent, as $U$ and $U'$ are, and differ by a null set, so their common probability equals
its own square and is zero or one. Hence $a(U,\pi)$ is almost surely a constant $c(\pi)$, the value $s$ with
$\Pr[a(U,\pi)=s]=1$, which is measurable in $\pi$ by Fubini's theorem, and $Z=c(\Pi)$ almost surely. \qed

In the language of $\sigma$-fields, the $\sigma$-fields of $(U,\Pi)$ and of $(U',\Pi)$ meet, up to null sets, in the
$\sigma$-field of $\Pi$.

\begin{corollary}[SmileCity's lattice]\label{cor:smile}
On SmileCity, let $X$ be as in Theorem~\ref{thm:shared}, $K$-valid almost surely, and suppose that its faces lie on one
lattice along rows, that is, $X\in\Omega_1$ almost surely. Then the offset $\Phi(X)$ is almost surely a function of the
coins shared over the whole plane, and the period of $\Lambda$ along rows is a multiple of eight. Equivalently, the
residue modulo eight of the columns where face runs start, counted from a cell $v$, is a function of those coins and of
$v$. A generator that reads no coin shared over the whole plane puts its faces at the same offset in every world.
\end{corollary}

\emph{Proof.} By Proposition~\ref{prop:order} along rows, $\Phi(X)=h_1(\mathcal G_{e_1})$ with $\sigma^{T_1}\circ h_1=h_1$, that is,
$h_1-T_1\equiv h_1$ modulo eight, where $T_1$ is the period of $\Lambda$ along rows, so eight divides $T_1$. By
Proposition~\ref{prop:order} along columns, $\Phi(X)=h_2(\mathcal G_{e_2})$. Every family of coins has infinite extent
along rows only, along columns only, along both, which makes its blocks the whole plane, or along neither. Let $U$,
$U'$ and $\Pi$ collect the coins of the families of the first three kinds. They are independent, $\mathcal G_{e_1}$ is
generated by $(U,\Pi)$ and $\mathcal G_{e_2}$ by $(U',\Pi)$, so Lemma~\ref{lem:meet} with $Z=\Phi(X)$ gives
$\Phi(X)=c(\Pi)$. The offset counted from $v$ is $\Phi(X)$ minus the column of $v$, modulo eight. Without coins shared over the plane, $\Pi$ is trivial and $\Phi(X)$ is a constant. \qed

The hypothesis that the faces lie on one lattice cannot be dropped, since validity alone does not put them there. The
example's black first and last rows span the whole width, so a window that meets the two black rows between
consecutive rows of faces sees at most one row of faces, and is a window of the tiled example at that row's shift.
Shifting each band of eight rows of the tiled example, the rows
$8b$ to $8b+7$, by its own amount along the rows therefore leaves every window allowed. Drawing the shift of each band
from a coin of the band gives a generator in the general form whose output is $K$-valid for every value of its coins
and has the example's statistics exactly, and whose offset varies from band to band with coins shared along rows.
Appendix~\ref{app:smile} finds with a complete solver the same freedom between strips of faces separated by two black
rows across the width, and another along a row, through a column of faces three cells wide that never closes. On the
colorings in which the face runs of row $y$ share one class modulo eight, that class is an order parameter along rows,
so for a generator whose worlds are almost surely such colorings it is a function of $\mathcal G_{e_1}$ and of $y$ by
Proposition~\ref{prop:order}, but it need not be the same on every row.

\paragraph{Rigid layouts.}
Part (a) and Proposition~\ref{prop:order} constrain a generator that shares nothing along $e$, and they constrain
phases whatever is shared.

\begin{corollary}[rigid layouts]\label{cor:rigid}
Let $X$ be as in Theorem~\ref{thm:shared}, $K$-valid almost surely. (a) If $X$ reads no coin shared along $e$, then at
every cell every transported or one-way quantity along $e$ almost surely takes a value fixed in advance, the same in
every world, and this layout is periodic under $\Lambda$; every order parameter along $e$ is almost surely a constant.
(b) Under the full labeling, almost surely every patch of $X$ lies in a component whose $d_c$ divides $T$. When $P=N$,
the total variation between the window statistics of $X$ and $f$ is therefore at least the example's share of windows
in the other components.
\end{corollary}

\emph{Proof.} (a) With $\mathcal G_e$ trivial the functions $h_v$ of part (a) and $h$ of Proposition~\ref{prop:order}
are constants. For $t\in\Lambda$, $\varphi(Q^P_{v+t}(X))=\varphi(Q^P_v(\tau_tX))$, and $\tau_tX=X(\theta_t\omega)$ has
the law of $X$, so $h_{v+t}=h_v$. (b) By part (a) of Theorem~\ref{thm:shared}, at every cell the value $(c,z)$ of the
full labeling is almost surely fixed by $\sigma^T$, which adds $T$ to the level modulo $d_c$, and $d_c\ge1$ on a line
free of forbidden windows; so $d_c$ divides $T$. When $P=N$ the window law of $X$ at every cell puts no mass on the
windows of the other components, while $f$ puts the example's share on them. \qed

Part (a) is why a generator given the phase channels and no coin shared along the phase's direction lays the phase
out the same way for every seed. Part (b) is Proposition~\ref{prop:blocks}(c) without a radius. On Fabric, whose phase period \phaseDFabric\ does not divide the octaves' period $T=\coinPeriod$, the components
whose period does not divide $T$ hold \fabricPhaseMass\ of the example's windows (Appendix~\ref{app:scopefloors}), so a
generator whose lattice $\Lambda$ has that period along the phase's direction and whose output is free of forbidden
windows, at any depth, has window statistics at least that far from the example's.

\subsection{\texorpdfstring{Theorem~\ref{thm:glue}: when the row types are enough}{Theorem 4}}
\label{app:cvglue}

Theorem~\ref{thm:glue} gives a generator free of forbidden windows that reads only the coin from which its band's row
types are drawn and its position.

\emph{Rows and strips.} Let the example have $H$ rows and $W$ columns, read with wrap-around, and let $Y_c$,
$c\in\Z/H$, be its rows, each a sequence of period $W$. The alphabet $\mathcal R$ consists of the rows $\tau^sY_c$ of
the example at every horizontal shift $s\in\Z/W$, at most $HW$ rows. A sequence of rows of $\mathcal R$ is \emph{valid}
if every $N$ consecutive rows form a strip all of whose windows lie in $K$, a finite condition since the rows share the
period $W$. In a valid sequence the strip of $P$ rows that starts at a given row is locally valid at every column, and
its $P\times P$ patches are consecutive patches along rows, so they lie in one weak component of $G_{P,e}$, the strip's
\emph{type}. As in Section~\ref{sec:cert}, the type of row $y$ of a coloring is the type of the strip of $P$ rows from
$y-j_0$, where $j_0=\lfloor P/2\rfloor$.

\emph{The sampler's sequences.} Band $b$ holds the rows $bL$ to $bL+L-1$. The sampler draws for it the types
$t_0,\dots,t_{L+1}$ of the rows $bL$ to $bL+L+1$, with $(t_0,t_1)$ and $(t_L,t_{L+1})$ both the anchor $\alpha$, every
three consecutive types shown by the example and every free run within the run limits (Appendix~\ref{app:procedure}).
Consecutive bands agree on the two rows they share, which both give the anchor. Let $\mathcal W$ be the finite set of
type sequences that the sampler can draw for a band.

\emph{Realizations.} A \emph{realization} of $w=(t_0,\dots,t_{L+1})\in\mathcal W$ is a valid sequence
$R_0,\dots,R_{L+P-2}$ of rows of $\mathcal R$ in which, for $j=0,\dots,L-1$, the strip of $P$ rows that starts at $R_j$
has type $t_{j+j_0}$. Placed at the rows $bL$ to $bL+L+P-2$, the strip that starts at row $bL+j$ is the strip from
$y-j_0$ for $y=bL+j+j_0$, so the realization gives every row $y$ from $bL+j_0$ to $bL+L-1+j_0$ the type $t_{y-bL}$ that
$w$ assigns it. The rows $bL$ to $bL+j_0-1$ take their types from the realization of band $b-1$, whose sequence gives
them the anchor's types, as $w$ does. This indexing needs $t_{L-1+j_0}$ to lie in the sequence, that is $j_0\le2$, which
holds for $P\le5$ and so on every example of the corpus, read at $P=3$ or $P=4$; a larger $P$ would need an anchor
that fixes $j_0$ consecutive types.

The example's rows \emph{glue} if there is a strip $Z=(Z_0,\dots,Z_{P-2})$ of rows of $\mathcal R$ such that every
$w\in\mathcal W$ has a realization with $R_0,\dots,R_{P-2}$ equal to $Z$ and $R_L,\dots,R_{L+P-2}$ equal to $Z$. The rows
of $Z$ carry their shifts, so $Z$ is the same at both ends at the same shift. Shifting a realization along the rows
gives a realization of the same sequence, since neither $K$ nor the types change, so pinning $Z$ at one end costs
nothing; pinning it at both ends is the condition.

\emph{Proof of Theorem~\ref{thm:glue}.} Let the rows glue with the strip $Z$, and let $L\ge P-1$, as on every example
of the corpus (Table~\ref{tab:glue}). The bands are the blocks of a lattice fixed in absolute
coordinates, groups of $L$ consecutive rows, so their coins form a family of infinite extent along rows. From the coin
$\omega_b$ of band $b$ the sampler draws $w^{(b)}\in\mathcal W$, and a fixed rule, deterministic or using further coins
of the same band, chooses a realization $\rho^{(b)}=(R^{(b)}_0,\dots,R^{(b)}_{L+P-2})$ of $w^{(b)}$ that begins and ends
with $Z$, which exists because the rows glue. The output is $X_{(x,y)}=R^{(b)}_{y-bL}(x)$ for $bL\le y<bL+L$. It
depends only on the coin of the band that holds the cell and on $(x\bmod W,\,y\bmod L)$, a function of the position
modulo any common multiple $p$ of $W$ and $L$, so $X$ is a random-access generator in the general form, of radius zero.

The rows $(b+1)L$ to $(b+1)L+P-2$ lie in band $b+1$, since $L\ge P-1$, and there the output is
$R^{(b+1)}_0,\dots,R^{(b+1)}_{P-2}$, which is $Z$, which is $R^{(b)}_L,\dots,R^{(b)}_{L+P-2}$. So the output agrees
with $\rho^{(b)}$ on all the rows $bL$ to $bL+L+P-2$. A window whose top row $y$ lies in band $b$ occupies the rows $y$
to $y+N-1\le bL+L+N-2\le bL+L+P-2$, so it is a window of the valid sequence $\rho^{(b)}$ and lies in $K$, for every
value of the coins. The same comparison applies to the strips of $P$ rows that start in band $b$, so every row $y$ from
$bL+j_0$ to $bL+L-1+j_0$ has the type that $w^{(b)}$ gives it, and the row types of the output are the sampled ones.
\qed

\paragraph{The statistics of a generator that glues rows.}
Suppose the rule records, with every row it outputs, the row $c$ of the example and the shift $s$ it copies, so that
the row is $\tau^sY_c$. A \emph{stretch} is a maximal run of output rows that copy consecutive rows $c,c+1,\dots$ of the
example at one shift, and a \emph{cut} is the boundary between two consecutive stretches. A cut is \emph{free} if every
strip of $N$ rows across it is nonetheless a strip of the tiled example up to a horizontal shift; a cut with $N-1$
copies of one row on each side is free when the example repeats that row $N$ times in a row, as the skylines repeat a
row of sky. Let $\rho_{\mathrm{cut}}$ be the expected share of rows $y$ whose strip of $N$ rows from $y$
crosses a cut that is not free. Every other row $y$ starts a strip of $N$ rows that is a strip of the tiled example up
to shift; let $\nu(c)$ be the expected share of rows whose strip is the example's strip from row $c$, taking the least
such $c$ when several qualify, and $\bar\nu=\nu/(1-\rho_{\mathrm{cut}})$ the \emph{row usage}. Shares are taken over
the rows of a band, and $U_H$ is the uniform law on the $H$ rows of the example.

\begin{proposition}[statistics]\label{prop:gluestats}
The window statistics $h_X$ of such a generator satisfy
$\|h_X-f\|_{\mathrm{TV}}\le\|\bar\nu-U_H\|_{\mathrm{TV}}+\rho_{\mathrm{cut}}$. When $P=N$, every generator whose output
is $K$-valid with the sampled row types satisfies $\|h_X-f\|_{\mathrm{TV}}\ge\|\pi_{\mathrm S}-\pi_f\|_{\mathrm{TV}}$,
where $\pi_{\mathrm S}$ gives the expected share of each type among the rows of a sampled band and $\pi_f$ its share
among the rows of the example.
\end{proposition}

\emph{Proof.} The law of $X$ is invariant under translation by $L$ rows and by $p$ columns, so its window statistics are
the window law averaged over the rows of one band and the columns of one period, a multiple of $W$. For $c\in\Z/H$ let
$f_c$ be the uniform law on the $W$ windows of the example's strip from row $c$, so that $f=\frac1H\sum_cf_c$. A row $y$
whose strip is the example's strip from row $c$ at some shift contributes, over the columns of a period, the windows of
that strip, each equally often, that is, the law $f_c$. Hence $h_X=\sum_c\nu(c)f_c+\rho_{\mathrm{cut}}\,g$ for some law
$g$, and
\[
\begin{aligned}
\|h_X-f\|_{\mathrm{TV}}&\le\frac12\sum_c\Bigl|\nu(c)-\frac1H\Bigr|+\frac{\rho_{\mathrm{cut}}}2\\
&\le(1-\rho_{\mathrm{cut}})\,\|\bar\nu-U_H\|_{\mathrm{TV}}+\rho_{\mathrm{cut}},
\end{aligned}
\]
using $|\nu(c)-\frac1H|\le|\nu(c)-\frac{1-\rho_{\mathrm{cut}}}H|+\frac{\rho_{\mathrm{cut}}}H$. When $P=N$, the strip of $P$
rows from $y$ is the strip of the windows with top row $y$, so the type of row $y+j_0$ is the weak component of any of
these windows, a function of the window. Under $h_X$ this function has the law that gives each type its expected share
among the rows of $X$, which is $\pi_{\mathrm S}$ when the row types are the sampled ones, and under $f$ it has the law
$\pi_f$. Total variation does not grow when a function is applied to both laws, which gives the lower bound. \qed

The distance of such a generator from the example's statistics thus has two sources: how often it uses each row of the
example, which the sampled types largely fix, and the cuts that are not free. Where the sampler's law of types differs
from the example's, as when the run limits draw free runs of lengths the example does not show, no generator that
follows the sampled types comes closer to $f$ than the type marginals allow, however it fills the rows.

\subsection{Deciding whether the rows glue}
\label{app:cvdecide}

Whether the rows glue with a given strip $Z$ is an inclusion between two regular languages, which a subset
construction decides; the rows glue if the inclusion holds for one of the finitely many strips. The
\emph{realization automaton} has as states the sequences of $P-1$ rows of $\mathcal R$; from a state it may append a row
when the last $N$ rows form a valid strip, and it then emits the type of the strip of $P$ rows that the appended row
completes. The sampler's automaton is the one the method runs, with its anchor and run limits (Table~\ref{tab:course}): its
states are the pairs of consecutive types, with a counter for the length of the current free run. For a candidate strip $Z$, the computation follows a band row by row. A configuration holds the
sampler's state and counter after the types drawn so far, and the set of states of the realization automaton that some
partial realization of those types, beginning with $Z$, can reach. It starts from the anchor and the set $\{Z\}$, and at
each row, for each configuration and each step the sampler can take, it keeps the states reached by appending a row
whose strip has the type that the sampler requires. Configurations that agree in every part evolve alike and are merged,
and the set in the configuration that a sequence of types reaches is exactly the set of last $P-1$ rows of the partial
realizations of that sequence that begin with $Z$. The rows glue with $Z$ if and only if the set of every
configuration at the end of the band contains $Z$; since an empty set stays empty, the computation stops with a
negative answer as soon as a configuration whose sampler state can still complete a sequence of $\mathcal W$ has an
empty set. A configuration consists of a sampler state, a counter bounded by the run limit and a subset of
a finite set, so at each of the $L+P-1$ rows of a band there are finitely many, and the computation ends with an exact
answer.

The candidates for $Z$ are the example's own $P-1$ rows at each place where the example shows the anchor, at shift
zero. A realization over the example's rows at shift zero is also a realization over $\mathcal R$ with the same $Z$, so
when the check holds with the smaller alphabet it holds with $\mathcal R$. A positive answer for one candidate settles the condition. A negative answer in general needs every strip of $\mathcal R$
checked as a candidate separately: one strip must begin and end a realization of every sequence the sampler can draw,
and a construction started from a set of strips forgets which strip each partial realization began with, so it cannot
tell whether a single strip serves every sequence. Started from the set
of all strips at once, it can still give a stronger negative witness: if some sequence the sampler can draw empties the
set, that sequence has no realization from any strip, and the rows glue with no strip. Any finite alphabet of rows can take the
place of $\mathcal R$, the rows of a solver's colorings for instance, without other change.

\begin{table*}[!htbp]\centering\footnotesize
\caption{The gluing check on the \rowInvCountWords\ examples with a row invariant, run after the copier's results were
known. For each example: the patch size $P$ of the row types, and the band length $L$ and the anchor of the row-type
sampler, as in Table~\ref{tab:course}; the alphabet on which the check was decided, the example's rows at shift zero
where they glue there, which implies that they glue over all shifts, and otherwise the rows at every horizontal shift;
whether the rows glue; and the size of the computation, namely the configurations reached over the band, the most at
any one row, and the largest set of states of the realization automaton in one configuration, each the largest over
the candidate strips $Z$. The last row is Wall with its shifts restricted to multiples of three, where every strip of
three rows that the check explores is a strip of the tiled example. On 3Bricks the sizes are those of the run from the
candidates, and the negative answer rests on the run from the set of all strips at once (see the text).}
\label{tab:glue}
\input{tables/glue_v16}
\end{table*}

We ran the check after the copier's results were known, so it explains them rather than predicting them. The rows glue
on \glueHoldsWords\ of the \rowInvCountWords\ examples with a row invariant (Table~\ref{tab:glue}), the examples on
which the copier exists, and on \glueShiftZeroWords\ of them they glue already at shift zero. Configurations stay few:
at most \glueConfigsMax\ at any one row, and at most \glueConfigsTotalMax\ over a whole band. On Wall the rows glue only with their shifts: over the rows at shift zero,
some sequences of courses cannot end on the strip they began with, while the shifts let every band return to it, as the
copier's offsets, chained within a band toward the next band's first offset, do (Appendix~\ref{app:setup}). They glue
on Wall even with the shifts restricted to multiples of three, and then every strip of three rows in a realization is a
strip of the tiled example, so a generator built on these realizations has only free cuts, and by
Proposition~\ref{prop:gluestats} its distance from the example's statistics comes from its row usage alone. On 3Bricks
the rows glue with no strip. Started from the set of all valid strips of three rows of $\mathcal R$ at once, the
construction finds that no realization carries more than \glueBricksFreeRunWords\ consecutive rows of the free type, the length of
the example's own free run, while the sampler draws free runs of up to \glueBricksRunLimitMax\ rows. The set empties on the
same beginning of a band as from the candidates, a free run one row longer than the example's, so that sequence has no
realization from any strip. Of the type sequences
the sampler can draw for a band, only \glueBricksBandWalksWithRealizationWords, whose free runs all have the example's
length, has a realization from some strip. The complete solver colors a free run of every length that the sampler draws there, as the record
of expectations notes (Appendix~\ref{app:record}), with rows that are not rows of the example, so free runs of every
drawn length exist over other rows; whether some larger finite set of rows glues, with one pinned strip, is open.

On the \glOneCountWords\ walls we drew or transposed for the held-out test (Appendix~\ref{app:heldout}), the check ran
first and its result was recorded before any copier was built. The rows glue on each, and the copier then left no
forbidden window on any of their \glOneCanvasesWords\ canvases, three seeds each, before any mending. On the wall drawn
with mortar two rows high, no window of three rows meets two courses, so every joint allows every offset, and there the
copier draws each stretch's offset from its own coin without chaining.

\subsection{Why mixing alone does not suffice}
\label{app:cvmixing}

A mixing condition on the sequence of rows cannot take the place of gluing for generators that commute with unit
shifts along the columns, which read neither their row nor coins on a fixed grid of row blocks. The proper colorings of $\Z$ with three
colors form a shift of finite type that is strongly irreducible, yet no block factor of i.i.d.\ coins is a proper
coloring of $\Z$; \citet{holroyd2016finitely} use this to separate block factors from finitely dependent processes, and
a block factor can make its errors rare as its radius grows but cannot remove them. If the constraint between consecutive rows were of this kind, a generator that commutes with unit shifts along the
columns and reads only independent coins of bounded extent along them would therefore leave defects at every radius. Reading the position, or coins on a fixed grid of blocks that reveal it, removes this obstruction but not the
cost of Theorem~\ref{thm:shared}: the coloring
$X_y=y \bmod 3$ is proper and uses no coins, and it lays the rows out the same way in every world. The generator of
Theorem~\ref{thm:glue} escapes both. It computes a whole band from one coin, which it may do because the
band's coin is shared along rows, so the rows can differ between worlds; and its bands sit on a fixed grid, each
beginning and ending with the pinned strip $Z$, which breaks the commutation with unit shifts along the columns that the
obstruction needs. Gluing is the condition under which bands computed separately agree where they meet. Along columns the band's coin is a block of extent $L$, so
Theorem~\ref{thm:shared}(b) along columns still binds every quantity carried along columns, with $\ell$ at least $L$.

%% file: tables/glue_v16.tex
\begin{tabular}{@{}lrrlllrrr@{}}\toprule
example & $P$ & $L$ & anchor & alphabet & glue & \begin{tabular}[b]{@{}r@{}}configu-\\rations\end{tabular} & \begin{tabular}[b]{@{}r@{}}most at\\one row\end{tabular} & \begin{tabular}[b]{@{}r@{}}largest\\set\end{tabular}\\\midrule
Wall & 3 & 32 & most frequent pair & all shifts & yes & 174 & 7 & 18\\
SimpleWall & 3 & 48 & start of a free run & shift 0 & yes & 2{,}970 & 120 & 3\\
3Bricks & 4 & 48 & end of a free run & all shifts & no & 64 & 3 & 992\\
Flowers & 3 & 112 & end of a free run & shift 0 & yes & 2{,}116 & 36 & 22\\
MoreFlowers & 3 & 112 & end of a free run & shift 0 & yes & 2{,}116 & 36 & 21\\
Platformer & 3 & 1{,}040 & end of a free run & shift 0 & yes & 17{,}425 & 33 & 80\\
Skyline & 3 & 176 & end of a free run & shift 0 & yes & 3{,}213 & 39 & 72\\
Skyline2 & 3 & 512 & end of a free run & shift 0 & yes & 8{,}449 & 33 & 38\\
\midrule
Wall, free joints & 3 & 32 & most frequent pair & shifts 0, 3, 6 & yes & 174 & 7 & 6\\
\bottomrule
\end{tabular}

%% file: app_procedure.tex
\section{The decision procedure and its constants}
\label{app:procedure}

The steps below run in this order on every example and fix every constant of the method; Table~\ref{tab:course} gives
what the row-type sampler reads from each example it serves.

\begin{enumerate}
\item \emph{Certificate search.} For every patch size $P$ from $N$ to \certPatchMax\ and each of four directions (rows,
  columns, the diagonal and the antidiagonal), the search of Section~\ref{sec:cert} looks for a quantity carried by the
  locally valid $P\times P$ patches. It stops at the first patch size whose locally valid patches number more than
  \certCap, and its reach is the patch size before. The first certificate, under a tie-break fixed before any result was
  read, has the smallest $P$, then the largest spread of the quantity, $\kappa_f=1-\max_s f(\varphi=s)$ of
  inequality~\eqref{eq:vfr}, where $f(\varphi=s)$ is the share of the example's positions at which the quantity takes
  the value $s$, then a phase before an invariant, then the directions in the order above.
\item \emph{Input.} An invariant along rows gives the row-type sampler, a phase of period $d$ the phase channels of
  period $d$, and a quantity that changes only one way, or no certificate, the octaves alone; the octaves are part of
  every input. The row-type sampler goes to \methodCourseNames, the phase channels to \methodPhaseNames, and the octaves
  alone to the other \methodNOct\ public examples. The sampler reads the row types at the certificate's patch size $P$.
  On 3Bricks, whose row invariant appears first at patch size \bricksCertP, every row read at patch size $N$ has the
  same type, so a sampled sequence of types would be constant; read at \bricksCertP, its rows take \bricksTypesWords\
  types.
\item \emph{Row types and states.} The type of row $y$ is the certificate's label of the strip of $P$ rows from
  $y-\lfloor P/2\rfloor$, where $P$ is the certificate's patch size (rows $y-1$ to $y+1$ when $P=3$). The states are the
  pairs of types of two consecutive rows that the periodic example shows, and a state $(a,b)$ steps to $(b,c)$ when the
  example shows the three types $a,b,c$ on consecutive rows. A type $a$ is free when the state $(a,a)$ steps to itself;
  $j$ consecutive states $(a,a)$ make a free run of $j+1$ rows of $a$. The rule reads the state of its row as one-hot
  channels.
\item \emph{Run limits.} The automaton read from the windows allows free runs that no coloring shows. On Skyline,
  Skyline2 and Platformer it allows runs of \courseAutomatonRun\ rows, while the complete solver colors no open strip
  with a free run shorter than \courseSolverRunSkyline\ rows on Skyline, \courseSolverRunSkylineTwo\ on Skyline2 and
  \courseSolverRunPlatformer\ on Platformer. Every free run is therefore held between the larger of the solver's least
  run and the example's shortest run, and the larger of the example's longest run and \courseRunMaxRows\ rows, with each
  run counted exactly in the states of the walk. The upper limit is \courseRunMaxPairs\ repeated states, $R-2$ for the
  rule's $R=\mainR$ rounds, so the two rows that bound a free run lie at most $R$ rows apart.
\item \emph{Anchor and band length.} The anchor is the example's most frequent state that does not step to itself, with
  ties going to the lowest index: on the skylines the last free row above the ground, on Wall the top row of a course.
  The band length $L$ is the smallest multiple of \courseBandMult, the octaves' coarsest period, at or above the
  example's height and at or above the least $n$ for which, under the run limits, every state reaches every state in
  exactly $n$ steps. $L$ has no cap, and drawing a band takes work in proportion to it: on Platformer, whose bands hold
  \platBand\ rows, the type of a row costs at most a walk of \platBand\ steps, and that walk gives the type of every row
  of its band at once.
\item \emph{Walks.} Band $b$ holds rows $bL$ to $bL+L-1$. Row $bL$ takes the anchor, and the next $L-1$ rows take a walk
  that steps back to the anchor at row $(b+1)L$, so the anchor falls on the same rows, the multiples of $L$, in every
  world of an example. A step between two states has weight equal to the number of times the example shows that step,
  divided by the number of times it shows the first state; the weights are held as integers after scaling by their least
  common denominator. A backward pass sums the weights of all walks of $k$ steps from each state to the anchor, and at
  each step the next state is drawn with probability proportional to its weight times the summed weight of its
  completions, by an exact comparison with a \courseHashBits-bit uniform number made from two outputs of the coins'
  integer hash of the seed, the band's index and the step. The state of any row is computed from the seed with work
  bounded by $L$.
\item \emph{Row-color term.} For the first \courseWarmFracWords\ of the \msSteps\ updates, that is, before update
  \courseWarmUntil, the loss gains, for every row of the training canvas, the cross entropy between the example's color
  histogram over the rows of that row's type and the rule's color distribution on the last round, averaged along the
  row, at the training temperature. Its weight is fixed at the first update so that the term equals the rest of the loss
  there. From update \courseWarmUntil\ on the term is absent.
\item \emph{Phase channels.} One-hot channels of $x \bmod d$ and of $y \bmod d$ in absolute coordinates, with
  $d=\phaseDFabric$ on Fabric, \phaseDRedMaze\ on RedMaze, \phaseDSimpleMaze\ on SimpleMaze and \phaseDSmileCity\ on
  SmileCity.
\item \emph{Octaves.} Each cell has \coinFlatWords\ coins of its own and \coinPerPairWords\ coins for each of the
  \coinPairs\ scale pairs $(i,j)$ with $i,j\in\{\coinScaleList,\infty\}$ other than $(0,0)$, \coinChannels\ channels in
  all.
\item \emph{Rule.} State width \ruleWidth, \mainR\ rounds to generate, and \msSteps\ updates, each drawing its number of
  rounds from \trainRoundsLo\ to \trainRoundsHi; the other weights and schedules are in Appendix~\ref{app:training}.
\item \emph{Variation term.} At every update its weight makes it \varFracWords\ of the loss without it, on every
  example. On the rigid examples (Appendix~\ref{app:rigid}), where every valid world is a translate of the tiling, the
  term cannot fall without a defect, and it is kept there at the same share.
\item \emph{Repair.} The rule's output passes through \repairRoundsWords\ coin-gated repair rounds.
\item \emph{Mending.} Mending radius $\rho_{\max}=\mendRadiusMain$ and blocks of side $B=\mendBlockMain$, in
  \mendPassesWords\ passes with the lattice shifted by half a block between them and \mendClassesWords\ classes of
  blocks in each pass. A block that cannot be completed as a whole is split into clusters, its defective cells each
  grown by $N-1$ cells.
\end{enumerate}

\begin{table*}[!htbp]\centering\footnotesize
\caption{The row-type sampler on the examples the procedure gives it: the patch size $P$ at which the row types are
read, the number of row types and of states, where the anchor sits, the limits of a free run in rows, the band length
$L$, the base-2 logarithm of the number of walks a band can take, and the anchor's spacing in the example and in sampled
bands (median, with the smallest and largest, over \courseSampleBands\ bands on each of \courseSampleSeedsWords\ seeds).
Wall has no free type.}
\label{tab:course}
\input{tables/course_v14}\end{table*}

Where the run limits leave few run lengths, the sampler leaves little more than an order: Platformer's band of
\platBand\ rows holds \platCoursesEachWords\ stretches of \platCourseShort\ rows from one visit of the anchor to the
next and \platCoursesEachWords\ of \platCourseLong, so its walks differ only in the order of these stretches. No sampled
free run is longer than \courseRunMaxRows\ rows on any example, and within that limit runs longer than the example's own
do appear: SimpleWall's example shows free runs of \simpleWallExampleRuns\ rows, and its sampled runs reach
\simpleWallRunMax.

\paragraph{How the search and the procedure were arrived at.}
The search as first fixed looked for invariants and phases only, under a coarser labeling that pooled every patch
outside the certified component into one value, and it certified the same examples except Town; the one-way quantity,
which certifies Town alone, and the full labeling were added after that run (Appendix~\ref{app:cert}). The examples
shaped parts of the procedure, and each part falls in one of two groups. Fixed before any training were
the reading of the row types from the certificate's strips, the automaton and its transition weights, the band length, and
the lower run limit set by the complete solver. Added after a diagnosed failure were the phase channels at the certified
period, after channels of the coordinates modulo powers of two left Fabric, whose period is \phaseDFabric, invalid; the
row-color term, after the lagged reading on Wall; the anchor at a state that does not step to itself, the upper run
limit and the example's shortest run as a second lower limit, after Skyline and Skyline2 failed at the feet of
buildings under free runs longer than the rule's radius, and Skyline also under runs shorter than the example's; and the
reading of the types at the certificate's own patch size, after the reading at patch size $N$ gave 3Bricks, certified at
patch size \bricksCertP, one row type, before any rule of 3Bricks was trained on the sampler. The procedure as it then
stood was applied unchanged to the other examples with a row invariant, to the other phase examples, and to three walls
and two mazes that had not shaped it (Section~\ref{sec:heldout}). Appendix~\ref{app:record} lists every step with what we
expected and what happened.

\paragraph{Why the input follows the certificate.}
Where the octaves could carry the quantity (Section~\ref{sec:theory}), a rule must learn to extract it, and a rule told
its position outright by channels of the coordinates extracted it on the mazes and on Wall but painted the same raw
canvas for every seed (Appendix~\ref{app:sharing}). The certificate names the quantity itself, so the input gives that
and leaves the rest to the coins. No input was built for a one-way quantity: the search had found one on Town before any of these parts
existed, and the octave rule was already valid there. In a stationary world free of forbidden windows a one-way
quantity is constant along the lines of its direction (proof of Theorem~\ref{thm:nogo}), so its values could be
supplied as row types are.

%% file: tables/course_v14.tex
\begin{tabular}{@{}lrrrlcrrcc@{}}\toprule
example & $P$ & \begin{tabular}[b]{@{}r@{}}row\\types\end{tabular} & \begin{tabular}[b]{@{}r@{}}pair\\states\end{tabular} & anchor & \begin{tabular}[b]{@{}c@{}}free run\\(rows)\end{tabular} & $L$ & \begin{tabular}[b]{@{}r@{}}$\log_2$ of walks\\per band\end{tabular} & \begin{tabular}[b]{@{}c@{}}spacing in\\the example\end{tabular} & \begin{tabular}[b]{@{}c@{}}sampled\\spacing\end{tabular}\\\midrule
Wall & 3 & 4 & 5 & most frequent pair & -- & 32 & 5.2 & 4, 5 & 5 (4--5)\\
SimpleWall & 3 & 4 & 5 & start of a free run & 2--31 & 48 & 16.5 & 5, 6 & 5 (5--14)\\
3Bricks & 4 & 2 & 3 & end of a free run & 7--31 & 48 & 10.8 & 8 & 12 (8--32)\\
Flowers & 3 & 5 & 6 & end of a free run & 20--31 & 112 & 9.7 & 24 & 28 (24--35)\\
MoreFlowers & 3 & 5 & 6 & end of a free run & 20--31 & 112 & 9.7 & 24 & 28 (24--35)\\
Platformer & 3 & 3 & 4 & end of a free run & 30--31 & 1{,}040 & 29.2 & 32 & 33 (32--33)\\
Skyline & 3 & 4 & 5 & end of a free run & 25--31 & 176 & 10.3 & 28 & 29 (28--34)\\
Skyline2 & 3 & 3 & 4 & end of a free run & 29--31 & 512 & 22.3 & 31 & 32 (31--33)\\
\bottomrule
\end{tabular}

%% file: app_method.tex
\section{The learned automaton and the mender}
\label{app:method}
\label{app:params}

\subsection{The rule and its objective}

\paragraph{The rule.}
The rule is a neural cellular automaton \citep{mordvintsev2020growing} with a radius-one update and a state $s$ of fixed
width per cell. A cell's input $\omega$ (its own coins, the octaves, and the row types or the phase channels where the
procedure gives them) enters through one learned $1\times1$ map $A$. The state starts at $s_0+a\tanh(A\omega)$, with
$s_0$ a learned vector, and each round sets $s\leftarrow s+\tanh\big(W_2\,\mathrm{silu}(W_1*s)\big)+b\tanh(A\omega)$,
where $W_1$ is a $3\times3$ convolution, silu the sigmoid-weighted linear unit, $W_2$ a $1\times1$ convolution, and $a$
and $b$ constants (Appendix~\ref{app:training}). A linear output layer gives $C$ color logits per cell. The update is
deterministic and unpadded, so each round shrinks the computed canvas by one cell on each side, and after $R$ rounds the
output on a region depends only on the input within $R$ cells of it: the rule is a random-access generator of radius $R$
on its input. The same weights run for any number of rounds, and we generate with \mainR\ rounds.

\paragraph{Objective.}
Training draws random seeds and random absolute positions, so that block boundaries fall anywhere on the canvas, and
asks the output of every round for a valid coloring with the example's statistics. Let $p_t$ be the distribution over
colors that round $t$ gives a cell, the softmax of its logits at a temperature lowered over training, and let the
\emph{hard output} of round $t$ give each cell its color of largest logit. Terms on the hard output use the
straight-through form $e_t+p_t-\operatorname{sg}(p_t)$, where $e_t$ is the one-hot vector of the cell's hard color and
$\operatorname{sg}$ stops the gradient, so they are exact in the forward pass and pass their gradient through $p_t$.
Update $u$ runs the rule for $R_u$ rounds, drawn uniformly from \trainRoundsLo\ to \trainRoundsHi, and minimizes
\begin{equation}\label{eq:objective}
\begin{aligned}
\mathcal L={}&\sum_{t=1}^{R_u}w_t\Big[S_t+\lambda_F\,\mathrm{KL}(f\,\|\,\hat f_t)+\lambda_H H_t\\
&\quad+\mathbf 1[u\ge u_0]\big(\lambda_E E_t+\lambda_M\,\mathrm{KL}(\nu\,\|\,\hat\nu_t)\big)\Big]\\
&+\lambda_{\mathcal V}\,\mathcal V+\mathbf 1[u<u_0]\,\lambda_{\mathcal C}\,\mathcal C,
\end{aligned}
\end{equation}
with round weights $w_t=t/\sum_{s=1}^{R_u}s$, which grow linearly and sum to one. The soft term $S_t$ is minus the mean
log-probability that a window is allowed under the product of the per-cell distributions, and $H_t$ is the mean entropy
of $p_t$; both stay on throughout. The two divergences are Kullback--Leibler divergences with the example's distribution
first: its window frequencies $f$ against the output's histogram $\hat f_t$ of the windows in $K$, and its color
marginal $\nu$ against the hard output's color histogram $\hat\nu_t$. The rate $E_t$ is the share of the hard output's
windows that are forbidden. In the \emph{warm-up}, the updates before $u_0$, $\hat f_t$ is the expected histogram under
the per-cell distributions and the hard terms are off; from $u_0$ on, $\hat f_t$ is the hard output's histogram and the
hard terms are on. The variation term $\mathcal V$ and the row-color term $\mathcal C$ are defined below, and
Appendix~\ref{app:training} gives the weights and schedules.

The soft term gives a dense gradient from the first update, but alone it is gamed: most rules trained on it collapsed to
one constant valid pattern that ignores the coins (Section~\ref{app:persample}). A spread distribution cannot lower the
hard terms, and the color-marginal term penalizes a constant output even when all of its windows are allowed. Weighting
the later rounds more and drawing the round count afresh at every update train the rule as a repair protocol over the
whole range of round counts.

\paragraph{The variation term.}
The variation term guards against a shortcut that shared coins open and the rest of the objective does not rule out:
the rule could copy the example on each coarse block and ignore its fine coins, which Proposition~\ref{prop:blocks}(b)
shows to be valid and faithful away from the blocks' boundaries, though it varies only from block to block. Under the
row types the copy can be exact, with stretches of the example between two anchors (Section~\ref{sec:learning}), and
Appendix~\ref{app:sharing} reports what a rule trained without the term does. To make the fine coins matter, each update
draws, for every canvas of the batch, a second input with the same seed and position, the coarse channels unchanged
and the fine channels hashed from a fresh seed, and runs the rule on it for the same number of rounds. The fine
channels are the cell's own coins and the \coinFinePairs\ pairs with one scale $0$ and the other finite, whose blocks
are one cell wide; the other \coinCoarsePairs\ pairs are coarse, and so are the row types and the phase channels where
the rule reads them. The \emph{variation term} $\mathcal V$ is the probability that the two copies give a cell the same
color, $\sum_c p(c)\,p'(c)$ under the last round's per-cell distributions at the current temperature, averaged over
cells and canvases, with gradients through both copies. Its weight is set at every update to \varFrac\ times the loss
without the term, divided by the term, with both values detached, so the term is always \varFracWords\ of that loss,
which differs in scale between examples and over training; it applies from the first update, through the warm-up.

\paragraph{The row-color term.}
Given the row types, the rule has to learn which colors each type means, and on the rest of the objective it can settle
in a stable wrong reading. On Wall the first rule trained on the row-type sampler read each row as the type of a row
ahead of it, by a lag that differed between the example's short and tall courses, and \csWallRowMatchWords\ of its rows
had the sampled type. Rules given the row types therefore carry, during the warm-up, the \emph{row-color term} $\mathcal
C$: for every row of the training canvas, the cross entropy between the example's color histogram over the rows of that
row's type and the rule's color distribution on the last round, averaged along the row, with a weight fixed at the first
update to equal the rest of the loss there. With the term, \wallRowsFollowWords\ row of the Wall rule follows the
sampled type on each of the eight canvases of the development and robustness seeds. Trained without it from training
seed 1, the rule on Wall is \wallNoColorTermValidWords\ after mending, while none of its rows has the type the sampler
drew, so at that seed the term decided whether the rule follows its input, and the world was valid either way.

\paragraph{The independent-coin rule and the octave rule.}
The independent-coin rule of Section~\ref{sec:theory} runs this construction on the cell's own coins alone: it reads
none of the octaves, the row types or the phase channels, and its objective has neither the variation term nor the
row-color term. The octave rule is the construction on the octaves alone, on every example; it is the method wherever
the procedure gives the octaves alone. The method and these two rules share the rest of the objective, its schedule,
repair, mending and the evaluation.

\subsection{Training}
\label{app:training}

\paragraph{Configuration.}
Each cell carries a state of width \ruleWidth\ and \coinFlat\ coins of its own, computed by an integer hash of the seed,
the two coordinates and the channel. In the update above, the learned vector $s_0$ starts at zero, $a$ is \initScale\
and $b$ is \reinjectScale. One rule is trained per example, on square canvases of \trainCanvas\ cells on a side plus the
margin the round count requires, each at an absolute origin drawn uniformly from a range far larger than any block, with
\msSteps\ Adam updates of batch \msBatch\ from a learning rate of \msLR\ on a cosine schedule that ends at \msLRFloor\ of
it, and the gradient norm clipped at \gradClip. The round count is drawn uniformly from \trainRoundsLo\ to
\trainRoundsHi\ at every update, and the rule is run for $R'=\mainR$ rounds unless stated otherwise. The softmax
temperature falls linearly from \tempStart\ to \tempEnd\ over the first \tempFallFracWords\ of the updates and stays
there. The weights of equation~\eqref{eq:objective} are $\lambda_F$ \lamF\ on the window divergence, $\lambda_H$ \lamH\
on the entropy, $\lambda_E$ \lamV\ on the hard forbidden-window rate $E_t$ and $\lambda_M$ \lamM\ on the color-marginal
divergence. The warm-up is the updates before $u_0$, \hardWarmUntil, the first \hardWarmFracWords\ of them. Every rule
given the row types carries the row-color term for the updates before \courseWarmUntil, the first
\courseWarmFracWords, the same window as the warm-up, at a weight $\lambda_{\mathcal C}$ fixed at the first update to
the rest of the loss there divided by the term, both values detached. The counts of the paper rest on training seed 0.
Besides it, the octave rule was trained from seeds 1 and 2 on every example, and the method's rule on each certified
example from seeds 1 and 2 with the same input as at seed 0, except on SmileCity, whose method rule was trained from
seed 1 but not from seed 2 (Table~\ref{tab:seeds}). Skyline and Skyline2 also have runs
at seed 1 made with the row-type sampler as it stood before its anchor and run limits were changed after the skylines
failed (Appendix~\ref{app:procedure}), so they are not rules of the method. A training run takes a median of
\msTrainMin\ minutes on one T4 for the octave rule, whose variation term adds a second forward and backward pass, and
\medTrainMin\ minutes for the independent-coin rule. Table~\ref{tab:variants} measures its variants against an
\emph{earlier configuration} of the independent-coin rule, with a fixed horizon of $R=\earlyConfigRounds$ rounds and
\earlyConfigUpdates\ updates.

\paragraph{The octaves' coins.}
For every scale pair $(i,j)$ with $i,j\in\{\coinScaleList,\infty\}$ other than $(0,0)$, taken in a fixed order, the
rule receives \coinPerPairWords\ channels. No finite scale equals $1$, so no block has a side of two cells. Each is the
hash of the cell's own coins with $\lfloor x/2^{i}\rfloor$ and $\lfloor y/2^{j}\rfloor$ in place of $x$ and $y$, a
coordinate left out of the hash when its scale is $\infty$, and a channel index offset past those of the cell's own
coins and of the code's other coin channels, so that no channel of the family coincides with another input. This gives \coinPairs\
pairs and \coinChannels\ channels: \coinSquarePairs\ square octaves; \coinLinePairs\ pairs with a dropped coordinate,
which give the row blocks, the column blocks and the pair hashed from the seed alone; and the elongated blocks between
them. The channels enter through the rule's input map with the cell's own coins, at the first round and at every round
after it. All coordinates are absolute, so chunks of any size and position see the same channels. Every training run
first checks the channels computed on the GPU against a reference computed on the CPU, on random regions with negative
and large coordinates; the integer hash agrees exactly, and the final division may differ only by rounding.

\paragraph{How the weights were set.}
The objective's weights were not searched. The variation term's was ablated: the octave rule was also trained without
the term on every example, and so were the method's rules on Wall and Skyline, in the configuration above otherwise
(Appendix~\ref{app:sharing}). The forbidden-window weight was raised once, together with a lower temperature, among the
objective variants of the earlier configuration (Table~\ref{tab:variants}), and the other weights were fixed before
training. The scales of the octaves were chosen by hand and not searched.

\paragraph{3Bricks.}
3Bricks has \bricksPatterns\ distinct windows, and at batch \msBatch\ its training runs out of the memory of a T4 when
the hard terms switch on. Every rule of 3Bricks, the independent-coin rule, the octave rule and the rule on the
row-type sampler, is therefore trained at batch \msBatchBricks\ from scratch, with everything else unchanged; it is the
only example trained at another batch.

\subsection{Repair, mending and the reach}
\label{app:mender}

\paragraph{Repair.}
The rule's output passes through \repairRoundsWords\ repair rounds. A cell covered by a forbidden window is
\emph{defective}. In each round every defective cell scores each color by the number of windows through it that the
color would make allowed, and the defective cells that win a fair coin take their best color, so that adjacent cells do
not all flip at once. These coins are hashed from the seed, the round and the position, so the rule with repair is still
a random-access generator (Definition~\ref{def:ra}), its radius larger by $N$ for each round.

\paragraph{Mending.}
Mending re-solves what repair leaves, on a lattice of square blocks of side $B$ fixed in absolute coordinates. A block
that contains forbidden windows frees the cells within distance $\rho$ of its defective cells, for $\rho=1,2,\dots$ up
to a cap $\rho_{\max}$, with the free region allowed to grow by at most $\min(\rho,\lfloor(B-N+1)/2\rfloor)$ cells
beyond the block, and re-solves the windows that touch the free region with Wave Function Collapse's own propagation and
minimum-entropy collapse, conditioned on every other cell. It makes two attempts at each radius with different
block-seeded randomness, skipping the second where propagation alone proves the instance unsolvable. The smallest $\rho$
at which the solver finds a completion bounds that defect's mending radius in the sense of \citet{balliu2022mending}
from above, and only from above, since the solver does not backtrack. When no radius up to the cap completes the
block's defects together, \emph{cluster completion} grows the block's defective cells by $N-1$ cells, takes the
connected components of the result as its clusters, and re-solves each cluster's defective cells alone with the same
ladder of radii, the same box and two attempts per radius, in scan order and on the canvas as the clusters before it
left it, so that a cluster without a completion keeps its defects without blocking its neighbors. Blocks are processed
in four classes per pass by the parity of their lattice coordinates, so blocks of one class are $2B$ apart and their
free regions cannot interact, over two passes with the lattice shifted by $B/2$ between them, and every solve is seeded
by the seed, the block's absolute coordinates, the pass and the cluster's index. We call $\rho_{\max}$ the
\emph{mending radius} and set $B=2\rho_{\max}$; the main results use $\rho_{\max}=\mendRadiusMain$ and
$B=\mendBlockMain$, and Table~\ref{tab:radii} adds $\rho_{\max}=\mendRadiusLo$ and $\rho_{\max}=\mendRadiusMid$. The
mender without cluster completion is the same procedure without the clusters: a block whose defects it cannot complete
together keeps its content. Because the lattice and the pass order depend on absolute coordinates, the mended output
commutes with shifts by multiples of $2B$ but not, in general, with other shifts.

\paragraph{What mending guarantees.}
Every window that touches a re-solved region is allowed by construction, so the residual after mending counts only
defects for which the solver, which can miss a completion that exists (Appendix~\ref{sec:forced}), found none within the
cap. How far a change can travel is counted from the mender's dependencies. In each of the eight class steps of the two
passes, a solve writes only cells within $m=\lfloor(B-N+1)/2\rfloor$ of its block, the largest growth of the free
region, and reads only cells within $m+N-1$, so one step carries a change at most $B+2m+N-2$ cells: \mendStepEight\ at
radius \mendRadiusLo\ and \mendStepCC\ at radius \mendRadiusMain. Over the eight steps, the mended value of a cell
depends on the repaired canvas within \mendBoundEight\ cells at radius \mendRadiusLo\ and \mendBoundCC\ at radius
\mendRadiusMain, so chunks of the repaired canvas mended separately agree when each carries a margin of that width.
Run for \mainR\ rounds, a rule has radius \floorRadiusRaw\ and, with the \repairRoundsWords\ repair rounds,
\floorRadiusRepair, $N$ cells for each round; a repair round reads only the windows that contain a cell, so $N-1$ cells
per round would suffice. For the independent-coin rule these are the radii at which Theorem~\ref{thm:floor} applies,
and counting $N$ only weakens the floor; a rule on shared coins falls outside the theorem. With the rule's rounds and
its repair rounds, no coin farther than \certRadiusCC\ cells from a cell can change its mended color at radius
\mendRadiusMain. That distance is the \emph{reach} of the whole pipeline, its margin of random access in the extended
sense of Section~\ref{sec:ra}. For a rule on shared coins, a coin read at a cell within the reach may be shared with
cells beyond it, so changing a coarse coin reaches every cell within \certRadiusCC\ cells of its block. The count is a
bound read off the dependency structure of the code, and the seam test of Appendix~\ref{app:seam} probes it from
below: chunked mending agrees with whole-canvas mending at margins of \seamAgreeMarginEight\ cells and more at radius
\mendRadiusLo, and at radius \mendRadiusMain\ on every tested canvas from a margin of \seamAllMargin\ cells, with one
canvas still differing at \seamOneDiffersMargin\ (Section~\ref{sec:access}).

\paragraph{Changes not adopted.}
Two further changes were tried at radius \mendRadiusMain\ and not adopted. A pass that put back, greedily in scan order,
every changed cell whose old color keeps all windows through it allowed changed no example's validity and left the
share of cells changed almost where it was (Appendix~\ref{app:rulemap}). Retrying a failed block with its clusters
grouped over \mendGroupWide\ cells, then over \mendGroupNarrow, before single clusters changed no example's validity
under the octave rule and left Wall's residual unchanged (Table~\ref{tab:record}).

%% file: app_setup.tex
\section{Protocols, references and measures}
\label{app:setup}

\subsection{The comparisons and their protocols}
\label{app:protocols}

\begin{table*}[!htbp]\centering\scriptsize\setlength{\tabcolsep}{3pt}\def\rrcell{\raggedright\hspace{0pt}}%
\caption{The comparisons and their protocols. Each mending radius comes with blocks of twice the radius. Canvases are
$256\times256$ and drawn on coin seeds that every rule shares, unless the row says otherwise. The mender with cluster
completion is that of Appendix~\ref{app:mender}; the mender without it leaves a block unchanged when it cannot complete
all of its defects together.}
\label{tab:protocols}
\begin{tabular}{@{}p{2.9cm}p{2.9cm}p{1.6cm}p{1.2cm}p{2.4cm}p{2.7cm}p{1.9cm}@{}}\toprule
\rrcell comparison & \rrcell rules & \rrcell mender & \rrcell radius & \rrcell canvases per example & \rrcell examples &
\rrcell reported in\tabularnewline\midrule
\rrcell validity, what learning carries, variation and edits & \rrcell the method, the octave rule, the
independent-coin rule & \rrcell with cluster completion & \rrcell \mendRadiusLo, \mendRadiusMid, \mendRadiusMain &
\rrcell three, on the coin seeds of development; eight in the robustness test & \rrcell all \nPublic & \rrcell
Section~\ref{sec:results} unless a row below applies\tabularnewline
\rrcell fresh coin seeds & \rrcell the method, the independent-coin rule & \rrcell with cluster completion & \rrcell
\mendRadiusMain & \rrcell three, on fresh coin seeds & \rrcell all \nPublic & \rrcell Section~\ref{sec:valid},
Table~\ref{tab:seeds}\tabularnewline
\rrcell seams & \rrcell the octave rule; the method's rules on the row types or the phase channels & \rrcell with
cluster completion & \rrcell \mendRadiusMain & \rrcell three, mended whole and in chunks, on larger canvases for the
wider margins & \rrcell the \nCCM\ the octave rule makes valid; \seamMethodAllKernelsWords\ certified examples given an
input, all but SmileCity & \rrcell Section~\ref{sec:access}, Appendix~\ref{app:seam}\tabularnewline
\rrcell the classes & \rrcell the independent-coin rule & \rrcell without cluster completion & \rrcell \mendRadiusLo &
\rrcell one & \rrcell all \nPublic & \rrcell Table~\ref{tab:mendmapeight}\tabularnewline
\rrcell the references & \rrcell the octave rule & \rrcell without cluster completion & \rrcell \mendRadiusLo & \rrcell
eight, shared with the references & \rrcell the \msVPopulation\ the octave rule makes valid at that radius & \rrcell
Section~\ref{sec:replay}, Appendix~\ref{app:replay}\tabularnewline
\rrcell complete solver, witnesses & \rrcell the independent-coin rule & \rrcell without cluster completion & \rrcell
\mendRadiusLo & \rrcell one & \rrcell named in the appendices & \rrcell Section~\ref{sec:resist},
Appendices~\ref{app:sat} and~\ref{app:witness}\tabularnewline
\rrcell window statistics & \rrcell the octave rule, the independent-coin rule, Wave Function Collapse & \rrcell none:
raw canvases & \rrcell -- & \rrcell eight at $64^2$, four at $128^2$ and $256^2$, the same for every method & \rrcell those
Wave Function Collapse completes &
\rrcell Section~\ref{app:stats}\tabularnewline
\bottomrule\end{tabular}\end{table*}

\paragraph{Generation and evaluation.}
We call the comparison of the first row of Table~\ref{tab:protocols} the \emph{corpus evaluation}; the method, the octave rule and the
independent-coin rule are evaluated there by the same code, at the same radii and coin seeds. Each canvas is generated
from the coins of its own cells and of a margin of \floorRadiusRepair\ cells around it, the radius of the rule with its
repair rounds, and every evaluation generates it again in chunks of $64\times64$ cells, each with the same margin, and
checks that the two agree cell for cell (Appendix~\ref{app:seam}). The canvases are then mended whole; mending in chunks
needs a wider margin and is tested separately (Section~\ref{sec:access}, Appendix~\ref{app:seam}). The evaluation at
radius \mendRadiusMain\ records every measure of Appendix~\ref{app:measures} after mending, and the robustness test
records the residual alone. Regions of $256^2$, $1024^2$ and $4096^2$ cells are generated in $256^2$ chunks, each with
its own margin, and the curve of the forbidden-window rate against the test-time round count is measured at $1024^2$
for $R'\in\{4,8,16,32,64\}$. The cell-coin edit of Appendix~\ref{app:measures} and the seam test hold for both rules on
every public example (Table~\ref{tab:persample}). For the octave rule, the first three of
the eight canvases of the comparison against the references are those of the corpus evaluation. The generation times
of the octave rule and of the independent-coin rule are compared on CPU with \msTimeThreads\ threads, on canvases of up
to $4096\times4096$ cells.

\paragraph{The classes.}
The classes depend on the independent-coin rule and its mender, while the certificate of Section~\ref{sec:cert} is
computed from the example alone, so comparing the two is not circular as a computation. On the three canvases of the
main protocol at radius \mendRadiusLo, the independent-coin rule is valid on the same examples as on the single canvas
that defines the classes, and Appendix~\ref{app:mendall} counts them.

\paragraph{Which rule.}
The octave rule and the independent-coin rule are one rule per example each, in every comparison of Table~\ref{tab:protocols} and in
the measures of connectivity, correlation and timing, with one exception: mending at $1024^2$
(Table~\ref{tab:mending}) uses three further independent-coin rules per example, trained in the same configuration from
three seeds, and the first of them is the rule of the complete-solver study on its five main examples and on Wall
(Table~\ref{tab:sat}). The same example's residual differs between tables, since a wider radius leaves less than a
narrower one on the examples it helps, the two menders leave different residuals, and these three further rules differ
from the main independent-coin rule as two training runs do.

\subsection{The references}
\label{app:references}

\paragraph{Shifted replay and the i.i.d.\ draw.}
\emph{Shifted replay} tiles the example from a uniformly random offset, so it is valid and faithful and varies only by
a translation. An \emph{i.i.d.\ draw} from the example's color frequencies copies nothing, so it sets the floor of the
measures of copying and the ceiling of divergence, and mended as the method is, it shows what mending does from noise.

\paragraph{The block reference.}
Block solvers keep every window allowed with no learned rule, by re-solving blocks of a copy of the example, in the
manner of model synthesis \citep{merrell2009model} and the block solvers of \citet{boris2021infinite} and
\citet{kleineberg2023infinite}. The \emph{block reference}
starts from shifted replay. In each sweep it shifts its lattice of $16\times16$ blocks by a fresh random offset, visits
the blocks in random order and replaces each by a completion under its frozen boundary in which every window is allowed,
found by the CaDiCaL SAT solver \citep{biere2020cadical} with random variable order and random initial phases, which is
randomized but not uniform; a block it cannot solve within a fixed budget keeps its content, so every window stays
allowed at every step. Its number of sweeps is set per example to the smallest whole number that brings its cost per
cell to at least \blockRefBudgetWords\ times the independent-coin rule's mean cost, estimated from its first sweep; the
octave rule costs as much per cell as the independent-coin rule. Its measured cost per cell is \replayCostRatioInvM\
times the octave rule's at the median, in single-thread CPU time, where the octave rule's cost counts generation,
\repairRoundsWords\ repair rounds and mending within radius \mendRadiusLo\ without cluster completion. Mending within
radius \mendRadiusMain\ costs more, so at that radius the ratio would be smaller. We call an example \emph{frozen for
the reference} when the reference changes no cell of shifted replay; these examples are \replayRigidNames, and none of
the \msVPopulation\ examples the octave rule makes valid is among them. This is a property of the reference's moves,
weaker than the rigidity of Appendix~\ref{app:rigid}.

\paragraph{The random-access block solver.}
The \emph{random-access block solver} makes the block reference random access. It starts from shifted replay on the
infinite plane and re-solves the $16\times16$ blocks of a lattice fixed in absolute coordinates in four classes, over
two passes with the lattice shifted by half a block between them, as our mender does, with each block's randomness
hashed from the seed, the pass and the block's coordinates. Any window is then computed from the blocks it depends on,
and that margin is counted in its cost. Two passes are the least it can run, and its outputs have no forbidden window by
construction (Appendix~\ref{app:replay}).

\paragraph{Block-shifted tiles, painted row types and the copier.}
Two references test what mending does without a rule. \emph{Block-shifted tiles} fill each $64\times64$ block of a
lattice fixed in absolute coordinates with the example tiled from an offset hashed from the seed and the block's
coordinates. Every window inside a block is allowed, so every defect lies on a block boundary, and the canvas is random
access without any learning; it is mended as the method is. On the examples with a row invariant, \emph{painted row
types} take the row types that the row-type sampler draws and draw each cell of a row from the example's color histogram
over the rows of that row's type, the histograms of the row-color term (Appendix~\ref{app:method}), before the same
mending. The \emph{copier} tests what the prescribed input carries alone: it fills the sampled sequence of row types
from the example alone. It cuts the sequence at its anchors, matches each stretch to a stretch of the example with the
same row types, lengthens a run by repeating one of its rows where every window stays a window of the example, and
copies each stretch with a horizontal offset hashed from the seed, chosen on the walls among the offsets that its joint
with the stretch above allows. On the walls the offsets chain only within a band of the row-type sampler: the first
stretch of a band takes an offset hashed from the seed and the band, and the others are drawn with coins hashed from the
seed and their first rows, uniformly over the sequences of allowed joints that lead to the next band's first offset, so
any offset costs work bounded by $L$. It is the method with a filler written by hand in place of the rule, and it reads
one coin per stretch and no fine coin. It exists on every example with a row invariant but 3Bricks, where no stretch of
rows can be repeated without a window that the example lacks, so that the copier cannot fill the longer runs the sampler
draws.

\paragraph{The Metropolis rules.}
Two priors written by hand stand in for the rule. The \emph{Metropolis rule} is the hand-written rule of
Appendix~\ref{app:floors}: hashed coins drive annealed Metropolis updates of the window energy defined there, in nine
parallel phases of radius 2. As a prior for mending it runs for \mhSweeps\ sweeps, and for \mhLongSweeps\ in a longer
test, before the method's \repairRoundsWords\ repair rounds and mending. The \emph{Metropolis filler} runs it on the
row types that the row-type sampler draws: a cell's proposals come from the colors of its row type, a window counts as
allowed only where the example has it centered on a row of the same type, and the energy adds the log frequencies of
the windows within their type. It runs for \mhTypedSweeps\ sweeps before the same mending, and each of the nine phases
of a sweep reads two cells farther, so its reach grows with every sweep. Appendix~\ref{app:floors} also defines the
second hand-written rule compared there, independent sampling followed by many rounds of our repair.

\paragraph{Wave Function Collapse.}
Our implementation of Wave Function Collapse, the overlapping model with minimum-entropy collapse and propagation, is
the reference for window statistics (Section~\ref{app:stats}). It restarts from scratch on a contradiction instead of
backtracking, and on a few examples it completes no large canvas; where it produced no $256^2$ canvas, the comparison
of window statistics uses the same solver with bounded chronological backtracking (Table~\ref{tab:wfcstats}).

\subsection{The measures}
\label{app:measures}

\paragraph{Defects and window statistics.}
The \emph{defect rate} is the density of forbidden windows among the $3\times3$ windows inside the canvas, and after
mending we call it the \emph{residual}. The total variation of Section~\ref{sec:ra} is bounded and needs no smoothing
for the windows a canvas lacks, so the body reports it; the appendix tables add the Kullback-Leibler divergence
$\mathrm{KL}(f\,\|\,h)$ from the example's frequencies $f$ to the canvas's frequencies $h$ of allowed windows, which the
objective and the comparison with Wave Function Collapse use. We estimate it by plugging in the counts of the canvas,
normalized over its allowed windows, with each frequency of $h$ floored at \klFloor\ inside the logarithm, so that a
window of the example that the canvas lacks adds a large but finite term. The estimator's bias depends on the size of
the canvas, so the comparison with Wave Function Collapse scores every method by the same estimator on canvases of one
size (Section~\ref{app:stats}).

\paragraph{The transported quantity.}
For the transported quantity of Section~\ref{sec:theory} we count how often it departs from the transport along its
direction, per hundred window positions of the raw output, its \emph{switch rate}; for an invariant along rows, that is
how often its value changes along a row. For a rule given row types we measure the share of rows of its raw output whose type, read from the
output's windows with the certificate's labels, is the type the row-type sampler drew for that row.

\paragraph{Copying and variation.}
Three measures separate copying from variation. Let $a$ be the largest share of cells equal to a translation of the
tiled example, and $a_0$ its value for the i.i.d.\ draw; the \emph{normalized agreement} $(a-a_0)/(1-a_0)$ is 0 for
the i.i.d.\ draw and 1 for a translate of the tiling. It uses one global translation of the tiled example.
\emph{Repetition} is the share of the canvas's $16\times16$ patches, taken at every position, that occur in the tiled
example. \emph{Aligned variation} is the share of cells that differ between two canvases of one method drawn from
different seeds, after the relative translation of at most 64 cells along each axis that makes it smallest, averaged
over the pairs of canvases of one method. Since it searches translations of up to 64 cells, it cannot align an example
whose period is longer, which matters only for shifted replay on Font.

\paragraph{Edits.}
Each edit overrides some or all of the coins read in a region, in the sense of Appendix~\ref{app:setting}, and leaves
every coin outside the region as it was (Table~\ref{tab:edits}). \emph{Local variation}, which we also call the
\emph{fine redraw} where a rule reads shared coins, edits one $32\times32$ region at the center of the canvas whose
edges lie on the coarsest block lattice. It redraws the cells' own coins and the shared coins whose block is one cell
wide in one axis and finite in the other, the fine coins of the variation term, which under independent coins are all
the cells' coins, with every coarser coin, the row types and the phase channels held; it regenerates the canvas with
\repairRoundsWords\ repair rounds, mends it with the same seed, and reports the share of the canvas that changes
together with the share of the changed cells that lie only in allowed windows. The \emph{full redraw} redraws every
coin that the region's cells read. For the independent-coin rule both are the redraw of its cells' own coins. The
cell-coin edit is the generator's own test of containment: on a raw canvas of side \editCanvas\ it resamples only the
cells' own coins on a $16\times16$ region, with every shared coin held, checks that no cell changes beyond the rule's radius, and counts the
cells that change.
Table~\ref{tab:variation} reports the fine and the full redraw, and Table~\ref{tab:persample} the cell-coin edit.

\begin{table}[!htbp]\centering\scriptsize\setlength{\tabcolsep}{3pt}\def\rrcell{\raggedright\hspace{0pt}}%
\caption{The edits we measure. Each overrides coins read in a square region and leaves every coin outside the region as
it was. Local variation and the full redraw are measured on mended canvases and report the share of the canvas that
changes and the share of the changed cells that lie only in allowed windows; for the independent-coin rule the two are
the same edit. The cell-coin edit is measured on a raw canvas of side \editCanvas\ and reports the cells that change.}
\label{tab:edits}
\begin{tabular}{@{}p{1.5cm}p{2.2cm}p{2.3cm}l@{}}\toprule
\rrcell edit & \rrcell coins redrawn & \rrcell coins held & region\tabularnewline\midrule
\rrcell local variation (fine redraw) & \rrcell the cells' own coins and the fine shared coins & \rrcell coarser coins, row types, phase
channels & $32\times32$\tabularnewline
\rrcell full redraw & \rrcell every coin the region's cells read & \rrcell none & $32\times32$\tabularnewline
\rrcell cell-coin edit & \rrcell the cells' own coins & \rrcell every shared coin & $16\times16$\tabularnewline
\bottomrule\end{tabular}\end{table}

\begin{figure}[!htbp]\centering\includegraphics[width=0.6\linewidth]{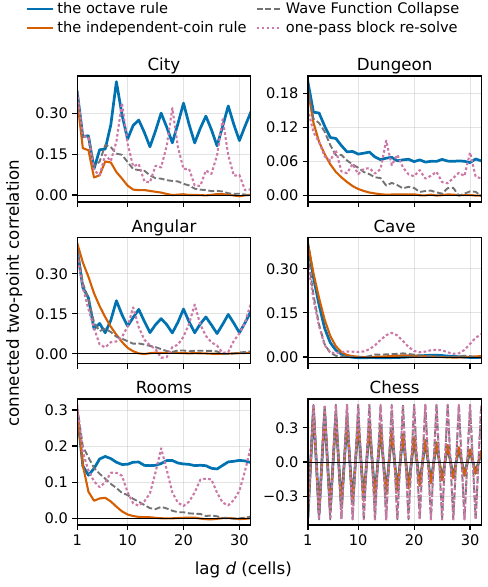}
\caption{Connected two-point correlation against the lag, one panel per example plotted: the octave rule and the
independent-coin rule on raw $512^2$ canvases, our Wave Function Collapse implementation on $64^2$ canvases, and a
one-pass block re-solve, which regenerates the $16\times16$ blocks of the tiled example once in scan order with Wave
Function Collapse, on $128^2$ canvases. None of the methods targets these statistics.}
\label{fig:corr}\end{figure}

\paragraph{Structure beyond the window.}
Two further measures look past the window, and no rule is trained on either. For connectivity, the passable color of an
example is the color, among those that cover at least a fifth of it, whose 4-connected components on shifted replay hold
the largest share of its cells in a single component; an example on which no such color reaches half is left out. A
generator keeps the example's connectivity when its largest component of that color holds at least 0.8 times the share
that shifted replay's does. For correlation we use the connected same-color correlation
$S(d)=\Pr[x_v=x_{v+d}]-\sum_c p_c^2$, averaged over the two axes, with $p_c$ the canvas's color frequencies, and the
correlation length is the first lag at which $S$ falls below a tenth of $S(1)$. The curves of the two rules in
Figure~\ref{fig:corr} come from raw $512\times512$ canvases at lags up to 32.

\paragraph{Examples whose worlds are fixed.}\phantomsection\label{app:rigid}
On \rigidNames, which we call \emph{rigid}, the locally valid patches of some size are exactly the patches of the tiled
example, each with one successor along each axis, so every $K$-valid world is a translate of the tiling. Fabric has
other such worlds, since a vertical seam can shift its tiling by one column and its background color can fill a region,
but its example shows a single column spacing and each such departure changes the window frequencies where it occurs,
so a world with Fabric's statistics can depart from a translate of the tiling only on a sparse set of cells. On these
examples validity is the whole claim, and the measures of copying and variation leave them out.

\subsection{Further controls and three dimensions}
\label{app:further}

\paragraph{Chess and Stripes.}
Two controls of known class are added to the public examples. Chess and Stripes are agreement constraints, whose only
$K$-valid colorings are the two phases of a period-two pattern: no factor of i.i.d.\ coins is almost surely $K$-valid
on them (Theorem~\ref{thm:nogo}), and a generator on independent coins of radius $r$ leaves a density of defects of
order at least $1/r$ (Proposition~\ref{prop:classes} in Appendix~\ref{app:proofs}).

\paragraph{Further controls.}
The per-example tables also list two proper $3$-coloring controls: proper3, whose constraint set is every proper
$3$-coloring of the window with uniform frequencies, and Col3, a random proper $3$-coloring of a torus used as an
example. Both count in no claim about the public examples. A random proper $4$-coloring of a torus is left out as a
control: its example contains only \colfourWindows\ of the \colfourProperWindows\ proper $4$-colored windows
(\colfourShare), so it does not stand for the proper $4$-coloring constraint.

\paragraph{One voxel example.}
The construction runs unchanged on $\Z^3$, with exact seams. Our one three-dimensional example is a $4^3$ maze at two
voxels per cell, with windows of $3\times3\times3$ voxels. The learned independent-coin rule stayed far from valid at
the training budgets we tried, although the statistics of its allowed windows came to match the example's, so we make
no three-dimensional claim.

%% file: app_tables.tex
\section{The certificate on every example}
\label{app:cert}

\begin{table*}[!htbp]\centering\footnotesize
\caption{Certified public examples and examples a constant coloring satisfies, by how far mending within radius 8 gets
under independent coins (the classes of Table~\ref{tab:mendmapeight}). Chess and Stripes are not counted.}
\label{tab:certclasses}
\input{tables/cert_classes_v12}\end{table*}

Whether a constant coloring exists does not predict which examples resist. One has no forbidden window on
\constSatCount\ of the \constSatOf\ public examples, spread over every class that mending gives under independent
coins, while the certificate falls on resistant examples only (Table~\ref{tab:certclasses}, Section~\ref{sec:resist}).
Among the resistant examples that no constant coloring satisfies, the search certifies \certCtwoCount\ of
\certCtwoOf, Wall among them. When the search widens for the resistant examples left without a certificate
(Section~\ref{sec:resist}), the patches grow to side \certSideTestPMin\ to \certSideTestPMax.

The first certificate found is an invariant along rows on \certInvariantCount\ of the certified examples and a phase
on \certPhaseCount\ (Table~\ref{tab:cert}). On Town, at $P=\townP$, the example's patches fall into two strongly
connected components according to the example row that holds a patch's top row, taken modulo \townRowPeriod, and the
program's bound $\kappa(0)=\townKappaZero$ lies below $\kappa_f=\townKappaFFrac$, because the program lets the shares of
the two components fall below the example's. On \certBigPNames, where $P>N$, the example's $3\times3$ statistics pin
the marginal of the quantity, and $\kappa(0)$ is one minus its largest entry, as when $P=N$. The constant stays
positive on every row when the statistics may differ from the example's by $\delta=0.05$, and Appendix~\ref{app:lp}
allows defects as well. The floor in Table~\ref{tab:cert} is $\kappa(0)/((2r+P+1)(P-N+1)^2)$ at radius
\floorRadiusRaw, and Appendix~\ref{app:floorcheck} says where it binds the trained independent-coin rule. The labeling
fixed before the search ran pooled every patch outside the certified component into one value, and on the phase
examples it gives smaller constants (Table~\ref{tab:record}).

\subsection{Dependence at a distance in a solver's worlds}
\label{app:longrange}

Could validity carry a quantity on the resistant examples without a certificate that the search's patches do not
reveal? A quantity carried along a direction leaves a trace that no patch size bounds: as long as its value varies
between worlds, two windows on that line stay dependent however far apart they lie. A generator of radius $r$ on
independent coins leaves cells more than $2r$ apart independent, but Wave Function Collapse, a global solver, can keep
such a dependence, so we looked for one in its worlds.

For every public example but \lrNonResNotMeasuredNames, which gave no valid world at either size, and for the controls
Chess, Stripes and Col3, our implementation of Wave Function Collapse (Appendix~\ref{app:references}), which restarts on a
contradiction and does not backtrack, made \lrWorlds\ worlds free of forbidden windows on a $\lrCanvas\times\lrCanvas$
canvas, or on a $\lrCanvasSmall\times\lrCanvasSmall$ canvas for the \lrCanvasSmallCountWords\ examples on which it could
not give them at the larger size, with the distances below scaled to \lrDistMidSmall\ and \lrDistMaxSmall. For each of
four directions (along rows, along columns, the diagonal and the antidiagonal) and a distance $k$, we take the plug-in
mutual information between the $3\times3$ window at a cell and the window $k$ cells farther along, pooled over cells and
worlds, and subtract the mean of the same estimate over \lrShuffles\ pairings in which the second window comes from
another world at the same place. The difference $D$ keeps the dependence and removes the estimator's bias and any effect
of position; by chance it can fall slightly below zero. The dependence persists along a direction if at \lrDistMax\
cells the estimate exceeds every one of the \lrShuffles\ pairings and $D$ there is at least half its value at
\lrDistMid. The rule and the counts we expected under it were fixed before the worlds were made.

\begin{table}[!htbp]\centering\scriptsize
\caption{Dependence between $3\times3$ windows at a distance in \lrWorlds\ worlds of Wave Function Collapse, in nats
above the pairings across worlds, at \lrDistMid\ cells ($D_{16}$) and \lrDistMax\ ($D_{64}$). The direction is the
certified one on a certified example and, on the others, the one in which $D_{16}$ is largest. The last column lists the
directions along which the dependence persists: at \lrDistMax\ cells it exceeds all \lrShuffles\ pairings and keeps at
least half its value at \lrDistMid. An asterisk marks a canvas of $\lrCanvasSmall\times\lrCanvasSmall$, read at
\lrDistMidSmall\ and \lrDistMaxSmall\ cells, and a dagger the one-way quantity of Town. Of the \lrNonResOf\ examples that
do not resist, only \lrNonResOrderNames\ shows a persistent dependence, along the \lrNonResOrderDir, and
\lrNonResNotMeasuredNames\ is not measured.}
\label{tab:longrange}
\input{tables/lr_v16}\end{table}

The dependence persists along the certified direction on \lrCertOrderWords\ of the \lrCertOfWords\ certified examples
whose quantity is not one-way, where we expected it on at least \lrCertBar, and along rows on Town, whose one-way
quantity runs along rows (Table~\ref{tab:longrange}). On \lrCertMissName, the one it misses, $D$ along rows keeps
\lrSkylineTwoRatio\ of its value, just under the half the rule asks, so the rule can miss a quantity that is there. Among
the examples that do not resist it persists only on \lrNonResOrderNames, where $D$ is close to zero at both distances, so
that the rule passes on noise. On the resistant examples without a certificate, where we expected it on at least
\lrUncertBar, it persists along no direction.

Read after the run, the scan shows that on these examples the dependence reaches well beyond the window but fades: where it is strongest it falls more than \lrUncertFallFloorWords{}fold between the two distances, and
on \lrUncertVanishNames\ it is gone at \lrDistMax\ in that direction. Dependence of this reach is not peculiar to them:
\lrNonResKeepMoreNames, which does not resist, keeps more of it at \lrDistMax\ cells than any of them. The test sees a
quantity only if a pair of $3\times3$ windows reads it and its value varies among the solver's worlds, and a solver's
worlds follow one law among those on valid worlds. Within these limits, a solver's worlds show no quantity carried along
a line on the resistant examples without a certificate.

A second test, fixed before its run, asked the same question of the min-coin copy of Appendix~\ref{app:mincoinrun}.
Where a quantity is carried beyond the reach of mending, the copy's share of forbidden windows after mending should fall
as one over the copy's radius, the order of Theorem~\ref{thm:floor}, as it does on the certified controls
\glTwoControlNames; where none is, we assumed it would fall faster. Over radii \tcRadiusMin\ to \tcLadderMax, the slope
of the logarithm of the share against that of the radius lies in the interval we read as carried, from \glTwoSlopeLo\
to \glTwoSlopeHi, on \glTwoCarriedNames; it lies below \glTwoSlopeNone\ on \glTwoNoneNames, and between the two on the
other three. The test agrees with the scan on \glTwoAgreeWords\ of the resistant examples without a certificate, where
we expected at least \glTwoAgreeBar.

We take the scan's reading on Cat and ColoredCity, for a reason found after both runs. A dependence that reaches beyond
the mending radius and then fades, as theirs does, can by itself make the share fall near one over the radius across
this ladder, so the slope cannot tell a carried quantity from a long but finite dependence. Their curves show it: the
share falls faster than the carried interval allows from radius \glTwoBendFrom\ to \glTwoBendTo\ and more slowly than
it allows after, so their slopes land in the interval as averages, while on the certified controls every step of the
ladder lies within it. The scan measures the dependence itself and was checked on both sides, on the certified
examples, where it persists on all but one, and on the examples that do not resist, where it fires once, on noise. The
two tests rank these examples in much the same order, with a rank correlation of \glTwoSpearman, and both find the
longest reach on Cat and ColoredCity.

%% file: tables/cert_classes_v12.tex
\begin{tabular}{@{}lrrr@{}}\toprule
class (mending at radius 8) & examples & certified & constant coloring valid\\\midrule
mendable within radius 8 & 21 & 0 & 20\\
partial & 10 & 0 & 9\\
resistant & 21 & 13 & 15\\
\midrule all public examples & 52 & 13 & 44\\
\bottomrule\end{tabular}

%% file: tables/lr_v16.tex
\setlength{\tabcolsep}{4pt}\begin{tabular}{@{}llrrrl@{}}\toprule
group & example & direction & $D_{16}$ & $D_{64}$ & persists along\\\midrule
certified & 3Bricks$^{*}$ & rows & 1.13 & 1.05 & rows\\
 & Fabric & columns & 2.42 & 2.42 & all four\\
 & Flowers & rows & 0.10 & 0.09 & rows\\
 & MoreFlowers & rows & 0.12 & 0.11 & rows\\
 & Platformer & rows & 0.84 & 0.54 & rows\\
 & RedMaze & rows & 2.51 & 2.51 & all four\\
 & SimpleMaze & rows & 2.54 & 2.54 & all four\\
 & SimpleWall & rows & 1.27 & 1.27 & rows\\
 & Skyline & rows & 0.42 & 0.29 & rows\\
 & Skyline2 & rows & 0.57 & 0.26 & none\\
 & SmileCity & columns & 1.76 & 0.88 & rows, columns\\
 & Town$^{\dagger}$ & rows & 1.18 & 0.62 & rows\\
 & Wall & rows & 1.58 & 1.58 & rows\\
\addlinespace
resistant, & Cat & rows & 0.79 & 0.10 & none\\
uncertified & ColoredCity & columns & 0.53 & 0.05 & none\\
 & Lake$^{*}$ & columns & 0.29 & 0.02 & none\\
 & Nested & rows & 0.16 & 0 & none\\
 & ScaledMaze & rows & 0.08 & 0.004 & none\\
 & Sewers & rows & 0.29 & 0.01 & none\\
 & Skew1 & rows & 0.13 & 0.003 & none\\
 & Skew2 & diagonal & 0.08 & $-$0.001 & none\\
\addlinespace
control & Chess & rows & 0.67 & 0.67 & all four\\
 & Stripes & rows & 0.68 & 0.68 & all four\\
 & Col3 & rows & 0.01 & 0 & none\\
\bottomrule\end{tabular}

%% file: app_evidence.tex
\section{More on where independent coins fail and where the method is valid}
\label{app:evidence}

The test of selectivity in Section~\ref{sec:resist} compares the share certified among the \certResistantOf\ examples
that resist independent coins with the share among the \certMendedOf\ that independent coins mend, leaving out the
partial ones. The search as first fixed, before we added the one-way quantity and the full labeling, certified the same
examples but Town, at $p=$~\certFisherP\ (Appendix~\ref{app:procedure}).

\begin{table*}[!htbp]\centering\footnotesize
\caption{The certificate on the certified public examples and on Chess and Stripes, two controls whose only valid
colorings are the two phases of a period-two pattern (Appendix~\ref{app:further}). Rows with an invariant or a phase
use the full labeling of Section~\ref{sec:theory}; Town's row uses its one-way quantity, whose constant is
$\kappa_f=1-\sum_{s\leadsto t}f_sf_t$ over pairs of components with $t$ reachable from $s$, and its $\kappa(\delta)$ is
the linear program's lower bound on $1-\sum_{s\leadsto t}p_sp_t$ (Appendix~\ref{app:lp}). Columns: the kind of quantity, its direction, the patch size $P$ at which the search found
it, the period $d$ of the phase (1 for an invariant, a dash for the one-way quantity), the number of values the example
visits and their marginal in the example, largest first, where $k{\times}m$ stands for $k$ values of share $m$ each,
$\kappa(\delta)$ at $\delta=0$ and $\delta=0.05$, the floor of
Theorem~\ref{thm:floor} at radius 32 for a generator whose window statistics match the example's, the independent-coin
rule's own distance $\delta$ from those statistics (raw, on one $1024^2$ canvas; an interval, since the run stored only
the $L_1$ distance over allowed windows and the rate of forbidden windows, printed as one value where its ends agree to
two digits), and whether a constant coloring is valid.}
\label{tab:cert}
\input{tables/cert_v12}\end{table*}

\begin{figure*}[!htbp]\centering\includegraphics[width=\textwidth]{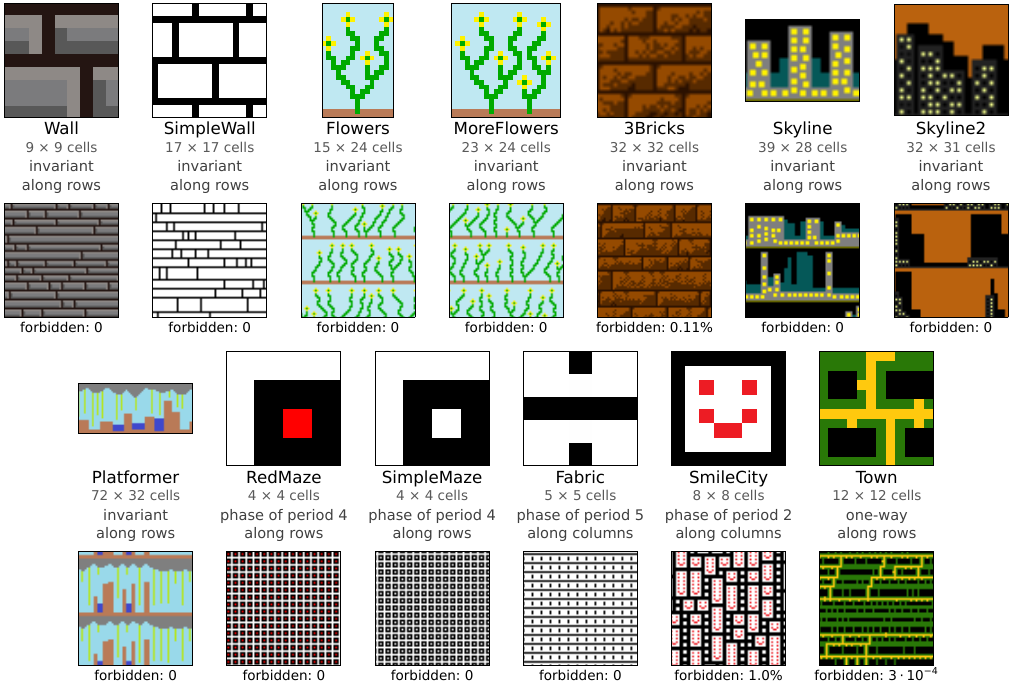}
\caption{The \certResistantWords\ certified examples in the order of Table~\ref{tab:cert}, each scaled to fill its panel, with
its size in cells and the kind of quantity its certificate finds: an invariant along rows, the row type of
Section~\ref{sec:theory}; a phase of period $d$ along rows or along columns; or, on Town, a one-way quantity along rows.
Under each example, the central $64\times64$ cells of one world of the method, mended within radius \mendRadiusMain\ and
drawn at one scale for every example, with the share of forbidden windows on the whole canvas. The method is invalid on
\certMethodInvalidWords\ of them: on SmileCity its faces come in mixed widths, and on 3Bricks the canvas shown is the
one of its three that keeps more than one window in a thousand forbidden.}
\label{fig:certified}\end{figure*}

\begin{table*}[!htbp]\centering\footnotesize
\caption{The method's worlds against the theorems on the certified examples, in the order of Table~\ref{tab:cert}, on
the \thmCanvasesWords\ canvases of the development and robustness seeds (on Wall also the fresh seeds), mended within
radius \mendRadiusMain. Columns: the kind of quantity and its patch size $P$; the total variation $\delta$ of the
canvases' $3\times3$ window statistics from the example's, smallest to largest; the program of Appendix~\ref{app:lp} at
the largest $\delta$; $\delta^*$, the least $\delta$ at which the program reaches zero, a dash where it stays positive
at every $\delta$, so that no generator on independent coins is free of forbidden windows on that example at all; the
share $\max_s p_s$ of the most common value of the quantity in the method's worlds; the floor of
Theorem~\ref{thm:floor} at the reach of the whole pipeline, $r=\certRadiusCC$, for a generator on independent coins with
that share, as forbidden windows per canvas in expectation; the largest number of forbidden windows on one mended canvas;
and the canvases with none. On Town, whose quantity is one-way, the program is the one-way program's lower bound, the
share is the largest component's and the floor is the one-way bound of Appendix~\ref{app:pfloor}.}
\label{tab:thm}
\input{tables/thm_v16}\end{table*}

\paragraph{The certified quantities.}
The quantities are of the kinds Section~\ref{sec:theory} describes (Table~\ref{tab:cert}, Figure~\ref{fig:certified}).
On the brick walls, the flowers, the skylines and the platformer, a line of one kind runs straight across the plane,
and the quantity is an invariant along rows, the row type. On the two mazes it is a phase along rows, on SmileCity and
Fabric a phase along columns, and on Town a one-way quantity along rows, which can pass from one strongly connected
component of the overlap graph to the other but never back. A constant coloring is valid on \certConstWords\ of the
\certResistantCount\ certified examples, so on these the constraints alone force no defect. The example's statistics
force them: the linear program of Section~\ref{sec:theory} gives $\kappa(0)>0$ on every certified example and on Chess
and Stripes, so no generator escapes the floor by concentrating the quantity on one value without moving away from the
example's statistics. Where the independent-coin rule has moved that far away, Theorem~\ref{thm:floor} still binds it
through its own $p_s$ (Appendix~\ref{app:floorcheck}).
Figure~\ref{fig:quantity} shows the mechanism on Wall: the independent-coin rule's forbidden windows sit where the row
type changes along a row, and the method, whose rule reads the row types, keeps each row's type across the canvas.

\paragraph{Mending under independent coins.}
With the mender of the main protocol, independent coins are valid on \ccFlatResistValidWords\ of the resistant
examples, certified or not (Figure~\ref{fig:corpus}). The bound does not require this failure after mending
(Section~\ref{sec:theory}, Appendix~\ref{app:scopefloors}), but on every certified example a witness accounts for most
of the mending instances at radii \wxRhoLo\ to \wxRhoHi\ that have no completion, and at radius \wxRhoHi\ for at least
\wxShareSixteenMin\ of them (Appendix~\ref{app:witness}). Chess and Stripes, whose only valid colorings are the two
phases of a period-two pattern, stay above one forbidden window in a hundred under independent coins: a mender of
bounded radius can remove a boundary between the two phases only by freeing all of the region it encloses, so the
boundaries the rule draws around larger regions survive mending (Appendix~\ref{app:proofs}). They stay above it under
the octaves too, although the octaves' blocks reveal the parity (Proposition~\ref{prop:blocks}), so there the octave rule
fails to learn what its input could carry.

\paragraph{The method on the certified examples.}
Under the row-type sampler the method is valid on \courseOtherValidWords\ of the \courseOtherOfWords\ examples with a
row invariant other than Wall, all but \courseOtherInvalidNames, and under the phase channels on Fabric and on
\mazeMethodValidWords\ mazes, whose valid worlds are essentially translates of the tiling (Appendix~\ref{app:rigid}).
Table~\ref{tab:thm} sets these worlds against the theorems on every certified example (Section~\ref{sec:valid}). On
\thmVfrNames, where $P=N$ and every canvas lies closer to the example's statistics than $\kappa_f$, inequality~\eqref{eq:vfr}
also says what the exclusion costs at a finite radius: a generator on independent coins with a canvas's statistics and
the reach of the whole pipeline would leave at least \thmVfrWindowsMin\ forbidden windows on it in expectation. On Platformer the canvases lie farther than $\kappa_f$, so the
inequality gives nothing, but the program, which constrains the whole distribution of windows and not only the
quantity's marginal, stays positive at their distance. On Fabric and the mazes these worlds are the tiled example, the
same canvas at every seed. The octaves alone carry SimpleWall's row type well enough that the octave rule's worlds lie
in the same region there (Figure~\ref{fig:bound}). On \thmSpreadNames\ the program reaches zero below the method's
distance, and only the floor from the method's own spread of the row type sets it apart, at the pipeline's reach and in
expectation. On \methodNotRobustCertNames, which the method makes valid on the development seeds but not on all
\thmCanvasesWords\ canvases, neither bound sets it apart: Skyline2's canvases keep forbidden windows, and on Town the floor
from its own spread expects fewer than one per canvas. There the evidence is the measured failure of independent coins
at every radius we tried. Of the certified examples that shaped no part of the procedure, the octave rule alone makes
\notShapedOctValidCountWords\ valid, \notShapedOctValidNames, and the prescribed input adds \notShapedGainedNames, where
reading the example at the coordinates the phase channels give already yields the tiled example. On Wall the rules
trained from training seeds 1 and 2 keep no forbidden window either (Appendix~\ref{app:results}).

\paragraph{The held-out examples.}\phantomsection\label{app:heldout}
The procedure was tested on \heldoutAllCountWords\ examples that did not shape it, three walls and two mazes. Wall and
SimpleWall transposed, whose invariant runs along columns, test the exchange of axes; a wall drawn for this test before
any training, with course heights that Wall does not have and thicker mortar, is a new example. The search certifies the
invariant on all three, along columns on the transposed walls, and the procedure, applied unchanged, gives each the
row-type sampler, under which the method is valid on all \heldoutValidCountWords. On every canvas of the three, each
row of the rule's output, or each column on the transposed walls, has the type the sampler drew. The octave rule alone,
trained the same way, behaves on the transposed walls as on the public ones. It is \heldOctWallTValidWords\ on
transposed Wall, where each of its three canvases keeps at least \heldOctWallTMendedLo\ of its windows forbidden after
mending, and it is \heldOctSimpleWallTValidWords\ on transposed SimpleWall. On the new wall it is also
\heldOctWallHTwoValidWords, but only \heldOctWallHTwoHeightsPct\% of its courses have one of the example's heights,
while \heldMethodWallHTwoHeightsWords\ course of the method has one.

The two mazes were drawn as the same spanning tree of rooms on a torus, with walls on a lattice of period \doorMazeD,
which does not divide the octaves' period. In the first, passages \heldMazePassageWords\ cells wide hide the lattice
from the $3\times3$ window, the search certifies nothing, and the procedure gives the octaves, under which the rule is
\heldMazeOctValidWords. In the second, openings \doorMazeOpeningWords\ cell wide let every window see the lattice, the
search certifies a phase of period \doorMazeD, and the procedure gives the phase channels of that period, under which the
method is \doorMazeMethodValidWords, with no forbidden window even before repair. The lattice fixes most of the cells,
so these worlds can differ only in which openings are open, and they do (aligned variation \doorMazeAlignedR; for the
same reason their normalized agreement is high, \doorMazeMethodNrep), while none of their $16\times16$ patches occurs in
the example. The octave rule on this maze is \doorMazeOctValidWords: before repair it leaves \doorMazeOctRawLo\ to
\doorMazeOctRawHi\ of its windows forbidden, above the floor that Proposition~\ref{prop:blocks} sets for its lattice
(\doorMazeOctFloor), and mending leaves up to \doorMazeOctMendedHi.

\paragraph{Examples from other implementations.}\phantomsection\label{app:freshex}
A search of public repositories, logged before any run, found \fxNExamples\ licensed example images in \fxNReposWords\
other open-source implementations of Wave Function Collapse and related generators; none is among the \nPublic\ examples
of the original repository. One more image was left out before any run: it has more colors than the rule draws coins
per cell, so its repair rounds would break ties with a global random generator, and its worlds would no longer be a
function of the seed and the position. The procedure ran on the rest unchanged, from the search and the input it
prescribes to training from training seed 0 and validity under the main protocol. The search certifies
\fxNCertifiedWords, \fxCertifiedNames, all from one repository, and gives \fxInputNames\ an input other than the octaves.
On \fxRefusedNames\ the certified invariant runs along a diagonal, for which the procedure has no input, so there the
method is the octave rule; we fixed before the run that it counts with that rule's result, as Town does.

The certificate again marks examples that resist independent coins: all \fxNCertifiedWords\ certified examples resist
them, \fxNCertResistantWords\ of the \fxNResistantWords\ that do. Given its input, the method is valid on
\fxMethodInputPhrase\ examples that receive one, \fxSeedsPhrase; the octave rule alone is valid on
\fxOctValidInputWords\ of them. \fxTiledNames\ are each one period of a pattern, with the period of their phase as their
side, and there the method's world is the tiled example. On \fxOctFragileNames\ the octave rule is valid under the main
protocol but not on all \thmCanvasesWords\ canvases, while the method is valid on all of them. Over all \fxNExamples\
examples the method is valid on \fxMethodValid\ and invalid on \fxMethodInvalidNames. On the \fxNUncert\ uncertified
examples the octave rule is valid on \fxOctValidUncert, block-shifted tiles of the example under the same mender on
\fxTileValidUncert, and i.i.d.\ colors on \fxIidValidUncertWords. As on the public examples, the min-coin copy of
Proposition~\ref{prop:mincoin} is valid after mending on none of the certified examples, and on each of them the
independent-coin rule's share of forbidden windows before repair lies above the floor of Theorem~\ref{thm:floor}, by a
factor of \fxCheckMinRatioWords\ or more. Table~\ref{tab:fresh} gives every generator on every example.

%% file: tables/cert_v12.tex
\setlength{\tabcolsep}{2.5pt}%
\begin{tabular}{@{}lllrrrlrrrrc@{}}\toprule
example & type & direction & $P$ & $d$ & values & \begin{tabular}[b]{@{}l@{}}example's\\marginal of $\varphi$\end{tabular} & $\kappa(0)$ & $\kappa(0.05)$ & \begin{tabular}[b]{@{}r@{}}floor\\$r{=}32$\end{tabular} & \begin{tabular}[b]{@{}r@{}}indep.-coin\\rule's $\delta$\end{tabular} & \begin{tabular}[b]{@{}c@{}}constant\\valid\end{tabular}\\\midrule
Wall & invariant & horizontal & 3 & 1 & 4 & (0.33,\,$3{\times}0.22$) & 0.67 & 0.65 & $9.8\cdot10^{-3}$ & 0.10--0.19 & n\\
SimpleWall & invariant & horizontal & 3 & 1 & 4 & (0.47,\,$3{\times}0.18$) & 0.53 & 0.48 & $7.8\cdot10^{-3}$ & 0.06--0.11 & y\\
Flowers & invariant & horizontal & 3 & 1 & 5 & (0.83,\,$4{\times}0.04$) & 0.17 & 0.12 & $2.5\cdot10^{-3}$ & 0.19--0.25$^\dagger$ & y\\
MoreFlowers & invariant & horizontal & 3 & 1 & 5 & (0.83,\,$4{\times}0.04$) & 0.17 & 0.12 & $2.5\cdot10^{-3}$ & 0.23--0.29$^\dagger$ & y\\
3Bricks & invariant & horizontal & 4 & 1 & 2 & (0.88,\,0.12) & 0.12 & 0.11 & $4.5\cdot10^{-4}$ & 0.33--0.40 & y\\
Skyline & invariant & horizontal & 3 & 1 & 4 & (0.89,\,$3{\times}0.04$) & 0.11 & 0.08 & $1.6\cdot10^{-3}$ & 0.27--0.35$^\dagger$ & y\\
Skyline2 & invariant & horizontal & 3 & 1 & 3 & (0.94,\,0.03,\,0.03) & 0.06 & 0.05 & $9.5\cdot10^{-4}$ & 0.14--0.22$^\dagger$ & y\\
Platformer & invariant & horizontal & 3 & 1 & 3 & (0.94,\,0.03,\,0.03) & 0.06 & 0.05 & $9.2\cdot10^{-4}$ & 0.12--0.16$^\dagger$ & y\\
RedMaze & phase & horizontal & 3 & 4 & 13 & (0.25,\,$12{\times}0.06$) & 0.75 & 0.75 & $1.1\cdot10^{-2}$ & 0.12--0.13 & n\\
SimpleMaze & phase & horizontal & 3 & 4 & 13 & (0.25,\,$12{\times}0.06$) & 0.75 & 0.75 & $1.1\cdot10^{-2}$ & 0.11 & n\\
Fabric & phase & vertical & 3 & 5 & 16 & (0.40,\,$15{\times}0.04$) & 0.60 & 0.55 & $8.8\cdot10^{-3}$ & 0.15--0.27 & y\\
SmileCity & phase & vertical & 5 & 2 & 2 & (0.50,\,0.50) & 0.50 & 0.50 & $7.9\cdot10^{-4}$ & 0.11--0.21 & n\\
Town & one-way & horizontal & 6 & -- & 2 & (0.67,\,0.33) & 0.15 & 0.12 & $1.3\cdot10^{-4}$ & 0.18--0.25$^\ddagger$ & y\\
\midrule \multicolumn{12}{@{}l}{\emph{controls}}\\
Chess & phase & horizontal & 3 & 2 & 2 & (0.50,\,0.50) & 0.50 & 0.50 & $7.4\cdot10^{-3}$ & 0.06 & n\\
Stripes & phase & vertical & 3 & 2 & 2 & (0.50,\,0.50) & 0.50 & 0.50 & $7.4\cdot10^{-3}$ & 0.06 & n\\
\bottomrule\multicolumn{12}{@{}l}{$^\dagger$ $P=N$, and the independent-coin rule's $\delta$ is at least $\kappa(0)=\kappa_f$: inequality~\eqref{eq:vfr} places no constraint on this rule (Appendix~\ref{app:cert}).}\\
\multicolumn{12}{@{}l}{$^\ddagger$ The independent-coin rule's $\delta$ is at least the lower bound $\kappa(0)$, and below $\delta=\townThmDelta$, where the program}\\
\multicolumn{12}{@{}l}{\phantom{$^\ddagger$}still gives \townThmKappa\ (Table~\ref{tab:thm}), so a generator on independent coins with the rule's statistics leaves}\\
\multicolumn{12}{@{}l}{\phantom{$^\ddagger$}forbidden windows; the program with defects was not solved for a one-way quantity.}\\
\end{tabular}

%% file: tables/thm_v16.tex
\setlength{\tabcolsep}{3pt}%
\begin{tabular}{@{}llrlrrrrrr@{}}\toprule
example & quantity & $P$ & $\delta$ & $\kappa(\delta_{\max})$ & $\delta^*$ & $\max_s p_s$ & \begin{tabular}[b]{@{}r@{}}expected\\at the reach\end{tabular} & \begin{tabular}[b]{@{}r@{}}forbidden\\windows\end{tabular} & clean\\\midrule
Wall & invariant & 3 & 0.35--0.38 & 0.60 & -- & 0.35 & 26.2 & 0 & 11/11\\
SimpleWall & invariant & 3 & 0.21--0.23 & 0.30 & 0.53 & 0.47 & 21.1 & 0 & 8/8\\
Flowers & invariant & 3 & 0.20--0.22 & 0 & 0.17 & 0.85 & 6.1 & 0 & 8/8\\
MoreFlowers & invariant & 3 & 0.18--0.20 & 0 & 0.17 & 0.85 & 6.1 & 0 & 8/8\\
3Bricks & invariant & 4 & 0.34--0.44 & 0.034 & 0.77 & 0.92 & 0.8 & 1069 & 1/8\\
Skyline & invariant & 3 & 0.26--0.35 & 0 & 0.25 & 0.89 & 4.3 & 0 & 8/8\\
Skyline2 & invariant & 3 & 0.25--0.36 & 0 & 0.24 & 0.94 & 2.5 & 107 & 2/8\\
Platformer & invariant & 3 & 0.087--0.138 & 0.028 & 0.30 & 0.94 & 2.5 & 0 & 8/8\\
RedMaze & phase & 3 & 0.004 & 0.75 & -- & 0.25 & 30.1 & 0 & 8/8\\
SimpleMaze & phase & 3 & 0.004 & 0.75 & -- & 0.25 & 29.9 & 0 & 8/8\\
Fabric & phase & 3 & 0.004 & 0.60 & 0.60 & 0.40 & 24.0 & 0 & 8/8\\
SmileCity & phase & 5 & 0.14--0.18 & 0.50 & -- & 0.47 & 2.3 & 692 & 0/8\\
Town & one-way & 6 & 0.29--0.33 & 0.023 & n/a & 0.65 & 0.7 & 96 & 3/8\\
\bottomrule\end{tabular}

%% file: app_mincoinrun.tex
\section{The failure of independent coins}
\label{app:indep}

\subsection{Witnesses}
\label{app:witness}

\begin{table*}[!htbp]\centering\footnotesize
\caption{Witnesses on every certified example and on the resistant examples without a certificate. The instances are
those of Appendix~\ref{app:sat} decided unsolvable at $\rho=\wxRhoLo$, \wxRhoMid\ and \wxRhoHi, single window and
cluster, drawn from the independent-coin rule of the main protocol and, on Wall, Sewers and Skew2, from the rule of
Table~\ref{tab:sat}. Columns: the unsolvable instances; those the solver refutes before any search; the share with a
witness, overall, per radius and among the instances the solver had to search; and the solvable instances searched,
none of which holds a witness, as none can. Without a certificate the labeling of patches of size 3 has one value, so no
witness can exist there, and the shares are those of the labeling of patches of size 4 along rows or columns.}
\label{tab:witness}
\input{tables/witness_v16}\end{table*}

A single pair of patches, the witness of Appendix~\ref{app:pwitness}, explains most of the unsolvable mending
instances on every certified example (Table~\ref{tab:witness}). An instance frees every cell within radius \wxRhoLo,
\wxRhoMid\ or \wxRhoHi\ of one forbidden window, or of a cluster of them, and freezes every other cell, as in
Appendix~\ref{app:sat}; these radii lie below the main mender's, \mendRadiusMain. The forbidden windows are those of
the independent-coin rule of the main protocol, and on Wall, Sewers and Skew2 those of the rule of
Table~\ref{tab:sat}. Each locally valid patch is labeled by the full labeling of the certificate, or for Town's one-way
quantity by its strongly connected component. A witness is a pair of frozen patches on one line along the
certificate's direction, on the two sides of the free region, whose labels no locally valid walk of the needed length
joins, or for Town whose downstream component the upstream one cannot reach.

The share of unsolvable instances with a witness is highest at the largest radius on every certified example: at radius
\wxRhoHi\ it is at least \wxShareSixteenMin\ on each, lowest on \wxShareSixteenMinName, and on Skyline it grows from
\witSkylineRhoFourAll\ at radius \wxRhoLo\ to \witSkylineRhoSixteenAll. A witness proves that its instance has no
completion, so the search finds none on the \wxSolvAll\ instances that have one, as it must; that count checks the
search, and so do two controls. The search finds no witness on \wxNegRegions\ free regions cut at random from colorings
without forbidden windows, the tiled examples and the outputs of Wave Function Collapse, and it finds one on every
instance of three seams built to break the quantity, \wxPosEach\ on each: of mixed phase on Chess and on Stripes, and in
the forbidden order on Town. A second search that uses no labels, a reachability search over the locally valid patches
along the witness's line, confirms every witness found, those on the seams included.

On Fabric and the mazes the solver refutes at least \wxTrivPhaseLo\ of the unsolvable instances before any search, and
the witnesses there mostly pair patches from two different weak components of the overlap graph: a pair of two levels
of one component appears in \wxPhaseLevelLo\ to \wxPhaseLevelHi\ of the witnessed instances. On SmileCity, which
visits a single component, all but \wxSmileLevelMissWords\ of the witnessed instances hold a pair of two levels of it.
On Town the pair the search reports is the forbidden order between the example's two strongly connected components on
\wxTownOrder\ of the witnessed instances, and on the others at least one side lies in a component the example never
visits.

An unsolvable instance often holds witnesses on several rows: the median witnessed instance has them on \witLinesWall\
rows on Wall and on \witLinesSkyline\ on Skyline, so freeing one frozen patch leaves the others in place. On Wall,
SimpleWall, Skyline and Flowers, freeing the nearer patch of one witness made no instance solvable, and neither did
freeing as many other frozen cells at random.

On the \wxUncertCountWords\ resistant examples without a certificate, where the labeling of patches of size 3 has one
value and no witness can exist, the labeling of patches of size 4 along rows or columns finds a witness in only
\wxUncertPfour\ of their unsolvable instances, and in at most \wxUncertPfourMax\ of them on any one example, so a
quantity at that patch size explains almost none of them. Each witness found there pairs the component of the labeling
that the example visits with one it never shows. On these examples a witness needs a frozen patch from such a
component, so the share is low partly for that reason.

\subsection{The min-coin copy on the corpus}
\label{app:mincoinrun}

We ran the min-coin copy of Proposition~\ref{prop:mincoin} on the certified examples, on Chess and Stripes, and on
uncertified examples for contrast. It runs on tori, so that no margin of width $r$ is needed, of side \tcTorusSmall\ up
to radius \tcContrastRadius, \tcTorusMid\ at radius \tcLadderMax\ and \tcTorusLarge\ beyond, with the priorities drawn
as a uniformly random order of the torus's cells, on the three coin seeds of Section~\ref{sec:setup}. The radii from
\tcRadiusMin\ to \tcLadderMax, each twice the one before, cover the \tcCertifiedWords\ certified examples, Chess, Stripes
and \tcControlsWords\ uncertified examples spread over the classes of Section~\ref{sec:setup}, and Wall alone runs at the
two radii beyond, up to \tcRadiusMax. Radius \tcContrastRadius, fixed before the run, and the independent-coin rule's
radius \tcRuleRadius, added before it, cover all \tcPublicOf\ public examples. At radius \tcContrastRadius\ the share of
the copy's windows that straddle two regions about equals the share of the windows of block-shifted tiles that straddle a
block boundary, so the two can be compared there. Each canvas is measured before mending and after mending as in the
main protocol.

The copy behaves as the proposition says. No forbidden window on any canvas has cells that all copy from one cell, the
share of windows whose cells copy from more than one cell lies within \tcShareDevMax\ of $q_r$ at every radius, and the
distance of the copy's window statistics from the example's is at most \tcDeltaOverQMax\ times $q_r$. At every radius,
on every certified example with a transported quantity, \tcFloorPairs\ pairs of example and radius in all,
$\varepsilon$ lies above the floor of Theorem~\ref{thm:floor} computed from the copy's own $p_s$, by a factor of
\tcFloorRatioAllLo\ to \tcFloorRatioAllHi. That factor grows with the radius where $\kappa(0)$ is small. Forbidden
windows carry no value of the quantity, so at small radii, where there are many, they raise $1-\max_s p_s$ and with it
the floor. Against the floor with the example's own statistics the factor changes by at most a factor of
\tcKappaSpreadMax\ over the radii of any one example, and at radius \tcContrastRadius\ it runs from \tcKappaRatioLo\ on
\tcKappaRatioLoName\ to \tcKappaRatioHi\ on \tcKappaRatioHiName. Over the certified examples and the radii,
$r\varepsilon$ has median \tcREpsMedian, and $\varepsilon/q_r$, the share of the windows where regions meet that are
forbidden, has median \tcEpsOverQMedian.

On Wall, mending removes few of the copy's forbidden windows, and the mended rate still falls as $1/r$: $r$ times it is
\tcWallRMendSixtyFour\ at radius \tcContrastRadius\ and \tcWallRMendFiveTwelve\ at radius \tcRadiusMax, against
\tcWallREpsSixtyFour\ and \tcWallREpsFiveTwelve\ before mending. Wall's floor with the example's statistics falls below
the threshold only from radius \tcWallFloorBelowThrR, so beyond that radius Theorem~\ref{thm:floor} alone does not
exclude a generator on independent coins with the example's statistics that is valid on Wall; at radius \tcRadiusMax\
each of the copy's three canvases still keeps more than one forbidden window in a thousand.

On the uncertified examples mending repairs far more of the copy's defects, in an analysis we made after the run. At
radius \tcContrastRadius, mending removes the largest share of the copy's forbidden windows among the certified
examples on \tcRemCertMaxName, \tcRemCertMax, and the smallest among the uncertified ones on \tcRemUncertMinName,
\tcRemUncertMin; the medians are \tcRemCertMedian\ and \tcRemUncertMedian. The uncertified side includes the
\tcResistInvalidSixtyFourWords\ uncertified examples that resist independent coins, although every one of them stays
invalid. At radius \tcRuleRadius, where the regions are smaller, the ranges overlap: mending removes up to
\tcRemCertMaxThirtyTwo\ on a certified example (\tcRemCertMaxThirtyTwoName) and as little as \tcRemUncertMinThirtyTwo\
on an uncertified one (\tcRemUncertMinThirtyTwoName).

At these two radii the copy makes nothing valid that the independent-coin rule or block-shifted tiles leave invalid.
At radius \tcRuleRadius\ it is valid after mending on \tcUncertValidThirtyTwo\ uncertified examples, all among the
\tcAutoUncertValidThirtyTwo\ that the rule makes valid, and at radius \tcContrastRadius\ on \tcUncertValidSixtyFour, all
among the \tcTilesValid\ that the tiles make valid (Section~\ref{sec:learning}). At radius \tcLadderMax, where the
regions are larger, it is valid on \tcControlsValidOneTwentyEightWords\ of the \tcControlsWords\ uncertified examples
of the ladder, among them \tcResistValidOneTwentyEightNames, which resists independent coins.

\subsection{The mending maps}
\label{app:mendall}

\begin{table*}[p]\centering\tiny
\caption{The mending map of the independent-coin rule: one $256^2$ canvas per example, run for \mainR\ rounds with
four repair rounds and mended as a whole, without cluster completion, within radius 4 with blocks of 8 and within radius
8 with blocks of 16. Columns: the number $|K|$ of distinct windows, the rule's raw forbidden-window rate, the residual
after each mender, the share of cells that mending within radius 8 changes, and the class, which records how far
mending gets under independent coins: mended within radius 4, mended once radius 8 is allowed, partial (at most one
forbidden window in a hundred at radius 8), or resistant. Figure~\ref{fig:corpus} groups the examples by this class.
The classes rest on this one canvas, and every certified example lies far from their boundaries, with at least
\certClassResidMin\ of its windows forbidden at radius 8. Sorted by the radius-8 residual.}
\label{tab:mendmapeight}
\input{tables/mendmap_r8_v12}\end{table*}

\begin{table}[!htbp]\centering\scriptsize\setlength{\tabcolsep}{4pt}
\caption{Wider mending of the independent-coin rule without cluster completion on the ten partial examples: the
forbidden-window rate after mending within radius 8, 12 and 16 (blocks of 16, 24 and 32) on the canvas of
Table~\ref{tab:mendmapeight}, and on
how many of three seeds radius 16 leaves at most one forbidden window in a thousand.}
\label{tab:widemend}
\input{tables/widemend_v12}\end{table}

\begin{table*}[!htbp]\centering\scriptsize
\caption{Mending at $1024^2$ on twelve public examples and the controls, with three independent-coin rules
per example, trained in the final configuration from three seeds and run for 32 rounds with four repair rounds; one
canvas per rule, generated and mended in $256^2$ chunks with 128-cell margins, within radius 4 with blocks of 8; means
over the three rules. Columns: the forbidden-window rate of the rule, after repair and after mending; the share of cells
changed; the share of defects mended within radius 2; and the pattern KL of the rule's output, of the mended output and
of Wave Function Collapse's $64^2$ output, which is not comparable at these sizes (Section~\ref{app:stats}).}
\label{tab:mending}
\input{tables/mending_v16}\end{table*}

Under independent coins mending makes some examples valid and leaves others far from valid. Of the \nMendEightRows\
public examples, \nMendFourOnly\ mend to at most one forbidden window in a thousand within radius 4, and \nMendEightOnly\
more, Dungeon, Angular, Office, Office2 and Cats, once radius 8 is allowed, for \nMendEight\ in all
(Table~\ref{tab:mendmapeight}). \nMendPartialEight\ are partial, at most one forbidden window in a hundred, and
\nMendWallEight\ are resistant, the skylines, the brick walls, most of the cities, the mazes \resistMazeNames\ and
the figurative Cat among them; a constant coloring satisfies \nConstSatWall\ of these. The knots and links, whose crossings
must stay consistent along a strand, are not among them. With blocks of 32 and radius 16 the independent-coin rule
becomes valid on \wideMendNames\ as well (Table~\ref{tab:widemend}), and with cluster completion at radius 24 on
\nCCFlat\ examples in all (Table~\ref{tab:radii}). From radius 8 to radius 24 it gains \ccNewFlatWords\ examples,
\ccNewFlatNames, every one of them partial, so it makes no resistant example valid at any radius we tried.

On Wall, Fabric, Flowers, Skyline and SmileCity the radius-8 residual is higher than the radius-4 one, because a block
for which the solver finds no completion is left as it was and the two maps use different block lattices. At $1024^2$ with three rules per
example and the radius-4 mender (Table~\ref{tab:mending}), Chess and Stripes stay at about \mendOneKRoundChess\ and
\mendOneKRoundStripes, Wall at \mendOneKRoundWall\ and Qud at \mendOneKRoundQud, while Col3 mends to
\mendOneKRoundColthree. The mending radii concentrate within 2 on the examples the independent-coin rule handles
best and spread to 3 and 4 on the harder ones. On Sand, Rooms, Maze and Sewers the divergence of the mended output
rises over the rule's own, since the solver's completions are not draws from the rule's distribution.

\subsection{The complete solver}
\label{app:sat}

\begin{table*}[!htbp]\centering\footnotesize
\caption{The complete solver on canvases of the independent-coin rule: on Angular, which mends within radius 8, on City
and Qud, which are partial, on Skew2 and Sewers, which are resistant and uncertified, and on Wall. That rule is the
first of the three independent-coin rules of Table~\ref{tab:mending} at 32 rounds, on one $\satCanvas^2$ canvas, and the mender works
within radius 8 on blocks of 16, without cluster completion. Columns: the blocks the mender failed on,
rebuilt exactly with their frozen surroundings, and how many of them a complete SAT solver completes; the density of
forbidden windows of that rule and of random errors placed on a canvas free of forbidden windows, both after four
repair rounds; the share of single-window instances at $\rho=8$ that are unsolvable, for its defects and for the random
errors,
\satSampledDefects\ sampled defects each; and the same share among instances with 9 to 20 other forbidden windows
nearby. Angular is left no failed block; Wall has no comparison with random errors.}
\label{tab:sat}
\input{tables/sat_v12}\end{table*}

\phantomsection\label{sec:forced}The mender's failures under independent coins are not an artifact of its capped solver. A complete
SAT solver, CaDiCaL 1.9.5 \citep{biere2020cadical}, given each block that the mender failed on within radius
\mendRadiusLo\ together with its frozen surroundings, completes none of them, on Wall or on the uncertified examples we
tested, and the rule's defects are harder to complete than random errors placed at the same density on a valid canvas
(Table~\ref{tab:sat}). This holds for each block as the rule left it: cluster completion within radius
\mendRadiusMain\ frees a wider region one cluster at a time and makes City and Qud valid under independent coins. On
Skew2 and Sewers, which resist independent coins and carry no certificate, the share of unsolvable instances hardly
changes as the radius grows, and the octave rule makes both valid.

Each test concerns one instance, a frozen surrounding and a free region, and says what no mender could do there; it
does not say that the example itself forces defects. When every cell within radius 8 of a single defect is freed, the
instance is unsolvable for \satRuleUnsatLo\ to \satRuleUnsatHi\ of the rule's defects on City, Qud, Skew2 and Sewers,
against \satRandomUnsatLo\ to \satRandomUnsatHi\ for random errors placed at the same density on a canvas free of
forbidden windows. The gap persists when the number of other forbidden windows nearby is kept fixed.

The SAT encoding is sound, and the solver decides what we ask of it. An instance frees every cell within distance
$\rho$ of one forbidden window, or of a cluster of them, fixes every other cell, and asks for a completion in which
every window that touches the free region is allowed. On free regions cut at random from canvases of Cave, Maze and City
free of forbidden windows the solver completes \satControlValidSolved\ of \satControlValidOf\ instances per example; on Chess and Stripes
with a boundary of mixed phase it refutes \satControlPhaseUnsat\ of \satControlPhaseOf\ per example; and a second
solver agrees on \satGlucoseAgree\ of the \satGlucoseOf\ instances re-solved.

The share of unsolvable defects changes little with the radius. A profile is the share of unsolvable instances at
$\rho=4$
and at $\rho=16$ for one set of defects: the independent-coin rule's at 16, 32 or 64 rounds, or the defects the mender
left or fixed, each as single windows and as clusters. On the five examples of Table~\ref{tab:sat}, \satProfileWithin\
of \satProfileOf\ profiles change by at most 10 points from $\rho=4$ to $\rho=16$, with changes from $\satProfileLo$ to
$+\satProfileHi$ points, and Wall's stays flat. On the other resistant examples, with the independent-coin rule and
pooled, \satWallsPooledSat\ of single-window instances at $\rho=8$ are solvable, and \satWallsNeverSat\ of
them have no solvable instance at any radius. An earlier design of the same test, which sampled defects without fixing
their local density, gave the same picture: Cave's defects are all solvable within radius 4 (\satOriginalCave), Maze's
mostly (\satOriginalMaze), Wall's not at radius 16 (\satOriginalWallUnsat\ unsolvable), and of the
\satOriginalLeftN\ defects the mender left on the five examples of Table~\ref{tab:sat} other than Wall,
\satOriginalLeftSat\ are solvable within radius 8.
That design left \satOriginalUnresolved\ of its instances at $\rho\le8$ undecided.

\subsection{The floor on the independent-coin rule}
\label{app:floorcheck}

\begin{table*}[!htbp]\centering\scriptsize
\caption{Theorem~\ref{thm:floor} with the independent-coin rule's own statistics, on the certified examples with an
invariant or a phase. Each row pools the rule's three $256^2$ canvases at the coin seeds of Section~\ref{sec:setup}.
$p_s$ is the share of cells whose $P\times P$ patch is locally valid with $\varphi=s$, under the full labeling. The
floor is $(1-\max_s p_s)/((2r+P+1)(P-N+1)^2)$, at radius \floorRadiusRaw\ on the raw canvases and at radius
\floorRadiusRepair\ after the four repair rounds, each beside the measured density $\varepsilon$ of forbidden windows.
The column ``spread'' is the floor at radius \floorRadiusRaw\ from the locally valid patches alone, their share minus
$\max_s p_s$, over the same denominator. The last two columns compare the rule's concentration among its locally valid
patches, $\max_s p_s$ over their share, with the example's largest occupancy.}
\label{tab:ownfloor}
\input{tables/ownfloor_v14}\end{table*}

Through the example's statistics the floor constrains the trained independent-coin rule only where that rule's
statistics lie within $\kappa_f$ of the example's, and the column $\delta$ of Table~\ref{tab:cert} gives that distance.
On \certDeltaAboveCount\ of the \certDeltaAboveOf\ examples with $P=N$, \certDeltaAboveNames, the rule's $\delta$ is at
least $\kappa_f$, so inequality~\eqref{eq:vfr} holds whatever its defect density. On \lpBigPBoundNames, where $P>N$,
its $\delta$ is also above $\kappa_f=\lpBigPBoundKappaF$, but the linear program at the lower end of its interval still
gives $\kappa(\delta)=\lpBigPBoundKappa$, so Theorem~\ref{thm:floor} still forces defects. The program over stationary
distributions is stricter than inequality~\eqref{eq:vfr}, and at the upper end of the rule's $\delta$, with defects
allowed at $\eta_0=\lpEtaRule$ (Appendix~\ref{app:lp}), it stays positive on \lpRuleDeltaStillNames. On Town the
rule's $\delta$, between \townDeltaLo\ and \townDeltaHi, lies above the program's bound $\kappa(0)$ and its interval
contains $\kappa_f$. The program over stationary distributions still gives \townThmKappa\ at $\delta=\townThmDelta$
(Table~\ref{tab:thm}), and its value can only grow as $\delta$ shrinks, so it also excludes a generator on independent
coins free of forbidden windows at the rule's $\delta$. We did not solve the program with defects for a one-way
quantity, so on Town we give no floor on the rule's defect density from the example's statistics. On the other examples of the table its $\delta$ is below $\kappa_f$, and the floor binds it. None of this
applies to the octave rule, whose coins are not independent.

Theorem~\ref{thm:floor} with the rule's own statistics needs no comparison with the example: it bounds $\varepsilon$ by
the rule's own $p_s$, so it binds the rule on \certDeltaAboveNames\ as well. Table~\ref{tab:ownfloor} measures $p_s$
on the rule's canvases, raw and after the four repair rounds, and every canvas lies above its floor, by a factor of
\ofRatioMedianRaw\ at the median on the raw ones. The floor at radius \floorRadiusRaw\ exceeds the threshold of
Section~\ref{sec:setup} on \ofAboveCountWords\ of the \ofExamplesWords\ examples, all but \ofBelowNames, as we
expected before measuring, but this count leans on the rule's own defects: when $P=N$ a patch is a single window, so
the share of patches that are not locally valid is $\varepsilon$ itself. The spread of $\varphi$ among the locally valid
patches alone gives a floor above the threshold on \ofSpreadCountWords\ of them, \ofSpreadNames, and there the rule
could be valid at its radius only by concentrating the quantity more than the example does. On every example but
\ofConcLessNames, where the two agree to three decimals, the rule concentrates $\varphi$ among its locally valid patches
at least as much as the example does. On \ofConcMoreNames\
it concentrates it more, which takes its statistics farther from the example's, and on \ofExampleBelowNames\ even a
generator that spread the quantity exactly as the example does would have a floor below the threshold at this radius
(Table~\ref{tab:cert}).

%% file: tables/witness_v16.tex
\begin{tabular}{@{}lrrrrrrrr@{}}\toprule
 & unsolvable & refuted before & \multicolumn{5}{c}{with a witness} & solvable\\
\cmidrule(lr){4-8}
example & instances & search & all & $\rho=4$ & $\rho=8$ & $\rho=16$ & searched & searched\\\midrule
\multicolumn{9}{@{}l}{\emph{a row invariant}}\\
Wall & 892 & 854 & 98\% & 94\% & 99\% & 100\% & 95\% & 8\\
SimpleWall & 900 & 689 & 100\% & 100\% & 100\% & 100\% & 100\% & 0\\
Flowers & 881 & 589 & 90\% & 79\% & 94\% & 97\% & 85\% & 19\\
Skyline & 895 & 716 & 65\% & 49\% & 60\% & 85\% & 57\% & 5\\
MoreFlowers & 843 & 583 & 72\% & 62\% & 76\% & 76\% & 68\% & 57\\
Platformer & 884 & 446 & 89\% & 87\% & 84\% & 96\% & 92\% & 16\\
Skyline2 & 899 & 733 & 73\% & 60\% & 70\% & 88\% & 73\% & 1\\
3Bricks & 899 & 599 & 94\% & 86\% & 96\% & 99\% & 93\% & 1\\
\midrule \multicolumn{9}{@{}l}{\emph{a phase}}\\
Fabric & 900 & 875 & 95\% & 88\% & 98\% & 100\% & 92\% & 0\\
RedMaze & 900 & 895 & 99\% & 97\% & 100\% & 100\% & 100\% & 0\\
SimpleMaze & 900 & 894 & 99\% & 97\% & 99\% & 100\% & 100\% & 0\\
SmileCity & 898 & 738 & 79\% & 63\% & 79\% & 95\% & 74\% & 2\\
\midrule \multicolumn{9}{@{}l}{\emph{a one-way quantity}}\\
Town & 900 & 691 & 68\% & 47\% & 70\% & 86\% & 65\% & 0\\
\midrule \multicolumn{9}{@{}l}{\emph{no certificate: the labeling at patch size 4, along rows or columns}}\\
Cat & 886 & 655 & 2\% & 1\% & 1\% & 3\% & 0.4\% & 14\\
ColoredCity & 892 & 729 & 0.3\% & 0\% & 0.3\% & 0.7\% & 1\% & 8\\
Lake & 852 & 533 & 0.6\% & 0.4\% & 0.7\% & 0.7\% & 0.9\% & 48\\
Nested & 899 & 493 & 0\% & 0\% & 0\% & 0\% & 0\% & 1\\
ScaledMaze & 818 & 291 & 0\% & 0\% & 0\% & 0\% & 0\% & 33\\
Skew1 & 876 & 579 & 0\% & 0\% & 0\% & 0\% & 0\% & 24\\
Sewers & 802 & 504 & 0.1\% & 0\% & 0\% & 0.4\% & 0\% & 98\\
Skew2 & 873 & 491 & 0\% & 0\% & 0\% & 0\% & 0\% & 27\\
\bottomrule
\end{tabular}

%% file: tables/mendmap_r8_v12.tex
\begin{tabular}{lrrrrrl}\toprule
example & $|K|$ & rule & mended $\rho\le4$ & mended $\rho\le8$ & cells changed $\rho\le8$ & class\\\midrule
Cave & 55 & 0.0015 & 0.00000 & 0.00000 & 0.4\% & mendable ($\rho\le4$)\\
Font & 210 & 0.0011 & 0.00000 & 0.00000 & 0.3\% & mendable ($\rho\le4$)\\
Forest & 77 & 0.0050 & 0.00000 & 0.00000 & 3.6\% & mendable ($\rho\le4$)\\
Hogs & 32 & 0.0053 & 0.00000 & 0.00000 & 3.2\% & mendable ($\rho\le4$)\\
Sand & 79 & 0.0077 & 0.00000 & 0.00000 & 4.6\% & mendable ($\rho\le4$)\\
Maze & 139 & 0.0136 & 0.00002 & 0.00000 & 8.7\% & mendable ($\rho\le4$)\\
Paths & 51 & 0.0076 & 0.00003 & 0.00000 & 4.7\% & mendable ($\rho\le4$)\\
Spirals & 78 & 0.0264 & 0.00005 & 0.00000 & 18.8\% & mendable ($\rho\le4$)\\
NotKnot & 47 & 0.0206 & 0.00006 & 0.00000 & 13.9\% & mendable ($\rho\le4$)\\
Rooms & 61 & 0.0162 & 0.00008 & 0.00000 & 15.4\% & mendable ($\rho\le4$)\\
LessRooms & 69 & 0.0318 & 0.00017 & 0.00000 & 25.1\% & mendable ($\rho\le4$)\\
Rule126 & 124 & 0.0320 & 0.00054 & 0.00000 & 26.5\% & mendable ($\rho\le4$)\\
MagicOffice & 96 & 0.0229 & 0.00056 & 0.00000 & 14.5\% & mendable ($\rho\le4$)\\
Wrinkles & 36 & 0.0196 & 0.00060 & 0.00000 & 16.3\% & mendable ($\rho\le4$)\\
Knot & 57 & 0.0173 & 0.00068 & 0.00000 & 12.9\% & mendable ($\rho\le4$)\\
Village & 50 & 0.0094 & 0.00085 & 0.00000 & 4.2\% & mendable ($\rho\le4$)\\
Dungeon & 57 & 0.0185 & 0.00129 & 0.00000 & 15.9\% & mendable ($\rho\le8$)\\
Angular & 49 & 0.0086 & 0.00143 & 0.00000 & 4.4\% & mendable ($\rho\le8$)\\
Office & 66 & 0.0103 & 0.00202 & 0.00000 & 7.8\% & mendable ($\rho\le8$)\\
Office2 & 145 & 0.0313 & 0.00806 & 0.00000 & 22.9\% & mendable ($\rho\le8$)\\
Cats & 67 & 0.0166 & 0.00349 & 0.00057 & 10.8\% & mendable ($\rho\le8$)\\
SimpleKnot & 50 & 0.0226 & 0.00524 & 0.00109 & 15.2\% & partial\\
Lines & 35 & 0.0151 & 0.00522 & 0.00144 & 11.3\% & partial\\
City & 41 & 0.0403 & 0.02141 & 0.00155 & 25.2\% & partial\\
Water & 12 & 0.0110 & 0.00569 & 0.00259 & 5.8\% & partial\\
TrickKnot & 78 & 0.0319 & 0.01125 & 0.00291 & 24.2\% & partial\\
Qud & 61 & 0.0570 & 0.02534 & 0.00505 & 32.7\% & partial\\
Mazelike & 76 & 0.0302 & 0.02144 & 0.00693 & 21.8\% & partial\\
Link & 87 & 0.0369 & 0.01722 & 0.00727 & 21.4\% & partial\\
Mountains & 31 & 0.0140 & 0.01082 & 0.00930 & 2.8\% & partial\\
Link2 & 71 & 0.0285 & 0.01807 & 0.00984 & 12.3\% & partial\\
Lake & 155 & 0.0605 & 0.02908 & 0.01156 & 31.4\% & resistant\\
ScaledMaze & 40 & 0.0267 & 0.01818 & 0.01370 & 11.1\% & resistant\\
Sewers & 105 & 0.0545 & 0.02551 & 0.01631 & 27.7\% & resistant\\
Skew2 & 59 & 0.0343 & 0.02570 & 0.01787 & 11.5\% & resistant\\
Skew1 & 48 & 0.0312 & 0.02365 & 0.02072 & 5.4\% & resistant\\
Nested & 41 & 0.0308 & 0.02534 & 0.02409 & 3.2\% & resistant\\
MoreFlowers & 108 & 0.0574 & 0.02513 & 0.02463 & 13.2\% & resistant\\
Platformer & 109 & 0.0376 & 0.02767 & 0.02551 & 3.7\% & resistant\\
Cat & 136 & 0.0717 & 0.04782 & 0.03777 & 15.6\% & resistant\\
Flowers & 92 & 0.0598 & 0.03506 & 0.03844 & 6.2\% & resistant\\
SimpleWall & 20 & 0.0521 & 0.04622 & 0.04394 & 1.5\% & resistant\\
ColoredCity & 93 & 0.0917 & 0.06591 & 0.05048 & 16.0\% & resistant\\
3Bricks & 423 & 0.0691 & 0.05284 & 0.05402 & 3.8\% & resistant\\
Skyline2 & 162 & 0.0770 & 0.06219 & 0.05859 & 4.6\% & resistant\\
Town & 87 & 0.0794 & 0.06755 & 0.06603 & 2.8\% & resistant\\
Skyline & 153 & 0.0897 & 0.06719 & 0.06963 & 4.9\% & resistant\\
SmileCity & 47 & 0.0985 & 0.08926 & 0.09068 & 0.0\% & resistant\\
Wall & 38 & 0.1098 & 0.08959 & 0.09522 & 2.6\% & resistant\\
SimpleMaze & 16 & 0.1071 & 0.10035 & 0.09949 & 0.2\% & resistant\\
Fabric & 19 & 0.1147 & 0.10613 & 0.10701 & 0.0\% & resistant\\
RedMaze & 16 & 0.1220 & 0.11346 & 0.11346 & 0.0\% & resistant\\
\midrule
\multicolumn{7}{l}{\textit{controls}, counted in no class and in no total above}\\
Col3 & 240 & 0.0087 & 0.00587 & 0.00398 & 4.0\% & --\\
Chess & 2 & 0.0575 & 0.05768 & 0.05768 & 0.0\% & --\\
Stripes & 2 & 0.0615 & 0.06126 & 0.05940 & 0.3\% & --\\
\bottomrule\end{tabular}

%% file: tables/widemend_v12.tex
\begin{tabular}{lrrrr}\toprule
example & $\rho\le8$ & $\rho\le12$ & $\rho\le16$ & seeds\\\midrule
City & 0.00155 & 0.00043 & 0.00000 & 3/3\\
Lines & 0.00144 & 0.00025 & 0.00012 & 3/3\\
SimpleKnot & 0.00109 & 0.00074 & 0.00016 & 3/3\\
Water & 0.00259 & 0.00087 & 0.00050 & 3/3\\
TrickKnot & 0.00291 & 0.00096 & 0.00050 & 3/3\\
Qud & 0.00505 & 0.00350 & 0.00271 & 0/3\\
Link & 0.00727 & 0.00522 & 0.00391 & 0/3\\
Mazelike & 0.00693 & 0.00186 & 0.00611 & 0/3\\
Mountains & 0.00930 & 0.00851 & 0.00817 & 0/3\\
Link2 & 0.00984 & 0.00756 & 0.00859 & 0/3\\
\bottomrule\end{tabular}

%% file: tables/mending_v16.tex
\begin{tabular}{lrrrrrrr}\toprule
example & seeds & rule & + repair & + mending & cells changed & $\rho\le2$ & KL (rule / mended / WFC)\\\midrule
Cave & 3 & 0.0019 & 0.0015 & 0 & 0.5\% & 98\% & 0.019 / 0.020 / 0.095\\
Paths & 3 & 0.0069 & 0.0072 & 0.00022 & 3.8\% & 63\% & 0.042 / 0.042 / 0.097\\
Sand & 3 & 0.0116 & 0.0097 & $<10^{-5}$ & 4.4\% & 89\% & 0.081 / 0.238 / 0.319\\
Maze & 3 & 0.0132 & 0.0107 & 0 & 6.6\% & 85\% & 0.088 / 0.161 / 0.162\\
Knot & 3 & 0.0181 & 0.0166 & 0.00089 & 9.2\% & 39\% & 0.029 / 0.029 / 0.144\\
Rooms & 3 & 0.0178 & 0.0161 & 0.00027 & 11.9\% & 61\% & 0.071 / 0.175 / 0.213\\
Angular & 3 & 0.0098 & 0.0077 & 0.00152 & 2.8\% & 41\% & 0.291 / 0.307 / 0.080\\
Skew2 & 3 & 0.0335 & 0.0338 & 0.02574 & 2.6\% & 51\% & 0.035 / 0.030 / 0.096\\
Qud & 3 & 0.0403 & 0.0361 & 0.01974 & 7.7\% & 45\% & 0.245 / 0.237 / 0.191\\
City & 3 & 0.0441 & 0.0375 & 0.02684 & 3.7\% & 29\% & 0.020 / 0.015 / 0.008\\
Sewers & 3 & 0.0430 & 0.0367 & 0.01036 & 15.4\% & 51\% & 0.192 / 0.314 / 0.229\\
Wall & 3 & 0.1074 & 0.0990 & 0.08833 & 3.1\% & 71\% & 0.040 / 0.050 / 0.136\\
Chess & 3 & 0.0612 & 0.0611 & 0.06093 & 0.0\% & 72\% & 0.000 / 0.000 / 0.000\\
Stripes & 3 & 0.0625 & 0.0623 & 0.06201 & 0.0\% & 70\% & 0.000 / 0.000 / --\\
Col3 & 3 & 0.0116 & 0.0114 & 0.00764 & 1.3\% & 73\% & 0.015 / 0.016 / --\\
\bottomrule\end{tabular}

%% file: tables/sat_v12.tex
\begin{tabular}{@{}lrrrrrrrr@{}}\toprule
 & \multicolumn{2}{c}{mender's failed blocks} & \multicolumn{2}{c}{density} & \multicolumn{2}{c}{unsolvable, $\rho=8$} & \multicolumn{2}{c}{9 to 20 defects nearby}\\
\cmidrule(lr){2-3}\cmidrule(lr){4-5}\cmidrule(lr){6-7}\cmidrule(lr){8-9}
example & blocks & solved & rule & random & rule & random & rule & random\\\midrule
Angular & 0 & 0 & 0.008 & 0.008 & 23\% & 23\% & 55\% & 34\%\\
City & 142 & 0 & 0.041 & 0.046 & 92\% & 59\% & 90\% & 50\%\\
Qud & 56 & 0 & 0.032 & 0.026 & 77\% & 43\% & 69\% & 41\%\\
Skew2 & 390 & 0 & 0.031 & 0.031 & 96\% & 69\% & 97\% & 61\%\\
Sewers & 112 & 0 & 0.041 & 0.041 & 90\% & 67\% & 88\% & 47\%\\
\midrule Wall & 509 & 0 & 0.101 & -- & 100\% & -- & -- & --\\
\bottomrule\end{tabular}

%% file: tables/ownfloor_v14.tex
\setlength{\tabcolsep}{4pt}%
\begin{tabular}{@{}lrrrrrrrrrr@{}}\toprule
 & & \multicolumn{4}{c}{raw canvases, $r{=}32$} & \multicolumn{2}{c}{repaired, $r{=}44$} & & \multicolumn{2}{c}{concentration}\\
\cmidrule(lr){3-6}\cmidrule(lr){7-8}\cmidrule(l){10-11}
example & $P$ & $\max_s p_s$ & \begin{tabular}[b]{@{}r@{}}locally\\valid\end{tabular} & floor & $\varepsilon$ & floor & $\varepsilon$ & spread & rule & example\\\midrule
Wall & 3 & 0.31 & 0.89 & $1.0\cdot10^{-2}$ & 0.11 & $7.5\cdot10^{-3}$ & 0.10 & $8.6\cdot10^{-3}$ & 0.34 & 0.33\\
SimpleWall & 3 & 0.45 & 0.95 & $8.1\cdot10^{-3}$ & 0.055 & $6.0\cdot10^{-3}$ & 0.052 & $7.3\cdot10^{-3}$ & 0.48 & 0.47\\
Flowers & 3 & 0.79 & 0.94 & $3.0\cdot10^{-3}$ & 0.058 & $2.2\cdot10^{-3}$ & 0.048 & $2.2\cdot10^{-3}$ & 0.84 & 0.83\\
MoreFlowers & 3 & 0.88 & 0.94 & $1.8\cdot10^{-3}$ & 0.058 & $1.2\cdot10^{-3}$ & 0.048 & $9.5\cdot10^{-4}$ & 0.93 & 0.83\\
3Bricks & 4 & 0.74 & 0.82 & $9.5\cdot10^{-4}$ & 0.072 & $6.3\cdot10^{-4}$ & 0.061 & $3.0\cdot10^{-4}$ & 0.90 & 0.88\\
Skyline & 3 & 0.87 & 0.91 & $1.9\cdot10^{-3}$ & 0.087 & $1.4\cdot10^{-3}$ & 0.077 & $6.6\cdot10^{-4}$ & 0.95 & 0.89\\
Skyline2 & 3 & 0.88 & 0.92 & $1.8\cdot10^{-3}$ & 0.080 & $1.2\cdot10^{-3}$ & 0.068 & $6.0\cdot10^{-4}$ & 0.96 & 0.94\\
Platformer & 3 & 0.91 & 0.96 & $1.3\cdot10^{-3}$ & 0.038 & $9.1\cdot10^{-4}$ & 0.032 & $7.5\cdot10^{-4}$ & 0.95 & 0.94\\
RedMaze & 3 & 0.22 & 0.88 & $1.1\cdot10^{-2}$ & 0.12 & $8.5\cdot10^{-3}$ & 0.11 & $9.7\cdot10^{-3}$ & 0.25 & 0.25\\
SimpleMaze & 3 & 0.22 & 0.89 & $1.1\cdot10^{-2}$ & 0.11 & $8.4\cdot10^{-3}$ & 0.10 & $9.8\cdot10^{-3}$ & 0.25 & 0.25\\
Fabric & 3 & 0.49 & 0.88 & $7.5\cdot10^{-3}$ & 0.12 & $5.5\cdot10^{-3}$ & 0.11 & $5.8\cdot10^{-3}$ & 0.55 & 0.40\\
SmileCity & 5 & 0.30 & 0.60 & $1.1\cdot10^{-3}$ & 0.10 & $8.1\cdot10^{-4}$ & 0.092 & $4.7\cdot10^{-4}$ & 0.51 & 0.50\\
\bottomrule\end{tabular}

%% file: app_smile.tex
\section{SmileCity's box lattice}
\label{app:smile}

SmileCity stays invalid under the input that the procedure prescribes: the rule does not keep its faces on one
lattice, and that lattice is not a quantity that a certificate over finite patches names. Its example is one face six cells wide inside a black border one cell wide,
so its tiling is a lattice of faces with gaps two cells wide, of period eight along both axes. Each face has rows of
eyes, a mouth row and white rows between them, and the mouth is the only place where the example puts two red cells
side by side. The search finds a phase of period two along columns, from patch size five, and on every canvas of both
training seeds the rule given the phase channels respects that phase at every patch, before and after mending. Almost
every forbidden window touches a face of the wrong width instead, at vertical walls where the rule switches between the
two lattices of faces modulo eight along the rows.

The widths a face can take explain why no patch fixes that lattice. A complete SAT solver, CaDiCaL 1.9.5
\citep{biere2020cadical}, run on the example's window set, with every coloring it returns checked again by our counter
of forbidden windows, finds that a face with a bottom, a run of non-black cells with black on both sides and black
below, is satisfiable only six cells wide among the widths from one to \smileWidthMax, while a face without a bottom,
running to the edge of the grid, is satisfiable at exactly the widths that are multiples of three, at every height from
two to \smileHeightMax, and so is a column of faces that never closes. Strips of eight rows that wrap along the rows are
periodic with periods \smilePeriodsNoMouth\ when they hold no mouth, and only with periods \smilePeriodsMouth\ when
they hold a mouth with the rows above and below it. Whether a face has a bottom is not a property of any finite patch: a
column of faces three cells wide, alternating rows of eyes and white rows with no mouth, fits beside the example's gap
columns at any vertical alignment, so strips that repeat the example's eight columns $m$ times and then such a column are
periodic along the rows with the odd period $8m+5$ and pass through the example's patches. Those patches therefore lie
in one strongly connected component whose cycle lengths share no divisor, and the search finds no invariant and no phase
along rows at any patch size up to \smileCertPmax, where the number of locally valid patches passes the cap, nor at
\smileCertPbeyond\ with the cap raised.

The solver also shows where a world free of forbidden windows can change the lattice: between two strips of faces
separated by two black rows across the whole width, at every shift, and along the rows through a column of faces three
cells wide that never closes, which shifts it by \smileShiftNarrow. In the instances we solved it never changes without
such rows, although black strips that do not span the width occur in them, and never along a strip bounded by black
rows. The rule's vertical walls
between the two lattices are neither, and mending within radius \mendRadiusMain\ cannot remove them, since a wall runs
through many faces. Within radius \mwRadiusA\ mending lowers the share of forbidden windows to \mwSmileRatioA\ of its
value, still above the threshold. Within radius \mwRadiusB, where the mender's first block covers nearly the whole canvas
and rewrites \mwSmileCellsB\ of its cells, it removes them on every canvas, and the total variation of the window
statistics from the example's rises from \mwSmileTvR\ to \mwSmileTvB\ (Appendix~\ref{app:mwide}). The lattice
is within the rule's reach, since the octaves' grid already carries the position modulo eight along the rows, eight
dividing \coinPeriod, and with the coordinates modulo \coinSquareSides\ given as channels the rule still leaves
\glSmileRate\ of its windows forbidden after mending (Table~\ref{tab:smile}), so it does not place its faces on the
lattice. The lattice is no obstacle to random access as such: since the example's first and last rows are black,
shifting each band of eight rows of the tiled example by an offset drawn from the band's coin leaves no forbidden window
for any value of the coins and keeps the example's statistics exactly (Appendix~\ref{app:converse}). The learned rule
does not find such worlds, and we report SmileCity as open for it.

%% file: app_diffusion.tex
\section{The diffusion model and its protocol}
\label{app:diffusion}
\begin{table*}[!htbp]\centering\scriptsize
\caption{The diffusion model of Section~\ref{sec:method} by input and class of example: the number of examples
($n$), the number valid after mending within radius \mendRadiusMain\ at each training seed, and over the models of both
seeds the median ratio of the model's share of forbidden windows before mending to a reference generator's on the same
example, with the median share of cells the mender changes for the model and for the reference. The reference is the
independent-coin rule, written ``the rule'' in the table, except with the input on the certified examples, where it is the
method.
The controls are uncertified: \ddTwoNames, which the rule makes valid, and \ddCtrlInvalidNames, which it leaves
invalid. Given the input, the valid models split into those whose every canvas has at most \ddRepBar\ of its
$16\times16$ patches in the tiled example and those whose canvases are the tiled example. On noise alone the mender
changes few of the model's cells on the certified examples because it removes few of their forbidden windows, a median
of \ddRemovedCertMedianA, against \ddRemovedCtrlMedianA\ on the controls the rule makes valid.}
\label{tab:diffusion}
\input{tables/diffusion_v15}
\end{table*}

The diffusion model of Section~\ref{sec:method} is a generic convolutional denoiser that draws each cell from
independent noise within a finite radius, so Theorem~\ref{thm:floor} applies to it. Of its settings only the training
budget and the number of levels were chosen on a trial run, described at the end of this appendix.

\paragraph{Denoiser.}
The denoiser is fully convolutional. A $3\times3$ convolution maps the noisy image to \ddChannels\ channels;
\ddBlocksWords\ residual blocks follow, each a $3\times3$ convolution, a per-channel scale and shift set by the noise
level \citep{perez2018film}, a SiLU and a $1\times1$ convolution added to the block's input; and a $1\times1$ convolution
gives the predicted noise of each color. No convolution is padded and nothing pools, strides, attends or normalizes over
space, so one step reaches \ddRadiusPerLevel\ cells, one for each $3\times3$ convolution, and commutes with
translations. The model has \ddParamsM\ million parameters, as the example's colors and input vary.

\paragraph{Examples and input.}
On noise alone the model is trained on the \certResistantWords\ certified examples, on Chess and Stripes, and on the
\tcControlsWords\ uncertified examples of Appendix~\ref{app:mincoinrun}, of which the independent-coin rule makes
\ddTwoNames\ valid. With the input, the first convolution also reads at every level the input
that the decision procedure of Section~\ref{sec:cert} gives the method's rule, without the octaves: the state of the
cell's row, drawn by the row-type sampler, where the certificate is an invariant, and the phase channels where it is a
phase. This covers Chess, Stripes and every certified example but Town, whose one-way quantity gets no input.

\paragraph{Training.}
Each model learns Gaussian diffusion \citep{ho2020ddpm} on one example's one-hot colors, scaled to $\pm1$, with
\ddTrainLevels\ levels on the cosine schedule of \citet{nichol2021improved} (offset \ddCosineOffset, the noise variance
added at each level capped at \ddBetaMax), and predicts the added noise. An update draws \ddBatch\ crops of \ddCrop\ by
\ddCrop\ cells from the tiled example at uniformly random offsets, each at a uniformly drawn level, and the loss is the
squared error of the predicted noise on the cells that the unpadded convolutions leave, so neither padding nor position
enters training. With the input, a crop carries the example's own input at its cells, the states of its rows or its
coordinates modulo the period, set to zero on \ddDropShare\ of the crops so that the network also predicts without it.
Training runs \ddUpdates\ updates of AdamW \citep{loshchilov2019adamw} at a constant learning rate of \ddLR, and
canvases are sampled from a moving average of the weights with decay \ddEMA. The training seed, 0 or 1, sets the
initial weights and every draw. The loss has no term of the rule's objective, equation~\eqref{eq:objective}, the
variation term included.

With the input, training shows the denoiser only the example's own sequence of row states, and on Wall, SimpleWall and
3Bricks most stretches of rows that one step reads, as the sampler draws them, never occur in it. On Wall at training
seed 0 the model is valid after mending on the sampler's input, and also on the example's own sequence repeated down
the canvas, on canvases kept out of every count.

\paragraph{Sampling.}
Canvases are drawn by ancestral sampling \citep{ho2020ddpm} over \ddLevels\ of the \ddTrainLevels\ levels, spaced evenly
with both ends kept \citep{nichol2021improved}. Each step clips the predicted clean image to $[-1,1]$ and draws the next
state from the Gaussian posterior given that image and the current state; a cell's color is the argmax of the last
predicted clean image, taken before clipping. The noise of every step is hashed from the coin seed, the step, the
cell's absolute coordinates and the channel \citep{salmon2011parallel}. The secondary sampler is the deterministic
sampler of \citet{song2021ddim}, over the same levels from the same initial noise. With the input, the primary sampler
is the plain conditional model, and guidance \citep{ho2022cfg} at weight $w=\ddGuidance$ replaces the predicted noise
$\hat\epsilon(x,c)$ by $(1+w)\,\hat\epsilon(x,c)-w\,\hat\epsilon(x,0)$.

\paragraph{Drawing a canvas.}
A canvas of side $S$ is the center of one of side $S+2r$, where $r=\ddRadius$ is the reach of \ddLevels\ steps. The
denoiser reads the larger canvas padded with zeros by \ddRadiusPerLevel\ cells at every level, and the padding's
influence moves inward by at most that much per level, so the kept cells equal the sampler's output on the whole plane.
A cell depends only on the noise and input within $r$ cells of it, both computed at absolute coordinates, so canvases
drawn at different origins agree where they overlap.

\paragraph{Evaluation.}
Canvases are measured as the rule's are. The primary sampler draws canvases on the development and robustness seeds,
the secondary sampler and guidance on the development seeds only. The sampler has no repair rounds, so the mender of
Section~\ref{sec:method} takes its output as it is, within radius \mendRadiusMain\ on blocks of \mendBlockMain, with
cluster completion, in \mendPassesWords\ passes. Each canvas is measured before mending, as the \emph{raw} canvas, and
after, by the defect rate, total variation, normalized agreement and repetition of Appendix~\ref{app:measures}, with the
share of cells the mender changes; aligned variation is taken over the mended canvases of the development seeds. On the
raw canvases of the examples with a transported quantity, $p_s$ pooled over the development seeds gives the floor of
Theorem~\ref{thm:floor} at radius \ddRadius, and the switch rate of Appendix~\ref{app:measures}, taken on patches of the
certificate's size, counts how often the quantity departs from its transport between consecutive locally valid patches
along its direction, per hundred patch positions. A sweep draws one raw canvas of side \ddSweepSize\ at each of
\ddSweepLevels\ levels on noise alone at training seed 0, on Wall, SimpleWall, Fabric and Cave.

\paragraph{What the model does with and without the input.}
Given the input, \ddCountedRepMaxWords\ of a canvas's $16\times16$ patches occur in the tiled example on
\ddThreeCountedNames, so those worlds are not copies. Nor do the mended worlds, which the model and the mender make together,
copy whole rows, as the copier does while keeping
few of the example's patches: in a test we wrote down before reading its canvases (Table~\ref{tab:record}), none of
the \ddRowTestRows\ rows with more than one color on their mended canvases, \ddRowTestCanvasesWords\ per example over both training seeds, is a row of the example up to a shift, against our
prediction that on Wall and SimpleWall at least half would be, while on \ddThreeEscapeNames\ every such row is. On Wall
\ddRowTestWallCourseLo\ to \ddRowTestWallCourseHi\ of the model's windows as tall as a course and 16 cells wide occur
in the example. On \ddFourZeroWords\ of the examples with the input the
quantity never switches value along its direction (Section~\ref{sec:diffusion}). On the two skylines, where the model
given the input stays invalid, the mender changes at least \ddThreeInvalidChangedMin\ of their cells. From noise
alone the model is a rougher generator than the independent-coin rule, whose objective penalizes forbidden windows: on
the certified examples it leaves \ddCertRawOverAutoMinA\ to \ddCertRawOverAutoMaxA\ times as many forbidden windows as
the rule does before its repair rounds, and on the controls a median of \ddCtrlRawOverAutoMedianA\ times as many. More
levels barely help: over \ddSixLevelsHigh\ levels its share on Wall, SimpleWall and Fabric is at least \ddSixMin\ of its
share over \ddSixLevelsLow. More levels also make each denoising step smaller, so the sweep does not isolate a cause, but the
min-coin copy does far better at the model's radius (Figure~\ref{fig:mincoin}), so the radius alone does not force
rates this high.

\paragraph{Seams.}
Chunks drawn separately agree with the whole canvas, and only the precision of the arithmetic limits how exactly a
canvas is reproduced. On Wall, under both conditions and at both training seeds, a chunk of \ddSeamChunk\ by
\ddSeamChunk\ cells drawn on its own larger canvas is compared with the same cells of the whole canvas. Canvases are
sampled in half precision, in which the two pass through convolutions of different shapes that the GPU library does not
promise to round alike, so each check is repeated in single precision and read there. In single precision
\ddSeamDiffSingleWords\ of the chunk's cells differ from the whole canvas, and in half precision \ddSeamDiffHalfWords.
The whole canvas drawn in half and again in single precision differs on \ddHalfSingleShareLo\ to \ddHalfSingleShareHi\
of its cells.

\paragraph{Budget and levels.}
The training budget and the number of levels were set on Cave before the study's other runs, against a bar fixed in
advance that no setting met. The setting was to be the smallest budget, then the fewer levels, at which the model on
noise alone is valid on Cave after mending and its raw rate at each development seed is at most \ddPilotBarWords\ times
the independent-coin rule's raw rate, before repair, at that seed, so that a failure on the certified examples could not
be put down to a poor sampler. Cave was trained at training seed 0 for \ddUpdates\ and \ddPilotUpdatesHigh\ updates and
sampled at \ddPilotLevelsLow\ and \ddLevels\ levels; Wall, trained alongside under both conditions, took no part in the
choice. Every setting was valid on Cave after mending and none met the bar: more updates raised the raw rate, and more
levels lowered it a little. We ran the study at the setting with the lowest raw rate, \ddUpdates\ updates and \ddLevels\
levels, and the models of Wall and Cave trained there for the choice are the study's models at training seed 0. At that
setting Cave's raw rate is \ddCaveRawOverRule\ times the independent-coin rule's, and the mender changes \ddCaveChanged\
of its cells against \ddCaveChangedRule\ for the rule. The model is thus rougher than the rule before mending, and
Table~\ref{tab:diffusion} gives its raw rate against the rule's and the share of cells the mender changes beside every
count.

\paragraph{The other samplers.}
The counts change little with the sampler. Given the input, the deterministic sampler is valid after mending on
\ddDdimValidB\ of the \ddDdimOfB\ models and guidance on \ddGuidValidB, against \ddPrimValidB\ for the primary sampler;
on noise alone the deterministic sampler is valid on \ddDdimValidA\ of the \ddDdimOfA\ models, and the primary sampler
on \ddPrimValidA.

%% file: tables/diffusion_v15.tex
\setlength{\tabcolsep}{4pt}%
\begin{tabular}{@{}llrrrlrrr@{}}\toprule
 & & & \multicolumn{2}{c}{valid after mending} & & & \multicolumn{2}{c}{cells the mender changes}\\
\cmidrule(lr){4-5}\cmidrule(l){8-9}
 & examples & $n$ & seed 0 & seed 1 & reference & \begin{tabular}[b]{@{}r@{}}raw rate over\\the reference's\end{tabular} & model & reference\\\midrule
noise alone & certified & 13 & 0 & 0 & the rule & 2.1 & 1.9\% & 10\%\\
 & controls the rule makes valid & 6 & 2 & 2 & the rule & 7.4 & 19\% & 18\%\\
 & controls the rule leaves invalid & 4 & 0 & 0 & the rule & 2.2 & 11\% & 31\%\\
 & Chess and Stripes & 2 & 0 & 0 & the rule & 1.7 & 3.7\% & --$^\ddagger$\\
\addlinespace
with the input & certified, with a transported quantity & 12 & 8 & 8 & the method & 3.0$^\dagger$ & 9.3\% & 11\%\\
 & \quad of these, repetition at most 0.05 on every canvas &  & 5 & 5 &  &  &  & \\
 & \quad of these, the tiled example &  & 3 & 3 &  &  &  & \\
 & Chess and Stripes & 2 & 2 & 2 & the rule & 0 & 0\% & --$^\ddagger$\\
\bottomrule
\multicolumn{9}{@{}l}{$^\dagger$ Leaves out Fabric, RedMaze and SimpleMaze, on which the method's rate before mending is zero.}\\
\multicolumn{9}{@{}l}{$^\ddagger$ The rule's rate on Chess and Stripes is that of its earlier evaluation, whose mender differs, so its share of changed cells is not given.}\\
\end{tabular}

%% file: app_results.tex
\section{Further results}
\label{app:results}

\subsection{Validity across radii, canvases and seeds}
\label{app:rulemap}

\begin{table}[!htbp]\centering\scriptsize\setlength{\tabcolsep}{3pt}
\caption{Validity at three mending radii, with blocks of twice the radius and cluster completion: the examples (of
\nMendMOf) valid on all three canvases for the octave rule, the method (radius 24 only), the independent-coin rule,
block-shifted tiles (radius 24 only) and i.i.d.\ colors drawn from the example's color frequencies (at radius 16 only on
the examples the octave rule leaves invalid at radius 8), and for the octave rule and the method on all eight canvases
of the robustness test; for the octave rule, the median share of cells that
mending changes on its valid examples and the smallest margin at which chunks mended separately agree with
whole-canvas mending on every canvas tested (Appendix~\ref{app:seam}). Entries marked $a$ use the mender without
cluster completion, which at radius 8 makes the same examples valid. A dash marks a test we did not run.}
\label{tab:radii}
\input{tables/radii_v14}\end{table}

\begin{table}[!htbp]\centering\scriptsize\setlength{\tabcolsep}{3pt}
\caption{Validity on other coin seeds and from other training seeds, within radius \mendRadiusMain. Upper part, at
training seed 0: the examples valid on the three canvases of the development seeds, and of these the examples valid on
the three canvases of the fresh seeds, which no decision used; on the examples without a certificate the method is the
octave rule. Lower part: the examples
valid at training seeds 0, 1 and 2 and at all three, for the method on the \methodSeedsCertOfWords\ certified examples
other than SmileCity, whose method rule was trained from seeds 0 and 1 but not from seed 2 (Table~\ref{tab:smile}), and for the octave rule on all
\nPublic\ public examples; and the range of the total variation (TV) of the method's $3\times3$ window frequencies from
the example's on Wall, where every canvas at every training seed is free of forbidden windows and
Theorem~\ref{thm:nogo} keeps every generator on independent coins that is free of them at least
$\kappa_f=\wallKappaFFrac$ away.}
\label{tab:seeds}
\begin{tabular}{@{}lcccc@{}}\toprule
 & \multicolumn{2}{c}{coin seeds} & & \\\cmidrule(lr){2-3}
 & development & fresh & & \\\midrule
the method & \nMethodAll & \freshMethodKept & & \\
independent coins & \nCCFlat & \freshFlatCount & & \\\midrule
 & \multicolumn{3}{c}{training seed} & \\\cmidrule(lr){2-4}
 & 0 & 1 & 2 & all three\\\midrule
the method, certified & \methodSeedZeroCertCount & \methodSeedOneCertCount & \methodSeedTwoCertCount
  & \methodAllSeedsCertCount\\
the method on Wall, TV & \ruleTVWallLo--\ruleTVWallHi & \pbWallSeedOneTVLo--\pbWallSeedOneTVHi
  & \pbWallSeedTwoTVLo--\pbWallSeedTwoTVHi & \\
the octave rule & \nCCM & \trainSeedOneCount & \trainSeedTwoCount & \trainSeedAllCount\\\bottomrule
\end{tabular}\end{table}

The octave rule's lead over independent coins grows with the mending radius, from \ccLeadEightWords\ examples at radius
8 to \ccLeadTwentyFourWords\ at radius 24 (Table~\ref{tab:radii}). At radius 8 that lead is net of one loss, \msLost,
which independent coins make valid and the octave rule only from radius 16 on; from radius 8 to radius 24 the octave
rule gains \ccNewMWords\ examples and loses \ccLostM. Cluster completion makes no example valid at radius 8 that
mending without it leaves invalid, under either rule. It gains at radii 16 and 24, whose blocks are large enough to
hold several clusters of defects.

Validity survives added canvases and fresh coin seeds on all but a few examples. On all eight canvases of the
robustness test the method is valid on \nMethodRobust\ examples, losing \methodValidNotRobustNames, and on the fresh
seeds, which no decision used, it loses only \freshMethodLostNames, while the independent-coin rule is valid on all
eight canvases and on the fresh seeds on the same \ccRobustFlat\ examples as on the development seeds
(Tables~\ref{tab:radii} and~\ref{tab:seeds}). Nor does validity lean
on the threshold's allowance. On \methodExactZero\ of the examples the method makes valid no canvas keeps a single
forbidden window, and the same holds on \ccExactZeroM\ of the octave rule's \nCCM\ valid examples and on
\ccExactZeroFlat\ of the independent-coin rule's \nCCFlat.

The one certified example that the method loses at another training seed (Section~\ref{sec:limits}),
\methodSeedOneLostNames, is invalid at seed 1 and valid at seed 2. On Wall every canvas at every seed is free of
forbidden windows and lies closer to the example's statistics than Theorem~\ref{thm:nogo} allows any generator on
independent coins free of them (Table~\ref{tab:seeds}). The octave rule, trained from seeds 1 and 2 on every example, is valid on \trainSeedOneCount\
and on \trainSeedTwoCount\ examples, against \nCCM\ at seed 0, and on \trainSeedAllCount\ at all three.

The mender's second pass, which re-solves the blocks of a lattice shifted by half a block, is part of the method. With
one pass at radius 24 the octave rule is valid on \onePassM\ examples and the independent-coin rule on \onePassFlat,
and the second pass takes \secondPassShareM\ of the mending time.

Wider mending rewrites more of the canvas. On the examples the octave rule makes valid at radius 8, mending changes a
median of \ccChangedEight\ of the cells at that radius and \ccChangedOld\ at radius 24; on the examples that only
radius 24 makes valid it changes \ccChangedNew, and up to \ccChangedMax\ on one example. The independent-coin rule's
median on its own valid examples at radius 24 is \ccChangedFlat, and on the examples the method makes valid the median
is \rewrittenMedianMethodFrac. The mender redraws every cell it frees, so these shares count cells rewritten whether or
not they were wrong. A pass that put back every changed cell whose old color keeps its windows allowed left the shares
almost where they were, so almost every rewritten cell would make a window forbidden if its old color returned, and we
did not adopt the pass.

\subsection{Examples from other implementations}
\label{app:freshres}
Table~\ref{tab:fresh} gives every generator of the test of Appendix~\ref{app:freshex} on each of the \fxNExamples\
examples, grouped by the repository each came from.

\begin{table*}[!htbp]\centering\small
\caption{The examples from other implementations (Appendix~\ref{app:freshex}), grouped by repository. Certificate and
input: what the search finds and what the procedure then gives the rule. Class: the independent-coin rule's class from
mending at radius $\rho$ of 4 and 8, as for the public examples. The method is the rule given that input; where the input is the
octaves it is the octave rule, and on \fxRefusedNames, whose invariant runs along a diagonal, the procedure has no input
to give and the method is the octave rule as well. Block-shifted tiles, i.i.d.\ colors and the min-coin copy are the floors, mended
in the same way. All rules are trained from training seed 0.}
\label{tab:fresh}
\input{tables/fresh_v16}\end{table*}

\subsection{Hand-written priors and the mender}
\label{app:floors}
\label{sec:floors}

\begin{table*}[!htbp]\centering\footnotesize
\caption{Hand-written rules against the learned independent-coin rule, on one $256^2$ canvas, the canvas of
Table~\ref{tab:mendmapeight}. The forbidden-window rate of the independent-coin rule at 32 rounds, raw and
after four repair rounds and mending within radius 4 with blocks of 8, of independent sampling followed by 32 or 64
rounds of our repair step, and of the Metropolis rule after 32 and 64 sweeps and after the same mending; the pattern KL
to the example of the independent-coin rule's raw output and of the Metropolis output after 64 sweeps.}
\label{tab:baselines}
\input{tables/baselines}\end{table*}

Random-access rules written by hand do worse than the learned independent-coin rule, in rate or in divergence, on
every example we ran them on. The first draws every cell independently from the example's color marginals with hashed
coins and then runs our own coin-gated repair step for 32 or 64 rounds. The second, the Metropolis rule, puts hashed
coins on the window energy $E=\Lambda\cdot\#\{\text{forbidden windows}\}-\sum_{\text{allowed}}\log f(w)$, with
$\Lambda=2\max(-\log f)$, updated in nine parallel phases of radius 2 and annealed from temperature 1 to 0.1 over 32 or
64 sweeps, so that both read far beyond the learned rule's radius. Both were run on the examples of
Table~\ref{tab:mending} with Chess, Stripes and Col3, at $256^2$ and $512^2$, and the two sizes give the same
counts.

The Metropolis rule's raw forbidden-window rate is at least twice the learned rule's on \nBone\ examples at 32 sweeps
and on \nBoneSixtyfour\ at 64, at median ratios of \medRatioMHrawThirtytwo\ and \medRatioMHraw. Independent sampling
followed by repair sits \medRatioFonerepThirtytwo\ times above the learned rule at the median after 32 rounds and
\medRatioFonerep\ times after 64, and at least ten times above it on \nBthree. The Metropolis rule's divergence is worse
than the learned rule's on \nBtwo. After mending, the learned rule's residual is at most half the Metropolis rule's on
\nBfour, at a median ratio of \medRatioMHmended. No hand-written rule comes within twice the learned rule on both the
rate and the divergence on any example (\nFalsifier). Sand is the one example where a hand-written rule reaches a lower
rate than the learned rule, and it does so by collapsing to nearly one color, which its energy rewards and its
divergence reports (Table~\ref{tab:baselines}). The term $-\sum\log f(w)$ is smallest when the most frequent window
fills the canvas, which is why our objective carries a color-marginal term.

As a prior for the mender of the main protocol, run for \mhSweeps\ sweeps before the method's repair rounds, the
Metropolis rule is valid on \mhCCValid\ examples, \mhCCCertValid\ of them certified, as many as the octave rule run for
\roundsFewest\ rounds. It reaches \mhCCOnlyNames, which neither learned rule makes valid
(Appendix~\ref{sec:open}). A longer run adds nothing. At \mhLongSweeps\ sweeps, \mhLongFactorWords\ times as many, it
stays invalid on all \mhLongOfWords\ uncertified examples it misses at \mhSweeps, and the octave rule makes
\mhLongOctValidCount\ of them valid, so the octave rule's lead there does not come from the budget of the sampler.

What the learned rule leaves decides how far the same mender gets. The octave rule's weights, run for \roundsList\
rounds before the same four repair rounds and the same mender within radius 24, are valid on \roundsCounts\ examples,
each set contained in the next, while the median share of forbidden windows before mending falls from \roundsRawFour\
at \roundsFewest\ rounds to \roundsRawThirtyTwo\ at \mainR.

Mending alone does not make a world from noise. Started from i.i.d.\ colors at the example's color frequencies and
mended within radius 24, each of the \iidAllOf\ public examples keeps at least \iidAllResidualMin\ of its windows
forbidden on every canvas. Nor do the row-type sampler and the mender make a wall on their own: from the painted row
types of Appendix~\ref{app:references}, mended within radius 24, between \samplerFloorResLo\ and \samplerFloorResHi\ of
the windows stay forbidden on Wall, Skyline and Skyline2. In both cases the mender frees regions bordered by noise on
every side, and almost none of them can be filled; on the painted row types it changes no cell.

Block-shifted tiles, whose forbidden windows all lie on block boundaries, are valid after mending on \tileCCValid\
examples, all of which the octave rule makes valid but \tileNotRuleNames, and on \tileCCCertValid\ of the certified
ones: there, two blocks whose courses or phases differ meet at a boundary that no completion within the mender's box
joins, as Proposition~\ref{prop:blocks}(a) leads one to expect. A median of \tileRepetitionMedian\ of their
$16\times16$ patches occur in the example, so their worlds are copies cut at the seams. The octave rule makes
\ruleNotTileCount\ examples valid that the tiles do not.

\subsection{The input alone, and composition under it}
\label{app:copier}

\begin{table}[!htbp]\centering\scriptsize\setlength{\tabcolsep}{2.3pt}%
\caption{Fillers of the row types on the \rowInvCountWords\ examples with a row invariant: the copier (none exists on
3Bricks), the re-solved copier, which re-solves the copier's canvas block by block, the method's own rule and the
Metropolis filler on the row types after \mhTypedSweeps\ sweeps, all mended within radius \mendRadiusMain. Valid means valid on all three canvases. The share of forbidden windows is that
of the canvas the mender receives, the median over the canvases; the other columns give the range over the canvases of
the share of cells mending changes, of rows equal to a row of the example up to a horizontal shift, of $16\times16$
patches found in the tiled example, and of the total variation (TV) of the $3\times3$ window frequencies from the
example's after mending. The column of cells changed counts only the mender's changes: the re-solved copier's own two
passes redraw \crRedrawnLo\ to \crRedrawnHi\ of the cells before the mender receives the canvas. On every example the
only rows of the re-solved copier and of the rule that equal a row of the example are rows of one color, and on Wall
those of the Metropolis filler are its mortar rows.}
\label{tab:copier}
\input{tables/copier_v14}
\end{table}

The copier is valid before mending wherever it exists (Section~\ref{sec:valid}), and Section~\ref{sec:learning}
compares it with the rule. Given the row types and only its own independent coins, without the octaves, a learned rule
is valid on \flatSamplerValidCountWords\ of the three examples we tried, \flatSamplerValidNames. On \phaseLookupNames,
each of them one period of its phase in size, painting the example's cell at the coordinates the phase channels give
yields the tiled example (Section~\ref{sec:valid}), and on the two mazes this world is the only valid one up to
translation. Given only the coordinate along the certificate's direction, the rule is valid on
\phaseOneCoordValidNames\ and on neither maze.

On Wall and SimpleWall, whose courses are short and whose offsets at the joints vary, fewer than \copierRepWallBelow\
of the copier's $16\times16$ patches occur in the tiled example, against \copierRepLongLo\ to \copierRepLongHi\ where
the free runs are long, and on Wall its window statistics lie \copierTVWallLo\ to \copierTVWallHi\ from the example's
in total variation (Table~\ref{tab:copier}). The method's rule on Wall leaves about
\wallRawMethodWords\ of its windows forbidden before mending, mending rewrites \ruleChangedWallLo\ to
\ruleChangedWallHi\ of its cells, and between \ruleRepHWallLo\ and \ruleRepHWallHi\ of its windows as tall as a course
and 16 cells wide occur in the example. On the \rowInvCountWords\ examples with a row invariant, the octave rule's share
of rows that are rows of the example, up to a horizontal shift, lies between \octRowCopyLo\ and \octRowCopyHi.

To generate a chunk, the re-solved copier draws the copier's canvas over every block the chunk depends on, re-solves
each block of the random-access block solver's lattice under its frozen surroundings with the same complete solver, in
two passes, and then mends the chunk with the main mender, which finds nothing to change. The passes redraw
\crRedrawnLo\ to \crRedrawnHi\ of the cells. One thread spends \crCpuLo\ to \crCpuHi\ CPU seconds per $256^2$ canvas
on the re-solved copier, the region its blocks depend on included. On the examples it makes valid, the octave rule
takes a median of \genCPUMedian\ seconds per canvas with its repair rounds and \mendCPUTwentyFour\ for mending within
radius \mendRadiusMain, and a $256^2$ chunk mended alone must generate and mend \chunkAreaMarginAll\ times its own area
at the measured margin (Appendix~\ref{app:seam}).

The re-solved copier's statistics are those of the solver. On Wall they lie \crTvWallLo\ to \crTvWallHi\ from the
example's, against \crTvBlockWall\ for the random-access block solver run from the tiled example
(Table~\ref{tab:replayra}). On each of the \crTvBlockExamplesWords\ examples where both were
measured, every example with a copier but Skyline2, its distance from the example's statistics is within
\crTvBlockDiffMax\ of the distance the same solver reaches when it starts from the tiled example; among them is
Platformer, the one example where the re-solved copier lies farther from the example's statistics than the rule. On Wall
its bricks are nearly the block reference's, with a median of \crBrickMedianRange\ cells and a 90th percentile of
\crBrickPNinetyRange, against \wallExBrickLength\ throughout the example, and so are its joints
(Table~\ref{tab:wallbricks}). It keeps less of the tiling than the copier on \crNrepBelowCopierWords\ examples, since
the passes redraw the copied rows while the sampled layout stays. On Wall its worlds vary between seeds at least as much
as the method's and keep as little of the tiling (Table~\ref{tab:wallvar}).

\begin{table*}[!htbp]\centering\scriptsize
\caption{Bricks, joints and copies on Wall. Each entry is the range over a generator's canvases ($256\times256$, mended
within radius 24 where the generator is mended; the method on the development and fresh seeds, the others on the
development seeds). Forbidden windows are a rate; joints are a share of course cells; the total variation counts all
forbidden windows as one outcome, and the next column is the share of it forced by the shortage of windows with a
joint. The octave rule and the independent-coin rule are not valid on Wall.}
\label{tab:wallbricks}
\input{tables/wallbricks_v15}\end{table*}

The rule's distance from the example's statistics on Wall lies in how it fills its rows: the distribution of its row
types is within \wallMethodTypeMarginal\ of the example's in total variation. Wall's example has courses
of bricks between rows of mortar, and inside a course a joint of the mortar color ends each brick. We call a cell a
course cell when its $3\times3$ window does not lie on a mortar line, a joint when it is a course cell of the mortar
color, and a brick a run of other course cells along a row between two joints; runs cut by the edge of the canvas are
left out. Every brick of the example is \wallExBrickLength\ cells long, and \wallExJointPct\% of its course cells are
joints. The method's bricks are longer than the example's already before mending, and after mending their median is
\wallMethodBrickMedianRange\ cells and their 90th percentile \wallMethodBrickPNinetyRange; mending removes a net
\wallMendJointLossPctLo\% to \wallMendJointLossPctHi\% of the joints, since it deletes more than it adds
(Table~\ref{tab:wallbricks}). A window that holds a joint is one outcome class of the window statistics, so the total variation is at least the difference between
two generators' masses on that class; for the method this difference is about \wallMethodJointBoundPct\% of its total
variation, and the same share holds for the block reference and the Metropolis filler. Wave Function Collapse's own
output lies \wfcTVWallLo\ from the example's statistics on Wall, and the two block solvers, which keep the tiled
example's layout of rows, lie closer still, as does the re-solved copier, which follows the sampled layout.

The Metropolis filler of Appendix~\ref{app:references}, after \mhTypedSweeps\ sweeps and the same mender, is not valid
on Platformer, Skyline, Skyline2 or 3Bricks, where at least \mhTypedFailMin\ of its windows remain forbidden after
mending. On \ruleBelowMHCountWords\ examples with a row invariant the rule's worlds lie closer to the example's
statistics than the filler's on every canvas (Table~\ref{tab:copier}). The filler also reads farther: after
\mhTypedSweeps\ sweeps its reach is \mhTypedReach\ cells on each side, against the rule's \mainR.

\paragraph{Composition before mending.}
A $16\times16$ patch that holds a forbidden window cannot occur in the tiled example, so the rule's repetition before
mending could be low because of its defects alone. Table~\ref{tab:clean} counts only the patches that hold none, and
the rows with more than one color that no forbidden window touches. On every example with a row invariant they are at
least \clCleanMin\ of the rule's patches, \clCleanPatchesTotal\ patches and \clCleanRowsTotal\ rows in all, and none of
them occurs in the example. Nor does any of the $16\times16$ patches that mending leaves untouched.

\begin{table*}[!htbp]\centering\scriptsize
\caption{The rule's composition before mending on the examples with a row invariant, counted after the run. Before
mending, after the rule's repair rounds: the share of the rule's $16\times16$ patches that hold no forbidden window
(range over the canvases), and how many of them occur in the tiled example (summed over the canvases); the rows with
more than one color that no forbidden window touches, summed over the canvases, and how many of them are rows of the
example up to a horizontal shift. After mending: the share of $16\times16$ patches none of whose cells mending changed,
and how many of them occur in the tiled example. The method's rule at training seed 0 on the canvases of
Table~\ref{tab:thm}.}
\label{tab:clean}
\input{tables/clean_v16}\end{table*}

\begin{table}[!htbp]\centering\scriptsize\setlength{\tabcolsep}{3pt}
\caption{Variation between seeds and copying of the tiling on Wall. Aligned variation is the share of cells that differ
between two canvases of one generator after the relative translation that makes it smallest, and normalized agreement
is the share of cells equal to the best translation of the tiled example, scaled to be 0 for an i.i.d.\ draw and 1 for
a translate of the tiling (Appendix~\ref{app:measures}). The method and the two copiers are measured on the three
canvases of the development seeds, mended within radius \mendRadiusMain, and the block solvers and shifted replay on
eight canvases, at the coin seeds of Table~\ref{tab:replayra}. The copier's aligned
variation was not measured.}
\label{tab:wallvar}
\input{tables/wallvar_v16}
\end{table}

\subsection{Copies and variation}
\label{app:replay}
\label{sec:pp}

\begin{figure}[!htbp]\centering\includegraphics[width=0.6\linewidth]{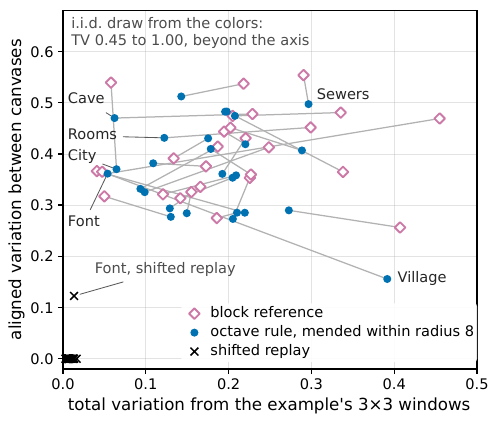}
\caption{Variation against window statistics on the \msVPopulation\ examples the octave rule makes valid at radius 8.
For each example, the variation between two canvases after the best alignment (higher means more varied) against the
total variation of their $3\times3$ windows from the example's (lower means closer statistics), for the octave rule
mended within radius 8 and for the block reference of Section~\ref{sec:setup}; a line joins the two. Means over eight
canvases. Shifted replay matches the statistics and does not vary; its value on Font reflects an alignment search
shorter than Font's period. An i.i.d.\ draw from the color frequencies lies beyond the axis.}
\label{fig:copies}\end{figure}

\begin{table*}[!htbp]\centering\scriptsize
\caption{The evaluation against replay for the octave rule, on the examples it makes valid at radius 8; means over eight
$256^2$ canvases (eight coin seeds, shared across methods). The octave rule is run for 32 rounds with four repair rounds
and mended within radius 8 with blocks of 16, without cluster completion; block is the block reference. Columns: the
total variation from the example's $3\times3$ frequencies, in which all forbidden windows together count as one
outcome; the share of cells that agree with the best translated tiling of the example, normalized as
$(a-a_{\mathrm{iid}})/(1-a_{\mathrm{iid}})$ with $a_{\mathrm{iid}}$ the agreement of an i.i.d.\ draw from the color
frequencies; the share of positions whose $16\times16$ patch occurs in the tiled example; the share of cells that
differ between two canvases after the best alignment; the share of cells changed by the fine redraw and by the full
redraw of a $32\times32$ region (Appendix~\ref{app:measures}); and the block reference's cost per cell over the octave
rule's, single-thread CPU time.}
\label{tab:variation}
\input{tables/variation_v13}\end{table*}

\begin{figure}[!htbp]\centering\includegraphics[width=0.6\linewidth]{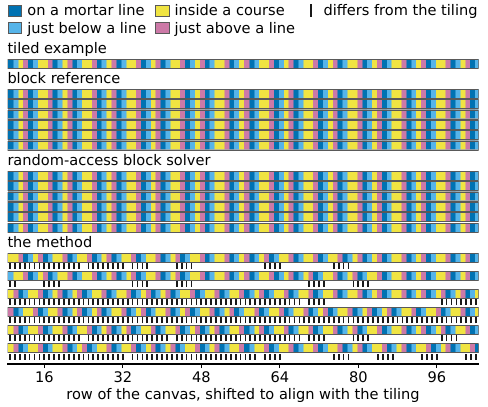}
\caption{The layout of rows on Wall. Each strip gives the types of \wallLayoutRowsPerStrip\ of the 256 rows of one
canvas, read downward, drawn left to right in the colors of Figure~\ref{fig:quantity} and shifted to align best with
the tiled example, the top strip, whose sequence both block solvers keep. The lower strips are the sequences that the
row-type sampler draws on \wallLayoutCanvasesWords\ of the method's canvases, those of the development and fresh seeds,
and that both copiers follow. A tick marks each row whose type differs from the aligned tiling.}
\label{fig:layout}\end{figure}

The block solvers keep the tiled example's layout of rows. On the \refLayoutExamplesWords\ examples whose row type is
read from a single $3\times3$ window, all but 3Bricks, every one of the \refLayoutCanvases\ canvases of the block
reference and of the random-access block solver has the tiled example's sequence of row types, shifted by the offset of
the replay it starts from. On Wall, where none of the method's canvases has the tiling's sequence
(Section~\ref{sec:learning}), only the rows around the fixed course tops where the sampler's bands meet have the same
type on every canvas (Figure~\ref{fig:layout}). On the other examples with a row invariant the method's worlds keep
more of the tiling than on Wall, up to a normalized agreement of \courseOtherNrepHi\ on \courseOtherNrepHiName, whose
bands differ only in the order of their runs (Appendix~\ref{app:procedure}).

The octave rule's valid worlds are not copies of the example either, and they vary as much as those of independent
coins. Over the examples each rule makes valid within radius 24, the octave rule's median aligned variation is
\ccAlignedMedian, against \ccAlignedFlatMedian\ for the independent-coin rule, and its median normalized agreement is
\ccNrepMedian, near that of an i.i.d.\ draw. On the \ccCommonOf\ examples both rules make valid, it is \ccNrepCommonM,
against \ccNrepCommonFlat\ for the independent-coin rule. Before mending, the window statistics of both rules are as
close to the example's as Wave Function Collapse's own output on most examples, and after mending the octave rule's lie
the closer at the median (Section~\ref{app:stats}).

Against the references at radius 8 each measure separates the methods example by example (Table~\ref{tab:variation}).
At the median over the \msVPopulation\ examples the octave rule makes valid there, none of a canvas's $16\times16$
patches occurs in the tiled example, while every patch of shifted replay does. The octave rule's normalized agreement
stays near the i.i.d.\ draw's on most examples, with a median of \msNrepMedian, and Village is the example closest to
the tiling and the one that varies least between canvases. The aligned variation searches shifts of up to 64 cells,
which cannot align Font's longer period, so shifted replay scores above zero there only. At radius \mendRadiusLo\ the
octave rule's median aligned variation is \msAlignedMedian, against \blockAlignedMedianM\ for the block reference,
which spends about \replayCostRatioInvM\ times its compute per cell, and its window statistics are the closer to the
example's on
\msTVBelowBlockCount\ of the \msVPopulation\ examples (Figure~\ref{fig:copies}).

\begin{table*}[!htbp]\centering\footnotesize
\caption{The octave rule after mending against the random-access block solver of Section~\ref{sec:setup}, which starts
from shifted replay and re-solves the $16\times16$ blocks of an absolute lattice class by class, each block seeded by
its coordinates, read after its first pass and after both. The octave rule is run under the protocol of the references
(Table~\ref{tab:protocols}): mended within radius \mendRadiusLo\ without cluster completion, on the \msVPopulation\
examples it makes valid at that radius. Means over eight $256^2$ canvases; the octave rule's values are those of
Table~\ref{tab:variation}.}
\label{tab:blocksolverrule}
\input{tables/blocksolver_rule_v13}\end{table*}

The random-access block solver after its first pass varies less than the octave rule, \bsOneAligned\ against
\msAlignedMedian, and keeps more of the tiling, with normalized agreement \bsOneNrep\ against \msNrepMedian; the octave
rule varies more on \bsOneRuleMoreVaried\ of the \msVPopulation\ examples (Table~\ref{tab:blocksolverrule}). After both
passes, the least it can run, the solver varies as much as the octave rule at the median, \bsTwoAligned, and still keeps
more of the tiling, \bsTwoNrep\ against \msNrepMedian. Example by example the two measures favor different methods: the
solver varies more than the octave rule on \bsTwoSolverMoreVaried\ of the \msVPopulation\ examples and on
\bsTwoSolverMoreVariedUncert\ of the \bsTwoUncertOf\ of them that carry no certificate, and the octave rule keeps less of
the tiling than the solver on \bsTwoRuleLowerNrep\ of the \msVPopulation\ and on \bsTwoRuleLowerNrepUncert\ of the
\bsTwoUncertOf\ (Table~\ref{tab:blocksolverrule}). Neither copies the example's patches: on none of these examples does
a $16\times16$ patch of the octave rule's canvases occur in the tiled example, and the solver's share of such patches is
at most \bsTwoRepMax, on \bsTwoRepMaxName. Costs per chunk were compared only with the octave rule mended within radius
\mendRadiusMain\ with cluster completion, as in the corpus evaluation: at the margin that mending needs, the solver's
cost per chunk over the octave rule's has a median of \raOverRuleChunkAtMargin\ (Appendix~\ref{app:seam}).

Edits are contained by construction, within the reach of Section~\ref{sec:ra}, and they change little of the world.
Redrawing the fine coins of a $32\times32$ region and mending again changes \ccLocalFineMedian\ of the canvas at the
median, against \ccLocalAllFlatMedian\ when the independent-coin rule's coins there are redrawn, and at least
\ccLocalValidMin\ of the changed cells lie in allowed windows only, on every example the octave rule makes valid. The
fine redraw changes few cells on examples such as Angular and City, where the coarse coins decide most of the layout,
and as many as the full redraw on examples such as Forest and Hogs. Resampling only the cells' own coins on a
$16\times16$ region of a raw canvas changes \msCoinEditCellsMedian\ cells at the median, against
\flatCoinEditCellsMedian\ under independent coins, so most of the variation of the octave rule's worlds comes from its
coarse coins.

No rule is trained on structure larger than the window, and the two rules keep the example's connectivity about
equally often. On the \msConnFlatSetOf\ examples of the independent-coin rule's valid set at radius 8 that have a
passable color, the largest region of that color holds at least four fifths of its share under shifted replay on
\msConnFlatSet\ under the octave rule and on \flatConnFlatSet\ under independent coins. On City, Dungeon and Angular
the octave rule's correlation stays above a tenth of its value at lag one at every lag we measured, up to \msCorrLagM,
while the independent-coin rule's falls below it at lag \msCorrLagFlatLo\ to \msCorrLagFlatHi, and part of what
persists follows the grid of the octaves (Figure~\ref{fig:corr}). Appendix~\ref{app:measures} defines both measures.

\subsection{The octaves and the variation term}
\label{app:sharing}
\label{sec:sharing}

Training finds the direction of the rows among the octaves without being told it. The octaves share coins along rows
and along columns alike, and on SimpleWall the octave rule's mended worlds keep no forbidden window on its three
canvases (Figure~\ref{fig:bound}), so the row type stays constant along every row. On Wall the octave rule lowers the
residual of independent coins but leaves Wall invalid (Figure~\ref{fig:corpus}); there the row types that the
certificate prescribes keep every row's type, as in Figure~\ref{fig:quantity}(c).

\begin{table}[!htbp]\centering\scriptsize\setlength{\tabcolsep}{4pt}
\caption{Block coins, the octaves of the method, against coins that vary smoothly between lattice points, as in octave
noise, each read by the octave rule's construction and trained on every public example. Columns: the examples valid on
all three canvases after mending within radius 8 without cluster completion and within radii 16 and 24 with it, and the
median of the grid statistic of Appendix~\ref{app:sharing} on the raw canvases of the examples both families make valid
at radius 8. The block coins' counts are those of Table~\ref{tab:radii}; the smooth coins' were measured with the same
menders and coin seeds.}
\label{tab:smooth}
\begin{tabular}{@{}lrrrr@{}}\toprule
 & \multicolumn{3}{c}{valid examples} & grid\\\cmidrule(lr){2-4}
coins & radius 8 & 16 & 24 & statistic\\\midrule
block (the octaves) & \nMendM & \nCCSixteenM & \nCCM & \latticeBlockStageTwo\\
smooth & \smoothEightM & \smoothSixteenM & \smoothCCM & \latticeSmoothStageTwo\\\bottomrule
\end{tabular}\end{table}

The octaves leave their grid in the output. On raw canvases we compared the $3\times3$ windows that straddle a boundary
of the blocks of side 4 or 16 with all other windows, by the total variation between their histograms, and took the
largest value over the two periods and the two axes. For the independent-coin rule, which commutes with every shift,
this grid statistic measures sampling noise alone. Over the \latticeOf\ examples the octave rule makes valid at radius
8, its median is \latticeMedianM\ for the octave rule and \latticeMedianFlat\ for the independent-coin rule. The same
statistic at periods 5 and 7, which a law periodic in 16 cannot favor, measures the octave rule's own longer-range
structure, and the value at the block periods is more than twice that on all but \latticeWithinOwnNull\ of the
examples. Coins that vary smoothly between lattice points, as in octave noise, keep random access and exact seams.
Trained in the same way on every public example, they lower the statistic to \latticeSmoothStageTwo, against
\latticeBlockStageTwo\ for the block coins, on the examples both families make valid at radius 8. They make fewer
examples valid at each radius (Table~\ref{tab:smooth}), so the method keeps the block coins.

\begin{table}[!htbp]\centering\scriptsize\setlength{\tabcolsep}{3pt}
\caption{The variation term left out, nothing else changed. Upper part: the octave rule on every public example,
mended within radius \mendRadiusLo\ without cluster completion; the examples valid on all three canvases, and the
median normalized agreement over the examples the independent-coin rule makes valid at that radius. Lower part: the
method's rules on Wall and Skyline, mended within radius \mendRadiusMain; their normalized agreement, and the ranges
over the three canvases of the total variation (TV) of the $3\times3$ window frequencies from the example's and of the
share of cells that mending changes.}
\label{tab:noterm}
\begin{tabular}{@{}lcccc@{}}\toprule
 & \multicolumn{2}{c}{with the term} & \multicolumn{2}{c}{without it}\\\cmidrule(lr){2-3}\cmidrule(lr){4-5}
\multicolumn{5}{@{}l}{\emph{the octave rule}}\\
examples valid & \multicolumn{2}{c}{\nMendM} & \multicolumn{2}{c}{\noTermMendCount}\\
normalized agreement & \multicolumn{2}{c}{\msNrepOnFlatSetMedian} & \multicolumn{2}{c}{\noTermNrepOnFlatSetMedian}\\
\addlinespace
\multicolumn{5}{@{}l}{\emph{the method}}\\
 & Wall & Skyline & Wall & Skyline\\
normalized agreement & \vZeroCmpWallWithNrep & \vZeroCmpSkylineWithNrep & \vZeroCmpWallNoNrep
  & \vZeroCmpSkylineNoNrep\\
window TV & \vZeroCmpWallWithTVLo--\vZeroCmpWallWithTVHi & \vZeroCmpSkylineWithTVLo--\vZeroCmpSkylineWithTVHi
  & \vZeroCmpWallNoTVLo & \vZeroCmpSkylineNoTVLo--\vZeroCmpSkylineNoTVHi\\
cells changed & \vZeroCmpWallWithChangedLo--\vZeroCmpWallWithChangedHi
  & \vZeroCmpSkylineWithChangedLo--\vZeroCmpSkylineWithChangedHi & \vZeroCmpWallNoChangedLo--\vZeroCmpWallNoChangedHi
  & \vZeroCmpSkylineNoChangedLo--\vZeroCmpSkylineNoChangedHi\\\bottomrule
\end{tabular}\end{table}

The variation term is not what keeps the rules from copying. Trained without it, nothing else changed, the octave rule
is valid on about as many examples, with the same median normalized agreement (Table~\ref{tab:noterm}), so it does not
take the shortcut of Proposition~\ref{prop:blocks}(b), a copy of the example on each coarse block. Under the row types
the shortcut is exact, as the copier shows, and the method's rules on Wall and Skyline trained without the term do not
take it either. The rule on Wall is then \vZeroWallValidWords\ and the rule on Skyline \vZeroSkylineValidWords, no
$16\times16$ patch of their canvases occurs in the example, and their normalized agreement is unchanged. On Wall their
rows match the example's only at the mortar, and at most \vZeroWallRepHHi\ of their windows as tall as a course occur
in the example. What the term changes is the division of labor with the mender. Without it the rules' window
statistics lie closer to the example's, on Wall as close as Wave Function Collapse's own output, but mending rewrites
more of their cells (Table~\ref{tab:noterm}). We keep the term because with it the rule leaves the mender less to
rewrite, and as a guard on the examples we did not train without it.

The term guards against the block copy under coins, not against a layout tied to coordinates the rule is given.
Given one-hot channels of its coordinates modulo the octaves' block sides, \coinSquareSides, the rule painted one raw
canvas for every seed on \glFixedCountWords\ of the \glTriedCountWords\ examples it was trained on, with the term at
its standard weight, and under the phase channels every seed gives the same canvas on Fabric
(Section~\ref{sec:limits}).

\subsection{Seams and the cost of mending}
\label{app:seam}

\begin{table}[!htbp]\centering\footnotesize
\caption{Chunked against whole-canvas mending within radius 24 with blocks of 48 and cluster completion, on
\seamCCCanvases\ canvases (three coin seeds for each of the \nCCM\ examples the octave rule makes valid at that radius): the
canvases on which every interior cell agrees, at each margin. Margins of 64 and 96 tile the interior of a $256^2$
canvas; the others mend one central chunk of a $512^2$ canvas, \seamChunkFiveTwelve\ cells wide at margins of 128 to 192 and \seamChunkTwoTwentyFour\ cells wide at 224, which
leaves \seamGapFiveTwelve\ cells of the canvas beyond the window on each side at a margin of 192 and
\seamGapTwoTwentyFour\ at 224, the only context such a test sees. Canvases that
agree at 128 were not rerun at wider margins and are counted as agreeing there; the row of canvases rerun says how many
were.}
\label{tab:seamcc}
\begin{tabular}{@{}lcccccc@{}}\toprule
margin & 64 & 96 & 128 & 160 & 192 & 224\\\midrule
chunks & \multicolumn{2}{c}{tiling $256^2$} & \multicolumn{4}{c}{one central chunk of $512^2$}\\
canvases rerun & \seamCCRunSixtyFour & \seamCCRunNinetySix & \seamCCRunOneTwentyEight & \seamCCRunOneSixty
  & \seamCCRunOneNinetyTwo & \seamCCRunTwoTwentyFour\\
canvases that agree & \seamCCAgreeSixtyFour & \seamCCAgreeNinetySix & \seamCCAgreeOneTwentyEight & \seamCCAgreeOneSixty$^b$
  & \seamCCAgreeOneNinetyTwo$^b$ & \seamCCAgreeTwoTwentyFour$^b$\\\bottomrule
\multicolumn{7}{@{}l}{$^b$ counts the \seamCCAgreeOneTwentyEight\ canvases that agree at 128 and were not rerun.}
\end{tabular}\end{table}

Seams are exact before mending, and after mending at the margins we tested. The generator's own seam test, chunked
against one-shot generation, is part of every evaluation: it holds cell for cell for the octave rule on all
\msSeamsExact\ public examples, raw and after repair, and for the independent-coin rule on every public example. For
the mended output we take each of the octave rule's repaired canvases, mend it once as a whole, and mend its interior
again in chunks, each chunk mended alone from a window that adds a margin on every side and that keeps its absolute
coordinates, so that the lattice, the classes, the passes and every seed are those of the whole canvas. The mender is
seeded by the seed, the absolute block coordinates and the cluster's index alone, so the test does not depend on which
coins the rule reads.

With cluster completion within radius 24 and blocks of 48 (Table~\ref{tab:seamcc}), chunked and whole-canvas mending
agree on every interior cell at a margin of 64 cells on only \seamCCAgreeSixtyFour\ of the \seamCCCanvases\ canvases,
at 128 on \seamCCAgreeOneTwentyEight, at 192 on \seamCCAgreeOneNinetyTwo\ and at 224 on \seamCCAgreeTwoTwentyFour. At
192 the canvases that still differ are \seamCCDisagreeNames, and at 224 one canvas of \seamCCWorstExample\ differs in
\seamCCWorstCells\ cells. At a margin of \seamAllMargin\ cells every canvas agrees. Each of these tests compares the
chunk with one whole canvas of finite size, so it sees only the context that this canvas adds beyond the window: at a
margin of \seamAllMargin, with a central chunk of \seamAllChunkWide\ cells on canvases \seamAllCanvasWide\ wide,
\seamAllGap\ cells. We therefore also mended \sxCanvasesWords\ whole canvases \sxCanvasWide\ cells wide, the octave rule
on \sxOctaveNames\ and the method on Wall at three coin seeds each, whose edge lies \sxEdgeGap\ cells from a central
chunk of \sxChunkWide\ cells, past the counted reach, so that on the chunk they have the values of the unbounded world.
On all \sxCanvasesWords, a chunk mended alone agrees with the whole canvas on every cell at margins of \sxMarginList.
The margin of \seamAllMargin\ remains a measurement on the canvases we tested, not a guarantee, and we do not call
chunked mending at this radius exact below the reach counted in Appendix~\ref{app:mender}, \mendBoundCC\ cells. The
reach grows with the block, since each of the eight class steps
reads and writes a box that grows with it. Without cluster completion, within radius 8 and blocks of 16, on
\seamCanvases\ canvases (three coin seeds for each of the octave rule's examples valid at radius 8 and for Wall,
Skyline and SimpleWall), no interior cell differs at margins of 64 and 96 cells, nor at a margin of 128 cells on a
chunked $512^2$ canvas, while at 32 cells the results differ on \seamDiagCount\ of the \seamExamples\ examples.

The method's own rules, which read the row types or the phase channels, keep the same seams. On
\seamMethodAllKernelsWords\ of the \certInputCountWords\ certified examples whose input is not the octaves, all but
SmileCity, whose rule this test did not include, chunks agree with whole-canvas mending at a margin of
\seamMethodMargin\ cells on all \seamMethodAllOf\ canvases. The rules on the phase channels of Fabric, RedMaze and
SimpleMaze leave nothing to mend; on the other \seamMethodAllMendedWords, mending changes up to \seamMethodAllChangedHiPct\ of the $64^2$ cells
of a chunk, on \seamMethodAllChangedHiName, and on Wall \seamMethodWallChangedLo\ to \seamMethodWallChangedHi, the same
cells whether the chunk is mended alone or with the whole canvas.

Mending is the larger part of the pipeline's cost. On one CPU thread the octave rule and its four repair rounds take a
median of \genCPUMedian\ seconds per $256^2$ canvas, and mending it whole within radius 24 a median of
\mendCPUTwentyFour\ seconds, \mendCPURatio\ times as long as at radius 8 and \mendShareTwentyFour\ of the pipeline's
time. Per cell the random-access block solver costs a median of \raCostRatioRuleTwentyFour\ times as much, and its time
already counts the blocks it depends on, so with the area that a $256^2$ chunk must generate and mend at the measured
margin (Section~\ref{sec:access}) the ratio of their costs per chunk has a median of \raOverRuleChunkAtMargin. At that
margin a chunk of $64^2$ cells carries \chunkAreaSmallAll\ times its own area.

\subsection{Wider mending}
\label{app:mwide}

\begin{table}[!htbp]\centering\scriptsize
\caption{The \mwExamplesWords\ public examples the method leaves invalid, mended again within radius \mwRadiusA\ (blocks of \mwBlockA) and
\mwRadiusB\ (blocks of \mwBlockB), from the same repaired canvases: the share of forbidden windows on the worst of the three
canvases that validity reads, the median total variation of the $3\times3$ window statistics from the example's, and the
median share of cells that mending at \mwRadiusB\ rewrites. An asterisk marks an example valid on all \mwSeedsWords\
canvases, a dagger a certified example given its input; the run on \mwKilledNames\ at \mwRadiusB\ reached the time limit
set in advance.}
\label{tab:mwide}
\input{tables/mwide_v16}\end{table}

Are the method's failures a matter of the mender's reach? For the \mwExamplesWords\ public examples the method leaves
invalid, we mended its repaired canvases again on \mwSeedsWords\ coin seeds, with the same mender within radius \mwRadiusA\ in blocks of
\mwBlockA\ and within radius \mwRadiusB\ in blocks of \mwBlockB, and wrote down beforehand what we expected
(Table~\ref{tab:record}). Each run first mended one canvas again at radius \mendRadiusMain\ and checked that it equals the
stored canvas cell for cell. The wider blocks lengthen the reach that random access must cover, from \mendBoundCC\ cells to
\mwBoundA\ at radius \mwRadiusA\ and \mwBoundB\ at \mwRadiusB.

Within radius \mwRadiusA\ the method becomes valid on \mwValidANames\ (Table~\ref{tab:mwide}), on all \mwSeedsWords\ canvases, and
does not copy: replay agreement and the share of the example's patches move by at most \mwFiveMax. On Skew1 the window
statistics move away from the example's, from a total variation of \mwSkewTvR\ to \mwSkewTvA. The other \mwInvalidAWords\
stay invalid. The rates of Lake, Link2 and Mountains do not fall as a group (their median ratio to radius
\mendRadiusMain's is \mwFarRatioMedian), SmileCity's falls to \mwSmileRatioA\ of its value, and on 3Bricks, whose rows do
not glue, \mwBricksOverA\ of the \mwSeedsWords\ canvases keep more than one forbidden window in a thousand, against
\mwBricksOverR\ at radius \mendRadiusMain.

Within radius \mwRadiusB\ the mender's first block covers nearly the whole canvas, so where its solve succeeds the mender
is close to a global solver. On \mwValidBWords\ examples one or a few block calls leave no forbidden window on any canvas
after rewriting \mwCellsBLo\ to \mwCellsBHi\ of the cells; \mwValidBNewNames\ become valid this way, with the window
statistics about \mwTvRatioBNew\ times as far from the example's as at radius \mendRadiusMain. On SmileCity this repairs
the lattice of faces that no input of the procedure fixes (Appendix~\ref{app:smile}). On \mwInvalidBNames\ the first
solve fails, the mender goes on block by block, and some canvases stay invalid.

\subsection{What stays open}
\label{sec:open}

\begin{table}[!htbp]\centering\scriptsize\setlength{\tabcolsep}{3pt}
\caption{SmileCity under inputs that carry the position of its faces, each given with the octaves and trained as the
method's rules are: the range over the three canvases of the share of forbidden windows after mending within radius
\mendRadiusMain, at training seeds 0 and 1, one value where its ends agree; a dash marks a rule we did not train. The
first row is the input the procedure prescribes; the others are not prescribed. None is valid. Figure~\ref{fig:corpus}
gives the octave rule's.}
\label{tab:smile}
\begin{tabular}{@{}p{3.6cm}cc@{}}\toprule
input & seed 0 & seed 1\\\midrule
phase channels of period \phaseDSmileCity\ along columns & \smileMethodResLo\ to \smileMethodResHi & \mpSmileSpanOne\\
coordinates modulo \coinSquareSides\ along both axes & \glSmileRate & --\\
phase channels of period eight along rows and \phaseDSmileCity\ along columns & \smileMPEightResLo\ to
  \smileMPEightResHi & \smileMPEightSeedOneResLo\ to \smileMPEightSeedOneResHi\\\bottomrule
\end{tabular}\end{table}

The certified examples that the method leaves invalid, \certMethodInvalidNames\ (Section~\ref{sec:valid}), are
those whose invariant or phase needs a patch larger than the window (Table~\ref{tab:cert}). The canvas of 3Bricks that
it misses keeps \threeBricksWorst\ of its windows forbidden. On SmileCity, inputs that carry the position of its faces
do not make it valid either: phase channels of period eight along the rows, which the procedure does not prescribe, leave more windows forbidden than the
prescribed input, and so do the coordinates modulo \coinSquareSides\ (Table~\ref{tab:smile}).

The octave rule leaves \uncertInvalidMCountWords\ uncertified examples invalid, \uncertInvalidMNames. The Metropolis
rule with the same mending makes \mhCCOnlyCountWords\ of them valid, \mhCCOnlyNames, so a random-access generator
exists for each of these, and the octave rule failed to learn one at each of its three training seeds. On
\uncertOpenNames\ neither is valid at training seed 0, and trained from seed 1 the octave rule is valid on
\uncertOpenSeedOneNames.

%% file: tables/radii_v14.tex
\setlength{\tabcolsep}{2pt}%
\begin{tabular}{@{}rrrrrrrrrr@{}}\toprule
 & \multicolumn{5}{c}{examples valid (of 52)} & \multicolumn{2}{c}{8 canvases} & & \\
\cmidrule(lr){2-6}\cmidrule(lr){7-8}
radius & \begin{tabular}[b]{@{}r@{}}octave\\rule\end{tabular} & method & \begin{tabular}[b]{@{}r@{}}indep.\\coins\end{tabular} & tiles & i.i.d. & \begin{tabular}[b]{@{}r@{}}octave\\rule\end{tabular} & method & \begin{tabular}[b]{@{}r@{}}cells\\changed\end{tabular} & \begin{tabular}[b]{@{}r@{}}chunked\\margin\end{tabular}\\\midrule
8 & 27 & -- & 21 & -- & -- & 26$^a$ & -- & 0.071$^a$ & 64$^a$\\
16 & 35 & -- & 27 & -- & 0/25 & -- & -- & 0.14 & --\\
24 & 40 & 45 & 28 & 23 & 0/52 & 36 & 41 & 0.20 & 256\\
\bottomrule\end{tabular}

%% file: tables/fresh_v16.tex
\setlength{\tabcolsep}{3pt}\begin{tabular}{@{}llllcccccc@{}}\toprule
 & & & \multicolumn{2}{c}{independent-coin rule} & \multicolumn{5}{c}{valid}\\
\cmidrule(lr){4-5}\cmidrule(lr){6-10}
example & certificate & input & class & valid & \begin{tabular}[b]{@{}c@{}}octave\\rule\end{tabular} & method & tiles & \begin{tabular}[b]{@{}c@{}}i.i.d.\\colors\end{tabular} & \begin{tabular}[b]{@{}c@{}}min-coin\\copy\end{tabular}\\\midrule
\multicolumn{10}{@{}l}{\emph{GeEom/greedytile}}\\
GreedyA & none & octaves & resistant & -- & -- & -- & -- & -- & --\\
GreedyB & none & octaves & mendable ($\rho\le8$) & \checkmark & -- & -- & -- & -- & --\\
GreedyC & none & octaves & resistant & -- & -- & -- & -- & -- & --\\
GreedyD & none & octaves & mendable ($\rho\le4$) & \checkmark & \checkmark & \checkmark & \checkmark & -- & \checkmark\\
GreedyE & none & octaves & resistant & -- & \checkmark & \checkmark & -- & -- & --\\
GreedyF & none & octaves & mendable ($\rho\le4$) & \checkmark & \checkmark & \checkmark & \checkmark & -- & \checkmark\\
GreedyG & none & octaves & partial & -- & \checkmark & \checkmark & \checkmark & -- & \checkmark\\
\addlinespace[1pt]\multicolumn{10}{@{}l}{\emph{gridbugs/wfc}}\\
GbDiamond & invariant (diagonal) & octaves & resistant & -- & -- & -- & -- & -- & --\\
GbMountain & invariant (rows) & row types & resistant & -- & \checkmark & \checkmark & -- & -- & --\\
GbTransparent & none & octaves & partial & \checkmark & \checkmark & \checkmark & \checkmark & -- & \checkmark\\
GbWater & none & octaves & partial & \checkmark & \checkmark & \checkmark & \checkmark & -- & \checkmark\\
GbLattice & phase ($d=4$, vertical) & phase channels & resistant & -- & -- & \checkmark & -- & -- & --\\
GbTriangles & phase ($d=4$, horizontal) & phase channels & resistant & -- & -- & \checkmark & -- & -- & --\\
GbZigzag & phase ($d=4$, horizontal) & phase channels & resistant & -- & -- & \checkmark & -- & -- & --\\
\addlinespace[1pt]\multicolumn{10}{@{}l}{\emph{krychu/wfc}}\\
KrCave & none & octaves & resistant & -- & -- & -- & -- & -- & --\\
KrCurl & none & octaves & mendable ($\rho\le4$) & \checkmark & \checkmark & \checkmark & \checkmark & -- & \checkmark\\
KrSquare & none & octaves & partial & \checkmark & \checkmark & \checkmark & \checkmark & -- & \checkmark\\
KrTwoLines & none & octaves & mendable ($\rho\le4$) & \checkmark & \checkmark & \checkmark & \checkmark & -- & \checkmark\\
\addlinespace[1pt]\multicolumn{10}{@{}l}{\emph{BorisTheBrave/DeBroglie}}\\
DbPathway & none & octaves & mendable ($\rho\le8$) & \checkmark & -- & -- & -- & -- & --\\
\addlinespace[1pt]\multicolumn{10}{@{}l}{\emph{sjcasey21/wavefunctioncollapse-clj}}\\
ScCityTemplate & none & octaves & partial & \checkmark & \checkmark & \checkmark & \checkmark & -- & --\\
\addlinespace[1pt]\multicolumn{10}{@{}l}{\emph{nascentlogic/WaveFunctionCollapse}}\\
NlBlob & none & octaves & mendable ($\rho\le4$) & \checkmark & \checkmark & \checkmark & \checkmark & -- & \checkmark\\
NlCaves & none & octaves & mendable ($\rho\le4$) & \checkmark & \checkmark & \checkmark & \checkmark & -- & \checkmark\\
NlRocks & none & octaves & mendable ($\rho\le8$) & \checkmark & \checkmark & \checkmark & \checkmark & -- & \checkmark\\
NlRocks2 & none & octaves & mendable ($\rho\le4$) & \checkmark & \checkmark & \checkmark & \checkmark & -- & \checkmark\\
NlRuins & none & octaves & mendable ($\rho\le4$) & \checkmark & \checkmark & \checkmark & \checkmark & -- & --\\
\addlinespace[1pt]\multicolumn{10}{@{}l}{\emph{mxgmn/MarkovJunior}}\\
MjRoom & none & octaves & mendable ($\rho\le8$) & \checkmark & \checkmark & \checkmark & \checkmark & -- & \checkmark\\
\addlinespace[1pt]\multicolumn{10}{@{}l}{\emph{finnbear/wfc}}\\
SwPattern & none & octaves & mendable ($\rho\le8$) & \checkmark & \checkmark & \checkmark & \checkmark & -- & \checkmark\\
\bottomrule
\multicolumn{10}{@{}l@{}}{\checkmark\ valid: after mending within radius \mendRadiusMain, at most one window in a thousand forbidden on each canvas; -- invalid}\\
\end{tabular}

%% file: tables/baselines.tex
\begin{tabular}{lrrrrrrrrr}\toprule
& \multicolumn{7}{c}{forbidden-window rate} & \multicolumn{2}{c}{pattern KL}\\\cmidrule(lr){2-8}\cmidrule(lr){9-10}
example & \multicolumn{2}{c}{the learned rule} & \multicolumn{2}{c}{independent sampling} & \multicolumn{3}{c}{Metropolis} & the rule & Metropolis\\
& & & \multicolumn{2}{c}{$+$ repair} & & & & &\\
& raw & mended & 32 rounds & 64 rounds & 32 sweeps & 64 sweeps & 64, mended & raw & 64 sweeps\\\midrule
Cave & 0.0015 & 0.00000 & 0.0552 & 0.0478 & 0.0367 & 0.0330 & 0.00000 & 0.028 & 0.073\\
Paths & 0.0076 & 0.00003 & 0.1204 & 0.1075 & 0.0194 & 0.0136 & 0.00102 & 0.031 & 1.037\\
Sand & 0.0077 & 0.00000 & 0.0038 & 0.0033 & 0.0008 & 0.0006 & 0.00000 & 0.056 & 3.235\\
Maze & 0.0136 & 0.00002 & 0.0345 & 0.0296 & 0.0477 & 0.0431 & 0.00000 & 0.084 & 0.293\\
Knot & 0.0173 & 0.00068 & 0.1251 & 0.1142 & 0.0464 & 0.0373 & 0.00214 & 0.037 & 0.411\\
Rooms & 0.0162 & 0.00008 & 0.1276 & 0.1154 & 0.1546 & 0.1491 & 0.09038 & 0.081 & 0.122\\
Angular & 0.0086 & 0.00143 & 0.2446 & 0.1791 & 0.2500 & 0.2216 & 0.19583 & 0.462 & 0.108\\
Skew2 & 0.0343 & 0.02570 & 0.1354 & 0.1173 & 0.0821 & 0.0720 & 0.05399 & 0.021 & 0.459\\
Qud & 0.0570 & 0.02534 & 0.2734 & 0.2005 & 0.2856 & 0.2559 & 0.24231 & 0.152 & 0.069\\
City & 0.0403 & 0.02141 & 0.3382 & 0.2540 & 0.3361 & 0.3075 & 0.30495 & 0.007 & 0.078\\
Sewers & 0.0545 & 0.02551 & 0.2587 & 0.2148 & 0.2980 & 0.2837 & 0.28139 & 0.191 & 0.137\\
Wall & 0.1098 & 0.08959 & 0.9202 & 0.5580 & 0.7151 & 0.3588 & 0.35743 & 0.042 & 0.019\\
\midrule\multicolumn{10}{l}{\itshape controls}\\
Chess & 0.0575 & 0.05768 & 0.2476 & 0.1934 & 0.1941 & 0.1261 & 0.12261 & 0.000 & 0.000\\
Stripes & 0.0615 & 0.06126 & 0.2417 & 0.1889 & 0.1897 & 0.1269 & 0.12346 & 0.000 & 0.000\\
Col3 & 0.0087 & 0.00587 & 0.0645 & 0.0501 & 0.0807 & 0.0721 & 0.01951 & 0.010 & 0.865\\
\bottomrule\end{tabular}

%% file: tables/copier_v14.tex
\begin{tabular}{@{}lcrrrrr@{}}\toprule
filler & valid & \begin{tabular}[b]{@{}r@{}}forbidden\\before\\mending\end{tabular} & \begin{tabular}[b]{@{}r@{}}cells\\changed\\(\%)\end{tabular} & \begin{tabular}[b]{@{}r@{}}row\\copy\end{tabular} & repetition & \begin{tabular}[b]{@{}r@{}}window\\TV\end{tabular}\\\midrule
\multicolumn{7}{@{}l}{\emph{Wall}}\\
copier & yes & 0 & 0 & 1.00 & 0--0.008 & 0.036--0.044\\
re-solved copier & yes & 0 & 0 & 0.22 & 0 & 0.13--0.14\\
rule & yes & 0.0085 & 3--6 & 0.22 & 0 & 0.35--0.38\\
Metropolis & yes & 0.0078 & 1--2 & 0.22 & 0 & 0.49--0.51\\
\addlinespace
\multicolumn{7}{@{}l}{\emph{SimpleWall}}\\
copier & yes & 0 & 0 & 1.00 & 0 & 0.008--0.011\\
re-solved copier & yes & 0 & 0 & 0.17--0.18 & 0 & 0.020--0.022\\
rule & yes & 0.0080 & 3--4 & 0.17--0.18 & 0 & 0.21\\
Metropolis & yes & 0.029 & 3 & 0.17--0.18 & 0 & 0.28--0.31\\
\addlinespace
\multicolumn{7}{@{}l}{\emph{Platformer}}\\
copier & yes & 0 & 0 & 1.00 & 0.64--0.68 & 0.021--0.028\\
re-solved copier & yes & 0 & 0 & 0.16--0.18 & 0--$2\cdot10^{-5}$ & 0.28\\
rule & yes & 0.0001 & 0 & 0.12--0.13 & 0 & 0.087--0.138\\
Metropolis & no & 0.13 & 3--12 & 0.07 & 0 & 0.30--0.34\\
\addlinespace
\multicolumn{7}{@{}l}{\emph{Skyline2}}\\
copier & yes & 0 & 0 & 1.00 & 0.69--0.70 & 0.026--0.034\\
re-solved copier & yes & 0 & 0 & 0.09 & 0 & 0.13--0.14\\
rule & yes & 0.016 & 31--39 & 0.09 & 0 & 0.30--0.32\\
Metropolis & no & 0.22 & 0--1 & 0.06--0.07 & 0 & 0.33\\
\addlinespace
\multicolumn{7}{@{}l}{\emph{Skyline}}\\
copier & yes & 0 & 0 & 1.00 & 0.55--0.58 & 0.039--0.046\\
re-solved copier & yes & 0 & 0 & 0.14 & 0 & 0.12--0.14\\
rule & yes & 0.015 & 26--44 & 0.14 & 0--$2\cdot10^{-5}$ & 0.32--0.35\\
Metropolis & no & 0.23 & 0--1 & 0.11 & 0 & 0.36--0.37\\
\addlinespace
\multicolumn{7}{@{}l}{\emph{Flowers}}\\
copier & yes & 0 & 0 & 1.00 & 0.41--0.46 & 0.098--0.104\\
re-solved copier & yes & 0 & 0 & 0.12--0.14 & 0 & 0.14\\
rule & yes & 0.022 & 17--19 & 0.11--0.12 & 0 & 0.21--0.22\\
Metropolis & yes & 0.048 & 17--18 & 0.11--0.12 & 0 & 0.26--0.29\\
\addlinespace
\multicolumn{7}{@{}l}{\emph{MoreFlowers}}\\
copier & yes & 0 & 0 & 1.00 & 0.41 & 0.098--0.103\\
re-solved copier & yes & 0 & 0 & 0.12--0.14 & 0 & 0.12--0.13\\
rule & yes & 0.018 & 19--20 & 0.11--0.12 & 0 & 0.18--0.19\\
Metropolis & yes & 0.050 & 17--19 & 0.11--0.12 & 0 & 0.26--0.30\\
\addlinespace
\multicolumn{7}{@{}l}{\emph{3Bricks}}\\
no copier & & & & & & \\
rule & no & 0.023 & 29--33 & 0.07--0.09 & 0 & 0.34--0.37\\
Metropolis & no & 0.092 & 7--11 & 0--0.004 & 0 & 0.49--0.50\\
\bottomrule\end{tabular}

%% file: tables/wallbricks_v15.tex
\begin{tabular}{@{}lrrrrrrrr@{}}\toprule
generator & \begin{tabular}[b]{@{}r@{}}forbidden\\windows\end{tabular} & \begin{tabular}[b]{@{}r@{}}median\\brick\end{tabular} & \begin{tabular}[b]{@{}r@{}}90th\\percentile\end{tabular} & \begin{tabular}[b]{@{}r@{}}joints\\(\%)\end{tabular} & \begin{tabular}[b]{@{}r@{}}windows with\\a joint (\%)\end{tabular} & \begin{tabular}[b]{@{}r@{}}window\\TV\end{tabular} & \begin{tabular}[b]{@{}r@{}}TV forced by\\joints (\%)\end{tabular} & \begin{tabular}[b]{@{}r@{}}$3\times16$\\repetition\end{tabular}\\\midrule
example & 0 & 8 & 8 & 11 & 41 & 0 & & 1.00\\
shifted replay & 0 & 8 & 8 & 11 & 41 & 0.005--0.006 & 24--37 & 1.00\\
block reference & 0 & 8 & 19--20 & 9 & 31--32 & 0.13 & 69--70 & 0.03--0.04\\
copier & 0 & 8 & 8 & 11 & 41 & 0.04 & 5 & 0.87--0.90\\
re-solved copier & 0 & 8 & 18--19 & 9 & 31--32 & 0.13--0.14 & 70 & 0.04--0.05\\
method, before mending & 0.008--0.01 & 12--13 & 35--37 & 5 & 16--18 & 0.32--0.35 & 71 & 0.01--0.02\\
method & 0 & 14--16 & 41--47 & 4 & 14--16 & 0.35--0.38 & 69--70 & 0.008--0.01\\
Metropolis filler & 0 & 26--33 & 92--102 & 2 & 6--7 & 0.49--0.51 & 69 & 0.0007--0.003\\
octave rule & 0.003--0.05 & 8--10 & 21--26 & 7--9 & 25--27 & 0.19--0.23 & 60--72 & 0.03\\
independent-coin rule & 0.09--0.10 & 6 & 12--13 & 9 & 29--30 & 0.12--0.14 & 81--87 & 0.01--0.02\\
\bottomrule\end{tabular}

%% file: tables/clean_v16.tex
\begin{tabular}{@{}lrrrrrrr@{}}\toprule
& & \multicolumn{2}{c}{patches before mending} & \multicolumn{2}{c}{multicolor rows} & \multicolumn{2}{c}{patches after mending}\\
\cmidrule(lr){3-4}\cmidrule(lr){5-6}\cmidrule(l){7-8}
example & canvases & \begin{tabular}[b]{@{}r@{}}clean\\(\%)\end{tabular} & \begin{tabular}[b]{@{}r@{}}in the\\example\end{tabular} & clean & \begin{tabular}[b]{@{}r@{}}rows of the\\example\end{tabular} & \begin{tabular}[b]{@{}r@{}}untouched\\(\%)\end{tabular} & \begin{tabular}[b]{@{}r@{}}in the\\example\end{tabular}\\\midrule
Wall & 11 & 46--54 & 0 & 1\,038 & 0 & 24--37 & 0\\
SimpleWall & 8 & 46--55 & 0 & 980 & 0 & 23--32 & 0\\
Platformer & 8 & 97--100 & 0 & 1\,674 & 0 & 96--100 & 0\\
Skyline2 & 8 & 33--55 & 0 & 161 & 0 & 5--29 & 0\\
Skyline & 8 & 21--39 & 0 & 259 & 0 & 1--8 & 0\\
Flowers & 8 & 7--10 & 0 & 4 & 0 & 0--1 & 0\\
MoreFlowers & 8 & 9--13 & 0 & 49 & 0 & 0--1 & 0\\
3Bricks & 8 & 5--14 & 0 & 13 & 0 & 0--17 & 0\\
\bottomrule\end{tabular}

%% file: tables/wallvar_v16.tex
\begin{tabular}{@{}lrrr@{}}\toprule
generator & \begin{tabular}[b]{@{}r@{}}aligned\\variation\end{tabular} & \begin{tabular}[b]{@{}r@{}}normalized\\agreement\end{tabular} & canvases\\\midrule
the method & 0.41 & 0.09 & 3\\
the re-solved copier & 0.48 & 0.09 & 3\\
the copier & -- & 0.13 & 3\\
the random-access block solver & 0.23 & 0.67 & 8\\
the block reference & 0.24 & 0.64 & 8\\
shifted replay & 0.00 & 1.00 & 8\\
\bottomrule\end{tabular}

%% file: tables/variation_v13.tex
\begin{tabular}{@{}lrrrrrrrrrr@{}}\toprule
 & \multicolumn{2}{c}{total variation} & \multicolumn{2}{c}{normalized replay} & repetition & \multicolumn{2}{c}{aligned variation} & \multicolumn{2}{c}{local redraw} & block cost\\
\cmidrule(lr){2-3}\cmidrule(lr){4-5}\cmidrule(lr){7-8}\cmidrule(lr){9-10}
example & rule & block & rule & block & rule & rule & block & fine & all & over rule\\\midrule
Angular & 0.143 & 0.218 & 0.057 & 0.132 & 0.000 & 0.512 & 0.537 & 0.06\% & 1.15\% & 10\\
Cave & 0.062 & 0.229 & 0.020 & 0.047 & 0.000 & 0.470 & 0.478 & 0.22\% & 1.02\% & 13\\
City & 0.065 & 0.058 & 0.094 & 0.330 & 0.000 & 0.370 & 0.539 & 0.05\% & 1.01\% & 24\\
Dungeon & 0.093 & 0.187 & 0.052 & -0.017 & 0.000 & 0.332 & 0.414 & 0.85\% & 1.53\% & 12\\
Flowers & 0.205 & 0.142 & 0.146 & 0.312 & 0.000 & 0.353 & 0.313 & 1.27\% & 1.52\% & 26\\
Font & 0.054 & 0.455 & 0.026 & -0.098 & 0.000 & 0.361 & 0.469 & 0.55\% & 0.69\% & 16\\
Forest & 0.149 & 0.155 & 0.073 & 0.133 & 0.000 & 0.284 & 0.326 & 0.69\% & 0.69\% & 12\\
Hogs & 0.130 & 0.050 & 0.064 & 0.008 & 0.000 & 0.277 & 0.317 & 0.76\% & 0.76\% & 8\\
Knot & 0.192 & 0.221 & 0.067 & 0.041 & 0.000 & 0.361 & 0.431 & 0.47\% & 1.15\% & 18\\
LessRooms & 0.220 & 0.195 & 0.051 & 0.083 & 0.000 & 0.419 & 0.444 & 1.69\% & 2.18\% & 6\\
Lines & 0.220 & 0.041 & 0.107 & -0.002 & 0.000 & 0.285 & 0.366 & 0.75\% & 0.86\% & 27\\
MagicOffice & 0.205 & 0.226 & 0.117 & 0.030 & 0.000 & 0.273 & 0.353 & 0.92\% & 1.11\% & 17\\
Maze & 0.196 & 0.205 & 0.006 & 0.102 & 0.000 & 0.482 & 0.474 & 1.26\% & 1.25\% & 19\\
Mazelike & 0.175 & 0.134 & 0.064 & 0.303 & 0.000 & 0.430 & 0.391 & 1.05\% & 1.99\% & 14\\
MoreFlowers & 0.209 & 0.121 & 0.132 & 0.319 & 0.000 & 0.358 & 0.321 & 1.35\% & 1.60\% & 24\\
NotKnot & 0.273 & 0.407 & 0.124 & 0.146 & 0.000 & 0.290 & 0.256 & 0.45\% & 0.83\% & 13\\
Office & 0.109 & 0.173 & 0.074 & 0.196 & 0.000 & 0.382 & 0.376 & 0.41\% & 0.97\% & 18\\
Office2 & 0.178 & 0.227 & 0.073 & 0.301 & 0.000 & 0.410 & 0.360 & 0.38\% & 1.21\% & 29\\
Paths & 0.099 & 0.249 & 0.038 & -0.104 & 0.000 & 0.325 & 0.413 & 0.85\% & 0.90\% & 11\\
Platformer & 0.198 & 0.338 & 0.118 & 0.352 & 0.000 & 0.483 & 0.365 & 0.06\% & 1.04\% & 43\\
Rooms & 0.122 & 0.299 & 0.063 & 0.057 & 0.000 & 0.431 & 0.452 & 0.31\% & 1.13\% & 11\\
Rule126 & 0.208 & 0.335 & 0.014 & 0.086 & 0.000 & 0.474 & 0.481 & 2.01\% & 2.14\% & 20\\
Sand & 0.129 & 0.166 & 0.058 & 0.063 & 0.000 & 0.294 & 0.336 & 0.78\% & 0.79\% & 8\\
Sewers & 0.297 & 0.290 & 0.100 & 0.119 & 0.000 & 0.497 & 0.554 & 0.24\% & 2.10\% & 26\\
Spirals & 0.289 & 0.202 & 0.059 & 0.055 & 0.000 & 0.407 & 0.451 & 1.26\% & 1.57\% & 16\\
Village & 0.392 & 0.047 & 0.265 & 0.267 & 0.000 & 0.156 & 0.365 & 0.22\% & 0.46\% & 8\\
Wrinkles & 0.210 & 0.186 & 0.075 & 0.082 & 0.000 & 0.285 & 0.275 & 1.08\% & 1.30\% & 12\\
\midrule median & 0.192 & 0.202 & 0.067 & 0.086 & 0.000 & 0.361 & 0.391 & 0.75\% & 1.13\% & 16\\
\bottomrule\end{tabular}

%% file: tables/blocksolver_rule_v13.tex
\begin{tabular}{lrrrrrr}\toprule
& \multicolumn{3}{c}{aligned variation} & \multicolumn{3}{c}{normalized agreement with the tiling}\\
example & rule & one pass & two passes & rule & one pass & two passes\\\midrule
Angular & 0.51 & 0.34 & 0.52 & 0.06 & 0.48 & 0.17\\
Cave & 0.47 & 0.43 & 0.48 & 0.02 & 0.25 & 0.06\\
City & 0.37 & 0.31 & 0.42 & 0.09 & 0.63 & 0.52\\
Dungeon & 0.33 & 0.30 & 0.40 & 0.05 & 0.36 & 0.07\\
Flowers & 0.35 & 0.20 & 0.29 & 0.15 & 0.67 & 0.43\\
Font & 0.36 & 0.47 & 0.47 & 0.03 & -0.09 & -0.16\\
Forest & 0.28 & 0.31 & 0.33 & 0.07 & 0.17 & 0.02\\
Hogs & 0.28 & 0.26 & 0.31 & 0.06 & 0.31 & 0.06\\
Knot & 0.36 & 0.34 & 0.42 & 0.07 & 0.33 & 0.10\\
LessRooms & 0.42 & 0.40 & 0.44 & 0.05 & 0.32 & 0.09\\
Lines & 0.29 & 0.31 & 0.37 & 0.11 & 0.30 & 0.03\\
MagicOffice & 0.27 & 0.25 & 0.34 & 0.12 & 0.44 & 0.11\\
Maze & 0.48 & 0.42 & 0.48 & 0.01 & 0.33 & 0.06\\
Mazelike & 0.43 & 0.24 & 0.31 & 0.06 & 0.66 & 0.50\\
MoreFlowers & 0.36 & 0.20 & 0.30 & 0.13 & 0.65 & 0.43\\
NotKnot & 0.29 & 0.24 & 0.26 & 0.12 & 0.36 & 0.17\\
Office & 0.38 & 0.28 & 0.35 & 0.07 & 0.45 & 0.27\\
Office2 & 0.41 & 0.27 & 0.37 & 0.07 & 0.47 & 0.27\\
Paths & 0.33 & 0.36 & 0.41 & 0.04 & 0.14 & -0.09\\
Platformer & 0.48 & 0.29 & 0.34 & 0.12 & 0.60 & 0.44\\
Rooms & 0.43 & 0.37 & 0.45 & 0.06 & 0.40 & 0.12\\
Rule126 & 0.47 & 0.37 & 0.47 & 0.01 & 0.42 & 0.10\\
Sand & 0.29 & 0.31 & 0.34 & 0.06 & 0.21 & 0.04\\
Sewers & 0.50 & 0.26 & 0.46 & 0.10 & 0.68 & 0.37\\
Spirals & 0.41 & 0.31 & 0.44 & 0.06 & 0.37 & 0.08\\
Village & 0.16 & 0.24 & 0.37 & 0.27 & 0.63 & 0.24\\
Wrinkles & 0.29 & 0.25 & 0.27 & 0.07 & 0.30 & 0.13\\
\bottomrule\end{tabular}

%% file: tables/mwide_v16.tex
\setlength{\tabcolsep}{3pt}\begin{tabular}{@{}lrrrrrrr@{}}\toprule
 & \multicolumn{3}{c}{forbidden windows} & \multicolumn{3}{c}{TV to the example} & rewritten\\
\cmidrule(lr){2-4}\cmidrule(lr){5-7}
example & $\rho=24$ & 48 & 96 & 24 & 48 & 96 & at 96\\\midrule
3Bricks$^{\dagger}$ & 0.0011 & 0.0031 & stopped & 0.35 & 0.35 & -- & --\\
SmileCity$^{\dagger}$ & 0.0102 & 0.0043 & 0$^{*}$ & 0.16 & 0.17 & 0.33 & 50\%\\
Lake & 0.0038 & 0.0101 & 0.0320 & 0.23 & 0.24 & 0.25 & 54\%\\
Link2 & 0.0114 & 0.0106 & 0.0047 & 0.20 & 0.21 & 0.24 & 40\%\\
Mountains & 0.0073 & 0.0032 & 0$^{*}$ & 0.10 & 0.14 & 0.21 & 31\%\\
ScaledMaze & 0.0017 & 0.0007$^{*}$ & 0$^{*}$ & 0.05 & 0.06 & 0.08 & 46\%\\
Skew1 & 0.0020 & 0.0003$^{*}$ & 0$^{*}$ & 0.21 & 0.30 & 0.32 & 37\%\\
\bottomrule\end{tabular}

%% file: app_record.tex
\section{Expectations and outcomes}
\label{app:record}

Which of our expectations did the experiments meet, and which did they miss? We wrote each expectation in a
version-controlled log before the experiment ran, with the threshold it had to meet and what methods without learning
score. Table~\ref{tab:record} gives, for each study in the order of the paper, what we expected and what happened, as it
fell. The log also holds the expectations for runs that the paper does not report.

Some analyses ran without an expectation written in advance and have no row, among them the mending maps of the whole
corpus (Tables~\ref{tab:mendmapeight} and~\ref{tab:mendall}), the correlations beyond the window
(Figure~\ref{fig:corr}), the check of inequality~\eqref{eq:vfr} on measured outputs (Section~\ref{supp:floorcheck}), the full labeling, the linear program with defects and the comparison of each rule's $\delta$
with $\kappa_f$ (Table~\ref{tab:cert}), the one-way quantity, added to the certificate search after its first run, which
certifies Town, the theorems set against the method's worlds (Table~\ref{tab:thm}), and the count of patches and rows
free of forbidden windows before mending (Table~\ref{tab:clean}). The forced edits (Section~\ref{app:forced}) were planned with the first runs and measured after them. The row-color term, the boundary anchor
and the run limits were set after a run had been seen, as was the decision that a certified example counts only with
the input its certificate prescribes. Cluster completion departs from the criterion we had fixed for adopting it, which
asked that it raise the residual on no canvas at radius \mendRadiusLo\ and make some example valid there: it raised the
residual slightly on a canvas of Wall and made no example valid. We kept it as a candidate because it never lost at
radius \mendRadiusMid, and it became our mender after comparisons with i.i.d.\ colors and with independent coins under
the same mender, written down before they ran. The comparison of the octave rule with the random-access block solver
was set in advance for the solver's first pass only; its reading after the second pass and its counts on the
uncertified examples came after the run (Table~\ref{tab:blocksolverrule}). The rows say what was changed before a run,
give both readings where our recount of an outcome differs from the reading made at the time, and say which
predictions were never scored.

Where an expectation missed, its row names the parts that did. The independent-coin rule fell short of the rates we
expected: with an earlier objective most of its rules of radius \mainR\ collapsed to a constant pattern, and the voxel
examples missed their bars. Mending, the certificate, the witnesses, complete search and training for local
completability each missed bars of their own. Several early generators for the certified examples missed, Fabric and
the mazes were not rigid at the patch size we expected, a floor without learning made Wall valid, and a held-out maze
was not certified. The octave rule missed most of its study on \msPilotExamplesWords\ examples and its divergence
before mending, the octaves leave their grid in its output, and cluster completion changed more cells than we allowed.
The hand-written rules trailed the learned rule on fewer examples than we expected, the random-access block solver kept
a higher aligned variation than the independent-coin rule, and the diffusion model missed its pilot, its controls and
validity without copying, so by the measure set in advance the input is not sufficient for that model at our training
budget. The scan of a solver's worlds found no persistent dependence on the resistant examples without a certificate,
where we expected one on most of them, and the slope of the min-coin copy agreed with the scan on fewer of them than we
expected.

{\setlength{\LTcapwidth}{\linewidth}
\renewcommand{\arraystretch}{1.0}
\setlength{\tabcolsep}{4pt}
\setlength{\defaultaddspace}{1pt}
\scriptsize
\begin{longtable}{@{}>{\raggedright\arraybackslash}p{0.12\linewidth}>{\raggedright\arraybackslash}p{0.45\linewidth}>{\raggedright\arraybackslash}p{0.39\linewidth}@{}}
\caption{What we expected from each study, in the order of the paper, and what happened. Counts of the form ``$a$ of
$b$'' are examples unless the row says otherwise; valid means valid on every canvas of the protocol the row names.}
\label{tab:record}\\
\toprule
Study & What we expected & What happened\\
\midrule
\endfirsthead
\multicolumn{3}{@{}l}{Table~\ref{tab:record}, continued}\\
\toprule
Study & What we expected & What happened\\
\midrule
\endhead
\bottomrule
\endfoot
\bottomrule
\endlastfoot
\multicolumn{3}{@{}l}{\itshape Section~\ref{sec:generators} and Appendix~\ref{app:further}: the independent-coin rule}\\
\addlinespace
An earlier run with a soft objective (Section~\ref{app:persample})
 & At radius \mainR, a rate of at most \recMilli\ raw or after repair on \softShareBar\ of the examples, falling with
 the radius; exact seams; contained coin edits; at least \softSpeedBar\ times faster per cell than our Wave Function
 Collapse implementation, with divergence within \recTwice\ its own; no decay on rigid examples; a finite radius for
 forced edits
 & Missed: \softCollapsed\ of the \softLegs\ rules of radius \mainR\ collapsed to a constant valid pattern; at smaller
 radii the median rate after repair was \softRepairEight\ and \softRepairSixteen. Held: seams, coin edits. Speed held
 as read at the time, over each method's range of times (at least \softSpeedRead\ times), and missed on our recount
 (a median of \softSpeedMedian\ times, under \softSpeedBar\ on \softSpeedSlowNames). The rest was not scored, the run being
 unusable\\
\addlinespace
The test-time curves, earlier and final configurations (Section~\ref{supp:persample})
 & At \curveRounds\ rounds, a rate of at most \recCenti\ raw or \recMilli\ after repair on \softShareBar\ of the
 examples, falling monotonically with the rounds; divergence within \recTwice\ Wave Function Collapse's; decay on the
 proper colorings, none exponential on the agreement controls; exact seams; contained coin edits. The final
 configuration at most \recHalf\ the earlier one's median rate without collapse (set before its run in place of
 \curveOldBar\ times), rank correlation at least \curveRankBar
 & Partly held. Missed: the rate, on \nPoneFirst\ and \nPone\ of \nPoneOf, monotone on \nMonotoneFirst\ and
 \nMonotone\ of \nRules\ rules; the final rate, \ratioFinalFirst\ times the earlier, read at the time, before the last
 rule was in, as held. Held: the divergence as read at the time, on \nPtwoFirst\ and \nPtwo\ of \nPtwoOf, though
 missed on the rest read example by example; controls; seams; coin edits; the ranking, \spearmanFinalFirst\\
\addlinespace
The trained radius alone (Figure~\ref{fig:radius})
 & Over trained radii up to \radLadderTop, a slope of log rate against log radius from $\radBandLo$ to $\radBandHi$ on
 Chess, Stripes, Wall and Town; at most $\radMendBar$, or a rate of \recMilli\ at the largest radius, on Cave, Font,
 Hogs and Paths, which mend; from $\radMendBar$ to $\radColBar$ on Col3. Refuting: $\radSteep$ or steeper on the first
 group, $\radColBar$ or shallower on the second
 & Partly held: Chess and Stripes ($\slopeSixtyfourChess$, $\slopeSixtyfourStripes$). Missed: Wall and Town
 ($\slopeSixtyfourWall$, $\slopeSixtyfourTown$); the examples that mend ($\slopeSixtyfourFont$ to
 $\slopeSixtyfourHogs$, none at \recMilli), whose refuting observation occurred on Font and Paths up to radius
 \mainR\ ($\slopeFont$, $\slopePaths$) and on all over the whole ladder; Col3 ($\slopeSixtyfourColthree$). The
 classes could not be told apart\\
\addlinespace
Variants of the objective; the pattern-space rule (Table~\ref{tab:variants})
 & A variant alone brings the earlier configuration's median rate to \recHalf\ without collapse; random training rounds
 make the curves fall monotonically to the largest round count; the changes combined reach \recHalf, with monotone
 curves; the pattern-space rule reaches \recHalf, or the free rule stays
 & Missed, except monotone curves under random rounds (\varRandomMono\ of \varRandomOf): no variant alone reaches
 \recHalf; combined, \variantFinalSixteen\ times, monotone on \varFinalMono; the pattern-space rule
 \variantPatternSixteen\ times, a negative result\\
\addlinespace
Voxel examples (Appendix~\ref{app:further})
 & Exact seams and at most \threeDBar\ of windows forbidden at the trained rounds on a voxel maze and on voxel pipes,
 then at most \threeDBarTwo\ on the maze with a larger budget; no decay of a voxel checkerboard from \threeDRoundsLo\ to
 \threeDRoundsHi\ rounds
 & Missed, except the seams: the maze at \threeDMaze, then \threeDMazeTwo; the pipes at \threeDPipes, under the bar
 only because the rule collapsed onto empty space. The checkerboard fell from \threeDChessLo\ to \threeDChessHi, read
 at the time as holding, since the fall is not exponential\\
\addlinespace
Window statistics against Wave Function Collapse at matched sizes (Section~\ref{app:stats})
 & Divergence at or below our implementation's on at least \statsSmallBar\ examples at the smallest size, within
 \statsSd\ standard deviations on \statsSmallTolBar; on \statsMidBar\ of those it completes at the middle size; on
 \statsBigBar\ at the largest, within \statsSd\ standard deviations on \statsBigTolBar. Refuting: above \recTwice\ its
 divergence on most examples. Backtracking completes \statsBtBar\ of the \nBtOf\ where restarts fail
 & Held, except the tolerant count at the largest size, \nPtwoT\ of the \nPtwoMOf\ completed, though a larger share
 than the bar: at the smallest size \nPtwoMSmall\ of \nPtwoMSmallOf\ and \nPtwoTSmall\ within, at the middle
 \nPtwoMMid\ of \nPtwoMMidOf, at the largest \nPtwoM\ of \nPtwoMOf; no refuting observation; backtracking completed
 \nBtSolved\ of the \nBtOf, and \nBtSolvedAll\ of the \nBtOfAll\ it ran on, which add 3Bricks, Cat and Lake, not run
 under restarts at $256^2$ (Table~\ref{tab:wfcstats})\\
\addlinespace
\multicolumn{3}{@{}l}{\itshape Section~\ref{sec:resist} and Appendices~\ref{app:indep} and~\ref{app:cert}: where
independent coins fail}\\
\addlinespace
The certificate (Table~\ref{tab:cert})
 & Chess and Stripes certified (a check); none that mends within radius \mendRadiusLo; at least \certCtwoBar\ of the
 \certCtwoOf\ resistant examples no constant coloring satisfies, Wall among them; certificates commoner among
 resistant examples by \certCthreeBar\ points; with what the complete solver clears, a cover of \certCfourBar\ of the
 \certResistantOf\ resistant examples; the unsolvable share flat in the radius on certified examples and falling on
 the rest, on \certCfiveBar\ of each
 & Partly held: Chess and Stripes; none of the \certMendedOf\ that mend or the \certPartialOf\ partial ones;
 \certCtwoCount\ of \certCtwoOf, Wall among them; \certResistantRegCount\ of \certResistantOf\ against none,
 \certCthreePoints\ points ($p=$~\certFisherP, a test made after the run). Missed: the cover, at
 \certResistantRegCount, as the complete solver clears none of the mender's failed blocks; the profiles, flat on the
 uncertified resistant examples too\\
\addlinespace
The certificate against the $3\times3$ target (Table~\ref{tab:cert})
 & Under the labeling fixed before the run, the example's window statistics force a spread of the quantity on
 \certTargetBar\ of the \certTargetOf\ certified examples and on each of Chess, Stripes, Wall, SimpleWall, RedMaze and
 SimpleMaze
 & Held: \certTargetForcedWalls\ of \certTargetOf, and all \certTargetForcedSix. That labeling forces
 \certSpecSimpleMaze\ on RedMaze and SimpleMaze and \certSpecFabric\ on Fabric; the full labeling, made after the run
 and used in Table~\ref{tab:cert}, \certFullSimpleMaze\ and \certFullFabric\\
\addlinespace
Witnesses on Wall, SimpleWall, Flowers and Skyline (Appendix~\ref{app:witness})
 & A witness on at least \witBar\ of the unsolvable repairs on \witCountBar\ of these examples; freeing the nearer
 witness patch solves more repairs than freeing as many other cells, by \witFreeBar\ points on \witCountBar
 & Partly held: witnesses on \witCoverageHeld\ of \witCoverageOf\ (Wall \witCoverageWall, SimpleWall
 \witCoverageSimpleWall, Flowers \witCoverageFlowers, Skyline \witCoverageSkyline), none on \witSoundSat\ solvable
 instances. Missed: freeing, on \witBtwoCount\ of them, with \witTargetedSolved\ instances solved against
 \witControlSolved, as most failed repairs carry many witness rows\\
\addlinespace
Witnesses on every certified example (Table~\ref{tab:witness})
 & Checks: earlier witnesses found again, every witness confirmed without labels, witnesses on seams of mixed phase
 and on Town's forbidden order, none on solvable instances or free regions. A witness on at least \recHalf\ of the
 unsolvable instances of each certified example with a row invariant; on the \wxUncertCountWords\ uncertified
 resistant examples, none at patch size \wxPatchSmall\ and at most \wxUncertBar\ at \wxPatchBig
 & Held: every check, with no witness on the \wxSolvAll\ solvable instances or \wxNegRegions\ free regions; all
 \wxRowInvCountWords, from \wxShareRowInvMinNear\ to \wxShareRowInvMax; none at \wxPatchSmall, \wxUncertPfour\ pooled
 at \wxPatchBig. Without an expectation, the phase examples and Town carry witnesses on \wxOtherShareLo\ to
 \wxOtherShareHi\\
\addlinespace
The min-coin copy (Appendix~\ref{app:mincoinrun})
 & Checks of the implementation (windows whose cells copy from several cells within \tcOneBand\ of $q_r$) and of
 Theorem~\ref{thm:floor}; invalid after mending on every certified example and on Chess and Stripes at radii
 \tcRadiusMin\ to \tcLadderMax; valid on at least \tcFourBar\ uncertified examples at radius
 \tcContrastRadius; Wall invalid at radii \tcRadiusLargeMin\ and \tcRadiusMax, $r$ times its mended rate within a
 factor of \tcFiveFactorWords\ of that at \tcContrastRadius
 & Held: within \tcShareDevMax; above the floor on all \tcFloorPairs\ pairs of example and radius, valid on none;
 valid on \tcUncertValidSixtyFourWords, none resistant; Wall \tcWallRMendTwoFiftySix\ and \tcWallRMendFiveTwelve\ against
 \tcWallRMendSixtyFour\\
\addlinespace
Exact local mending at the largest canvas (Table~\ref{tab:mending}, Section~\ref{supp:mendall})
 & Divergence within \mendDivBar\ of the rule's own; a mended rate of at most \recMilli, with at most \mendCellsBar\ of
 cells changed, on \mendNineBar\ of the \nMoneOf\ examples; radii of at most \mendRadiusBar\ for \mendDefectsBar\ of the
 defects on the best \mendBestWords; failure on independent sampling; exact seams; Chess and Stripes keeping at least
 \recCenti, Col3 and the random proper $4$-coloring mending; fewer cells changed by the final rules. Refuting: failure
 on more than \mendBlocksBar\ of the blocks of an example in the $\log^*$ class. Changed before the main runs, after a
 test on Knot where the divergence bar failed and \mendTestFailed\ of block attempts failed: that bar became Wave
 Function Collapse's own divergence, and the refuting observation a mended rate above \recMilli
 & Partly held. Missed: the divergence, on that test; the rate, on \mendEarlyCount\ with the earlier rules and \nMone\
 with the final; Col3 and the $4$-coloring (\mendedColthree, \mendedColfour). Held: radii, independent sampling,
 seams, Chess and Stripes, and fewer cells changed (\medChanged\ against \mendEarlyChanged). The restated refuting
 observation was read at the time as having occurred, on Angular, Skew2, Qud, City and Sewers, but a constant coloring
 satisfies them all, which places them outside the class it names\\
\addlinespace
Wider mending (Table~\ref{tab:widemend})
 & At radius \mendRadiusMid, at least \wideNewBar\ of the \widePartialOf\ partial examples valid, their median residual
 at most \recHalf\ that at radius \mendRadiusLo, without copies; no certified example valid, their residual at least
 \recHalf\ of it; at most \wideUncertBar\ uncertified resistant examples valid; row-coin rules valid on SimpleWall and
 Skyline, not on Wall
 & Held, except Skyline's row-coin rule, at \wideRowSkyline: \wideMendNew\ partial examples valid on every seed, ratio
 \wideRatio, no reused patch; no certified example, ratio \wideCertRatio; \wideUncertMendedWords\ uncertified;
 SimpleWall at most \wideRowSimpleMax, Wall \wideRowWall\\
\addlinespace
Complete local search on the rule's defects (Table~\ref{tab:sat})
 & Defects completed within radius \wxRhoLo\ on Cave and Maze (\satCaveBar), within \mendRadiusLo\ on Angular, City,
 Qud, Skew2 and Sewers (\recHalf, pooled and, as changed before the run, each), not within \mendRadiusMid\ on Wall
 (\satWallBar); few undecided (\satUndecidedBar); independent sampling's defects, and the rule's against random errors,
 unsolvable more often (\satDiffBar\ points, on \satDiffCount). Changed before the run: encoding controls; the
 mender's failed blocks mostly solvable; the unsolvable share falling by \satFallBar\ points with the radius on
 Angular, Qud and City, flat on Wall
 & Partly held. Missed: Maze (\satOriginalMaze; Cave \satOriginalCave); the mender's leftovers, \satOriginalLeftSat\
 of \satOriginalLeftN\ pooled, nearly all on Skew2 (City \satLeftCity, Skew2 \satLeftSkewtwo, Sewers
 \satLeftSewers); the failed blocks, \satFailedBlocksSolved\ of \satFailedBlocks\ solvable; the fall, with
 \satProfileWithin\ of \satProfileOf\ profiles within \satFlatBar\ points. Held: Wall (\satOriginalWallUnsat, flat);
 \satOriginalUnresolved\ undecided; controls; independent sampling on \satPfourCount, on its letter, as the strata did
 not fix the defect count; random errors on \satRuleMinusRandomCount\\
\addlinespace
Theorem~\ref{thm:floor} with the rule's own statistics (Table~\ref{tab:ownfloor})
 & On each certified example with an invariant or a phase, the rate at least the floor from the rule's own shares of
 the quantity's values, at radius \floorRadiusRaw\ and \floorRadiusRepair; that floor above the threshold of validity
 on at least \thrOwnFloorCount\ of them
 & Held, the first without power, since no shares could have made it fail; the floor exceeds the threshold on all
 but \ofBelowNames. A departure during the run: a 3Bricks canvas made again differed from the
 stored one at a near tie of the rule's scores, so the stored canvases were measured; neither outcome changes\\
\addlinespace
Training for local completability (Table~\ref{tab:completability})
 & Changed before the run, after trials on City: the reward became our mender, block by block. It lowers the mended
 residual to \recThird\ of the starting rule's, and to \recHalf\ of what the defect reward reaches, on \complBar\ of
 the \complOf\ examples, and the unsolvable share of its defects at radius \mendRadiusLo\ by \complDthreeBar\ points
 & Missed: \complDoneCount\ and \complDtwoCount\ of \complOf; the second, read at the time as holding on Qud and
 Sewers, holds on our recount on Sewers alone, a miss either way. Not scored: the unsolvable share\\
\addlinespace
Dependence at a distance in a solver's worlds (Table~\ref{tab:longrange})
 & A dependence that persists from \lrDistMid\ to \lrDistMax\ cells along the certified direction on at least
 \lrCertBar\ of the \lrCertOf\ certified examples whose quantity is not one-way, along some direction on at least \lrUncertBar\ of
 the \certUncertifiedResistantWords\ resistant examples without a certificate, and on at most \lrNonResBar\ of the
 \lrNonResOf\ that do not resist
 & Partly held: \lrCertOrder\ of \lrCertOf, \lrNonResOrder\ of \lrNonResOf. Missed: persistence on none of the
 resistant examples without a certificate, so the targeted search for a quantity on them did not run\\
\addlinespace
The slope of the min-coin copy against the radius (Appendix~\ref{app:longrange})
 & On the certified controls \glTwoControlNames, a slope from \glTwoSlopeLo\ to \glTwoSlopeHi; the resistant
 examples without a certificate read as carrying a quantity in that interval and as carrying none below
 \glTwoSlopeNone; agreement with the scan above on at least \glTwoAgreeBar\ of them
 & Held on the controls. Read as carrying: \glTwoCarriedNames; none: \glTwoNoneNames; neither: \glTwoUndecidedNames.
 Missed: the agreement, on \glTwoAgreeWords\ of them\\
\addlinespace
\multicolumn{3}{@{}l}{\itshape Section~\ref{sec:valid} and Appendices~\ref{app:procedure} to~\ref{app:method},
\ref{app:evidence} and~\ref{app:results}: local generators given the certificate's input}\\
\addlinespace
The certified quantity given by hand (Table~\ref{tab:sharing306})
 & With row coins or phase channels, the mended residual at most \recTenth\ of the independent-coin rule's on
 \shareBar\ of the \shareEOf\ certified examples, with divergence within \shareKLBar\ times its on as many, and no
 harm on Cave
 & Partly held: the residual, on \shareEoneCount, the phase examples by producing the coordinate tiling; Cave
 (\shareCaveKLRow\ against \shareCaveKLCtrl). Missed: the divergence, within the bar on \shareEtwoCount\ only, the
 phase examples; with row coins \shareRowKLLo\ to \shareRowKLHi\ times\\
\addlinespace
Shared coins on SimpleWall (Table~\ref{tab:sharing308})
 & A row-coin rule with a heavier statistical weight reaches the independent-coin rule's divergence (at most
 \pilotKLstar), keeping at most \recFifth\ of its residual and at most \recHalf\ that of column coins; it switches
 value along a row at most \recFifth\ as often; a local coin redraw changes at least \recHalf\ as many cells
 & Partly held: the switches, \pilotSwitchRowLo\ to \pilotSwitchRowHi\ against \pilotSwitchOrdLo\ to
 \pilotSwitchOrdHi. Missed: no row-coin rule reached the divergence (the best \pilotRowKLBest); the redraw, at
 \pilotRedrawRatio\ of the independent-coin rule's share. Not decided: column coins, whose fine-tuning was canceled
 once the first part failed; without it they leave \pilotColOverRow\ times the row-coin residual\\
\addlinespace
The grid channels (Appendix~\ref{app:procedure})
 & With one-hot channels of the coordinates modulo the octaves' block sides, at least \thrGridPhase\ of the phase
 examples (Fabric, RedMaze, SimpleMaze and SmileCity) valid, and Skyline; Cave, Rooms and City still valid, their grid
 statistic at most \recTwice\ as high
 & Missed: valid only on \glPhaseValidNames, translates of the tiling (Fabric \glFabricRate, SmileCity
 \glSmileRate); Skyline \glSkylineSpan; Cave, Rooms and City valid, but a fine coin redraw changes no cell, and on
 \glFixedCountWords\ of the \glTriedCountWords\ examples the raw canvas is the same at every coin seed. After the run
 we saw that Fabric's period, \phaseDFabric, lies outside the octaves' grid\\
\addlinespace
The row-type sampler, first form (Appendix~\ref{app:procedure})
 & Changed before the run: row types read from the certificate's strip labels, walks from the example's own
 transitions. Wall's mended rate falling at least \thrTenfold\ on each canvas, Wall valid on at least
 \thrWallValidTwo\ canvases; Skyline valid; raw rows following their types on at least \thrRowsFollow
 & Missed: Wall at \csWallSpan\ (the octave rule \ccWallResidualLo\ to \ccWallResidualHi), with \csWallRowMatch\ of
 its rows following; Skyline at \csSkylineSpan; rows not scored on the skylines. After the run: the rule read the
 types with a lag that depends on the course height\\
\addlinespace
The row-color term and the phase channels (Appendix~\ref{app:method})
 & Set after a run had been seen: the row-color term; changed before its run: its form, and a run at another training
 seed. With it, Wall's raw rows following their types on at least \thrRowsFollow, Wall valid, Skyline and Skyline2
 valid. Phase channels make Fabric valid, and SmileCity, with raw canvases that differ between coin seeds
 & Partly held: Wall, with \wallRowsFollow\ of rows following and no forbidden window after mending on any canvas at
 both training seeds; Fabric, each canvas a translate of the tiling. Missed: Skyline (\cwSkylineSeeds\ after mending)
 and Skyline2 (\cwSkylineTwoSeeds); SmileCity, at \mpSmileSpan, though its raw canvases differ\\
\addlinespace
The method on every certified example, with the boundary anchor and run limits (Appendix~\ref{app:procedure})
 & Phase channels make RedMaze and SimpleMaze valid. Set after a run had been seen: the boundary anchor and run
 limits, with which Skyline and Skyline2 are valid, rows following their types on at least \thrRowsFollow. Flowers,
 MoreFlowers, Platformer and SimpleWall, retrained under the anchor (changed before the run), stay valid. Changed
 before the run: 3Bricks and Platformer's shortest runs left out
 & Held: both mazes, every raw canvas a translate of the tiling; both skylines, with \baSkylinesFollow\ of rows
 following (at another training seed, read beside, \baSkylinesSeedOneInvalidNames\ is invalid, at
 \baSkylinesSeedOneWorst\ on its worst canvas); \baRetrainValidWords\ retrained, with \baRetrainFollow\ following\\
\addlinespace
SmileCity's lattice (Appendix~\ref{app:smile})
 & The search finds a phase of period \thrSmilePeriod\ along SmileCity's rows at some patch size up to
 \thrSmilePatch; channels of that lattice make it valid, with raw canvases that differ between seeds
 & Missed: no invariant or phase along rows up to the search's reach, \smileCertPmax, nor at \smileCertPbeyond; the
 channels leave \smileMPEightResLo\ to \smileMPEightResHi\ of windows forbidden after mending (\smileMethodResLo\ to
 \smileMethodResHi\ under the prescribed input); the raw canvases differ\\
\addlinespace
3Bricks under the row-type sampler (Appendix~\ref{app:procedure})
 & Its row types read at the certificate's patch size, \bricksCertP, decide its input; if they are not constant, its
 rule retrained under the row-type sampler stays valid
 & Missed: the types are not constant (\bricksTypesWords\ of them), and the retrained rule keeps \threeBricksWorst\ of
 its windows forbidden after mending on a canvas, so 3Bricks leaves the certified count\\
\addlinespace
Rigid examples (Appendix~\ref{app:rigid})
 & Fabric, RedMaze and SimpleMaze rigid at the certificate's patch size; SmileCity and the examples with a row
 invariant not
 & Missed: RedMaze and SimpleMaze are rigid only from patch size \rigidFromP, and at the certificate's own size
 \mazeBranching\ of their \mazeWindows\ windows branch along each axis; Fabric is rigid at no size up to \rigidReach.
 The rest held\\
\addlinespace
The sampler's floors and the copier of courses (Appendix~\ref{app:copier})
 & Row types sampled, each cell drawn from the example's colors for its type, then mended: invalid on Wall, Skyline and
 Skyline2. Set after that floor was read: Wall's sampled courses filled from the example's course at a hashed offset,
 then mended, invalid on some canvas. Set after that was read: this copier valid on Wall, SimpleWall, Platformer and
 Skyline2, with no raw defect and repetition at least \thrCopierRep\ on Platformer and Skyline2
 & Partly held: the first floor (\fOneWall, \fOneSkyline, \fOneSkylineTwo); the copier, valid at every radius run,
 repetition \copierRepLo\ to \copierRepHi. Missed: the stronger floor is valid on Wall (raw \replayRawSpan). Our
 reading that no copier exists on Skyline, Flowers and MoreFlowers was withdrawn after the run\\
\addlinespace
Further copiers, and painting a period (Appendix~\ref{app:copier})
 & Every run length the sampler draws has a coloring, filled by copied rows on all but 3Bricks; copiers without a
 forbidden window on Skyline, Flowers and MoreFlowers (repetition at least \thrUniformCopierRep) and on Wall and
 SimpleWall (at most \thrJointRep, at least \thrJointOffsets\ offsets at each kind of joint); at least
 \thrCourseCoverage\ of each wall canvas in windows of the example as tall as a course, at most \thrRuleRepH\ of the
 rule's; painting a period of a phase example leaves no forbidden window
 & Partly held: the run lengths; both copiers without a forbidden window, repetition \uniformCopierRepLo\ to
 \uniformCopierRepHi\ and below \copierRepWallBelow; coverage \courseCoverageMin, the rule's at most \ruleRepHWallHi;
 painting, by construction. Missed: the offsets, on a canvas of \jointFewExampleNames\ where \jointFewKindsWords\
 kinds of joint never repeat; we kept this copier and withdrew our claim of cleaner joints for the rule, departing
 from the condition set in advance\\
\addlinespace
The held-out walls (Section~\ref{sec:heldout})
 & Wall and SimpleWall transposed are certified along columns and valid under the sampler along $x$; a wall drawn for
 the test, with new course heights and a thicker mortar, is certified along rows and valid
 & Held: \heldoutTransposedValidWords\ transposed walls, every column following its type; the drawn wall
 \heldoutWallHTwoValidWords, with \heldoutWallHTwoFollow\ of its rows following\\
\addlinespace
Whether the rows glue on the held-out walls (Section~\ref{sec:cert})
 & Decided before any copier was built on them: on each wall, a copier valid before mending on all three canvases if
 its rows glue, and no valid copier if they do not
 & Held where it was tested: the rows glue on all \glOneCountWords\ walls, and the copier left no forbidden window on
 any of their \glOneCanvasesWords\ canvases; no wall failed to glue, so the second case did not arise\\
\addlinespace
The octave rule alone on the held-out walls (Appendix~\ref{app:heldout})
 & Invalid on transposed Wall, as on Wall, and valid on transposed SimpleWall, as on SimpleWall; no expectation on the
 drawn wall, whose courses the octaves can lay out
 & Held: \heldOctWallTValidWords\ on transposed Wall, each canvas keeping at least \heldOctWallTMendedLo\ of its
 windows forbidden after mending; \heldOctSimpleWallTValidWords\ on transposed SimpleWall. On the drawn wall,
 \heldOctWallHTwoValidWords, with \heldOctWallHTwoHeightsPct\% of its courses at the example's heights\\
\addlinespace
The held-out mazes (Section~\ref{sec:heldout})
 & On a maze with walls on a lattice of period \thrHeldMazePeriod, the search finds that phase at patch size
 \thrHeldMazePatch\ and no invariant, and the method given it is valid at aligned variation above
 \thrHeldMazeAligned; the same on a maze with openings \doorMazeOpeningWords\ cell wide, at patch size
 \thrDoorMazePatch
 & Partly held. Missed on the first: nothing certified at patch sizes \heldMazeReachLo\ to \heldMazeReach, its
 passages being \heldMazePassageWords\ cells wide, so the procedure gives the octaves; the octave rule, read as set in
 advance, is valid at aligned variation \heldMazeAligned; the control, now the method's own run, was not launched.
 Held on the other: certified as expected, the method \doorMazeMethodValidWords\ at aligned variation
 \doorMazeAligned\\
\addlinespace
Examples from other implementations (Section~\ref{sec:heldout}, Table~\ref{tab:fresh})
 & The procedure, unchanged, on \fxNExamples\ example images from other repositories: every certified example
 resistant; on the certified examples given an input, the method valid on at least \recHalf\ of them and on at least as
 many as the octave rule; the method valid on at least \fxThreeBar\ of the \fxNExamples; on the uncertified examples the
 octave rule valid on more than block-shifted tiles and than i.i.d.\ colors, and over all \fxNExamples\ on more than
 independent coins; the min-coin copy valid on no certified example; at least \recThird\ of the resistant examples
 certified. Changed before the run: \fxRefusedNames, certified along a diagonal, receives no input and counts with the
 octave rule
 &
 Held: all \fxNCertifiedWords\ certified examples resistant; the method valid on \fxMethodInputPhrase\ given an
 input (its world the tiled example on \fxTiledNames), the octave rule on
 \fxOctValidInputWords; the method on \fxMethodValid\ of
 \fxNExamples; the octave rule on \fxOctValidUncert\ uncertified examples, tiles on \fxTileValidUncert\ and i.i.d.\
 colors on \fxIidValidUncertWords, and on \fxOctValidAll\ in all against \fxCtrlValidAll\ for independent coins; the copy
 valid on \fxTcValidCertWords; \fxNCertResistantWords\ of the \fxNResistantWords\ resistant examples certified\\
\addlinespace
The method at other training seeds (Table~\ref{tab:seeds})
 & On the \methodSeedsCertOfWords\ certified examples other than SmileCity, each seed's count within
 \thrMethodSeedsWithin\ of the first seed's; every example valid at the first seed valid at another (refuted if
 \thrMethodSeedsFalsifier\ or more fail at both)
 & Held: \methodSeedZeroCertCount, \methodSeedOneCertCount\ and \methodSeedTwoCertCount; \methodSeedOneLostNames\
 fails at the second seed only, and \methodSeedsInvalidAllNames\ is invalid at every seed\\
\addlinespace
Coin seeds never used in a decision (Table~\ref{tab:seeds})
 & Each rule's count within \thrFresh\ of its count on the development seeds
 & Held: the independent-coin rule valid on \freshFlatCount\ of \nPublic\ on both, the same examples; the method keeps
 \freshMethodKept\ of its \nMethodTwentyFour, losing \freshMethodLostNames\\
\addlinespace
The method's worlds against Theorem~\ref{thm:nogo} (Figure~\ref{fig:bound})
 & Past the bound, valid with total variation below $\kappa_f$, on SimpleWall, RedMaze and SimpleMaze, and on Wall at
 each training seed where it is valid; on none of Skyline, Skyline2, Flowers, MoreFlowers and Platformer (refuted by
 \thrPastFalsifier\ or more)
 & Held: SimpleWall at most \pbSimpleWallTV\ against \simpleWallKappaF, the mazes \pbMazeTV\ against \pbMazeKappaF;
 Wall at both other seeds; none of the others, the closest \pbFiveClosestName\ at \pbFiveClosestTV\ against
 \pbFiveClosestKappa\\
\addlinespace
\multicolumn{3}{@{}l}{\itshape Section~\ref{sec:learning} and Appendix~\ref{app:results}: what a learned rule adds}\\
\addlinespace
The re-solved copier (Table~\ref{tab:copier})
 & Valid on all \copierCountWords\ examples with a copier at every seed; at most \recHalf\ of its rows equal to a row
 of the example (the copier's share is \copierRowCopyMin); on Wall and SimpleWall a total variation between the
 copier's and the rule's on every canvas. Set before the run: if it is valid everywhere, copies a share of rows at
 most \recTenth\ above the rule's on each and lies closer than the rule on at least \crFiveBar, the paper says
 learning is not needed there
 & Held, validity by construction: below \recHalf\ on every canvas, with \crMultiRowWords\ of its rows that mix colors
 equal to a row of the example; between them on every canvas (Wall \crTvWallLo\ to \crTvWallHi); closer than the
 rule on \crTvBelowRuleCountWords, all but \crTvAboveRuleNames, so the paper gives that account\\
\addlinespace
The Metropolis filler and the longer Metropolis rule (Appendix~\ref{app:floors})
 & Annealed Metropolis on the prescribed input, each row held to its type, valid on at least \thrTypedValid\ of the
 \rowInvCountWords\ examples with a row invariant after \mhTypedSweeps\ sweeps; without the input, at \mhLongFactorWords\ times the sweeps, valid on
 at most \thrLongValid\ of the uncertified examples it leaves invalid (refuted by \thrLongFalsifier)
 & Held: on \mhTypedValidCountWords, \mhTypedValidNames, though nowhere as close to the example's statistics as the
 rule, which is valid where the filler is not on \mhRuleOnlyNames; the longer rule on \mhLongValidWords\ of them\\
\addlinespace
The variation term under the row types (Table~\ref{tab:noterm})
 & Wall's rule trained without the term is valid and copies no blocks (on each canvas at most \thrVZeroRep\ of its
 patches in the example, agreement at most \thrVZeroReplay); set before the run, it copies rows if more than
 \thrVZeroRows\ of its rows or of its windows as tall as a course occur in the example on \thrVZeroCanvases\ canvases
 & Held: valid on every coin seed, repeating \vZeroWallRepWords\ of its patches at agreement \vZeroWallNrep; its rows
 equal the example's only at the mortar (\vZeroWallRowCopyHi), so it does not copy rows\\
\addlinespace
The octave rule on \msPilotExamplesWords\ examples (Table~\ref{tab:pilotablation})
 & Changed before the run: the variation term's weight reset at each update to a fixed share of the loss. On at least
 \msWallsBar\ walls a residual at most \recTwice\ the row-coin rule's and \recThird\ of the independent-coin rule's; on
 \msVarBar\ of SimpleWall, Wall, Skyline, Cave and Rooms, aligned variation at most \msAlignedSlack\ below the
 independent-coin rule's, agreement at most \msNrepSlack\ above and divergence within \msKLBar\ times, and on
 \msWallsBar\ walls a divergence at most the row-coin rule's; correlation length at least \msCorrBar\ times its on Cave
 and Rooms; at least \msUncertBar\ uncertified resistant examples
 valid; the term's removal lowering variation on \msVarBar
 & Partly held: the walls, on \msWallsFound\ of \msWallsFoundOf\ (Skyline above \recTwice). Missed: variation, on
 \msVariationKeptCount; divergence \msWallKLRatioLo\ to \msWallKLRatioHi\ times higher on SimpleWall and Wall, and
 above the row-coin rule's on both; Cave's
 correlation length, \msCaveCorrLength\ against \flatCaveCorrLength\ (Rooms held); the uncertified examples, only
 \msPilotUncertValidNames\ valid (only \strongTermPilotUncertValidNames\ under the stronger term); the term's
 removal, on \msTermEffectCount\\
\addlinespace
The octave rule on every public example (Tables~\ref{tab:radii} and~\ref{tab:variation})
 & At radius \mendRadiusLo, at least \msCountBar\ valid, keeping \msKeepBar\ of the independent-coin rule's
 \nMendEight; resistant residuals cut at least \msCutBar; on that rule's valid examples divergence within
 \msKLMendBar\ times its and agreement at most \msNrepBar\ above. Then at least \msSixteenBar\ valid at radius
 \mendRadiusMid; validity on at least \msSeedsBar\ of \robustSeeds\ seeds; agreement, repetition, aligned variation
 and local redraws within the bars of the evaluation against replay; divergence before mending at or below Wave
 Function Collapse's on at least as many examples as the independent-coin rule
 & Held, except the divergence before mending, on \msWfcCount\ of \msWfcOf\ against \wfcCountFlat: valid on
 \nMendM\ of \nMendMOf; \msKept\ kept; residuals cut \msCutMedian\ times; divergence \msKLMendedMedian\ against
 \flatKLMendedMedian; \nMendMSixteen\ valid at radius \mendRadiusMid; \msRobustBelowSeven\ below the bar on seeds;
 agreement \msNrepMedian, repetition \msRepetitionMedian, aligned variation \msAlignedRatio\ of the block reference's,
 local redraws on \msLocalAllValidCount\ of \msVPopulation\\
\addlinespace
The octave rule at other training seeds (Table~\ref{tab:seeds})
 & At each seed at least \tsCountBar\ examples valid at radius \mendRadiusLo, at least \tsKeepBar\ of the first seed's
 \nMendM\ kept, and a count within \tsDiffBar\ of the variant without the variation term (changed before the run: the
 seeds ran beside the smooth coins); at radius \mendRadiusMain, counts within \thrSeedsWithin\ of \nCCM, with at least
 \thrSeedsUncert\ of \mhCCOnlyNames\ invalid
 & Partly held: \trainSeedCounts; at radius \mendRadiusMain, \trainSeedOneCount\ and \trainSeedTwoCount, with all of
 \mhCCOnlyNames\ invalid. Missed: the variant, at \trainSeedCountsNoTerm, differs by more than \tsDiffBar\ at the last
 seed; its sum does not exceed the rule's, so the rule stays\\
\addlinespace
The octaves' grid in the output (Appendix~\ref{app:sharing})
 & On at least \latticeBar\ of the \latticeOf\ examples the octave rule makes valid, a grid statistic at most
 \recTwice\ the independent-coin rule's; if not, the paper says that the grid is visible
 & Missed: on \latticeWithinTwice\ only, at medians of \latticeMedianM\ and \latticeMedianFlat, so the paper says so.
 Measured beside it without an expectation, at periods the grid cannot favor: within \recTwice\ the rule's own value
 there on \latticeWithinOwnNull\\
\addlinespace
Coins that vary smoothly between lattice points (Table~\ref{tab:smooth})
 & Changed before the run: City, Angular, Flowers, Platformer and Village added. A median grid statistic at most
 \recHalf\ the block coins'; valid as often as block coins on the first study's examples, less \smLessBar; residuals on
 the walls at most \recTwice\ theirs. At radius \mendRadiusMid, Wall's mean residual at most \recHalf\ its value at
 \mendRadiusLo, and at least \smTwoBar\ of SimpleWall, Flowers and Platformer valid; written though we expected neither
 to hold: Wall valid on every canvas, as many examples valid as block coins. With cluster completion at radius
 \mendRadiusMain, fewer valid than block coins, which then stay
 & Partly held: the grid statistic, \smoothPilotLatticeM\ against \smoothPilotLatticeBlock; the count; at radius \mendRadiusMid, all of
 SimpleWall, Flowers and Platformer, and neither statement we expected not to hold held; \smoothCCM\ against \nCCM,
 so the block coins stay. Missed: SimpleWall's and Skyline's residuals, above \recTwice\ the block coins' (Wall's
 fell); Wall at radius \mendRadiusMid, only from \smoothWallEight\ to \smoothWallSixteen\\
\addlinespace
Cluster completion and its controls (Table~\ref{tab:radii})
 & On the smooth coins' \ccSmoothOf\ examples invalid at radius \mendRadiusLo, a lower residual on at least
 \ccSmoothLowerBar\ and a higher one on no canvas, at least \ccValidEightBar\ valid at radius \mendRadiusLo\ and
 \ccValidSixteenBar\ at \mendRadiusMid. On the octave rule, at least \ccSixteenBar\ valid at radius \mendRadiusMid\ and
 \ccTwentyFourBar\ at \mendRadiusMain, Wall invalid, under \recTenth\ of cells changed on every valid canvas; i.i.d.\
 colors valid on fewer; the independent-coin rule on at most \ccFlatBar, its lead growing with the radius; the counts
 at radius \mendRadiusLo\ of both rules
 & Partly held. Missed: on the smooth coins, a lower residual on all but Wall and Platformer, unchanged there and
 higher on a canvas of Wall, and none valid at radius \mendRadiusLo; on the rule, more than \recTenth\ of cells changed
 on every newly valid example, up to \ccChangedMax. Held: SimpleWall, Flowers, Platformer and ColoredCity valid at
 radius \mendRadiusMid; on the rule \nCCSixteenM\ and \nCCM, Wall invalid; i.i.d.\ colors valid on \iidCCValid\ of \iidCCOf; the
 independent-coin rule on \nCCFlat, leads \ccLeadEight, \ccLeadSixteen\ and \ccLeadTwentyFour; the counts at
 \mendRadiusLo\\
\addlinespace
Mending i.i.d.\ colors (Appendix~\ref{app:results})
 & On none of the \nMendM\ examples the octave rule makes valid at radius \mendRadiusLo\ do i.i.d.\ colors at the
 example's frequencies become valid under the same mender at radius \mendRadiusMain
 & Held: none; with the earlier control, valid on none of the \iidAllOf\ public examples, every canvas keeping at least
 \iidAllResidualMin\ of its windows forbidden\\
\addlinespace
Variants of the mender, not adopted (Table~\ref{tab:radii})
 & A pass that restores changed cells where their windows allow cuts the share of cells changed \mvTwofold\ and moves
 either rule's count by at most \mvOneBar; grouping a failed block's clusters cuts Wall's residual \mvTwofold\ on
 \mvWallBar\ canvases and brings the rule to at least \mvRuleBar\ examples, the independent-coin rule staying at most
 \mvFlatBar
 & Partly held: the pass leaves both counts unchanged; the independent-coin rule stays within its bar on the examples
 whose run finished, before the run stopped. Missed: the share hardly moves; Wall's residual is unchanged on every
 canvas; the rule stays at \nCCM\\
\addlinespace
The octave rule re-evaluated at radius \mendRadiusMain\ (Table~\ref{tab:radii}, Appendix~\ref{app:replay})
 & Counts reproduced on the development canvases; on \robustSeeds\ canvases at least \reRuleBar\ for the rule and at
 most \reFlatBar\ for the independent-coin rule; agreement with the tiling at most \reNrepBar\ and aligned variation at
 least \reAlignedBar
 & Partly held: the counts reproduced; \ccRobustM\ for the rule; agreement \ccNrepMedian, aligned variation
 \ccAlignedMedian. Missed: the
 independent-coin rule, at \ccRobustFlat\\
\addlinespace
Block-shifted tiles, a floor without learning (Table~\ref{tab:radii})
 & Tiles of the example shifted block by block, mended at radius \mendRadiusMain, valid on at least \tileValidBar\ of
 the \nPublic\ examples and on at most \tileCertBar\ of the \certResistantCount\ certified ones, at a median repetition
 of at least \recHalf
 & Partly held: valid on \tileCCCertValid\ of the certified examples. Missed: \tileCCValid\ valid, at a median
 repetition of \tileRepetitionMedian\\
\addlinespace
Checks raised by an outside reading (Appendix~\ref{app:results})
 & On Wall at least \thrWallMisplaced\ of the octave rule's forbidden windows on rows misplaced along $y$; on Wall and
 Skyline fewer of its unsolvable clusters with a witness than the independent-coin rule's; with the mender's first
 pass alone, at least \thrOnePassM\ valid for it, at most \thrOnePassFlat\ for the independent-coin rule; its count
 rising with the rounds, at most \roundsFourBar\ at \roundsFewest; Metropolis then the mender valid on at most
 \thrMetropolis\ (refuted by \thrMetropolisFalsifier); a smaller radius of completion than the independent-coin
 rule's; the block solver keeping more cells equal to the cell a period away on at least \periodSolverAboveRuleBar\ of
 the \periodSolverAboveRuleOf\ examples
 & Partly held: misplaced rows, pooled \wallMisplacedPooled\ (by canvas \wallMisplacedSeeds); witnesses on Wall,
 \witRuleWall\ against \witCoverageWall; the independent-coin rule at \onePassFlat; the rounds, \roundsCounts;
 Metropolis, \mhCCValid. Missed: Skyline's witnesses, \witRuleSkyline\ against \witCoverageSkyline; the octave rule at
 \onePassM; the radius of completion, equal at the median and smaller on \radCompSmallerOn\ of \radCompOf; the block
 solver ahead on \periodSolverAboveRule\\
\addlinespace
Hand-written rules (Table~\ref{tab:baselines})
 & On \hwBar\ of the \hwOf\ examples, the Metropolis rule's raw rate at least \recTwice\ the independent-coin rule's and
 its divergence worse, and after mending the rule's residual at most \recHalf\ the Metropolis rule's; independent
 sampling with repair \hwTenfold\ above the rule. Refuting: a hand-written rule within \recTwice\ the learned one's
 rate and divergence on most examples
 & Partly held: the raw rate on \nBone\ (\nBoneSixtyfour\ with the longer run); the mended residual on \nBfour, at a
 median ratio of \medRatioMHmended; the refuting observation did not occur. Missed: the divergence, worse on \nBtwo;
 independent sampling, \hwTenfold\ above on \nBthree, at median factors of \medRatioFonerepThirtytwo\ and
 \medRatioFonerep\\
\addlinespace
The evaluation against replay (Table~\ref{tab:replay})
 & On the independent-coin rule's valid examples: normalized agreement with the tiling at most \replayBar\ (changed
 before the run from raw agreement), patch reuse at most \recHalf\ of shifted replay's, aligned variation at least
 \recHalf\ of the block reference's, local redraws in valid windows, and better than the block reference on total or
 aligned variation on \recHalf\ of the examples (a cost bar dropped before the run)
 & Held: agreement \replayNormG\ (block reference \replayNormB), reuse \replayRepetitionG\ against
 \replayRepetitionShift; aligned variation \replayAlignedRatio\ of the reference's; local redraws on \replayLocalCount\
 of \replayLocalOf; better on \replayVfourCount\ of \replayPopulation. The raw agreement the bar first named is
 \replayRawG, against \replayIid\ for an i.i.d.\ draw\\
\addlinespace
The random-access block solver against the independent-coin rule (Table~\ref{tab:replayra})
 & Changed before the run: the solver runs both its passes whatever their cost. The rule's median aligned variation at
 least the solver's, its normalized agreement at most the solver's, and better on either measure on \recHalf\ of the
 examples
 & Partly held: agreement \raNormG\ against \raNormA; better on \raBetterCount\ of \raOf. Missed: aligned variation,
 \raAlignedG\ against \raAlignedA. Its passes cost \raCostRatioLo\ to \raCostRatioHi\ times the rule's compute\\
\addlinespace
The octave rule against the random-access block solver (Table~\ref{tab:blocksolverrule})
 & After the solver's first pass (read, as changed before the run, from its run with both passes), the octave rule's
 median aligned variation at least \bsBar\ times the solver's and its normalized agreement below the solver's
 & Held: aligned variation \msAlignedMedian\ against \bsOneAligned, a ratio of \bsOneAlignedRatio; agreement
 \msNrepMedian\ against \bsOneNrep.
 After both passes, read after the run, the ratio is \bsTwoAlignedRatio\ and the solver's agreement \bsTwoNrep\\
\addlinespace
Connectivity of the passable color (Appendix~\ref{app:replay})
 & On at least \recHalf\ of the examples kept, the independent-coin rule's largest passable component at least
 \connBar\ times shifted replay's
 & Held: \flatConnFlatSet\ of \msConnFlatSetOf; an i.i.d.\ draw passes on \iidConnFlatSet\\
\addlinespace
\multicolumn{3}{@{}l}{\itshape Section~\ref{sec:access} and Appendix~\ref{app:seam}: random access after mending}\\
\addlinespace
Chunked against whole-canvas mending (Table~\ref{tab:seamcc})
 & Without cluster completion, at radius \mendRadiusLo, agreement on every interior cell of every canvas (changed
 before the run: the test, which as first set could not fail). With it, agreement at a margin of \seamCCMarginBar\ on
 at least \seamCCShareBar\ of the \seamCCCanvases\ canvases; every canvas that differed at \seamMarginBase\ agreeing at
 \thrSeamWide; on the method's rules other than the octave rule, agreement on every canvas at \thrSeamMethod
 & Held: without cluster completion on all \seamCanvases\ canvases, while at a narrower margin, tested without an
 expectation, they differ on \seamDiagCount\ of the \seamExamples\ examples; with it, \seamCCAgreeOneNinetyTwo\ of
 \seamCCCanvases\ at \seamCCMarginBar\ (\seamCCWorstExample\ still differs at \seamCCWorstMargin); all at
 \seamMarginsWide; \seamMethodAgree\ of
 \seamMethodOf\ and \seamRetrainsAgree\ of \seamRetrainsOf\ canvases, with no cell differing; SmileCity's rule not
 run\\
\addlinespace
Wider mending of the method (Table~\ref{tab:mwide})
 & Within radius \mwRadiusA, at least two of 3Bricks, Skew1 and ScaledMaze valid, and at most one of Lake, Link2 and
 Mountains, with their median rate at most \mwTwoBar\ of radius \mendRadiusMain's; SmileCity invalid at \mwRadiusA\ and
 \mwRadiusB, keeping at least half its rate at \mwRadiusB; within \mwRadiusB, at least four of the other six valid;
 wherever an example becomes valid, replay agreement and the share of the example's patches within \mwFiveBar\ of radius
 \mendRadiusMain's
 & Partly held: \mwValidANames\ valid at \mwRadiusA, none of the other three, and no copying (at most \mwFiveMax).
 Missed: the median rate of Lake, Link2 and Mountains, \mwFarRatioMedian\ of radius \mendRadiusMain's; SmileCity, valid
 at \mwRadiusB; at \mwRadiusB, \mwFourN\ of the six, with \mwKilledNames\ stopped by the time limit set in advance and
 counted as invalid\\
\addlinespace
\multicolumn{3}{@{}l}{\itshape Section~\ref{sec:diffusion} and Appendix~\ref{app:diffusion}: a diffusion model given the
input}\\
\addlinespace
The diffusion model (Table~\ref{tab:diffusion})
 & Pilot: a budget at which, on noise alone, the model is valid on Cave and leaves before mending at most
 \ddPilotBarWords\ times the independent-coin rule's forbidden windows. On noise alone, invalid on every certified
 example and valid on at least \ddTwoBar\ of the \ddTwoOfWords\ uncertified controls, else its failures are not read as
 specific; with the input, valid without copying on at least \ddThreeBar\ certified examples with a
 transported quantity, else the input is not sufficient for this model. Also: reading the input, Chess and Stripes
 valid only with it, no gain from more levels, and the floor and seam checks
 & Partly held. Missed: the pilot, at no setting within the bar (the study ran at the one with the fewest forbidden
 windows on Cave, \ddCaveRawOverRule\ times the rule's); the controls, valid on \ddTwoValidNames\ only, so its failures
 are not read as specific; validity without copying, on \ddThreeCountedWords\ of the \ddThreeValidWords\ valid, so by
 the measure fixed in advance the input is not sufficient for this architecture at our training budget. The bar
 overlooked that the valid worlds of \ddThreeEscapeNames\ are essentially translates of the tiling, which left
 \ddThreeBar\ to find among the other \ddTransNonPhaseWords\ (the method makes \ddMethodValidNonPhaseWords\ valid).
 Held: the rest, invalid on every certified example from noise among them\\
\addlinespace
The diffusion model's rows (Appendix~\ref{app:diffusion})
 & Written after the run, before this test read its canvases: on Wall and SimpleWall at least \recHalf\ of the model's
 rows that mix colors are rows of the example, on Flowers, MoreFlowers and Platformer fewer, on Fabric, RedMaze and
 SimpleMaze all (a check); an example counted without copying where at least \recHalf\ are leaves the count
 & Partly held: none on Flowers, MoreFlowers and Platformer, all on the tiled examples, so the count stays at
 \ddThreeCountedWords. Missed: none of the rows of Wall and SimpleWall that mix colors is a row of the example\\
\end{longtable}}

%% file: supp_tables.tex
\section{Additional tables}
\label{supp:tables}

These tables and figures carry measurements that the paper states without showing, and those that the record of
what we expected (Appendix~\ref{app:record}) cites. Every table here that mends was made before cluster completion, within
radius \mendRadiusLo\ and blocks of 16, except Table~\ref{tab:mendall}, which mends within radius 4 with blocks of 8,
Table~\ref{tab:forced}, which mends around the pinned block within radius 6, and the wider radii of
Table~\ref{tab:widemendrule}.

\subsection{The octave rule at radius 8}
\label{supp:octave8}

The corpus evaluation without cluster completion, on the same canvases at radius \mendRadiusLo, also mended the octave
rule's canvases within radius 12 and 16 with blocks of 24 and 32 (Table~\ref{tab:widemendrule}).

\begin{table*}[p]\centering\tiny
\caption{Both rules on the corpus at radius 8, evaluated by the same code: three $256^2$ canvases per example (coin seeds
shared), 32 rounds, four repair rounds and mending within radius 8 with blocks of 16, without cluster completion.
Columns: the residual
after mending, mean and worst of the three seeds, for the independent-coin rule and for the octave rule; on how many
seeds the octave rule leaves at most one forbidden window in a thousand; and the $3\times3$ KL of the mended canvas for
both. The class is the one mending gives under independent coins (Table~\ref{tab:mendmapeight}); an asterisk marks a
certified example. Figure~\ref{fig:corpus} gives the residuals after mending within radius 24.}
\label{tab:mendmapboth}
\input{tables/mendmap_both_v13}\end{table*}

\begin{table*}[!htbp]\centering\scriptsize
\caption{The octave rule mended wider without cluster completion, on every example where it leaves a
defect at radius 8 on some seed: the mean residual over the three canvases of Table~\ref{tab:mendmapboth} after mending
within radius 8, 12 and 16 (blocks of 16, 24 and 32), on how many of the three each radius leaves at most one forbidden window in a thousand, and the share of cells that
mending within radius 16 changes. Class and asterisk as in Table~\ref{tab:mendmapboth}.}
\label{tab:widemendrule}
\input{tables/widemend_v13}\end{table*}

\begin{table}[p]\centering\tiny\setlength{\tabcolsep}{4pt}
\caption{The octave rule's residual after mending within radius 8, without cluster completion, on the \robustSeeds\ canvases per
example of
Table~\ref{tab:variation}: on how many it is at most one forbidden window in a thousand, the mean and the worst. A
dagger marks the examples the octave rule makes valid on the three canvases of Table~\ref{tab:mendmapboth}; class and
asterisk as there.}
\label{tab:robust}
\input{tables/robust_v13}\end{table}

At radius 8 the octave rule is valid on \nMendM\ examples (Table~\ref{tab:mendmapboth}), and its worst canvas tells
the same story as its mean on almost every example. Without
cluster completion, mending within radius 12 makes it valid on \nMendMTwelve\ and within radius 16 on \nMendMSixteen\
(Table~\ref{tab:widemendrule}), against \nCCSixteenM\ at radius 16 with cluster completion (Table~\ref{tab:radii}).
The larger radius changes more cells, and on a few examples, Village among them, the residual at radius 16 is slightly
above the one at radius 12. That mender leaves a block as it was when it finds no completion of all its defects
together, so one defect it cannot complete keeps every other defect of its block, and a larger block keeps more. This
is the effect that cluster completion removes. The \certPhaseCount\ phase examples, \msPhaseNames, keep nearly the same residual at every
radius, and the radius-16 mender changes almost none of their cells.

Validity at radius 8 holds beyond the three canvases. On \robustSeeds\ canvases per example (Table~\ref{tab:robust}),
\msRobustAllEight\ of the \nMendM\ examples are valid on every canvas, and \msRobustNotAll\ on all but one. The
\msRobustCertifiedAbove\ phase examples stay above one forbidden window in a hundred on every canvas.

\subsection{The independent-coin rule against the references}
\label{supp:replay}

\begin{table*}[!htbp]\centering\scriptsize
\caption{The same evaluation for the independent-coin rule on the \nMendEight\ examples it makes valid, with the same
seeds, mender and block reference. Columns: the total variation from the example's $3\times3$ frequencies; the share of
cells that agree with the best translated tiling of the example, for an i.i.d.\ canvas drawn from the color
frequencies, for the independent-coin rule and for block, and the same normalized as $(a-a_{\mathrm{iid}})/(1-a_{\mathrm{iid}})$; the
share of positions whose $16\times16$ patch occurs in the tiled example; the share of cells that differ between two
canvases after the best alignment; and the share of the cells changed by a local coin redraw that lie in allowed windows
only.}
\label{tab:replay}
\input{tables/replay_valid_v12}\end{table*}

Table~\ref{tab:replay} is the same evaluation for the independent-coin rule, whose normalized agreement has a median of
\replayNormG\ against \replayNormB\ for the block reference.

\begin{table*}[!htbp]\centering\footnotesize
\caption{The independent-coin rule after mending against the random-access block solver of Section~\ref{sec:setup},
which starts from shifted replay and re-solves the $16\times16$ blocks of an absolute lattice class by class, in two
passes with the lattice shifted by half a block, each block seeded by its coordinates. Means over eight $256^2$ canvases
with the seeds of Table~\ref{tab:replay}; the last column is the solver's cost per cell over the independent-coin
rule's, single-thread CPU time.}
\label{tab:replayra}
\input{tables/replay_ra_v12}\end{table*}

Against the independent-coin rule the random-access block solver runs its two passes, with the lattice shifted by half
a block between them, which are the least it can run, and they cost \raCostRatioLo\ to \raCostRatioHi\ times
the independent-coin rule's compute per cell, whose cost counts generation, repair and mending within radius 8, so we
compared at that cost (Table~\ref{tab:replayra}). Its outputs have no forbidden window by construction, and they keep
more of the replay they start from: normalized agreement \raNormA\ against \raNormG\ at the median, and the
independent-coin rule's is lower on \raBetterReplay\ of the \raOf\ examples. Its aligned variation is higher,
\raAlignedA\ against \raAlignedG, and the independent-coin rule's is higher on \raBetterAligned\ of them. Its total
variation is above the independent-coin rule's on \raTVBetterG\ examples. On the
certified examples it is valid and close to replay, with normalized agreement \raWallNormALo\ to \raWallNormAHi\ against
the independent-coin rule's \raWallNormGLo\ to \raWallNormGHi.

\subsection{The octaves, shared input given by hand, and completability}
\label{supp:sharing}

\paragraph{The study of the variation term.}
The study of the octaves on \msPilotExamplesWords\ examples, SimpleWall, Wall, Skyline, Cave, Rooms and the
\msPilotUncertOf\ uncertified resistant examples (Table~\ref{tab:pilotablation}), trains the same family with no term
and with the weight set to \varFracStrong\ of the loss. Its rules at the standard weight are the octave rules of those
examples, and the rules of the other examples were trained afterwards in the same configuration. The octave rule was
also trained without the term on every example, in the configuration above otherwise (Appendix~\ref{app:sharing}).

\paragraph{Rules given the direction.}
The row-coin rule, on SimpleWall, Wall and Skyline, and the column-coin rule, on SimpleWall, are trained from scratch in
the configuration of Appendix~\ref{app:method}, with \dirExtraChannelsWords\ extra channels in place of the coin family
and no variation term. Row coins are hashed from the seed, the row and the channel only, so they are equal along every
row; column coins are the same channels hashed from the column. On the examples whose certified quantity is a phase of
period $d$, the extra input is one-hot channels of $x\bmod d$ and $y\bmod d$. Each rule is compared with the
independent-coin rule of the same example, evaluated in the same way. The study of the statistical weight on SimpleWall
fine-tunes the row-coin rule for \statWeightUpdates\ further updates with the weight on the window divergence multiplied
by \statWeightRowList, and the independent-coin rule for the same budget with the weight multiplied by
\statWeightOrdList, at a constant learning rate of \statWeightLR\ and at the temperature where training ends, \tempEnd,
from the start.

\begin{table*}[!htbp]\centering\footnotesize
\caption{The study of the octaves on \msPilotExamplesWords\ examples (three $256^2$ canvases each, coin seeds shared, mended within
radius 8 without cluster completion): the walls whose certificate is an invariant along rows, Cave and Rooms, and the \msPilotUncertOf\ uncertified
resistant examples. Columns: the residual and the $3\times3$ KL after mending of the independent-coin rule, of the
row-coin rule (walls only), of the octave rule without the variation term, of the octave rule, and of the octave rule
with the term at the heavier weight of Appendix~\ref{app:params} (``term 0.3'', three tenths of the loss).}
\label{tab:pilotablation}
\input{tables/pilot_ablation_v13}\end{table*}

The study of the octaves on \msPilotExamplesWords\ examples separates what the octaves give from what the variation term gives
(Table~\ref{tab:pilotablation}). The octave rule without the term already reaches or passes the row-coin rule on
SimpleWall and Wall (\noTermSimpleWallResidual\ and \noTermWallResidual), and on Skyline it stays above it
(\noTermSkylineResidual). The term at its standard weight changes little, and at the heavier weight it returns Wall to
the level of independent coins (\strongTermWallResidual). On the \msPilotUncertOf\ uncertified resistant examples the
octave rule cuts the residual by \msPilotUncertCutLo\ to \msPilotUncertCutHi\ times, a median of \msPilotUncertCutMedian, at a
median KL \msPilotUncertKLRatio\ times that of independent coins, and makes \msPilotUncertValidNames\ valid; without the
term the median cut is \noTermPilotUncertCutMedian\ and \noTermPilotUncertValidNames\ become valid. On Cave and Rooms
the octave rule's divergence falls below that of independent coins (\msCaveKL\ against \flatCaveKL, \msRoomsKL\ against
\flatRoomsKL).

\begin{table*}[!htbp]\centering\footnotesize
\caption{The certified quantity as an extra input given by hand. The independent-coin rule, and the same construction
trained with row coins (SimpleWall, Skyline, Wall) or with the phase of its coordinates as one-hot channels (Chess,
Stripes, RedMaze, SimpleMaze), at 32 rounds with four repair rounds, on three $256^2$ canvases (coin seeds shared)
mended within radius 8 with blocks of 16, without cluster completion; means over the canvases. Cave, which carries no certificate, checks
that row coins do no harm. The last two columns count the cells changed by redrawing the coins of one $16\times16$
region.}
\label{tab:sharing306}
\input{tables/sharing306_v12}\end{table*}

\begin{table*}[!htbp]\centering\footnotesize
\caption{Coins shared by hand on SimpleWall: the independent-coin rule (``ordinary''), the row-coin rule and the
column-coin rule, with the weight on the window statistics each was trained or fine-tuned with; means over the three
$256^2$ canvases of Table~\ref{tab:sharing306}, mended as there. Columns: the raw forbidden-window rate, the
residual and the $3\times3$ KL after mending, how often the transported quantity switches value per hundred window
positions along a row of the raw canvas, the share of the mended canvas that changes when every per-cell coin of the
canvas is redrawn with the shared coins held (a different measure from the local variation of
Section~\ref{sec:setup}), and the agreement with the best translated tiling.}
\label{tab:sharing308}
\input{tables/sharing308_v12}\end{table*}

\begin{table*}[!htbp]\centering\footnotesize
\caption{Training for local completability: the mended residual and KL of the starting rule, of the rule fine-tuned
with a reward for each defect it removes, and of the rule fine-tuned with the mender's own success as the reward. The
starting rule is the first of the three independent-coin rules of Table~\ref{tab:mending}; every rule is run
for 32 rounds with four repair rounds on three $256^2$ canvases and mended within radius 8 with blocks of 16, without
cluster completion.}
\label{tab:completability}
\input{tables/completability_v12}\end{table*}

\paragraph{Shared input given by hand.}
With row coins the residual falls on all three walls (Table~\ref{tab:sharing306}), by about \shareRowCutHi\ times on
SimpleWall, and the coin edit changes fewer cells. With the phase of the coordinates as input, the residual, the
divergence and the cells changed by a coin edit are all zero,
which is the signature of a rule that outputs the coordinate tiling, the escape through coordinates of
Section~\ref{sec:theory}. On Cave the row coins leave the divergence at \shareCaveKLRow, against \shareCaveKLCtrl\ for
the independent-coin rule.

On SimpleWall (Table~\ref{tab:sharing308}) a heavier statistical weight lowers the divergence of the independent-coin
rule below its starting value without lowering its residual, and lowers that of the row-coin rule from
\pilotRowKLStart\ to \pilotRowKLBest\ at the heaviest weight, which is still above the independent-coin rule's starting
\pilotKLb. The fine-tuned row-coin rules keep residuals from \pilotRowResLo\ to \pilotRowResHi, about
\pilotRowResRatioLoR\ to \pilotRowResRatioHiR\ times below the independent-coin rule's. The quantity switches along a
row about as often for the column-coin rule as for the independent-coin rule, and redrawing every per-cell coin of the
canvas, with the shared coins held, changes \pilotRedrawRowLo\ to \pilotRedrawRowHi\ of the cells for the row-coin
rules against \pilotRedrawOrd\ for the independent-coin rule. The column-coin rule was not fine-tuned.

\paragraph{Training for completability.}
Rewarding a rule on independent coins for leaving defects that the mender completes did not help consistently. We
fine-tuned it on City, Qud, Skew2 and Sewers with two rewards: one for each forbidden window removed, and one for each
block the mender completes (Table~\ref{tab:completability}). The mender's reward cut the residual to a third of the
starting rule's or less on Qud alone, and to below half of what the defect reward reaches on Sewers alone, where both
rewards left more defects than the starting rule.

\subsection{Inequality~(\ref{eq:vfr}) on measured outputs}
\label{supp:floorcheck}

\begin{table*}[p]\centering\scriptsize
\caption{Inequality~\eqref{eq:vfr}, $\varepsilon(2r+4)+\delta\ge\kappa_f$, evaluated on measured outputs for the
certified examples with $P=3$, every example but SimpleWall. $\varepsilon$ is the density of forbidden windows,
$\delta$ the total variation from the example's $3\times3$ frequencies, in which all forbidden windows together count as
one outcome, $\kappa_f$ the constant under the full labeling, $r$ the total radius, and the slack is
$\varepsilon(2r+4)+\delta-\kappa_f$. Unmarked rows ``indep.\ coins'' are the independent-coin rule on
one $1024^2$ canvas, raw at radius \floorRadiusRaw\ and with four repair rounds at radius \floorRadiusRepair; rows
marked $^c$ pool the eight $256^2$ canvases of Table~\ref{tab:replay}, mended within radius 8 with blocks of 16; the
rows with row coins or a phase input are the rules of Table~\ref{tab:sharing306}, three $256^2$ canvases each. Rows
marked $^c$ are computed from stored canvases; the others come from run summaries that store $\varepsilon$ and the
$L_1$ distance over allowed windows, so $\delta$ is an interval and so is the slack. The last column says whether the
inequality covers the output: it covers the independent-coin rule, raw and with repair, which is
translation-equivariant and reads independent coins. Mended outputs, rules with shared coins and rules that read the phase of their coordinates are
outside it and are given for context.}
\label{tab:floorchecka}
\input{tables/floorcheck_v12a}\end{table*}

\begin{table*}[p]\centering\scriptsize
\caption{The same as Table~\ref{tab:floorchecka} for SimpleWall, with every rule of Table~\ref{tab:sharing308}, each on
its three $256^2$ canvases (mended within radius 8 with blocks of 16); the number in parentheses is the weight on the
window statistics a rule was fine-tuned with.}
\label{tab:floorcheckb}
\input{tables/floorcheck_v12b}\end{table*}

Every measured output that inequality~\eqref{eq:vfr} covers satisfies it. Tables~\ref{tab:floorchecka}
and~\ref{tab:floorcheckb} evaluate it on every output we stored for the certified examples with $P=3$, at the radius of
the raw independent-coin rule, \floorRadiusRaw, and of that rule with four repair rounds, \floorRadiusRepair. All
\floorCheckCovered\ outputs the inequality covers satisfy it, with a smallest slack of \floorCheckMinSlackR.
The slack is large because the raw rules leave a few percent of their windows forbidden at these radii, so the defect
term alone exceeds $\kappa_f$. The check is therefore a consistency check of the measurements, the labeling and the
code, and it says nothing about how tight the inequality is. Outputs outside the inequality behave as its scope
predicts. The rules with row coins still satisfy it, with a smallest slack of \floorRowMinSlack, since they still leave
more than one percent of their windows forbidden before mending. The rules that read the phase of their coordinates
fall below it on all \floorPhaseNegative\ of their rows, with no defect and nearly exact statistics, which is the
coordinate escape of Section~\ref{sec:theory}. Mended outputs are listed without a slack, since the mender works on an
absolute lattice.

\subsection{The independent-coin rule, example by example}
\label{supp:persample}

\begin{table*}[p]\centering\tiny
\caption{The independent-coin rule, example by example, one rule per example run for \mainR\
rounds, \nSweepSeeds\ evaluation seed each. Columns: the number $|K|$ of distinct $N\times N$ windows and of colors
$C$; the forbidden-window rate at $1024^2$ after 16, \mainR\ and 64 rounds, and after four repair rounds at 64; the
pattern KL of the rule's $1024^2$ output and of our Wave Function Collapse implementation's $64^2$ output, for contrast
only (the size-matched comparison is Table~\ref{tab:wfcstats}); whether chunked and one-shot generation agree cell for
cell (seam) and whether redrawing the coins of a $16^2$ region changed cells only within the rule's radius (coin edit);
and the training time in minutes on one GPU. Rates are measured on one $1024^2$ canvas generated in $256^2$ chunks. A
dash marks a canvas our Wave Function Collapse implementation never produced. Chess, Stripes, Col3 (a random proper
3-coloring of a torus) and proper3 (all proper 3-colored windows) are controls.}
\label{tab:persample}
\input{tables/per_sample_v12}\end{table*}

\begin{figure}[!htbp]\centering\includegraphics[width=0.6\linewidth]{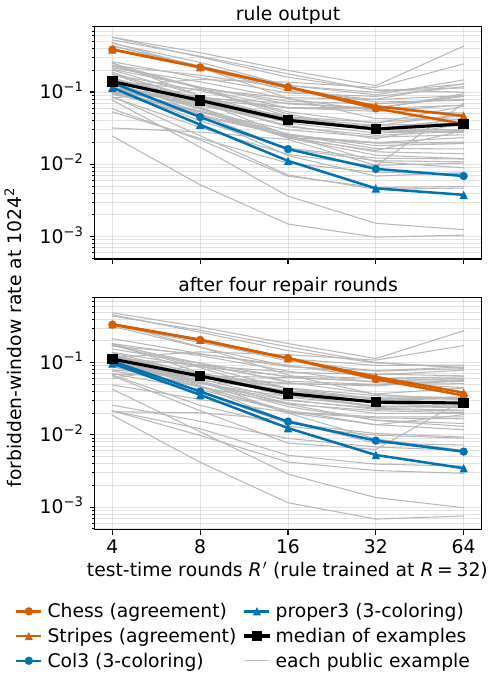}
\caption{The independent-coin rule's forbidden-window rate at $1024^2$ against the number of test-time rounds, on the
public examples (gray, with the median in black) and on the controls the legend names: the raw output above, and the
output after repair below. Decay alone is not evidence of a class, since Chess and Stripes decay too, at about the rate
$1/r$ of Proposition~\ref{prop:classes}(b).}
\label{fig:curves}\end{figure}

\begin{figure}[!htbp]\centering\includegraphics[width=0.6\linewidth]{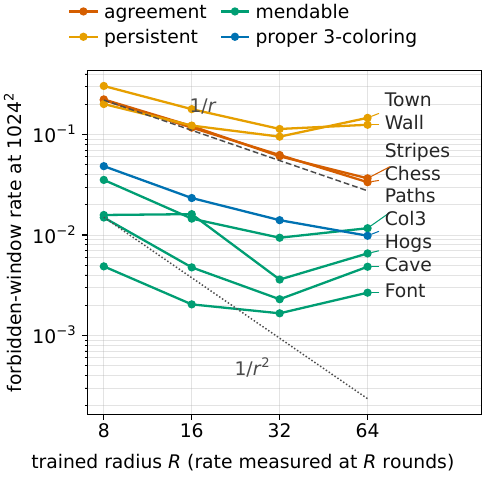}
\caption{The radius varied alone, from 8 to 64 rounds at one fixed training configuration of the independent-coin
rule: the forbidden-window rate at the trained radius against the radius, log-log, with guide lines at $1/r$ and
$1/r^2$. The groups separate by level; their slopes overlap.}
\label{fig:radius}\end{figure}

Before mending, the rule on independent coins leaves forbidden windows on most examples, fewer with more rounds within
the trained range (Table~\ref{tab:persample}). At \mainR\ rounds the median forbidden-window rate is \medVsixteenRaw\ raw and \medVsixteenRep\ after four
repair rounds, with \nVsixteenLeFive\ of the \nPoneOf\ public examples below 5\% and \nVsixteenLeOne\ below 1\%, from
\bestVsixteen\ on \bestSample\ to \worstVsixteen\ on \worstSample. The rate falls steeply over the trained range of
round counts (Figure~\ref{fig:curves}) and is monotone out to 64 rounds on \nMonotone\ of \nRules\ rules, controls
included; beyond the trained range some curves flatten or rise. Chess and Stripes decay at about the rate $1/r$ of
Proposition~\ref{prop:classes}(b), with Stripes at \ctrlStripesVsixteen\ at \mainR\ rounds and \ctrlStripesVsixtyfour\
at 64; the random 3-coloring Col3 is at \ctrlColthreeVsixteen\ and \ctrlColthreeVsixtyfour. Chunked and one-shot
generation agree cell for cell on every rule, and a coin edit stays within the rule's radius on every rule.

Varying the radius alone (Figure~\ref{fig:radius}), Chess and Stripes fall with log-log slopes of $\slopeChess$ and
$\slopeStripes$ over radii 8 to 32, about the rate $1/r$. Four mendable examples, Cave, Hogs, Paths and Font, fall at
slopes from $\slopeFont$ to $\slopeCave$, which overlap the controls, but at a lower level at every radius up to 32. Wall
and Town, whose residual persists under mending, fall more slowly ($\slopeWall$ and $\slopeTown$), and Col3 falls at
$\slopeColthree$. At radius 64 all but the controls rise again, which reflects the fixed training budget, so over 8 to
64 the slopes of the four mending examples range from $\slopeSixtyfourFont$ to $\slopeSixtyfourHogs$. Slopes therefore do
not separate the classes at these radii.

\subsection{The mending map at radius 4}
\label{supp:mendall}

Table~\ref{tab:mendall} gives the mending map of the independent-coin rule on every example, mended as a whole within
radius 4.

\begin{table}[p]\centering\tiny
\caption{The mending map of the independent-coin rule, run for \mainR\ rounds with four
repair rounds, one $256^2$ canvas per example, mended as a whole within radius 4 with blocks
of 8: its forbidden-window rate after repair and after mending; the share of cells changed; and the share
of mended defects found within radius 2. Sorted by the mended rate.}
\label{tab:mendall}
\setlength{\tabcolsep}{4pt}%
\input{tables/mendall}\end{table}

\subsection{Forced edits}
\label{app:forced}

\begin{table*}[!htbp]\centering\scriptsize
\caption{Forced edits with the independent-coin rule (\mainR\ rounds, four repair rounds): a
$192^2$ canvas per trial, mended within radius 4 with blocks of 8, then mending within radius 6 around a pinned
$6\times6$ block of the example; medians over the trials whose canvas had no forbidden window before the pin.
Columns: pins tried, canvases free of forbidden windows before the pin, pins mended without a net defect, defects a pin creates, the greatest distance of a
changed cell from the block, cells changed, and the residual defects of the canvas before the pin.}
\label{tab:forced}
\input{tables/forced}\end{table*}

A designer can pin content into a generated world wherever the canvas around it is valid. A coin edit is contained by
construction, but a pinned block is a
mending problem: the block is frozen at a random position near the center of a mended $192^2$ canvas, and the defects
along its boundary are mended around it within radius 6. We tested this with the independent-coin rule. Where the canvas had no forbidden window before the pin, which holds on Cave, Font, Hogs, Sand, Maze and
Forest, the pin
is absorbed without a net defect on nearly every trial, within a few cells of the block (Table~\ref{tab:forced}). On
Angular and Paths the canvas is rarely free of forbidden windows before the pin, so the changed cells cannot be charged to the pin alone. The
resistant examples of Table~\ref{tab:mendmapeight}, whose canvases keep forbidden windows after mending, were not tested with this rule. The forced edits use the independent-coin rule of the corpus evaluation, run for \mainR\
rounds.

\subsection{Window statistics against Wave Function Collapse}
\label{app:stats}
\label{sec:stats}

\begin{table*}[p]\centering\tiny
\caption{Size-matched pattern KL to the example of the raw canvas (before repair and mending): the independent-coin
rule (``indep.''), the octave rule, and our Wave Function Collapse implementation (``WFC''), at $64^2$ (eight seeds), $128^2$
and $256^2$ (four seeds); means over the canvases each method completed. Both rules are run for 32 rounds. All
methods are scored at matched canvas sizes by the same plug-in estimator over the allowed $N\times N$ windows of the
canvas, taken without wrap-around. The restart solver of Wave
Function Collapse starts from scratch on a contradiction; a dash marks a canvas it never produced. At $256^2$, where
the restart solver produced no canvas or, on 3Bricks, Cat and Lake, was not run, a double dagger marks the same solver
with bounded chronological backtracking, which completed \nBtSolvedAll\ of these \nBtOfAll\ examples. The
last row counts the examples on which each rule's divergence is at or below Wave Function Collapse's, of those with a
Wave Function Collapse canvas. Chess and Stripes are left out.}
\label{tab:wfcstats}
\input{tables/wfcstats_v13}\end{table*}

Before mending, both rules match the example's window statistics at least as well as Wave Function Collapse on most
examples, the independent-coin rule more often than the octave rule. The comparison is made on canvases of the same
size, scored by the same estimator, since its bias depends on the size (Table~\ref{tab:wfcstats}). The
independent-coin rule's divergence is at or below Wave Function Collapse's on \wfcCountFlat\ of the \msWfcOf\ examples Wave Function Collapse completes at $128^2$, and the octave
rule's on
\msWfcCount; the counts are \nPtwoMSmall\ and \msWfcCountSmall\ of \msWfcCountSmallOf\ at $64^2$, and \nPtwoMExt\ and
\msWfcCountExt\ of \msWfcCountExtOf\ at $256^2$. At the median the octave rule's raw divergence is \msRawKLRatioSmall,
\msRawKLRatioMid\ and \msRawKLRatioExt\ times the independent-coin rule's at the three sizes. Mending changes the
comparison, since its completions are draws from Wave Function Collapse's own propagation: after mending within radius
8, the octave rule's median divergence on the examples the independent-coin rule makes valid is \msKLMendedMedian, against
\flatKLMendedMedian.

\subsection{Variants of the objective}
\label{app:persample}

\begin{table*}[!htbp]\centering\scriptsize
\caption{Objective variants on five examples, one seed each, for the rule on independent coins: the median over
examples of the variant's forbidden-window rate at 16, 32 and 64 test-time rounds, divided by the earlier configuration's
rate at its trained 16 rounds; how many curves fall monotonically from 8 to 64 rounds; and collapses to a constant
pattern. The configuration the paper uses is marked.}
\label{tab:variants}
\input{tables/variants_v12}\end{table*}

Each part of the objective removes a failure we saw without it. An earlier run used the soft objective alone, and at radius 32 most of its rules
collapsed to a constant pattern with no forbidden window, with a large divergence to the example and no cell changed by a coin edit. The
straight-through objective with the color-marginal term and training at every round removed the collapse at 16 rounds
on every example (\nCollapsedFirst\ of \nRulesFirst\ rules collapsed), with a median rate of \medFirstSixteen\ at 16
rounds. Trained at a fixed 32 rounds, the same objective still collapsed on \nRthirtytwoCollapsedFirst\ of
\nRthirtytwoLegsFirst\ rules: their output ignored the coins or sat far from the example's statistics, and on several
of them it was a constant coloring with no forbidden window. Drawing the round count afresh at every update removes this.
Table~\ref{tab:variants} reports the variants we ran. More optimization moves the rate furthest of any single change, a
harder temperature helps less, and a score-function reward on the hard windows makes the curves fall monotonically from
8 to 64 rounds, beyond the trained horizon. The final configuration combines longer training, the hard-window settings and round counts
drawn afresh, and roughly halves the residual again. The octave rule uses the same objective, with the variation term added, and the same
schedule (Appendix~\ref{app:params}).

%% file: tables/mendmap_both_v13.tex
\begin{tabular}{@{}llrrrrrrr@{}}\toprule
 & & \multicolumn{2}{c}{independent coins} & \multicolumn{3}{c}{the octave rule} & \multicolumn{2}{c}{KL after mending}\\
\cmidrule(lr){3-4}\cmidrule(lr){5-7}\cmidrule(lr){8-9}
example & class & mean & worst seed & mean & worst seed & seeds valid & indep. & octave\\\midrule
Cave & mended & 0.00000 & 0.00000 & 0.00000 & 0.00000 & 3/3 & 0.026 & 0.014\\
Dungeon & mended & 0.00000 & 0.00000 & 0.00005 & 0.00005 & 3/3 & 0.033 & 0.032\\
Font & mended & 0.00000 & 0.00000 & 0.00000 & 0.00000 & 3/3 & 0.115 & 0.083\\
Forest & mended & 0.00000 & 0.00000 & 0.00000 & 0.00000 & 3/3 & 0.071 & 0.091\\
Hogs & mended & 0.00000 & 0.00000 & 0.00000 & 0.00000 & 3/3 & 0.049 & 0.064\\
Knot & mended & 0.00000 & 0.00000 & 0.00000 & 0.00000 & 3/3 & 0.042 & 0.160\\
LessRooms & mended & 0.00000 & 0.00000 & 0.00002 & 0.00005 & 3/3 & 0.220 & 0.187\\
Maze & mended & 0.00000 & 0.00000 & 0.00000 & 0.00000 & 3/3 & 0.156 & 0.156\\
NotKnot & mended & 0.00000 & 0.00000 & 0.00000 & 0.00000 & 3/3 & 0.415 & 0.298\\
Office & mended & 0.00000 & 0.00000 & 0.00000 & 0.00000 & 3/3 & 0.064 & 0.060\\
Office2 & mended & 0.00000 & 0.00000 & 0.00005 & 0.00012 & 3/3 & 0.239 & 0.136\\
Paths & mended & 0.00000 & 0.00000 & 0.00001 & 0.00002 & 3/3 & 0.037 & 0.052\\
Rooms & mended & 0.00000 & 0.00000 & 0.00000 & 0.00000 & 3/3 & 0.190 & 0.087\\
Rule126 & mended & 0.00000 & 0.00000 & 0.00000 & 0.00000 & 3/3 & 0.169 & 0.186\\
Sand & mended & 0.00000 & 0.00000 & 0.00000 & 0.00000 & 3/3 & 0.151 & 0.086\\
Spirals & mended & 0.00000 & 0.00000 & 0.00000 & 0.00000 & 3/3 & 0.235 & 0.254\\
Village & mended & 0.00000 & 0.00000 & 0.00006 & 0.00019 & 3/3 & 0.336 & 0.426\\
Wrinkles & mended & 0.00000 & 0.00000 & 0.00003 & 0.00008 & 3/3 & 0.117 & 0.112\\
Angular & mended & 0.00002 & 0.00006 & 0.00000 & 0.00000 & 3/3 & 0.423 & 0.087\\
MagicOffice & mended & 0.00005 & 0.00012 & 0.00023 & 0.00036 & 3/3 & 0.630 & 0.369\\
Cats & mended & 0.00033 & 0.00057 & 0.00351 & 0.00504 & 0/3 & 0.365 & 0.249\\
SimpleKnot & partial & 0.00110 & 0.00121 & 0.00133 & 0.00141 & 0/3 & 0.093 & 0.074\\
Lines & partial & 0.00133 & 0.00144 & 0.00000 & 0.00000 & 3/3 & 0.031 & 0.122\\
City & partial & 0.00159 & 0.00251 & 0.00000 & 0.00000 & 3/3 & 0.002 & 0.011\\
TrickKnot & partial & 0.00205 & 0.00291 & 0.00104 & 0.00118 & 2/3 & 0.117 & 0.062\\
Water & partial & 0.00258 & 0.00259 & 0.00293 & 0.00301 & 0/3 & 0.004 & 0.017\\
Qud & partial & 0.00508 & 0.00649 & 0.00345 & 0.00601 & 1/3 & 0.227 & 0.156\\
Mazelike & partial & 0.00556 & 0.00693 & 0.00015 & 0.00026 & 3/3 & 0.041 & 0.086\\
Link & partial & 0.00656 & 0.00727 & 0.00280 & 0.00353 & 0/3 & 0.075 & 0.094\\
Mountains & partial & 0.01024 & 0.01139 & 0.01135 & 0.01175 & 0/3 & 0.010 & 0.011\\
Link2 & partial & 0.01065 & 0.01178 & 0.01469 & 0.01522 & 0/3 & 0.077 & 0.086\\
ScaledMaze & resistant & 0.01259 & 0.01370 & 0.00660 & 0.00704 & 0/3 & 0.003 & 0.006\\
Lake & resistant & 0.01366 & 0.01623 & 0.00612 & 0.00719 & 0/3 & 0.228 & 0.226\\
Sewers & resistant & 0.01544 & 0.02344 & 0.00014 & 0.00022 & 3/3 & 0.261 & 0.382\\
Skew1 & resistant & 0.01804 & 0.02072 & 0.00755 & 0.01017 & 0/3 & 0.043 & 0.117\\
Skew2 & resistant & 0.01982 & 0.02164 & 0.00674 & 0.00995 & 0/3 & 0.033 & 0.093\\
Nested & resistant & 0.02148 & 0.02409 & 0.01281 & 0.01411 & 0/3 & 0.084 & 0.067\\
MoreFlowers & resistant$^{*}$ & 0.02444 & 0.02672 & 0.00027 & 0.00082 & 3/3 & 0.252 & 0.163\\
Platformer & resistant$^{*}$ & 0.02632 & 0.02799 & 0.00024 & 0.00047 & 3/3 & 0.053 & 0.141\\
Flowers & resistant$^{*}$ & 0.03657 & 0.03844 & 0.00000 & 0.00000 & 3/3 & 0.158 & 0.187\\
Cat & resistant & 0.03849 & 0.04582 & 0.00874 & 0.01085 & 0/3 & 0.191 & 0.290\\
SimpleWall & resistant$^{*}$ & 0.04574 & 0.04693 & 0.00055 & 0.00118 & 2/3 & 0.016 & 0.092\\
ColoredCity & resistant & 0.05143 & 0.05479 & 0.01033 & 0.01256 & 0/3 & 0.086 & 0.103\\
3Bricks & resistant$^{*}$ & 0.05485 & 0.05558 & 0.00295 & 0.00431 & 0/3 & 0.955 & 0.962\\
Skyline2 & resistant$^{*}$ & 0.06115 & 0.06253 & 0.00261 & 0.00369 & 0/3 & 0.111 & 0.102\\
Skyline & resistant$^{*}$ & 0.06585 & 0.06963 & 0.01101 & 0.01327 & 0/3 & 0.236 & 0.348\\
Town & resistant$^{*}$ & 0.06636 & 0.06859 & 0.00128 & 0.00307 & 2/3 & 0.136 & 0.516\\
SmileCity & resistant$^{*}$ & 0.09111 & 0.09151 & 0.02968 & 0.03810 & 0/3 & 0.043 & 0.065\\
Wall & resistant$^{*}$ & 0.09448 & 0.09843 & 0.02620 & 0.04495 & 0/3 & 0.048 & 0.152\\
SimpleMaze & resistant$^{*}$ & 0.10059 & 0.10191 & 0.09593 & 0.10734 & 0/3 & 0.000 & 0.002\\
Fabric & resistant$^{*}$ & 0.10856 & 0.11552 & 0.05429 & 0.06209 & 0/3 & 0.053 & 0.039\\
RedMaze & resistant$^{*}$ & 0.11339 & 0.11393 & 0.07554 & 0.08131 & 0/3 & 0.000 & 0.001\\
\midrule\multicolumn{9}{@{}l}{\textit{controls}, counted in no total}\\
Chess & -- & 0.05521 & 0.06003 & 0.06144 & 0.06958 & 0/3 & 0.000 & 0.000\\
Stripes & -- & 0.05876 & 0.05940 & 0.03446 & 0.04698 & 0/3 & 0.000 & 0.000\\
\bottomrule\end{tabular}

%% file: tables/widemend_v13.tex
\begin{tabular}{@{}llrrrrrrr@{}}\toprule
 & & \multicolumn{3}{c}{mean forbidden windows} & \multicolumn{3}{c}{seeds at most $10^{-3}$} & changed\\
\cmidrule(lr){3-5}\cmidrule(lr){6-8}
example & class & $\rho\le8$ & $\rho\le12$ & $\rho\le16$ & 8 & 12 & 16 & at 16\\\midrule
Paths & mended & 0.00001 & 0.00000 & 0.00000 & 3 & 3 & 3 & 9\%\\
LessRooms & mended & 0.00002 & 0.00000 & 0.00000 & 3 & 3 & 3 & 32\%\\
Wrinkles & mended & 0.00003 & 0.00000 & 0.00000 & 3 & 3 & 3 & 18\%\\
Dungeon & mended & 0.00005 & 0.00000 & 0.00000 & 3 & 3 & 3 & 19\%\\
Office2 & mended & 0.00005 & 0.00000 & 0.00000 & 3 & 3 & 3 & 8\%\\
Village & mended & 0.00006 & 0.00000 & 0.00031 & 3 & 3 & 3 & 20\%\\
MagicOffice & mended & 0.00023 & 0.00014 & 0.00002 & 3 & 3 & 3 & 14\%\\
Cats & mended & 0.00351 & 0.00302 & 0.00654 & 0 & 0 & 0 & 20\%\\
Mazelike & partial & 0.00015 & 0.00000 & 0.00002 & 3 & 3 & 3 & 27\%\\
TrickKnot & partial & 0.00104 & 0.00035 & 0.00032 & 2 & 3 & 3 & 36\%\\
SimpleKnot & partial & 0.00133 & 0.00058 & 0.00032 & 0 & 3 & 3 & 23\%\\
Link & partial & 0.00280 & 0.00139 & 0.00119 & 0 & 1 & 1 & 24\%\\
Water & partial & 0.00293 & 0.00151 & 0.00053 & 0 & 0 & 3 & 16\%\\
Qud & partial & 0.00345 & 0.00015 & 0.00001 & 1 & 3 & 3 & 46\%\\
Mountains & partial & 0.01135 & 0.01060 & 0.01026 & 0 & 0 & 0 & 7\%\\
Link2 & partial & 0.01469 & 0.01679 & 0.01715 & 0 & 0 & 0 & 16\%\\
Sewers & resistant & 0.00014 & 0.00004 & 0.00000 & 3 & 3 & 3 & 24\%\\
Platformer & resistant$^{*}$ & 0.00024 & 0.00018 & 0.00003 & 3 & 3 & 3 & 4\%\\
MoreFlowers & resistant$^{*}$ & 0.00027 & 0.00022 & 0.00000 & 3 & 3 & 3 & 15\%\\
SimpleWall & resistant$^{*}$ & 0.00055 & 0.00003 & 0.00000 & 2 & 3 & 3 & 4\%\\
Town & resistant$^{*}$ & 0.00128 & 0.00090 & 0.00125 & 2 & 2 & 2 & 16\%\\
Skyline2 & resistant$^{*}$ & 0.00261 & 0.00280 & 0.00173 & 0 & 0 & 0 & 16\%\\
3Bricks & resistant$^{*}$ & 0.00295 & 0.00172 & 0.00164 & 0 & 0 & 0 & 16\%\\
Lake & resistant & 0.00612 & 0.00402 & 0.00741 & 0 & 0 & 0 & 37\%\\
ScaledMaze & resistant & 0.00660 & 0.00618 & 0.00557 & 0 & 0 & 0 & 22\%\\
Skew2 & resistant & 0.00674 & 0.00335 & 0.00369 & 0 & 0 & 0 & 23\%\\
Skew1 & resistant & 0.00755 & 0.00479 & 0.00626 & 0 & 0 & 0 & 18\%\\
Cat & resistant & 0.00874 & 0.00627 & 0.00525 & 0 & 1 & 0 & 36\%\\
ColoredCity & resistant & 0.01033 & 0.00473 & 0.00535 & 0 & 0 & 0 & 33\%\\
Skyline & resistant$^{*}$ & 0.01101 & 0.00823 & 0.00622 & 0 & 0 & 0 & 29\%\\
Nested & resistant & 0.01281 & 0.01014 & 0.00505 & 0 & 0 & 0 & 20\%\\
Wall & resistant$^{*}$ & 0.02620 & 0.03219 & 0.03738 & 0 & 0 & 0 & 13\%\\
SmileCity & resistant$^{*}$ & 0.02968 & 0.03043 & 0.03205 & 0 & 0 & 0 & 15\%\\
Fabric & resistant$^{*}$ & 0.05429 & 0.05435 & 0.05443 & 0 & 0 & 0 & 0\%\\
RedMaze & resistant$^{*}$ & 0.07554 & 0.07562 & 0.07565 & 0 & 0 & 0 & 0\%\\
SimpleMaze & resistant$^{*}$ & 0.09593 & 0.09593 & 0.09593 & 0 & 0 & 0 & 0\%\\
\midrule\multicolumn{9}{@{}l}{\textit{controls}}\\
Stripes & -- & 0.03446 & 0.03382 & 0.03544 & 0 & 0 & 0 & 0\%\\
Chess & -- & 0.06144 & 0.05815 & 0.05550 & 0 & 0 & 0 & 2\%\\
\bottomrule\end{tabular}

%% file: tables/robust_v13.tex
\begin{tabular}{@{}llrrr@{}}\toprule
example & class & seeds $\le10^{-3}$ & mean & worst\\\midrule
Angular$^{\dagger}$ & mended & 8 & 0.00000 & 0.00000\\
Cave$^{\dagger}$ & mended & 8 & 0.00000 & 0.00000\\
City$^{\dagger}$ & partial & 8 & 0.00000 & 0.00000\\
Font$^{\dagger}$ & mended & 8 & 0.00000 & 0.00000\\
Forest$^{\dagger}$ & mended & 8 & 0.00000 & 0.00000\\
Hogs$^{\dagger}$ & mended & 8 & 0.00000 & 0.00000\\
Knot$^{\dagger}$ & mended & 8 & 0.00000 & 0.00000\\
Lines$^{\dagger}$ & partial & 8 & 0.00000 & 0.00000\\
Maze$^{\dagger}$ & mended & 8 & 0.00000 & 0.00000\\
NotKnot$^{\dagger}$ & mended & 8 & 0.00000 & 0.00000\\
Office$^{\dagger}$ & mended & 8 & 0.00000 & 0.00000\\
Rule126$^{\dagger}$ & mended & 8 & 0.00000 & 0.00000\\
Sand$^{\dagger}$ & mended & 8 & 0.00000 & 0.00000\\
Spirals$^{\dagger}$ & mended & 8 & 0.00000 & 0.00000\\
Paths$^{\dagger}$ & mended & 8 & 0.00000 & 0.00002\\
Rooms$^{\dagger}$ & mended & 8 & 0.00000 & 0.00002\\
LessRooms$^{\dagger}$ & mended & 8 & 0.00001 & 0.00005\\
Wrinkles$^{\dagger}$ & mended & 8 & 0.00001 & 0.00008\\
Village$^{\dagger}$ & mended & 8 & 0.00002 & 0.00019\\
Office2$^{\dagger}$ & mended & 8 & 0.00003 & 0.00012\\
Dungeon$^{\dagger}$ & mended & 8 & 0.00003 & 0.00011\\
Mazelike$^{\dagger}$ & partial & 8 & 0.00009 & 0.00026\\
Sewers$^{\dagger}$ & resistant & 8 & 0.00014 & 0.00039\\
MagicOffice$^{\dagger}$ & mended & 8 & 0.00018 & 0.00036\\
Flowers$^{\dagger}$ & resistant$^{*}$ & 7 & 0.00018 & 0.00144\\
Platformer$^{\dagger}$ & resistant$^{*}$ & 8 & 0.00026 & 0.00047\\
MoreFlowers$^{\dagger}$ & resistant$^{*}$ & 8 & 0.00028 & 0.00082\\
SimpleWall & resistant$^{*}$ & 6 & 0.00100 & 0.00512\\
TrickKnot & partial & 5 & 0.00101 & 0.00118\\
SimpleKnot & partial & 0 & 0.00143 & 0.00171\\
Town & resistant$^{*}$ & 5 & 0.00156 & 0.00611\\
Link & partial & 0 & 0.00262 & 0.00353\\
Water & partial & 0 & 0.00291 & 0.00360\\
Cats & mended & 0 & 0.00343 & 0.00591\\
Skyline2 & resistant$^{*}$ & 0 & 0.00377 & 0.00744\\
3Bricks & resistant$^{*}$ & 0 & 0.00396 & 0.00916\\
Qud & partial & 1 & 0.00470 & 0.00908\\
Skew2 & resistant & 0 & 0.00519 & 0.00995\\
Lake & resistant & 0 & 0.00591 & 0.00961\\
ScaledMaze & resistant & 0 & 0.00716 & 0.00848\\
Skew1 & resistant & 0 & 0.00780 & 0.01017\\
Skyline & resistant$^{*}$ & 0 & 0.00983 & 0.01327\\
ColoredCity & resistant & 0 & 0.01043 & 0.01507\\
Nested & resistant & 0 & 0.01058 & 0.01411\\
Mountains & partial & 0 & 0.01074 & 0.01175\\
Cat & resistant & 0 & 0.01092 & 0.01680\\
Link2 & partial & 0 & 0.01551 & 0.01905\\
Wall & resistant$^{*}$ & 0 & 0.02099 & 0.04495\\
SmileCity & resistant$^{*}$ & 0 & 0.02732 & 0.03810\\
Fabric & resistant$^{*}$ & 0 & 0.05813 & 0.07592\\
RedMaze & resistant$^{*}$ & 0 & 0.07961 & 0.09892\\
SimpleMaze & resistant$^{*}$ & 0 & 0.09124 & 0.10734\\
\midrule\multicolumn{5}{@{}l}{\textit{controls}}\\
Chess & -- & 0 & 0.06118 & 0.06958\\
Stripes & -- & 0 & 0.03475 & 0.04752\\
\bottomrule\end{tabular}

%% file: tables/replay_valid_v12.tex
\begin{tabular}{@{}lrrrrrrrrrrr@{}}\toprule
 & \multicolumn{2}{c}{total variation} & \multicolumn{3}{c}{replay agreement} & \multicolumn{2}{c}{normalized} & 16$\times$16 reuse & \multicolumn{2}{c}{aligned variation} & redraw in\\
\cmidrule(lr){2-3}\cmidrule(lr){4-6}\cmidrule(lr){7-8}\cmidrule(lr){10-11}
example & ind.\ coins & block & i.i.d. & ind.\ coins & block & ind.\ coins & block & ind.\ coins & ind.\ coins & block & valid windows\\\midrule
Angular & 0.40 & 0.22 & 0.36 & 0.40 & 0.45 & 0.05 & 0.13 & 0.00 & 0.50 & 0.54 & 100.0\%\\
Cats & 0.38 & 0.03 & 0.54 & 0.61 & 0.73 & 0.15 & 0.41 & 0.00 & 0.24 & 0.32 & 100.0\%\\
Cave & 0.10 & 0.23 & 0.51 & 0.52 & 0.53 & 0.02 & 0.05 & 0.00 & 0.46 & 0.48 & 100.0\%\\
Dungeon & 0.09 & 0.19 & 0.63 & 0.65 & 0.62 & 0.06 & $-$0.02 & 0.00 & 0.32 & 0.41 & 100.0\%\\
Font & 0.06 & 0.45 & 0.62 & 0.63 & 0.58 & 0.03 & $-$0.10 & 0.00 & 0.36 & 0.47 & 100.0\%\\
Forest & 0.14 & 0.15 & 0.66 & 0.68 & 0.70 & 0.06 & 0.13 & 0.00 & 0.29 & 0.33 & 100.0\%\\
Hogs & 0.09 & 0.05 & 0.68 & 0.69 & 0.68 & 0.04 & 0.01 & 0.00 & 0.30 & 0.32 & 100.0\%\\
Knot & 0.12 & 0.22 & 0.58 & 0.59 & 0.59 & 0.04 & 0.04 & 0.00 & 0.38 & 0.43 & 100.0\%\\
LessRooms & 0.26 & 0.19 & 0.52 & 0.54 & 0.56 & 0.04 & 0.08 & 0.00 & 0.40 & 0.44 & 100.0\%\\
MagicOffice & 0.20 & 0.23 & 0.63 & 0.67 & 0.64 & 0.09 & 0.03 & 0.00 & 0.28 & 0.35 & 99.9\%\\
Maze & 0.19 & 0.20 & 0.50 & 0.51 & 0.56 & 0.01 & 0.10 & 0.00 & 0.48 & 0.47 & 100.0\%\\
NotKnot & 0.38 & 0.41 & 0.60 & 0.65 & 0.66 & 0.13 & 0.15 & 0.00 & 0.26 & 0.26 & 100.0\%\\
Office & 0.13 & 0.17 & 0.57 & 0.57 & 0.65 & 0.00 & 0.20 & 0.00 & 0.42 & 0.38 & 100.0\%\\
Office2 & 0.27 & 0.23 & 0.52 & 0.54 & 0.66 & 0.04 & 0.30 & 0.00 & 0.44 & 0.36 & 99.9\%\\
Paths & 0.09 & 0.25 & 0.65 & 0.66 & 0.61 & 0.04 & $-$0.10 & 0.00 & 0.32 & 0.41 & 100.0\%\\
Rooms & 0.25 & 0.30 & 0.51 & 0.51 & 0.53 & 0.01 & 0.06 & 0.00 & 0.47 & 0.45 & 100.0\%\\
Rule126 & 0.20 & 0.34 & 0.51 & 0.51 & 0.55 & 0.00 & 0.09 & 0.00 & 0.48 & 0.48 & 100.0\%\\
Sand & 0.15 & 0.17 & 0.66 & 0.68 & 0.68 & 0.06 & 0.06 & 0.00 & 0.29 & 0.34 & 100.0\%\\
Spirals & 0.27 & 0.20 & 0.53 & 0.55 & 0.56 & 0.04 & 0.06 & 0.00 & 0.41 & 0.45 & 100.0\%\\
Village & 0.35 & 0.05 & 0.57 & 0.67 & 0.69 & 0.23 & 0.27 & 0.00 & 0.19 & 0.37 & 100.0\%\\
Wrinkles & 0.21 & 0.19 & 0.66 & 0.69 & 0.69 & 0.08 & 0.08 & 0.00 & 0.27 & 0.27 & 100.0\%\\
\midrule median & 0.20 & 0.20 & 0.57 & 0.61 & 0.62 & 0.04 & 0.08 & 0.00 & 0.36 & 0.41 & --\\
\bottomrule\end{tabular}

%% file: tables/replay_ra_v12.tex
\begin{tabular}{@{}lrrrrrrr@{}}\toprule
 & \multicolumn{2}{c}{aligned variation} & \multicolumn{2}{c}{normalized agreement} & \multicolumn{2}{c}{total variation} & cost per cell\\
\cmidrule(lr){2-3}\cmidrule(lr){4-5}\cmidrule(lr){6-7}
example & ind.\ coins & blocks & ind.\ coins & blocks & ind.\ coins & blocks & blocks / ind.\ coins\\\midrule
Angular & 0.50 & 0.52 & 0.05 & 0.17 & 0.40 & 0.21 & 10\\
Cats & 0.24 & 0.21 & 0.15 & 0.59 & 0.38 & 0.04 & 10\\
Cave & 0.46 & 0.48 & 0.02 & 0.06 & 0.10 & 0.23 & 17\\
Dungeon & 0.32 & 0.40 & 0.06 & 0.07 & 0.09 & 0.18 & 10\\
Font & 0.36 & 0.47 & 0.03 & $-$0.16 & 0.06 & 0.45 & 56\\
Forest & 0.29 & 0.33 & 0.06 & 0.02 & 0.14 & 0.16 & 19\\
Hogs & 0.30 & 0.31 & 0.04 & 0.06 & 0.09 & 0.05 & 9\\
Knot & 0.38 & 0.42 & 0.04 & 0.10 & 0.12 & 0.21 & 11\\
LessRooms & 0.40 & 0.44 & 0.04 & 0.09 & 0.26 & 0.20 & 18\\
MagicOffice & 0.28 & 0.34 & 0.09 & 0.11 & 0.20 & 0.22 & 11\\
Maze & 0.48 & 0.48 & 0.01 & 0.06 & 0.19 & 0.21 & 24\\
NotKnot & 0.26 & 0.26 & 0.13 & 0.17 & 0.38 & 0.40 & 9\\
Office & 0.42 & 0.35 & 0.00 & 0.27 & 0.13 & 0.16 & 13\\
Office2 & 0.44 & 0.37 & 0.04 & 0.27 & 0.27 & 0.24 & 11\\
Paths & 0.32 & 0.41 & 0.04 & $-$0.09 & 0.09 & 0.25 & 12\\
Rooms & 0.47 & 0.45 & 0.01 & 0.12 & 0.25 & 0.28 & 19\\
Rule126 & 0.48 & 0.47 & 0.00 & 0.10 & 0.20 & 0.33 & 11\\
Sand & 0.29 & 0.34 & 0.06 & 0.04 & 0.15 & 0.16 & 26\\
Spirals & 0.41 & 0.44 & 0.04 & 0.08 & 0.27 & 0.21 & 12\\
Village & 0.19 & 0.37 & 0.23 & 0.24 & 0.35 & 0.06 & 13\\
Wrinkles & 0.27 & 0.27 & 0.08 & 0.13 & 0.21 & 0.18 & 6\\
\midrule median & 0.36 & 0.40 & 0.04 & 0.10 & 0.20 & 0.21 & 12\\
\midrule\multicolumn{8}{@{}l}{\emph{resistant certified examples, where our rule leaves defects}}\\
Wall & 0.70 & 0.23 & 0.03 & 0.67 & 0.14 & 0.12 & 5\\
SimpleWall & 0.38 & 0.15 & 0.02 & 0.61 & 0.09 & 0.03 & 2\\
Flowers & 0.38 & 0.29 & 0.09 & 0.43 & 0.22 & 0.13 & 6\\
Skyline & 0.62 & 0.30 & 0.04 & 0.63 & 0.30 & 0.13 & 7\\
\bottomrule\end{tabular}

%% file: tables/pilot_ablation_v13.tex
\begin{tabular}{@{}lrrrrrrrrrr@{}}\toprule
 & \multicolumn{5}{c}{forbidden windows after mending} & \multicolumn{5}{c}{KL after mending}\\
\cmidrule(lr){2-6}\cmidrule(lr){7-11}
example & indep. & row & no term & rule & strong term & indep. & row & no term & rule & strong term\\\midrule
SimpleWall & 0.0457 & 0.0005 & 0.0004 & 0.0005 & 0.0007 & 0.016 & 0.053 & 0.110 & 0.092 & 0.073\\
Wall & 0.0945 & 0.0208 & 0.0119 & 0.0262 & 0.1015 & 0.048 & 0.138 & 0.142 & 0.152 & 0.061\\
Skyline & 0.0658 & 0.0053 & 0.0159 & 0.0110 & 0.0142 & 0.236 & 0.576 & 0.293 & 0.348 & 0.338\\
Cave & 0.0000 & -- & 0.0000 & 0.0000 & 0.0000 & 0.026 & -- & 0.028 & 0.014 & 0.020\\
Rooms & 0.0000 & -- & 0.0000 & 0.0000 & 0.0000 & 0.190 & -- & 0.117 & 0.087 & 0.090\\
Cat & 0.0385 & -- & 0.0096 & 0.0087 & 0.0109 & 0.191 & -- & 0.278 & 0.290 & 0.228\\
ColoredCity & 0.0514 & -- & 0.0084 & 0.0103 & 0.0000 & 0.086 & -- & 0.157 & 0.103 & 0.358\\
Lake & 0.0137 & -- & 0.0081 & 0.0061 & 0.0089 & 0.228 & -- & 0.236 & 0.226 & 0.224\\
Nested & 0.0215 & -- & 0.0099 & 0.0128 & 0.0181 & 0.084 & -- & 0.115 & 0.067 & 0.086\\
ScaledMaze & 0.0126 & -- & 0.0050 & 0.0066 & 0.0099 & 0.003 & -- & 0.005 & 0.006 & 0.004\\
Sewers & 0.0154 & -- & 0.0002 & 0.0001 & 0.0073 & 0.261 & -- & 0.591 & 0.382 & 0.237\\
Skew1 & 0.0180 & -- & 0.0035 & 0.0076 & 0.0076 & 0.043 & -- & 0.158 & 0.117 & 0.127\\
Skew2 & 0.0198 & -- & 0.0008 & 0.0067 & 0.0015 & 0.033 & -- & 0.133 & 0.093 & 0.138\\
\bottomrule\end{tabular}

%% file: tables/sharing306_v12.tex
\begin{tabular}{@{}llrrrrrr@{}}\toprule
 & & \multicolumn{2}{c}{mended residual} & \multicolumn{2}{c}{mended KL} & \multicolumn{2}{c}{cells changed by a coin edit}\\
\cmidrule(lr){3-4}\cmidrule(lr){5-6}\cmidrule(lr){7-8}
example & extra input & ind.\ coins & with input & ind.\ coins & with input & ind.\ coins & with input\\\midrule
SimpleWall & row coins & 0.0457 & 0.00047 & 0.016 & 0.053 & 229 & 78\\
Skyline & row coins & 0.0658 & 0.0053 & 0.236 & 0.576 & 380 & 42\\
Wall & row coins & 0.0945 & 0.0208 & 0.048 & 0.138 & 401 & 130\\
Chess & phase, $d=2$ & 0.0552 & 0 & 0.000 & 0.000 & 232 & 0\\
Stripes & phase, $d=2$ & 0.0588 & 0 & 0.000 & 0.000 & 660 & 0\\
RedMaze & phase, $d=4$ & 0.1134 & 0 & 0.000 & 0.000 & 1479 & 0\\
SimpleMaze & phase, $d=4$ & 0.1006 & 0 & 0.000 & 0.000 & 69 & 0\\
\midrule
Cave & row coins (harm check) & 0 & 0 & 0.026 & 0.028 & 155 & 189\\
\bottomrule\end{tabular}

%% file: tables/sharing308_v12.tex
\begin{tabular}{@{}llrrrrrrr@{}}\toprule
rule & \begin{tabular}[b]{@{}l@{}}statistical\\weight\end{tabular} & \begin{tabular}[b]{@{}r@{}}fine-\\tuned\end{tabular} & \begin{tabular}[b]{@{}r@{}}raw\\rate\end{tabular} & \begin{tabular}[b]{@{}r@{}}mended\\residual\end{tabular} & \begin{tabular}[b]{@{}r@{}}mended\\KL\end{tabular} & \begin{tabular}[b]{@{}r@{}}$\phi$ switches\\per 100\end{tabular} & \begin{tabular}[b]{@{}r@{}}redraw\\share\end{tabular} & \begin{tabular}[b]{@{}r@{}}tiling\\agreement\end{tabular}\\\midrule
ordinary & 1 & no & 0.0545 & 0.0457 & 0.016 & 3.14 & 0.40 & 0.62\\
ordinary & 1 & yes & 0.0545 & 0.0466 & 0.019 & 3.24 & 0.40 & 0.61\\
ordinary & 3 & yes & 0.0571 & 0.0468 & 0.011 & 3.13 & 0.41 & 0.61\\
ordinary & 10 & yes & 0.0660 & 0.0543 & 0.004 & 3.23 & 0.40 & 0.61\\
\midrule
row coins & 1 & no & 0.0179 & 0.00047 & 0.053 & 0.09 & 0.13 & 0.64\\
row coins & 3 & yes & 0.0184 & 0.00004 & 0.062 & 0.02 & 0.12 & 0.63\\
row coins & 10 & yes & 0.0233 & 0.00051 & 0.044 & 0.08 & 0.13 & 0.65\\
row coins & 30 & yes & 0.0349 & 0.00017 & 0.036 & 0.07 & 0.13 & 0.64\\
\midrule
column coins & 1 & no & 0.0428 & 0.0385 & 0.063 & 2.45 & 0.31 & 0.63\\
\bottomrule\end{tabular}

%% file: tables/completability_v12.tex
\begin{tabular}{@{}lrrrrrr@{}}\toprule
 & \multicolumn{3}{c}{mended residual} & \multicolumn{3}{c}{mended KL}\\
\cmidrule(lr){2-4}\cmidrule(lr){5-7}
example & start & defect reward & mender reward & start & defect reward & mender reward\\\midrule
City & 0.0025 & 0.0007 & 0.0022 & 0.002 & 0.002 & 0.001\\
Qud & 0.0018 & 0.0004 & 0.0003 & 0.266 & 0.424 & 0.347\\
Skew2 & 0.0184 & 0.0185 & 0.0132 & 0.033 & 0.038 & 0.053\\
Sewers & 0.0023 & 0.0528 & 0.0243 & 0.365 & 0.330 & 0.189\\
\bottomrule\end{tabular}

%% file: tables/floorcheck_v12a.tex
\setlength{\tabcolsep}{3pt}%
\begin{tabular}{@{}lllrrrrrl@{}}\toprule
example & generator & output & $r$ & $\varepsilon$ & $\delta$ & $\kappa_f$ & slack & covered\\\midrule
Chess & indep. coins & rule & 32 & 0.0602 & 0.060 & 0.500 & 3.65 & yes\\
Chess & indep. coins & + repair & 44 & 0.0599 & 0.060 & 0.500 & 5.07 & yes\\
Chess & indep. coins$^{c}$ & + mending & -- & 0.0575 & 0.058 & 0.500 & -- & no: absolute lattice\\
Chess & phase input & rule & 32 & 0.0000 & 0.000 & 0.500 & $-$0.50 & no: coordinate input\\
Chess & phase input & + repair & 44 & 0.0000 & 0.000 & 0.500 & $-$0.50 & no: coordinate input\\
\midrule
Stripes & indep. coins & rule & 32 & 0.0623 & 0.062 & 0.500 & 3.80 & yes\\
Stripes & indep. coins & + repair & 44 & 0.0621 & 0.062 & 0.500 & 5.27 & yes\\
Stripes & indep. coins$^{c}$ & + mending & -- & 0.0607 & 0.061 & 0.500 & -- & no: absolute lattice\\
Stripes & phase input & rule & 32 & 0.0000 & 0.000 & 0.500 & $-$0.50 & no: coordinate input\\
Stripes & phase input & + repair & 44 & 0.0000 & 0.000 & 0.500 & $-$0.50 & no: coordinate input\\
\midrule
Wall & indep. coins & rule & 32 & 0.1049 & 0.105--0.189 & 0.667 & 6.57 to 6.66 & yes\\
Wall & indep. coins & + repair & 44 & 0.0973 & 0.097--0.180 & 0.667 & 8.38 to 8.46 & yes\\
Wall & indep. coins$^{c}$ & + mending & -- & 0.0933 & 0.136 & 0.667 & -- & no: absolute lattice\\
Wall & row coins & rule & 32 & 0.0503 & 0.118--0.168 & 0.667 & 2.87 to 2.92 & no: shared coins\\
Wall & row coins & + repair & 44 & 0.0433 & 0.117--0.160 & 0.667 & 3.43 to 3.47 & no: shared coins\\
\midrule
Skyline & indep. coins & rule & 32 & 0.0851 & 0.268--0.353 & 0.107 & 5.95 to 6.03 & yes\\
Skyline & indep. coins & + repair & 44 & 0.0752 & 0.266--0.341 & 0.107 & 7.07 to 7.15 & yes\\
Skyline & indep. coins$^{c}$ & + mending & -- & 0.0653 & 0.302 & 0.107 & -- & no: absolute lattice\\
Skyline & row coins & rule & 32 & 0.0344 & 0.350--0.384 & 0.107 & 2.58 to 2.62 & no: shared coins\\
Skyline & row coins & + repair & 44 & 0.0276 & 0.348--0.376 & 0.107 & 2.78 to 2.81 & no: shared coins\\
\midrule
Skyline2 & indep. coins & rule & 32 & 0.0804 & 0.141--0.221 & 0.065 & 5.54 to 5.62 & yes\\
Skyline2 & indep. coins & + repair & 44 & 0.0688 & 0.155--0.224 & 0.065 & 6.42 to 6.49 & yes\\
Skyline2 & indep. coins$^{c}$ & + mending & -- & 0.0623 & 0.181 & 0.065 & -- & no: absolute lattice\\
\midrule
Flowers & indep. coins & rule & 32 & 0.0579 & 0.189--0.246 & 0.167 & 3.96 to 4.02 & yes\\
Flowers & indep. coins & + repair & 44 & 0.0471 & 0.190--0.237 & 0.167 & 4.36 to 4.40 & yes\\
Flowers & indep. coins$^{c}$ & + mending & -- & 0.0369 & 0.214 & 0.167 & -- & no: absolute lattice\\
\midrule
MoreFlowers & indep. coins & rule & 32 & 0.0613 & 0.234--0.295 & 0.167 & 4.24 to 4.30 & yes\\
MoreFlowers & indep. coins & + repair & 44 & 0.0497 & 0.237--0.287 & 0.167 & 4.64 to 4.69 & yes\\
MoreFlowers & indep. coins$^{c}$ & + mending & -- & 0.0233 & 0.288 & 0.167 & -- & no: absolute lattice\\
\midrule
Platformer & indep. coins & rule & 32 & 0.0373 & 0.124--0.161 & 0.062 & 2.60 to 2.64 & yes\\
Platformer & indep. coins & + repair & 44 & 0.0318 & 0.123--0.154 & 0.062 & 2.98 to 3.02 & yes\\
Platformer & indep. coins$^{c}$ & + mending & -- & 0.0262 & 0.136 & 0.062 & -- & no: absolute lattice\\
\midrule
SimpleMaze & indep. coins & rule & 32 & 0.1087 & 0.109--0.112 & 0.750 & 6.75 to 6.75 & yes\\
SimpleMaze & indep. coins & + repair & 44 & 0.1030 & 0.103--0.106 & 0.750 & 8.82 to 8.83 & yes\\
SimpleMaze & indep. coins$^{c}$ & + mending & -- & 0.1020 & 0.102 & 0.750 & -- & no: absolute lattice\\
SimpleMaze & phase input & rule & 32 & 0.0000 & 0.004 & 0.750 & $-$0.75 & no: coordinate input\\
SimpleMaze & phase input & + repair & 44 & 0.0000 & 0.004 & 0.750 & $-$0.75 & no: coordinate input\\
\midrule
RedMaze & indep. coins & rule & 32 & 0.1227 & 0.123--0.126 & 0.750 & 7.72 to 7.72 & yes\\
RedMaze & indep. coins & + repair & 44 & 0.1133 & 0.113--0.116 & 0.750 & 9.79 to 9.79 & yes\\
RedMaze & indep. coins$^{c}$ & + mending & -- & 0.1144 & 0.114 & 0.750 & -- & no: absolute lattice\\
RedMaze & phase input & rule & 32 & 0.0000 & 0.004 & 0.750 & $-$0.75 & no: coordinate input\\
RedMaze & phase input & + repair & 44 & 0.0000 & 0.004 & 0.750 & $-$0.75 & no: coordinate input\\
\midrule
Fabric & indep. coins & rule & 32 & 0.1130 & 0.152--0.265 & 0.600 & 7.23 to 7.35 & yes\\
Fabric & indep. coins & + repair & 44 & 0.1056 & 0.155--0.261 & 0.600 & 9.27 to 9.37 & yes\\
Fabric & indep. coins$^{c}$ & + mending & -- & 0.1062 & 0.200 & 0.600 & -- & no: absolute lattice\\
\bottomrule\end{tabular}

%% file: tables/floorcheck_v12b.tex
\setlength{\tabcolsep}{3pt}%
\begin{tabular}{@{}lllrrrrrl@{}}\toprule
example & generator & output & $r$ & $\varepsilon$ & $\delta$ & $\kappa_f$ & slack & covered\\\midrule
SimpleWall & indep. coins & rule & 32 & 0.0556 & 0.056--0.110 & 0.529 & 3.31 to 3.36 & yes\\
SimpleWall & indep. coins & + repair & 44 & 0.0531 & 0.055--0.108 & 0.529 & 4.42 to 4.47 & yes\\
SimpleWall & indep. coins$^{c}$ & rule & 32 & 0.0545 & 0.081 & 0.529 & 3.26 & yes\\
SimpleWall & indep. coins$^{c}$ & + repair & 44 & 0.0520 & 0.081 & 0.529 & 4.33 & yes\\
SimpleWall & indep. coins$^{c}$ & + mending & -- & 0.0457 & 0.082 & 0.529 & -- & no: absolute lattice\\
SimpleWall & indep. coins$^{c}$ & + mending & -- & 0.0474 & 0.083 & 0.529 & -- & no: absolute lattice\\
SimpleWall & indep. coins, fine-tuned (1)$^{c}$ & rule & 32 & 0.0545 & 0.103 & 0.529 & 3.28 & yes\\
SimpleWall & indep. coins, fine-tuned (1)$^{c}$ & + repair & 44 & 0.0528 & 0.102 & 0.529 & 4.43 & yes\\
SimpleWall & indep. coins, fine-tuned (1)$^{c}$ & + mending & -- & 0.0466 & 0.094 & 0.529 & -- & no: absolute lattice\\
SimpleWall & indep. coins, fine-tuned (3)$^{c}$ & rule & 32 & 0.0571 & 0.092 & 0.529 & 3.45 & yes\\
SimpleWall & indep. coins, fine-tuned (3)$^{c}$ & + repair & 44 & 0.0534 & 0.091 & 0.529 & 4.48 & yes\\
SimpleWall & indep. coins, fine-tuned (3)$^{c}$ & + mending & -- & 0.0468 & 0.089 & 0.529 & -- & no: absolute lattice\\
SimpleWall & indep. coins, fine-tuned (10)$^{c}$ & rule & 32 & 0.0660 & 0.071 & 0.529 & 4.03 & yes\\
SimpleWall & indep. coins, fine-tuned (10)$^{c}$ & + repair & 44 & 0.0604 & 0.067 & 0.529 & 5.10 & yes\\
SimpleWall & indep. coins, fine-tuned (10)$^{c}$ & + mending & -- & 0.0543 & 0.064 & 0.529 & -- & no: absolute lattice\\
SimpleWall & row coins & rule & 32 & 0.0179 & 0.082--0.100 & 0.529 & 0.77 to 0.79 & no: shared coins\\
SimpleWall & row coins & + repair & 44 & 0.0165 & 0.083--0.099 & 0.529 & 1.08 to 1.09 & no: shared coins\\
SimpleWall & row coins$^{c}$ & rule & 32 & 0.0179 & 0.090 & 0.529 & 0.78 & no: shared coins\\
SimpleWall & row coins$^{c}$ & + repair & 44 & 0.0165 & 0.090 & 0.529 & 1.08 & no: shared coins\\
SimpleWall & row coins$^{c}$ & + mending & -- & 0.0005 & 0.120 & 0.529 & -- & no: absolute lattice\\
SimpleWall & row coins, fine-tuned (3)$^{c}$ & rule & 32 & 0.0184 & 0.091 & 0.529 & 0.81 & no: shared coins\\
SimpleWall & row coins, fine-tuned (3)$^{c}$ & + repair & 44 & 0.0158 & 0.091 & 0.529 & 1.01 & no: shared coins\\
SimpleWall & row coins, fine-tuned (3)$^{c}$ & + mending & -- & 0.0000 & 0.132 & 0.529 & -- & no: absolute lattice\\
SimpleWall & row coins, fine-tuned (10)$^{c}$ & rule & 32 & 0.0233 & 0.061 & 0.529 & 1.11 & no: shared coins\\
SimpleWall & row coins, fine-tuned (10)$^{c}$ & + repair & 44 & 0.0194 & 0.059 & 0.529 & 1.32 & no: shared coins\\
SimpleWall & row coins, fine-tuned (10)$^{c}$ & + mending & -- & 0.0005 & 0.123 & 0.529 & -- & no: absolute lattice\\
SimpleWall & row coins, fine-tuned (30)$^{c}$ & rule & 32 & 0.0349 & 0.074 & 0.529 & 1.92 & no: shared coins\\
SimpleWall & row coins, fine-tuned (30)$^{c}$ & + repair & 44 & 0.0288 & 0.076 & 0.529 & 2.19 & no: shared coins\\
SimpleWall & row coins, fine-tuned (30)$^{c}$ & + mending & -- & 0.0002 & 0.112 & 0.529 & -- & no: absolute lattice\\
SimpleWall & column coins$^{c}$ & rule & 32 & 0.0428 & 0.133 & 0.529 & 2.51 & no: shared coins\\
SimpleWall & column coins$^{c}$ & + repair & 44 & 0.0415 & 0.132 & 0.529 & 3.42 & no: shared coins\\
SimpleWall & column coins$^{c}$ & + mending & -- & 0.0385 & 0.133 & 0.529 & -- & no: absolute lattice\\
\bottomrule\end{tabular}

%% file: tables/per_sample_v12.tex
\begin{tabular}{lrrrrrrrrccr}\toprule
example & $|K|$ & $C$ & raw $R'{=}16$ & $R'{=}32$ & $R'{=}64$ & repaired $R'{=}64$ & KL & KL$_{\mathrm{WFC}}$ & seam & coin edit & train (min)\\\midrule
3Bricks & 423 & 3 & 0.1139 & 0.0727 & 0.0767 & 0.0565 & 1.753 & 0.927 & y & y & 45\\
Angular & 49 & 3 & 0.0221 & 0.0097 & 0.0104 & 0.0058 & 0.411 & 0.080 & y & y & 56\\
Cat & 136 & 4 & 0.0743 & 0.0686 & 0.0794 & 0.0635 & 0.186 & 0.262 & y & y & 73\\
Cats & 67 & 3 & 0.0235 & 0.0200 & 0.0235 & 0.0178 & 0.146 & -- & y & y & 55\\
Cave & 55 & 2 & 0.0036 & 0.0015 & 0.0012 & 0.0010 & 0.027 & 0.095 & y & y & 54\\
Chess & 2 & 2 & 0.1174 & 0.0592 & 0.0363 & 0.0350 & 0.000 & 0.000 & y & y & 52\\
City & 41 & 3 & 0.0820 & 0.0428 & 0.0423 & 0.0303 & 0.006 & 0.008 & y & y & 53\\
Col3 & 240 & 3 & 0.0162 & 0.0086 & 0.0069 & 0.0059 & 0.008 & -- & y & y & 90\\
ColoredCity & 93 & 5 & 0.1237 & 0.0914 & 0.1473 & 0.0825 & 0.056 & 0.275 & y & y & 59\\
Dungeon & 57 & 2 & 0.0264 & 0.0181 & 0.0194 & 0.0198 & 0.021 & 0.090 & y & y & 55\\
Fabric & 19 & 3 & 0.1368 & 0.1158 & 0.2453 & 0.1721 & 0.048 & 0.046 & y & y & 52\\
Flowers & 92 & 4 & 0.0611 & 0.0584 & 0.0948 & 0.0585 & 0.121 & 4.031 & y & y & 58\\
Font & 210 & 2 & 0.0015 & 0.0010 & 0.0010 & 0.0008 & 0.100 & 0.493 & y & y & 86\\
Forest & 77 & 2 & 0.0068 & 0.0048 & 0.0049 & 0.0037 & 0.028 & 0.207 & y & y & 58\\
Hogs & 32 & 2 & 0.0071 & 0.0045 & 0.0047 & 0.0029 & 0.020 & 0.034 & y & y & 54\\
Knot & 57 & 2 & 0.0260 & 0.0165 & 0.0164 & 0.0143 & 0.033 & 0.144 & y & y & 55\\
Lake & 155 & 3 & 0.0685 & 0.0594 & 0.0656 & 0.0559 & 0.180 & 0.558 & y & y & 76\\
LessRooms & 69 & 2 & 0.0391 & 0.0320 & 0.0344 & 0.0323 & 0.101 & 0.102 & y & y & 54\\
Lines & 35 & 2 & 0.0231 & 0.0134 & 0.0120 & 0.0131 & 0.018 & 0.028 & y & y & 51\\
Link & 87 & 3 & 0.0532 & 0.0370 & 0.0637 & 0.0445 & 0.062 & 0.083 & y & y & 61\\
Link2 & 71 & 3 & 0.0421 & 0.0308 & 0.0429 & 0.0335 & 0.060 & 0.218 & y & y & 58\\
MagicOffice & 96 & 3 & 0.0301 & 0.0238 & 0.0256 & 0.0220 & 0.106 & 0.291 & y & y & 61\\
Maze & 139 & 2 & 0.0168 & 0.0122 & 0.0118 & 0.0090 & 0.087 & 0.162 & y & y & 70\\
Mazelike & 76 & 2 & 0.0435 & 0.0316 & 0.0363 & 0.0276 & 0.033 & 0.232 & y & y & 55\\
MoreFlowers & 108 & 4 & 0.0605 & 0.0609 & 0.0868 & 0.0551 & 0.209 & 3.987 & y & y & 63\\
Mountains & 31 & 2 & 0.0242 & 0.0149 & 0.0140 & 0.0118 & 0.008 & 0.207 & y & y & 51\\
Nested & 41 & 2 & 0.0497 & 0.0289 & 0.0276 & 0.0240 & 0.079 & 0.623 & y & y & 53\\
NotKnot & 47 & 2 & 0.0252 & 0.0196 & 0.0203 & 0.0167 & 0.126 & 0.393 & y & y & 51\\
Office & 66 & 2 & 0.0252 & 0.0109 & 0.0090 & 0.0061 & 0.021 & 0.159 & y & y & 57\\
Office2 & 145 & 4 & 0.0458 & 0.0308 & 0.0361 & 0.0245 & 0.066 & 0.386 & y & y & 72\\
Paths & 51 & 2 & 0.0132 & 0.0069 & 0.0077 & 0.0073 & 0.031 & 0.097 & y & y & 52\\
Platformer & 109 & 5 & 0.0455 & 0.0380 & 0.0467 & 0.0317 & 0.054 & 1.134 & y & y & 62\\
Qud & 61 & 3 & 0.0702 & 0.0594 & 0.0900 & 0.0612 & 0.138 & 0.191 & y & y & 53\\
RedMaze & 16 & 3 & 0.1992 & 0.1227 & 0.4322 & 0.2755 & 0.000 & 0.001 & y & y & 53\\
Rooms & 61 & 2 & 0.0230 & 0.0156 & 0.0293 & 0.0191 & 0.080 & 0.213 & y & y & 56\\
Rule126 & 124 & 2 & 0.0416 & 0.0314 & 0.0300 & 0.0266 & 0.137 & 0.357 & y & y & 64\\
Sand & 79 & 2 & 0.0104 & 0.0077 & 0.0074 & 0.0061 & 0.059 & 0.319 & y & y & 56\\
ScaledMaze & 40 & 2 & 0.0373 & 0.0271 & 0.0294 & 0.0237 & 0.020 & 0.017 & y & y & 53\\
Sewers & 105 & 3 & 0.0604 & 0.0541 & 0.0662 & 0.0511 & 0.182 & 0.229 & y & y & 63\\
SimpleKnot & 50 & 3 & 0.0345 & 0.0227 & 0.0365 & 0.0254 & 0.034 & 0.149 & y & y & 56\\
SimpleMaze & 16 & 2 & 0.1782 & 0.1059 & 0.1136 & 0.0853 & 0.000 & 0.001 & y & y & 57\\
SimpleWall & 20 & 2 & 0.0808 & 0.0553 & 0.0512 & 0.0450 & 0.014 & 0.021 & y & y & 50\\
Skew1 & 48 & 2 & 0.0398 & 0.0301 & 0.0284 & 0.0277 & 0.030 & 0.454 & y & y & 54\\
Skew2 & 59 & 2 & 0.0441 & 0.0346 & 0.0372 & 0.0326 & 0.021 & 0.096 & y & y & 55\\
Skyline & 153 & 5 & 0.0971 & 0.0869 & 0.1189 & 0.0870 & 0.259 & 1.486 & y & y & 72\\
Skyline2 & 162 & 5 & 0.0835 & 0.0795 & 0.1306 & 0.0784 & 0.116 & 2.604 & y & y & 77\\
SmileCity & 47 & 3 & 0.1232 & 0.0990 & 0.1293 & 0.0925 & 0.043 & 0.098 & y & y & 52\\
Spirals & 78 & 2 & 0.0331 & 0.0252 & 0.0286 & 0.0214 & 0.082 & 0.123 & y & y & 58\\
Stripes & 2 & 2 & 0.1175 & 0.0636 & 0.0468 & 0.0383 & 0.000 & -- & y & y & 55\\
Town & 87 & 3 & 0.1040 & 0.0785 & 0.0893 & 0.0655 & 0.108 & 0.194 & y & y & 58\\
TrickKnot & 78 & 3 & 0.0385 & 0.0342 & 0.0494 & 0.0319 & 0.057 & 0.126 & y & y & 56\\
Village & 50 & 3 & 0.0137 & 0.0094 & 0.0702 & 0.0304 & 0.211 & 0.290 & y & y & 51\\
Wall & 38 & 4 & 0.1604 & 0.1069 & 0.0997 & 0.0846 & 0.045 & 0.136 & y & y & 56\\
Water & 12 & 2 & 0.0163 & 0.0112 & 0.0108 & 0.0093 & 0.003 & 0.006 & y & y & 56\\
Wrinkles & 36 & 2 & 0.0251 & 0.0200 & 0.0196 & 0.0202 & 0.046 & 0.047 & y & y & 55\\
proper3 & 246 & 3 & 0.0112 & 0.0046 & 0.0038 & 0.0035 & 0.010 & -- & y & y & 96\\
\bottomrule\end{tabular}

%% file: tables/mendall.tex
\begin{tabular}{lrrrrrr}\toprule
example & $|K|$ & rule & + repair & + mending & cells changed & $\rho\le2$\\\midrule
Cave & 55 & 0.0015 & 0.0010 & 0.00000 & 0.3\% & 100\%\\
Font & 210 & 0.0011 & 0.0007 & 0.00000 & 0.3\% & 100\%\\
Hogs & 32 & 0.0053 & 0.0037 & 0.00000 & 2.8\% & 90\%\\
Sand & 79 & 0.0077 & 0.0068 & 0.00000 & 3.8\% & 83\%\\
Forest & 77 & 0.0050 & 0.0043 & 0.00000 & 3.1\% & 79\%\\
Maze & 139 & 0.0136 & 0.0115 & 0.00002 & 7.1\% & 87\%\\
Paths & 51 & 0.0076 & 0.0081 & 0.00003 & 3.9\% & 67\%\\
Spirals & 78 & 0.0264 & 0.0232 & 0.00005 & 12.9\% & 69\%\\
NotKnot & 47 & 0.0206 & 0.0181 & 0.00006 & 10.2\% & 57\%\\
Rooms & 61 & 0.0162 & 0.0138 & 0.00008 & 10.7\% & 65\%\\
LessRooms & 69 & 0.0318 & 0.0323 & 0.00017 & 18.3\% & 69\%\\
Rule126 & 124 & 0.0320 & 0.0282 & 0.00054 & 18.0\% & 51\%\\
MagicOffice & 96 & 0.0229 & 0.0224 & 0.00056 & 10.7\% & 62\%\\
Wrinkles & 36 & 0.0196 & 0.0220 & 0.00060 & 11.2\% & 53\%\\
Knot & 57 & 0.0173 & 0.0165 & 0.00068 & 8.9\% & 39\%\\
Village & 50 & 0.0094 & 0.0063 & 0.00085 & 2.4\% & 21\%\\
Dungeon & 57 & 0.0185 & 0.0196 & 0.00129 & 10.1\% & 41\%\\
Angular & 49 & 0.0086 & 0.0070 & 0.00143 & 1.9\% & 35\%\\
Office & 66 & 0.0103 & 0.0085 & 0.00202 & 3.9\% & 32\%\\
Cats & 67 & 0.0166 & 0.0154 & 0.00349 & 5.9\% & 23\%\\
Lines & 35 & 0.0151 & 0.0168 & 0.00522 & 4.0\% & 53\%\\
SimpleKnot & 50 & 0.0226 & 0.0232 & 0.00524 & 5.8\% & 50\%\\
Water & 12 & 0.0110 & 0.0104 & 0.00569 & 1.6\% & 34\%\\
Office2 & 145 & 0.0313 & 0.0250 & 0.00806 & 9.2\% & 41\%\\
Mountains & 31 & 0.0140 & 0.0134 & 0.01082 & 0.7\% & 68\%\\
TrickKnot & 78 & 0.0319 & 0.0287 & 0.01125 & 8.7\% & 36\%\\
Link & 87 & 0.0369 & 0.0383 & 0.01722 & 6.7\% & 50\%\\
Link2 & 71 & 0.0285 & 0.0295 & 0.01807 & 3.0\% & 56\%\\
ScaledMaze & 40 & 0.0267 & 0.0249 & 0.01818 & 3.3\% & 70\%\\
City & 41 & 0.0403 & 0.0343 & 0.02141 & 4.8\% & 36\%\\
Mazelike & 76 & 0.0302 & 0.0281 & 0.02144 & 3.6\% & 42\%\\
Skew1 & 48 & 0.0312 & 0.0317 & 0.02365 & 2.0\% & 62\%\\
MoreFlowers & 108 & 0.0574 & 0.0476 & 0.02513 & 9.8\% & 54\%\\
Nested & 41 & 0.0308 & 0.0288 & 0.02534 & 0.9\% & 45\%\\
Qud & 61 & 0.0570 & 0.0528 & 0.02534 & 10.8\% & 43\%\\
Sewers & 105 & 0.0545 & 0.0501 & 0.02551 & 13.9\% & 50\%\\
Skew2 & 59 & 0.0343 & 0.0348 & 0.02570 & 3.3\% & 58\%\\
Platformer & 109 & 0.0376 & 0.0314 & 0.02767 & 0.6\% & 80\%\\
Lake & 155 & 0.0605 & 0.0564 & 0.02908 & 12.3\% & 56\%\\
Flowers & 92 & 0.0598 & 0.0490 & 0.03506 & 6.3\% & 52\%\\
SimpleWall & 20 & 0.0521 & 0.0494 & 0.04622 & 0.5\% & 61\%\\
Cat & 136 & 0.0717 & 0.0627 & 0.04782 & 7.1\% & 63\%\\
3Bricks & 423 & 0.0691 & 0.0597 & 0.05284 & 2.0\% & 85\%\\
Skyline2 & 162 & 0.0770 & 0.0660 & 0.06219 & 1.5\% & 65\%\\
ColoredCity & 93 & 0.0917 & 0.0759 & 0.06591 & 4.0\% & 35\%\\
Skyline & 153 & 0.0897 & 0.0793 & 0.06719 & 4.1\% & 79\%\\
Town & 87 & 0.0794 & 0.0706 & 0.06755 & 1.4\% & 60\%\\
SmileCity & 47 & 0.0985 & 0.0908 & 0.08926 & 0.4\% & 59\%\\
Wall & 38 & 0.1098 & 0.1023 & 0.08959 & 3.7\% & 76\%\\
SimpleMaze & 16 & 0.1071 & 0.1003 & 0.10035 & 0.0\% & 0\%\\
Fabric & 19 & 0.1147 & 0.1073 & 0.10613 & 0.1\% & 67\%\\
RedMaze & 16 & 0.1220 & 0.1135 & 0.11346 & 0.0\% & 0\%\\
\midrule
\multicolumn{7}{l}{\textit{theory controls} ($^\ast$): counted in no class and in no total above}\\
Col3$^\ast$ & 240 & 0.0087 & 0.0084 & 0.00587 & 0.9\% & 74\%\\
Chess$^\ast$ & 2 & 0.0575 & 0.0577 & 0.05768 & 0.0\% & 0\%\\
Stripes$^\ast$ & 2 & 0.0615 & 0.0615 & 0.06126 & 0.0\% & 100\%\\
\bottomrule\end{tabular}

%% file: tables/forced.tex
\begin{tabular}{lrrrrrrr}\toprule
example & trials & valid & no net defect & defects/pin & max dist. & cells & residual\\\midrule
Angular & 12 & 0 & 10 & 34 & -- & -- & 78\\
Cave & 12 & 12 & 12 & 13 & 5 & 36 & 0\\
Font & 12 & 12 & 12 & 3 & 3 & 10 & 0\\
Forest & 12 & 11 & 12 & 12 & 7 & 37 & 0\\
Hogs & 12 & 12 & 12 & 14 & 5 & 20 & 0\\
Maze & 12 & 12 & 12 & 21 & 7 & 70 & 0\\
Paths & 12 & 2 & 12 & 13 & 4 & 20 & 2\\
Sand & 12 & 12 & 11 & 8 & 4 & 16 & 0\\
\bottomrule\end{tabular}

%% file: tables/wfcstats_v13.tex
\begin{tabular}{@{}lrrr@{\hskip 0.9em}rrr@{\hskip 0.9em}rrr@{}}\toprule
 & \multicolumn{3}{c}{$64^2$} & \multicolumn{3}{c}{$128^2$} & \multicolumn{3}{c}{$256^2$}\\
\cmidrule(lr){2-4}\cmidrule(lr){5-7}\cmidrule(lr){8-10}
example & indep. & rule & WFC & indep. & rule & WFC & indep. & rule & WFC\\\midrule
3Bricks & 4.427 & 4.541 & 1.316 & 3.644 & 3.848 & -- & 2.937 & 3.272 & --\\
Angular & 0.489 & 0.056 & 0.082 & 0.447 & 0.047 & 0.079 & 0.423 & 0.042 & 0.098\\
Cat & 0.305 & 0.473 & 0.281 & 0.182 & 0.259 & 0.334 & 0.187 & 0.241 & 0.247$^\ddagger$\\
Cats & 0.173 & 0.126 & -- & 0.139 & 0.097 & -- & 0.140 & 0.091 & 0.086$^\ddagger$\\
Cave & 0.048 & 0.026 & 0.082 & 0.033 & 0.020 & 0.078 & 0.026 & 0.013 & 0.081\\
City & 0.010 & 0.020 & 0.015 & 0.010 & 0.018 & 0.015 & 0.008 & 0.012 & 0.013\\
ColoredCity & 0.088 & 0.181 & 0.232 & 0.061 & 0.045 & 0.245 & 0.058 & 0.030 & 0.275\\
Dungeon & 0.039 & 0.097 & 0.094 & 0.025 & 0.046 & 0.139 & 0.021 & 0.037 & 0.145$^\ddagger$\\
Fabric & 0.052 & 0.048 & 0.076 & 0.038 & 0.038 & 0.062 & 0.048 & 0.041 & 0.060\\
Flowers & 0.282 & 0.454 & 3.648 & 0.146 & 0.098 & 3.179 & 0.136 & 0.084 & 4.042\\
Font & 0.418 & 0.410 & 0.418 & 0.182 & 0.145 & 0.114 & 0.139 & 0.093 & 0.067\\
Forest & 0.044 & 0.039 & 0.432 & 0.028 & 0.031 & 0.165 & 0.028 & 0.031 & 0.207\\
Hogs & 0.022 & 0.063 & 0.045 & 0.019 & 0.018 & 0.046 & 0.018 & 0.017 & 0.051\\
Knot & 0.057 & 0.340 & 0.242 & 0.038 & 0.203 & 0.139 & 0.035 & 0.155 & 0.147\\
Lake & 0.440 & 0.427 & 0.631 & 0.202 & 0.166 & 0.271 & 0.179 & 0.150 & 0.294$^\ddagger$\\
LessRooms & 0.146 & 0.165 & 0.151 & 0.109 & 0.091 & 0.108 & 0.096 & 0.087 & 0.117\\
Lines & 0.021 & 0.115 & 0.042 & 0.019 & 0.116 & 0.049 & 0.019 & 0.107 & 0.049\\
Link & 0.073 & 0.398 & 0.091 & 0.059 & 0.134 & 0.062 & 0.058 & 0.082 & 0.065\\
Link2 & 0.079 & 0.070 & 0.240 & 0.054 & 0.075 & 0.211 & 0.064 & 0.068 & --\\
MagicOffice & 0.195 & 0.403 & 0.754 & 0.117 & 0.142 & 0.330 & 0.110 & 0.125 & 0.375$^\ddagger$\\
Maze & 0.119 & 0.134 & 0.159 & 0.092 & 0.095 & 0.123 & 0.089 & 0.088 & 0.112\\
Mazelike & 0.067 & 0.146 & 0.256 & 0.043 & 0.098 & 0.161 & 0.033 & 0.068 & 0.134$^\ddagger$\\
MoreFlowers & 0.618 & 0.804 & 3.540 & 0.285 & 0.161 & 1.256 & 0.253 & 0.130 & 2.197$^\ddagger$\\
Mountains & 0.017 & 0.020 & 0.114 & 0.010 & 0.013 & 0.104 & 0.008 & 0.011 & 0.077\\
Nested & 0.093 & 0.144 & 0.554 & 0.082 & 0.064 & 0.553 & 0.079 & 0.089 & 0.481\\
NotKnot & 0.129 & 0.126 & 0.388 & 0.123 & 0.115 & 0.460 & 0.124 & 0.118 & 0.521$^\ddagger$\\
Office & 0.033 & 0.088 & 0.191 & 0.023 & 0.064 & 0.206 & 0.022 & 0.058 & 0.217\\
Office2 & 0.158 & 0.541 & 0.444 & 0.089 & 0.126 & 0.496 & 0.069 & 0.095 & 0.432$^\ddagger$\\
Paths & 0.055 & 0.051 & 0.067 & 0.031 & 0.047 & 0.068 & 0.032 & 0.044 & 0.073\\
Platformer & 0.146 & 0.565 & 1.187 & 0.060 & 0.204 & 0.741 & 0.052 & 0.145 & 0.314\\
Qud & 0.179 & 0.290 & 0.166 & 0.155 & 0.237 & 0.234 & 0.144 & 0.190 & 0.168\\
RedMaze & 0.001 & 0.004 & 0.001 & 0.000 & 0.001 & 0.000 & 0.000 & 0.001 & 0.000\\
Rooms & 0.089 & 0.176 & 0.215 & 0.089 & 0.063 & 0.194 & 0.081 & 0.052 & 0.185\\
Rule126 & 0.321 & 0.363 & 0.290 & 0.225 & 0.256 & 0.338 & 0.162 & 0.205 & 0.218$^\ddagger$\\
Sand & 0.104 & 0.102 & 0.275 & 0.059 & 0.037 & 0.257 & 0.060 & 0.038 & 0.279\\
ScaledMaze & 0.019 & 0.025 & 0.019 & 0.023 & 0.018 & 0.017 & 0.020 & 0.017 & 0.013\\
Sewers & 0.231 & 1.093 & 0.385 & 0.189 & 0.286 & 0.173 & 0.186 & 0.270 & 0.292\\
SimpleKnot & 0.043 & 0.054 & 0.134 & 0.034 & 0.034 & 0.128 & 0.036 & 0.031 & 0.135\\
SimpleMaze & 0.000 & 0.003 & 0.001 & 0.000 & 0.002 & 0.000 & 0.000 & 0.002 & 0.000\\
SimpleWall & 0.017 & 0.108 & 0.034 & 0.015 & 0.066 & 0.046 & 0.015 & 0.066 & 0.038\\
Skew1 & 0.056 & 0.094 & 0.210 & 0.038 & 0.074 & 0.241 & 0.033 & 0.059 & 0.443\\
Skew2 & 0.039 & 0.230 & 0.163 & 0.025 & 0.108 & 0.127 & 0.022 & 0.073 & 0.081\\
Skyline & 0.994 & 1.429 & 0.794 & 0.417 & 0.587 & 0.342 & 0.262 & 0.421 & 0.620\\
Skyline2 & 0.454 & 1.082 & 1.983 & 0.144 & 0.238 & 1.542 & 0.122 & 0.097 & 0.463\\
SmileCity & 0.046 & 0.185 & 0.166 & 0.046 & 0.179 & 0.114 & 0.044 & 0.167 & 0.277\\
Spirals & 0.095 & 0.127 & 0.155 & 0.082 & 0.127 & 0.169 & 0.079 & 0.116 & 0.172$^\ddagger$\\
Town & 0.131 & 1.315 & 0.267 & 0.124 & 0.371 & 0.320 & 0.108 & 0.306 & 0.351\\
TrickKnot & 0.070 & 0.099 & 0.235 & 0.064 & 0.066 & 0.112 & 0.058 & 0.062 & 0.121\\
Village & 0.331 & 0.347 & 0.221 & 0.195 & 0.230 & 0.094 & 0.209 & 0.229 & 0.095\\
Wall & 0.057 & 0.104 & 0.176 & 0.047 & 0.085 & 0.282 & 0.047 & 0.074 & 0.156\\
Water & 0.006 & 0.019 & 0.006 & 0.003 & 0.015 & 0.004 & 0.004 & 0.015 & --\\
Wrinkles & 0.047 & 0.081 & 0.053 & 0.042 & 0.072 & 0.055 & 0.044 & 0.074 & 0.063$^\ddagger$\\
\midrule at or below WFC & 42/51 & 27/51 & & 43/50 & 33/50 & & 43/49 & 37/49 \\
\bottomrule\end{tabular}

%% file: tables/variants_v12.tex
\begin{tabular}{llrrrrrr}\toprule
variant & change from the first configuration & $n$ & @16 & @32 & @64 & monotone 8$\to$64 & collapsed\\\midrule
longer training & 16\,000 updates, cosine schedule & 5 & 0.59 & 0.53 & 0.86 & 0/5 & 0\\
hard windows & temperature 0.2, forbidden-window weight 4 & 5 & 0.83 & 0.65 & 0.65 & 1/5 & 0\\
score-function reward & per-window credit to the $N^2$ cells & 5 & 1.11 & 0.84 & 0.82 & 5/5 & 0\\
random train rounds & rounds drawn from 8--32 & 4 & 1.09 & 0.75 & 0.73 & 3/4 & 0\\
hard windows + random rounds & the two above together & 4 & 0.84 & 0.66 & 0.71 & 1/4 & 0\\
hard + score function + random rounds & the three above together & 4 & 1.23 & 0.90 & 0.89 & 2/4 & 0\\
longer training + hard + random rounds (final) & the configuration used in the paper & 4 & 0.50 & 0.31 & 0.49 & 0/4 & 0\\
pattern-space rule & soft, parallel propagation & 5 & 2.55 & 2.09 & 1.83 & 5/5 & 0\\
\bottomrule\end{tabular}